%% file: pinn_ntk_arxiv.tex
\documentclass[11pt]{article}
\usepackage[T1]{fontenc}
\usepackage{lmodern}
\usepackage[margin=1in]{geometry}
\usepackage[round]{natbib}

\input{paper_preamble}

\newcommand{\coderepo}{\url{https://github.com/zifanlyu/pinn-ntk-certificates}}

\input{paper_title}

\author{Zifan Lyu\\
ETH Z\"urich\\
\texttt{ziflyu@ethz.ch}}
\date{}

\begin{document}
\maketitle

\input{paper_body}

\section*{Acknowledgments}
The author thanks Prof.\ Dr.\ Siddhartha Mishra for helpful discussion.

\bibliographystyle{plainnat}
\bibliography{myReferences}

\clearpage
\input{paper_appendix}

\end{document}

%% file: paper_preamble.tex
\usepackage{microtype}
\usepackage{graphicx}
\usepackage{float}
\usepackage{subfigure}
\usepackage{booktabs}
\usepackage{array}
\usepackage{longtable}
\usepackage{enumitem}
\usepackage{amsmath}
\usepackage{amssymb}
\usepackage{mathtools}
\usepackage{amsthm}
\usepackage{hyperref}
\usepackage{url}
\usepackage[capitalize,noabbrev]{cleveref}
\hypersetup{colorlinks=true,allcolors=black}

\makeatletter
\newcommand{\appendixcontentsmark}{}
\newcommand{\appendixcontents}{%
  \begingroup
  \renewcommand{\contentsname}{Appendix Contents}%
  \setcounter{tocdepth}{2}%
  \let\appendix@contentsline\contentsline
  \renewcommand{\contentsline}[4]{}%
  \renewcommand{\appendixcontentsmark}{%
    \let\contentsline\appendix@contentsline}%
  \fontsize{8}{8.5}\selectfont
  \tableofcontents
  \endgroup
}
\makeatother

\DeclareMathOperator{\rank}{rank}
\DeclareMathOperator{\linspan}{span}
\DeclareMathOperator{\ReLU}{ReLU}

\DeclareMathOperator{\supp}{supp}
\DeclareMathOperator{\Unif}{Unif}
\DeclareMathOperator{\Range}{Range}
\DeclareMathOperator{\spec}{spec}
\DeclareMathOperator{\sech}{sech}
\DeclareMathOperator{\Var}{Var}
\DeclareMathOperator*{\esssup}{ess\,sup}

\theoremstyle{plain}
\newtheorem{theorem}{Theorem}[section]
\newtheorem{proposition}[theorem]{Proposition}
\newtheorem{lemma}[theorem]{Lemma}
\newtheorem{corollary}[theorem]{Corollary}
\theoremstyle{definition}
\newtheorem{definition}[theorem]{Definition}
\newtheorem{assumption}[theorem]{Assumption}
\theoremstyle{remark}
\newtheorem{remark}[theorem]{Remark}
\newtheorem{example}[theorem]{Example}
\theoremstyle{plain}
\newtheorem{conjecture}[theorem]{Conjecture}
\crefname{conjecture}{conjecture}{conjectures}
\Crefname{conjecture}{Conjecture}{Conjectures}
\crefname{assumption}{assumption}{assumptions}
\Crefname{assumption}{Assumption}{Assumptions}

%% file: paper_title.tex
\title{Checkable NTK Positivity and Finite-Width Gradient Descent
for Scalar- and Vector-Valued PINNs with Strong-Form,
Weak-Form, and Nonlocal Linear Constraints}

%% file: paper_body.tex
\begin{abstract}
We give checkable positive-definiteness criteria for the limiting neural
tangent kernel (NTK) and high-probability finite-width gradient-descent
guarantees for scalar- or vector-valued physics-informed neural networks
(PINNs) with linear constraints. The constraints may be strong-form
differential rows of any fixed finite order, including coupled systems, with
any linear initial or boundary conditions; weak-form residual and boundary
functionals; or finite-measure nonlocal observations such as integral,
nonlocal-diffusion, and Dirac-data rows, all in any dimension. For each class, positive
definiteness of the limiting NTK is equivalent to a rank condition on a
finite coefficient, functional, or moment matrix of the fixed design: a
certificate computed before training that detects structural zero modes. The
model hypotheses are those of an ordinary two-layer PINN, a smooth
nonpolynomial activation with bounded symmetric initialization, and are met
by standard choices such as $\tanh$ with uniform initialization. Given a
certificate, explicit width and step-size
conditions ensure, with high probability, that the empirical NTK retains at
least half the limiting gap and that full-batch gradient descent decreases
the training loss geometrically at every iteration. Controlled
experiments check the certificates and the finite-width mechanism.
\end{abstract}

\section{Introduction}
Physics-informed neural networks incorporate sampled differential equations
and boundary data into a training
loss~\citep{Dissanayake1994,Lagaris1998,Lagaris2000,Raissi2019}.
The solution error comprises approximation, discretization or
generalization, and optimization components. Universal approximation supports
expressivity~\citep{Hornik1989}, and PINN error and partial differential
equation (PDE) stability analyses address approximation and discretization
under problem-dependent assumptions~\citep{DeRyck2024,Shin2020,Mishra2023},
but the nonconvex training problem remains
difficult~\citep{Wang2021,Wang2022}.

When the width is large relative to the number of sampled constraints, NTK
theory reduces this optimization to a spectral question: a strictly positive
smallest tangent-kernel eigenvalue makes the residual dynamics of gradient
descent (GD) contractive, provided the finite-width Gram matrix remains close
to its initialization~\citep{Jacot2018,Lee2019,Du2019}. In a PINN this Gram
matrix encodes sampled constraints rather than input examples: it is indexed
by point--operator pairs for strong-form losses and by bounded linear
functionals for weak-form or nonlocal losses; positivity must hold jointly
across PDE, initial, and boundary constraints; and one physical site may carry
several scalar equations. Postulating a spectral gap without verifying it therefore
leaves a central premise of the convergence argument unresolved.

We replace the unverified qualitative nondegeneracy premise with exact
structural criteria for finite collections of strong-form local differential,
weak-form functional, and nonlocal linear constraints. The resulting gap is positive for each fixed
nonredundant design, but its magnitude remains problem-dependent.
Under the hypotheses stated below, our argument follows the chain
\begin{equation*}
\begin{aligned}
&\text{nonredundancy}
\Longleftrightarrow \text{positive-definiteness certificate}\\
&\Longrightarrow \text{positive-definite finite-width kernel during GD}
\Longrightarrow \text{geometric loss decay}.
\end{aligned}
\end{equation*}
The exact verification in the first step is a \emph{positive-definiteness
certificate} (\Cref{def:positive_definiteness_certificate}). The final
implication uses only such a certificate and local row regularity of the
one-neuron prediction maps, so one finite-width GD argument covers strong-form
differential rows of arbitrary fixed finite order together with admissible
weak-form and nonlocal rows. Nonredundancy means independence of the
co-located coefficient vectors, of the combined residual and boundary
functionals, or of the loss-scaled signed-measure tuples, for the three
classes respectively. Thus,
for each fixed finite design in our setting, the limiting kernel is singular
exactly when the sampled constraints have a structural dependency; compatible
targets for such singular designs are treated by exact active-subspace
compression in~\Cref{sec:compatible_singular_kernels}.
Our contributions are:
\begin{itemize}[leftmargin=*,nosep]
    \item \textbf{Checkable positive-definiteness certificates.} For
    strong-form rows of any fixed finite order, scalar or coupled, and for
    weak-form and finite-measure nonlocal rows, all in any dimension, we derive
    exact limiting-NTK nullspace and rank formulas through sitewise
    coefficient vectors, joint functional dependencies, or signed-measure
    dependencies. They cover repeated sites and reduce positive definiteness
    to the rank of a finite coefficient, functional, moment, or Dirac-atom
    matrix of the fixed design.
    \item \textbf{Finite-width GD for broad linear observations.} For an
    ordinary two-layer PINN, a smooth nonpolynomial activation with bounded
    symmetric initialization, a certificate plus explicit width and step-size
    conditions preserve at least half of the limiting spectral gap and give
    geometric empirical-loss decay under full-batch GD at every iteration,
    for differential rows of any fixed finite order and for admissible weak
    and nonlocal rows.
    \item \textbf{Experiments consistent with the theorems.} In six matched
    two-row constraint families and a 770-run study over eleven strong-form,
    weak-form, and nonlocal PDE losses, every certificate is fixed before
    training, measured ranks match the predicted nullspaces, at every
    tabulated width above 16 every run keeps at least half the reference gap
    at every checkpoint, and a separate per-step audit finds the loss
    nonincreasing at every scored iterate.
\end{itemize}
The appendix contains full proofs, the constructive certificate procedures
with worked examples, a finite-witness lower bound on the smallest eigenvalue,
the compatible-singular extension, and fallback criteria outside the biased
analytic setting.

\paragraph{Related work}
\citet{Wang2022} derive the limiting PINN tangent kernel and identify loss
imbalance. The closest finite-width, full-batch GD analysis of strong-form PINNs
is that of~\citet{Gao2023}, for biased scalar cubic-ReLU networks with
pointwise second-order constraints; our argument
depends only on the induced linear observations, not their differential
order. Operator-oriented analyses establish transformed-kernel limits while
assuming a positive limiting Gram~\citep{Gan2025} or study admissible operators
and random-feature training~\citep{Zhao2025}; neither gives the joint exact
nullspace used here. For positivity of the differential NTK (DNTK),
\citet{Jin2026} prove the cross-location feature independence used in our
pointwise argument, under distinct physical points and scalar operators acting
componentwise; we identify the full nullspace for repeated
sites, coupled outputs, joint residual--boundary systems, and dependent rows.
\citet{Jin2026b} bound both extreme eigenvalues for a bias-free two-layer
rectified-power-unit (RePU) Poisson model with separated sites in dimension at
least three; our
finite-witness estimate applies more generally. Related optimization results treat projected
GD~\citep{Niessen2026}, stochastic wide-network training~\citep{Jin2027}, or
limits without a uniform spectral gap~\citep{Jiang2026}. For weak-form PINNs,
\citet{Kouma2026} give a conditional, frequency-localized DNTK analysis;
variational and nonlocal PINNs motivate our functional
rows~\citep{Kharazmi2021,Pang2020}, whose kernel treatment draws on
collocation, operator observations, and measure
embeddings~\citep{Schaback2006,Pfortner2022,Sriperumbudur2011}.
\Cref{sec:prior_work_comparison} compares assumptions and scope in detail.

\section{Setting}
Let \(\Omega\subset\mathbb R^{d_0}\) be the physical domain and write
\([M]=\{1,\ldots,M\}\) for any positive integer \(M\). Let \(N\) denote the
number of scalar constraint rows.

\begin{definition}[Two-layer network]
\label{def:ntk_parameterization}
For input dimension $d_0$, output dimension $d_2$, and width $d_1$, let
\begin{equation}
 \begin{aligned}
 \mathbf u(\mathbf x;\theta)
 &=\frac1{\sqrt{d_1}}\sum_{m=1}^{d_1}
 \sigma(\mathbf w_m^\top\mathbf x+\gamma b_m)\mathbf a_m,\\
 \boldsymbol\xi_m&=(\mathbf w_m,b_m),\qquad
 \theta_m=(\boldsymbol\xi_m,\mathbf a_m),\\
 \theta&=(\theta_1,\ldots,\theta_{d_1})\in\mathbb R^{d_1d_\theta},
 \end{aligned}
 \label{eqn:network_main}
\end{equation}
where $\sigma:\mathbb R\to\mathbb R$ is the scalar activation,
$\mathbf w_m\in\mathbb R^{d_0}$, $b_m\in\mathbb R$,
$\mathbf a_m\in\mathbb R^{d_2}$, $d_\theta=d_0+d_2+1$, and $\gamma\geq0$.
With the ridge functions
$\varphi_{\boldsymbol\xi,q}=\mathbf e_q\sigma(\mathbf w^\top\!\cdot+\gamma b)$
of a neuron $\boldsymbol\xi=(\mathbf w,b)$ in output $q\in[d_2]$,
$\mathbf u(\cdot;\theta)=d_1^{-1/2}\sum_{m=1}^{d_1}\sum_{q=1}^{d_2}
a_{m,q}\varphi_{\boldsymbol\xi_m,q}$.
\end{definition}

\begin{definition}[Sampled linear constraints]
\label{def:pde}
\label{def:linear_pde}
\label{def:scalar_collocation_constraints}
After decomposing every vector-valued PDE, initial condition, and boundary
condition into scalar rows, each constraint $i\in[N]$ is a linear observation
of $\mathbf u(\cdot;\theta)$ with a loss scale $s_i>0$, of one of three types.
\begin{enumerate}[label=(\roman*),leftmargin=*,nosep]
\item \emph{Strong-form pointwise.} For $\mathbf x_i\in\overline\Omega$,
\begin{equation}
 \begin{aligned}
 P_i(\theta)&=s_i(\mathcal A_i\mathbf u)(\mathbf x_i;\theta),\qquad
 (\mathcal A_i\mathbf u)(\mathbf x)
 =\sum_{q=1}^{d_2}(\mathcal A_{iq}u_q)(\mathbf x),\\
 (\mathcal A_{iq}v)(\mathbf x)
 &=\sum_{\alpha\in I_{iq}}
 a_{iq\alpha}(\mathbf x)\,\partial^\alpha v(\mathbf x),
 \end{aligned}
 \label{eqn:scalar_constraint_main}
\end{equation}
where every \(I_{iq}\subset\mathbb N_0^{d_0}\) is finite and the
coefficients are finite at $\mathbf x_i$; Dirichlet, Neumann, Robin, mixed,
and variable-coefficient conditions all have this form.
\item \emph{Weak-form functional.}
$P_i(\theta)=\lambda_i(\mathbf u(\cdot;\theta))$ with
$\lambda_i(f)=s_i a_i(f,v_i)$, where $\mathcal X$ is a trial space
containing the candidate solutions, $\mathcal V_i$ is normed,
$a_i:\mathcal X\times\mathcal V_i\to\mathbb R$ is a continuous bilinear map,
and $v_i\in\mathcal V_i$ is a fixed test function.
\item \emph{Finite-measure nonlocal.}
$P_i(\theta)=\lambda_i(\mathbf u(\cdot;\theta))$ with
$\lambda_i(f)=s_i\sum_{q=1}^{d_2}\int_D f_q\,d\nu_{iq}$ for a compact
$D\subseteq\overline\Omega$ and finite signed regular Borel measures
$\nu_{iq}$ on $D$.
\end{enumerate}
The scale $s_i$ carries the loss normalization, for example
$s_i=N_r^{-1/2}$ for a group of $N_r$ residual rows, and all three types
enter the same loss and tangent Gram matrix.
\end{definition}

\begin{definition}[Loss and tangent Gram matrix]
\label{def:pinn_loss}
\label{def:ntk}
Let $\mathbf v_P(\theta)=(P_i(\theta))_{i=1}^N$. If $y_i$ denotes the unscaled
target for row $i$, set $[\mathbf v_T]_i=s_i y_i$, so predictions and targets
use the same row scaling. Define
\begin{equation}
\begin{aligned}
 \mathbf r(\theta)&=\mathbf v_T-\mathbf v_P(\theta),
 &\mathcal L(\theta)&=\|\mathbf r(\theta)\|_2^2,\\
 [\mathbf J(\theta)]_{i,:}&=\nabla_\theta P_i(\theta)^\top,
 &i&\in[N],\\
 \mathbf K(\theta)&=\mathbf J(\theta)\mathbf J(\theta)^\top.
\end{aligned}
 \label{eqn:loss_kernel_main}
\end{equation}
Here $\mathbf J(\theta)\in\mathbb R^{N\times d_1d_\theta}$. We write
$\mathbf K_{(k)}=\mathbf K(\theta_{(k)})$.
\end{definition}

\begin{definition}[Gradient descent]
\label{def:gradient_descent}
For a step size $\eta>0$,
\begin{equation*}
 \theta_{(k+1)}=\theta_{(k)}-\eta\nabla_\theta\mathcal L(\theta_{(k)}).
\end{equation*}
We abbreviate $\mathbf r_{(k)}=\mathbf r(\theta_{(k)})$.
\end{definition}

\begin{assumption}[Initialization]
\label{assump:main}
The vectors $\theta_{m,(0)}$ are i.i.d.\ from a width-independent,
compactly supported law $\mu$. The constraint design and target are fixed
before initialization or sampled independently of it; in the latter case all
probability statements are conditional on their realization. Let $\mu_h$
denote the $\boldsymbol\xi$-marginal of $\mu$.
\end{assumption}

\begin{assumption}[Classwise observation and feature conditions]
\label{assump:pde_classes}
The fixed finite family of linear rows satisfies the clauses for one of the
following classes. A subscript ``reg'' marks sufficient regularity for
population-NTK existence and GD; ``pd'' marks the additional feature richness
used for the exact positive-definiteness criterion.
\begin{enumerate}[leftmargin=*,nosep]
\item[\textup{(\(\mathsf P_{\mathrm{reg}}\))}]
The rows are of type (i) in~\Cref{def:linear_pde}, with largest order
$r_{\max}<\infty$, and $\sigma\in C_{\mathrm{loc}}^{r_{\max}+1,1}(\mathbb R)$.
\item[\textup{(\(\mathsf P_{\mathrm{pd}}\))}]
In addition, $\sigma$ is real analytic and nonpolynomial, $\gamma>0$, and
$\operatorname{int}(\supp\mu_h)\neq\varnothing$.
\item[\textup{(\(\mathsf W_{\mathrm{reg}}\))}]
The rows are of type (ii) in~\Cref{def:linear_pde}, the trial space
$\mathcal X$ is a real Banach space of $\mathbb R^{d_2}$-valued functions
embedded continuously in $L^1_{\mathrm{loc}}(\Omega;\mathbb R^{d_2})$,
so that $\mathcal X$-valued derivatives and limits may be computed pointwise,
and every loss-scaled row $\lambda_i$ belongs to $\mathcal X^*$.
The one-neuron map
$\mathcal G(\boldsymbol\xi,\mathbf a)=\sum_{q=1}^{d_2}a_q\varphi_{\boldsymbol\xi,q}$
is twice continuously Fr\'echet differentiable into $\mathcal X$ on an open
neighborhood of every fixed parameter tube generated by $\supp\mu$.
\item[\textup{(\(\mathsf W_{\mathrm{pd}}\))}]
In addition, $\operatorname{int}(\supp\mu_h)\neq\varnothing$, every
$\boldsymbol\xi\mapsto\varphi_{\boldsymbol\xi,q}$ is real analytic into
$\mathcal X$, and
\begin{equation*}
 \overline{\linspan\{\varphi_{\boldsymbol\xi,q}:
 \boldsymbol\xi\in\mathbb R^{d_0+1},\ q\in[d_2]\}}^{\mathcal X}
 =\mathcal X.
\end{equation*}
\item[\textup{(\(\mathsf N_{\mathrm{reg}}\))}]
The rows are of type (iii) in~\Cref{def:linear_pde}, and
$\sigma\in C_{\mathrm{loc}}^2(\mathbb R)$.
\item[\textup{(\(\mathsf N_{\mathrm{pd}}\))}]
In addition, $\sigma$ is real analytic and nonpolynomial, $\gamma>0$, and
$\operatorname{int}(\supp\mu_h)\neq\varnothing$.
\end{enumerate}
Each ``pd'' label denotes its displayed condition together with the
corresponding ``reg'' clause; the bias-free case $\gamma=0$ admitted by
\Cref{def:ntk_parameterization} is used only in the appendix, chiefly for the
counterexamples of \Cref{sec:hypothesis_sharpness}. Row independence is not
assumed here: it is the design-side certificate below. Conditional
output-weight centering is imposed only for GD.
\end{assumption}

These clauses model the deterministic empirical loss obtained after fixing a
finite test or collocation family and any quadrature, including standard
variational PINN (VPINN) and hp-VPINN formulations and forward nonlocal PINNs
with fixed operator
parameters. The
analyticity and density clauses are sufficient conditions for the exact NTK
characterization, not conditions normally checked in implementations.
\Cref{tab:pde_coverage} summarizes the three classes. The only restriction is
linearity of the rows, not their order, dimension, or form: any fixed
finite-order linear operator, scalar or coupled, in any dimension, with any
linear initial or boundary rows, is covered, as are fixed-test variational and
finite-measure nonlocal observations, with all rows certified jointly.

\begin{table}[!b]
\centering
\footnotesize
\setlength{\tabcolsep}{4pt}
\renewcommand{\arraystretch}{1.1}
\caption{The three observation classes, each a fixed finite family of linear
rows under~\Cref{assump:main}. In every class the ``reg'' clause gives
existence of $\mathbf K_{(0)}^\infty$ and
$\mathbf K_{(0)}\to\mathbf K_{(0)}^\infty$; the ``pd'' clause identifies
$\ker\mathbf K_{(0)}^\infty$ with the listed relations, so row independence
certifies $\mathbf K_{(0)}^\infty\succ0$ through the output block; and that
certificate, conditional centering,
and~\cref{eqn:linear_pde_convergence_conditions} give geometric decay of the
empirical training loss, not of the PDE solution error
(\Cref{cor:classwise_gd_convergence}).}
\label{tab:pde_coverage}
\begin{tabular}{@{}
 >{\raggedright\arraybackslash}p{.14\linewidth}
 >{\raggedright\arraybackslash}p{.40\linewidth}
 >{\raggedright\arraybackslash}p{.22\linewidth}
 >{\raggedright\arraybackslash}p{.18\linewidth}@{}}
\toprule
Class & Rows and representative PDEs
& $\ker\mathbf K_{(0)}^\infty$ under ``pd''
& Result; certificate \\
\midrule
Strong-form pointwise
& Fixed finite-order linear differential rows, scalar or coupled: Poisson and
variable-coefficient elliptic, transport, heat, wave, biharmonic, Stokes,
linear elasticity, and linear initial or boundary rows
& Sitewise relations among the loss-scaled coefficient vectors
& \Cref{prop:limiting_ntk_rank}; \Cref{sec:pointwise_positivity_certificate} \\
\addlinespace
Weak-form functional
& Fixed bounded residual and boundary functionals on one trial space:
fixed-test variational elliptic or parabolic systems, tested traces, and the
discontinuous-coefficient weak Poisson problem of
\Cref{cor:continuum_weak_interface_poisson}
& Joint relations among the functionals in $\mathcal X^*$
& \Cref{prop:functional_ntk_rank}; \Cref{sec:weak_positivity_certificate} \\
\addlinespace
Finite-measure nonlocal
& Finite signed-measure tuples on compact $D$: averages and moments,
bounded-kernel integral and nonlocal-diffusion rows, Fredholm or Volterra
rows, Dirac data, and atomic quadrature
& Joint relations among the loss-scaled measure tuples
& \Cref{prop:nonlocal_ntk_rank}; \Cref{sec:nonlocal_positivity_certificate} \\
\bottomrule
\end{tabular}
\end{table}

The abstract GD theorem needs only the applicable ``reg'' clause and any
positive-definiteness certificate; the ``pd'' clauses make that certificate
equivalent to the checkable independence criterion in the table. %

For every fixed linear row considered below, linearity and
\Cref{def:ntk_parameterization} give
$P_i(\theta)=d_1^{-1/2}\sum_{m=1}^{d_1}h_i(\theta_m)$ with
$h_i(\boldsymbol\xi,\mathbf a)=\mathbf a^\top\mathbf f_i(\boldsymbol\xi)$,
where $\mathbf f_i:\mathbb R^{d_0+1}\to\mathbb R^{d_2}$ applies row $i$ to
the ridge functions of one neuron:
$[\mathbf f_i(\boldsymbol\xi)]_q
=s_i(\mathcal A_{iq}\sigma(\mathbf w^\top\!\cdot+\gamma b))(\mathbf x_i)$
for rows of type (i) and
$[\mathbf f_i(\boldsymbol\xi)]_q=\lambda_i(\varphi_{\boldsymbol\xi,q})$ for
types (ii) and (iii).
Define the single-neuron Jacobian
$\boldsymbol\Psi(z)\in\mathbb R^{N\times d_\theta}$ by
$[\boldsymbol\Psi(z)]_{ip}=\partial h_i(z)/\partial z_p$ and the output
feature matrix
$\boldsymbol\Phi_{\mathrm{out}}(\boldsymbol\xi)\in\mathbb R^{N\times d_2}$ by
$[\boldsymbol\Phi_{\mathrm{out}}(\boldsymbol\xi)]_{i,:}
=\mathbf f_i(\boldsymbol\xi)^\top$. Whenever these derivatives are
square-integrable, define the population full and output-weight NTKs
\begin{equation*}
 \mathbf K_{(0)}^\infty
 =\mathbb E_{z\sim\mu}\!\left[\boldsymbol\Psi(z)\boldsymbol\Psi(z)^\top\right],
 \qquad
 \mathbf K_{\mathrm{out}}^\infty
 =\mathbb E_{\boldsymbol\xi\sim\mu_h}\!\left[
   \boldsymbol\Phi_{\mathrm{out}}(\boldsymbol\xi)
   \boldsymbol\Phi_{\mathrm{out}}(\boldsymbol\xi)^\top\right].
\end{equation*}
Writing $\boldsymbol\Psi=(\boldsymbol\Psi_{\mathrm{hid}},
\boldsymbol\Phi_{\mathrm{out}})$ according to the hidden affine and output
parameters gives
\begin{equation}
 \mathbf K_{(0)}^\infty
 =\mathbf K_{\mathrm{out}}^\infty+\mathbf K_{\mathrm{hid}}^\infty,
 \qquad
 \mathbf K_{\mathrm{hid}}^\infty
 =\mathbb E_{z\sim\mu}\!\left[
 \boldsymbol\Psi_{\mathrm{hid}}(z)
 \boldsymbol\Psi_{\mathrm{hid}}(z)^\top\right]\succeq\mathbf0,
 \label{eqn:full_output_hidden_decomposition}
\end{equation}
and since both summands are positive semidefinite,
\begin{equation}
 \mathbf K_{(0)}^\infty\succeq\mathbf K_{\mathrm{out}}^\infty,
 \qquad
 \mathbf K_{(0)}^\infty\succeq\mathbf K_{\mathrm{hid}}^\infty,
 \qquad
 \lambda_{\min}(\mathbf K_{(0)}^\infty)
 \geq \lambda_{\min}(\mathbf K_{\mathrm{out}}^\infty)
       +\lambda_{\min}(\mathbf K_{\mathrm{hid}}^\infty).
 \label{eqn:block_pd_certificate_bound}
\end{equation}

\begin{definition}[Positive-definiteness certificate]
\label{def:positive_definiteness_certificate}
For a fixed loss-scaled row design, a \emph{positive-definiteness certificate}
is an exact proof that $\mathbf K_{(0)}^\infty\succ0$. By
\cref{eqn:block_pd_certificate_bound}, it suffices to prove either
$\mathbf K_{\mathrm{out}}^\infty\succ0$ or
$\mathbf K_{\mathrm{hid}}^\infty\succ0$; the other block need not be checked.
\end{definition}

Under the ``pd'' clauses of~\Cref{assump:pde_classes} the proofs give
$\ker\mathbf K_{(0)}^\infty=\ker\mathbf K_{\mathrm{out}}^\infty$, so the output
block is then also necessary. Our constructive certificates
(\Cref{sec:constructive_positivity_certificates}) use it because its features
expose the row relations directly.

\section{Positive definiteness from constraint independence}
Here \(\mathbf K_{(0)}\to\mathbf K_{(0)}^\infty\) means almost-sure
convergence along nested widths and convergence in probability under fresh
resampling. Each class follows the same pattern: the ``reg'' clause gives
existence and convergence, the ``pd'' clause identifies
$\ker\mathbf K_{(0)}^\infty$ with the linear relations among the rows in
their natural representation, and a finite rank computation then checks that
no relation exists. The proofs establish the same nullspace for the output
block en route, so a trivial nullspace proves
$\mathbf K_{\mathrm{out}}^\infty\succ0$ and hence certifies
$\mathbf K_{(0)}^\infty\succ0$.
\subsection{Pointwise differential constraints}
Let $\mathcal S=\{\mathbf x_i:i\in[N]\}$ be the set of sampled
physical locations. For $\mathbf x\in\mathcal S$, let
$I_{\mathbf x}=\{i:\mathbf x_i=\mathbf x\}$. Define the common coordinate
set
$\mathcal C_{\mathbf x}=\{(q,\alpha):q\in[d_2],
 \alpha\in\bigcup_{i\in I_{\mathbf x}}I_{iq}\}$ and, extending each row's
coefficient list by zeros in the derivative slots it does not use, the
loss-scaled vectors
\begin{equation*}
 \mathbf b_i(\mathbf x)
 =\bigl(s_i a_{iq\alpha}(\mathbf x)\bigr)_{(q,\alpha)\in
 \mathcal C_{\mathbf x}},
 \qquad i\in I_{\mathbf x}.
\end{equation*}

\begin{proposition}[Pointwise NTK: existence, nullspace, and rank]
\label{prop:limiting_ntk_rank}
Under~\Cref{assump:main} and clause
$\mathsf P_{\mathrm{reg}}$ of~\Cref{assump:pde_classes},
$\mathbf K_{(0)}^\infty$ exists and
$\mathbf K_{(0)}\to\mathbf K_{(0)}^\infty$. If clause
$\mathsf P_{\mathrm{pd}}$ also holds, then
\begin{equation}
\begin{aligned}
 \ker\mathbf K_{(0)}^\infty
 &=\left\{\mathbf c\in\mathbb R^N:
 \sum_{i\in I_{\mathbf x}}c_i\mathbf b_i(\mathbf x)=\mathbf0
 \ \forall\mathbf x\in\mathcal S\right\},\\
 \rank\mathbf K_{(0)}^\infty
 &=\sum_{\mathbf x\in\mathcal S}
 \dim\linspan\{\mathbf b_i(\mathbf x):i\in I_{\mathbf x}\}.
\end{aligned}
 \label{eqn:exact_dntk_rank}
\end{equation}
Consequently, $\mathbf K_{(0)}^\infty\succ0$ if and only if every
local family $\{\mathbf b_i(\mathbf x):i\in I_{\mathbf x}\}$ is linearly
independent.
\end{proposition}

The output index $q$ in $\mathbf b_i(\mathbf x)$ already accounts for coupled
vector-valued systems: several scalar equations at one point preserve positive
definiteness exactly when their local coefficient vectors are independent.
The check is a finite computation at each site: assemble
$\mathbf A_{\mathbf x}=[\mathbf b_i(\mathbf x)^\top]_{i\in I_{\mathbf x}}$
and verify $\rank\mathbf A_{\mathbf x}=|I_{\mathbf x}|$. The conclusion is
qualitative for each fixed design; \Cref{prop:quantitative_dntk_bound} bounds
$\lambda_{\min}(\mathbf K_{(0)}^\infty)$ from below using finitely many
parameter values and positive-measure sets on which the output-weight feature
matrix has controlled variation.

\subsection{Weak functional constraints}
For rows of type (ii) in~\Cref{def:linear_pde}, with $\mathcal X$ as in
clause $\mathsf W_{\mathrm{reg}}$, define
$\mathcal R_\lambda:\mathbb R^N\to\mathcal X^*$,
$\mathcal R_\lambda\mathbf c=\sum_{i=1}^N c_i\lambda_i$.

\begin{proposition}[Weak-form NTK: existence, nullspace, and rank]
\label{prop:functional_ntk_rank}
Under~\Cref{assump:main} and clause $\mathsf W_{\mathrm{reg}}$ of
\Cref{assump:pde_classes},
$\mathbf K_{(0)}^\infty$ exists and
$\mathbf K_{(0)}\to\mathbf K_{(0)}^\infty$. If clause
$\mathsf W_{\mathrm{pd}}$ also holds, then
\begin{equation*}
 \ker\mathbf K_{(0)}^\infty
 =\ker\mathcal R_\lambda,
 \qquad
 \rank\mathbf K_{(0)}^\infty
 =\rank\mathcal R_\lambda
 =\dim\linspan\{\lambda_1,\ldots,\lambda_N\}.
\end{equation*}
Thus $\mathbf K_{(0)}^\infty\succ0$ if and only if
$\lambda_1,\ldots,\lambda_N$ are linearly independent in $\mathcal X^*$.
\end{proposition}

If $\lambda_i(f)=s_i a(f,v_i)$ for a common continuous coercive bilinear form
$a$ on $\mathcal X$, independence of $v_1,\ldots,v_N$ suffices. Separate
residual and boundary independence does not: the combined family must be
independent (\Cref{thm:weak_constraint_positivity,cor:weak_boundary_joint_positivity}).
Independence in $\mathcal X^*$ is decided by finitely many numbers. Under
$\mathsf W_{\mathrm{pd}}$ each
$F_{iq}(\boldsymbol\xi)=\lambda_i(\varphi_{\boldsymbol\xi,q})$ is real
analytic, so $\mathbf K_{(0)}^\infty\succ0$ exactly when finitely many
derivatives of the $F_{iq}$ at one fixed center have row rank $N$
(\Cref{prop:analytic_functional_rank}); and since, for real analytic
nonpolynomial $\sigma$ and $\gamma>0$, differentiating
$\sigma(\mathbf w^\top\mathbf x+\gamma b)$ in $\mathbf w$ and $b$ at the
origin yields the monomials $p_\alpha(\mathbf x)=\mathbf x^\alpha$ up to
nonzero constants, derivatives may be replaced by moments:
$\mathbf K_{(0)}^\infty\succ0$ exactly when, for some finite $K$, the matrix
$[s_i a_i(\mathbf e_q p_\alpha,v_i)]_{i\in[N],\,(q,\alpha):|\alpha|\leq K}$
has row rank $N$ (\Cref{cor:monomial_moment_rank}). For a weak loss
evaluated by quadrature, the rows are combinations
$\lambda_i=\sum_jB_{ij}\mathcal Q_j$ of point evaluations
$\mathcal Q_j(f)=\partial^{\alpha_j}f_{q_j}(\mathbf x_j)$; once the retained
$\mathcal Q_j$ are shown independent in $\mathcal X^*$, positivity is
equivalent to $\rank\mathbf B=N$.

\subsection{Nonlocal integral constraints}
For rows of type (iii) in~\Cref{def:linear_pde}, set
$\mathcal X=C(D;\mathbb R^{d_2})$,
$\bar{\boldsymbol\nu}_i=s_i(\nu_{i1},\ldots,\nu_{id_2})$, and
$\mathcal R_\nu\mathbf c=\sum_i c_i\bar{\boldsymbol\nu}_i$ for
$\mathbf c\in\mathbb R^N$, with equality between measure tuples understood
componentwise.

\begin{proposition}[Nonlocal NTK: existence, nullspace, and rank]
\label{prop:nonlocal_ntk_rank}
Under~\Cref{assump:main} and clause $\mathsf N_{\mathrm{reg}}$ of
\Cref{assump:pde_classes},
$\mathbf K_{(0)}^\infty$ exists and
$\mathbf K_{(0)}\to\mathbf K_{(0)}^\infty$. If clause
$\mathsf N_{\mathrm{pd}}$ also holds, then
\begin{equation}
 \ker\mathbf K_{(0)}^\infty
 =\ker\mathcal R_\nu,
 \qquad
 \rank\mathbf K_{(0)}^\infty
 =\rank\mathcal R_\nu
 =\dim\linspan\left\{
 \bar{\boldsymbol\nu}_1,\ldots,\bar{\boldsymbol\nu}_N\right\}.
 \label{eqn:nonlocal_rank_main}
\end{equation}
Consequently, $\mathbf K_{(0)}^\infty\succ0$ if and only if
$\bar{\boldsymbol\nu}_1,\ldots,\bar{\boldsymbol\nu}_N$ are linearly independent.
\end{proposition}
\begin{proof}[Proof of \Cref{prop:functional_ntk_rank,prop:nonlocal_ntk_rank}]
\Cref{lem:bounded_functional_row_regularity} under $\mathsf W_{\mathrm{reg}}$,
respectively \Cref{prop:finite_measure_row_regularity} under
$\mathsf N_{\mathrm{reg}}$, verifies the abstract row bounds, so
\Cref{thm:linearNTK} gives existence and convergence. The nullspace follows
from \Cref{thm:weak_constraint_positivity}, respectively from
\Cref{thm:nonlocal_constraint_positivity} applied to the combined residual and
boundary tuples.
\end{proof}

The same moment test applies here with entries
$s_i\int_D p_\alpha\,d\nu_{iq}$ in place of $s_i a_i(\mathbf e_q p_\alpha,v_i)$
(\Cref{cor:monomial_moment_rank}); for measures with finitely many atoms,
such as Dirac data or quadrature rows, independence of the tuples is the rank
of a finite coefficient matrix over the distinct atoms.
\Cref{sec:constructive_positivity_certificates} gives the exact procedures
with worked examples.

\section{Finite-width gradient descent across operator classes}
The abstract transfer uses one local row-regularity condition.

\begin{definition}[Local row regularity]
\label{def:local_row_regularity}
Fix $R>0$. The rows satisfy $(\mathsf G_R)$ if there are width-independent
constants $A_{0,i},B_i,L_i<\infty$ such that, for every $z_0\in\supp\mu$
and all $z,z'\in\overline{\mathbb B}_R(z_0)$,
\begin{equation}
 |h_i(z_0)|\leq A_{0,i},\qquad
 \|\nabla h_i(z)\|_2\leq B_i,\qquad
 \|\nabla h_i(z)-\nabla h_i(z')\|_2\leq L_i\|z-z'\|_2.
 \label{eqn:gd_row_condition}
\end{equation}
\end{definition}

This restates the row bounds in~\Cref{assump:general} and is a proof condition,
not an extra PDE assumption: the ``reg'' clauses of~\Cref{assump:pde_classes}
imply it classwise. Set $A_{0,*}^2=\sum_iA_{0,i}^2$, $B_*^2=\sum_iB_i^2$,
$L_*^2=\sum_iL_i^2$,
$R_K=\esssup_{z\sim\mu}\|\boldsymbol\Psi(z)\|_2^2\leq B_*^2$, and, for
$0<\delta<1$ and $\mathbf h(z)=(h_i(z))_{i=1}^N$,
\begin{equation*}
 H_0^2=\mathbb E_{z\sim\mu}\|\mathbf h(z)\|_2^2\leq A_{0,*}^2,
 \qquad
 \overline r_\delta=\|\mathbf v_T\|_2+H_0
 +A_{0,*}\sqrt{2\log\tfrac{2}{\delta}}.
\end{equation*}
Under the conditional-centering hypothesis below,
\Cref{lem:initial_residual_bound} shows that $\overline r_\delta$ bounds the
initial residual with probability at least $1-\delta/2$, independently of
width.

\begin{theorem}[Finite-width GD for fixed linear observations]
\label{thm:linearPDEsconvergence}
Under~\Cref{assump:main}, suppose $(\mathsf G_R)$ of
\Cref{def:local_row_regularity} holds and
$\mathbb E[\mathbf a\mid\boldsymbol\xi]=\mathbf0$. Let a
positive-definiteness certificate give
\begin{equation*}
 0<\lambda_0\leq\lambda_{\min}(\mathbf K_{(0)}^\infty),
 \qquad \Lambda:=B_*^2,
 \qquad c_{\mathrm{Ch}}:=\frac14+\frac34\log\frac34.
\end{equation*}
A lower bound on either NTK block in \cref{eqn:block_pd_certificate_bound}
may serve as $\lambda_0$. For $0<\delta<1$, suppose
\begin{equation}
 d_1\geq\max\!\left\{
 \frac{R_K}{c_{\mathrm{Ch}}\lambda_0}\log\frac{2N}{\delta},
 (320+128\sqrt6)\frac{L_*^2\Lambda\overline r_\delta^2}{\lambda_0^3},
 \frac{16B_*^2\overline r_\delta^2}{R^2\lambda_0^2}
 \right\},
 \qquad 0<\eta\Lambda\leq\frac12.
 \label{eqn:linear_pde_convergence_conditions}
\end{equation}
Then, with probability at least $1-\delta$, simultaneously for every
$k\in\mathbb N_0$,
\begin{equation}
\begin{aligned}
 \max_{m\in[d_1]}
 \|\theta_{m,(k)}-\theta_{m,(0)}\|_2&\leq R,\\
 \mathbf K_{(k)}&\succeq\frac{\lambda_0}{2}\mathbf I_N,\\
 \mathcal L(\theta_{(k)})
 &\leq(1-\eta\lambda_0)^k\mathcal L(\theta_{(0)}).
\end{aligned}
 \label{eqn:linear_convergence_main}
\end{equation}
\end{theorem}

With a fixed number of loss groups, $N_g^{-1/2}$ row scaling for a group of
$N_g$ rows, and uniform unscaled row and target bounds, all constants above
except $\lambda_0$ are
group-size independent.
\Cref{ex:fully_numerical_gd_certificate} evaluates every constant for an exact
continuum mean observation and obtains $d_1\geq1{,}356$ at confidence $0.95$.

\begin{corollary}[Classwise GD transfer]
\label{cor:classwise_gd_convergence}
\label{cor:pointwise_full_rank_gd_convergence}
\label{cor:functional_gd_convergence}
\label{cor:nonlocal_gd_convergence}
Under~\Cref{assump:main}, suppose
$\mathbb E[\mathbf a\mid\boldsymbol\xi]=\mathbf0$ and one of the following
certificate conditions from~\Cref{assump:pde_classes} holds:
\begin{enumerate}[label=(\roman*),nosep,leftmargin=*]
\item $\mathsf P_{\mathrm{pd}}$ holds and every co-located coefficient family
is linearly independent;
\item $\mathsf W_{\mathrm{pd}}$ holds and the combined functional family is
linearly independent in $\mathcal X^*$; or
\item $\mathsf N_{\mathrm{pd}}$ holds and the combined measure tuples are
linearly independent.
\end{enumerate}
Then $(\mathsf G_R)$ holds and
$\lambda_0:=\lambda_{\min}(\mathbf K_{(0)}^\infty)>0$. Consequently, for
every $0<\delta<1$,~\cref{eqn:linear_pde_convergence_conditions} implies
\cref{eqn:linear_convergence_main} with probability at least $1-\delta$.
\end{corollary}
\begin{proof}
In each branch the cited rank condition proves
$\mathbf K_{\mathrm{out}}^\infty\succ0$, hence
$\mathbf K_{(0)}^\infty\succ0$ by \cref{eqn:block_pd_certificate_bound};
$(\mathsf G_R)$ follows from \Cref{prop:linearNTK},
\Cref{lem:bounded_functional_row_regularity}, or
\Cref{prop:finite_measure_row_regularity}, and
\Cref{thm:linearPDEsconvergence} applies.
\end{proof}

\paragraph{Scope of the transfer}
\Cref{lem:bounded_functional_row_regularity} transfers $C^2$ (or $C^{1,1}$)
regularity of the complete one-neuron map into the trial space to
$(\mathsf G_R)$ for arbitrary bounded rows, with explicit constants;
\Cref{cor:weak_Hk_row_regularity,prop:finite_measure_row_regularity}
instantiate it for weak rows in $H^k$ on a bounded Lipschitz domain in any
dimension under $\sigma\in C^{k+2}_{\mathrm{loc}}$, with pointwise Dirichlet
rows admissible once $k>d_0/2$ and pointwise rows of derivative order $r$ once
$k>d_0/2+r$, and for finite signed measures on compact $D$
under $\sigma\in C^2_{\mathrm{loc}}$; \Cref{lem:ridge_density_sobolev}
verifies the density clause of $\mathsf W_{\mathrm{pd}}$ on the same spaces.
Joint independence remains a separate check;
\Cref{cor:continuum_weak_interface_poisson,cor:weak_poisson_2d} verify
everything for weak Poisson problems with discontinuous coefficients in one
and two dimensions and, under conditional centering, give end-to-end
weak-form GD instances.
For singular designs whose targets satisfy the exact row relations,
\Cref{sec:compatible_singular_kernels} proves exact active-subspace
compression and the corresponding GD guarantee.

\paragraph{Proof roadmap}
Matrix Chernoff gives
$\mathbf K_{(0)}\succeq3\lambda_0\mathbf I_N/4$ with high probability. Inside
the radius-$R$ tube, singular-value perturbation controls the gap through
$(L_*/\sqrt{d_1})\|\theta-\theta_{(0)}\|_2$, the loss induction gives
$\max_m\|\theta_{m,(k)}-\theta_{m,(0)}\|_2
\leq4B_*\|\mathbf r_{(0)}\|_2/(\sqrt{d_1}\lambda_0)$ and closes localization,
and an integral Taylor remainder gives one-step contraction; the three width
terms in~\cref{eqn:linear_pde_convergence_conditions} enforce exactly these
three steps. Full proofs are in~\cref{sec:ntk_proof,sec:convergenceproof}.

\section{Numerical checks}
\label{sec:empirical_validation}
Three campaigns check the certificate route and the finite-width mechanism;
designs, tolerances, full tallies, and provenance are
in~\cref{sec:completed_empirical_appendix}. All use~\cref{eqn:network_main}
with $\sigma=\tanh$, $\gamma=1$, i.i.d.\ $\Unif[-1,1]$ parameters, both layers
trained, and $\eta=0.2/\lambda_{\max}(\mathbf K_{(0)})$. The broad study
trains the eleven fixed linear PDE designs of~\cref{tab:completed_pde_benchmarks}
at seven widths and ten seeds for $3{,}000$ full-batch steps (770 runs). The
control study takes six two-row constraint families---co-located pointwise
Poisson and vector Stokes rows, weak Poisson and nonlocal diffusion rows, and
weak or nonlocal boundary--residual pairs---and gives each five row designs:
the independent normalized pair $(\ell_1,\ell_2)$, three mixings that replace
$\ell_2$ by $(\ell_1+\varepsilon\ell_2)/\sqrt{1+\varepsilon^2}$ with
$\varepsilon\in\{1,0.1,0.01\}$, and the exact duplicate $\varepsilon=0$. At
$d_1\in\{16,256,4{,}096,16{,}384\}$ with five seeds this gives 600 kernels;
each of the 480 nonsingular kernels is trained on a manufactured and a
contrast compatible target and each of the 120 duplicates on the manufactured
target alone, giving $1{,}080$ runs. A three-row audit
uses designs with an exact row relation at $d_1\in\{32,128\}$ with three
seeds. For each of the eleven designs the applicable certificate matrix
of~\Cref{sec:constructive_positivity_certificates} is built, rank-checked, and
stored before any network is initialized; all eleven pass.

\begin{figure}[t]
  \centering
  \includegraphics[width=.62\linewidth,clip]{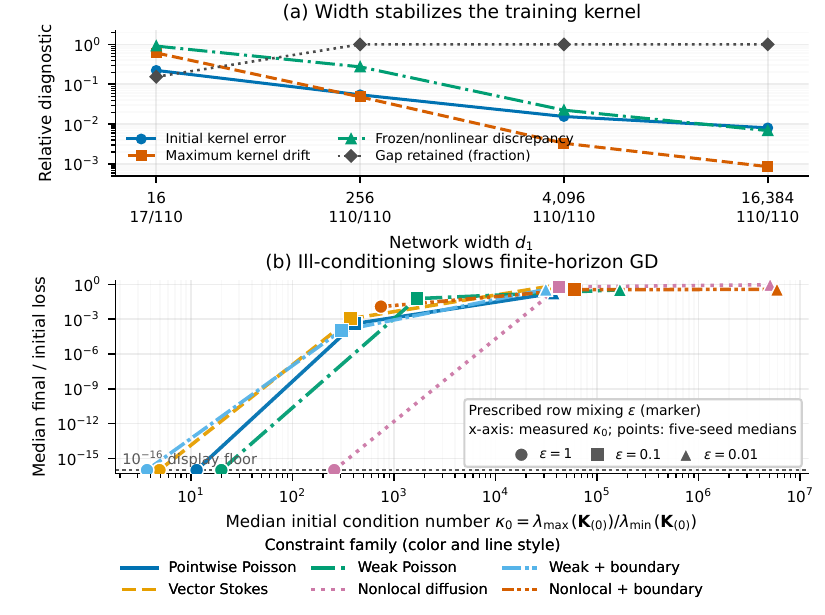}
  \caption{(a) Pooled medians over the 110 runs (eleven PDE designs, ten
  seeds) at each width of the initial relative kernel error against the
  randomized quasi-Monte Carlo (QMC) population reference
  $\widehat{\mathbf K}^{\infty}$, the maximum relative kernel drift, and the
  endpoint discrepancy from the frozen-kernel trajectory
  (\cref{tab:completed_width_results}); the gray curve is the fraction of runs
  with $\lambda_{\min}(\mathbf K_{(k)})\geq
  \lambda_{\min}(\widehat{\mathbf K}^{\infty})/2$ at every saved checkpoint.
  (b) Five-seed median final-to-initial loss ratio at $d_1=16{,}384$ for the
  contrast target of each two-row family at row mixing
  $\varepsilon\in\{1,0.1,0.01\}$ (markers), against the measured initial
  condition number
  $\kappa_0=\lambda_{\max}(\mathbf K_{(0)})/\lambda_{\min}(\mathbf K_{(0)})$;
  five medians at $\varepsilon=1$ lie below the $10^{-16}$ display floor.}
  \label{fig:theorem_supportive_evidence}
\end{figure}

\paragraph{Width stabilizes the kernel and GD decreases the loss}
Endpoint training loss decreases in all 770 runs
(\cref{tab:completed_operator_class_gd}). Pooled over the 110 runs at each
displayed width, the initial relative kernel error, the maximum kernel drift, and the
endpoint discrepancy from the frozen-kernel trajectory all decrease with width
(\cref{fig:theorem_supportive_evidence}a), and the fraction of runs holding
$\lambda_{\min}(\mathbf K_{(k)})$ at or above half the numerical
population-reference gap at every saved checkpoint rises from $17/110$ at
$d_1=16$ to $110/110$ at each larger displayed width. A per-step audit over
thirty two-row runs finds the loss
nonincreasing, up to a relative tolerance of $10^{-12}$, at all $65{,}860$
iterations that retain double-precision significance, with the largest per-step ratio matching the frozen-kernel
factor $(1-2\eta\lambda_{\min}(\mathbf K_{(0)}))^2$ to a relative discrepancy
of at most
$1.2\times10^{-5}$ (\Cref{sec:per_iterate_monotonicity}). These finite-horizon
diagnostics use a numerical reference, not a certified $\lambda_0$, and do not
verify the constants in~\cref{eqn:linear_pde_convergence_conditions}.

\paragraph{Near-dependence shrinks the gap without breaking rank}
At the registered $10^{-12}$ relative tolerance all 480 independent two-row
Gram matrices are classified nonsingular and all 120 duplicates singular
(\cref{tab:completed_rank_results}), an implementation check. For every
$\varepsilon>0$ the mixed design keeps full rank while
$\lambda_{\min}(\mathbf K_{(0)})/\lambda_{\max}(\mathbf K_{(0)})$ shrinks, and
at $d_1=16{,}384$ the contrast target's final-to-initial loss ratio increases
across $\varepsilon\in\{1,0.1,0.01\}$ in all six families
(\cref{fig:theorem_supportive_evidence}b,
\cref{tab:completed_family_conditioning}). The certified gap fixes only the
guaranteed factor $1-\eta\lambda_0$, not the observed finite-horizon loss.

\paragraph{Singular designs with compatible targets}
In the three-row audit, which exercises
\Cref{sec:compatible_singular_kernels}, a compatible target and its
null-displaced pair share one parameter path, so their losses
differ by exactly $0.0625$; that offset holds at all 19 saved checkpoints of
18 paired settings to within $4.44\times10^{-16}$. As
\Cref{cor:active_subspace_convergence} predicts,
the incompatible target's loss converges to the floor $0.0625$ rather than to
zero, and the shared
active loss falls to roundoff in median
(\cref{fig:joint_nullspace_loss_decomposition}).

\section{Scope and conclusion}
For each fixed finite design, structural nonredundancy determines whether the
limiting gap is positive, and this premise is checkable through a finite
coefficient, functional, or moment/Dirac matrix rather than assumed. The
network and initialization hypotheses are easy to satisfy and easy to verify:
each is a property of the chosen activation, architecture, and initialization
law, read off from those choices without any computation, not a condition on
the data or on the trajectory. An ordinary two-layer PINN meets all of them, with a smooth
nonpolynomial activation such as $\tanh$, sigmoid, softplus, or GELU, a bias,
all parameters trained, and i.i.d.\ bounded symmetric initialization such as
the uniform law of \Cref{sec:empirical_validation}; excluded are ReLU-type
activations, unbounded initialization laws, and output-weight laws that vary
with the width instead of the explicit
$d_1^{-1/2}$ scaling of~\cref{eqn:network_main}. Necessary conditions are
in \Cref{sec:necessary_conditions}.
\Cref{sec:hypothesis_sharpness} shows
that the classwise clauses of~\Cref{assump:main,assump:pde_classes} are not
decorative: six constructions, each deleting or weakening one clause, admit
independent rows whose limiting NTK is singular or fails to exist.

\paragraph{From training loss to solution error}
PINN error analyses built on stability of the underlying PDE bound the error
between the trained network and the exact solution by a stability constant
times the training residual plus a quadrature
term~\citep{Mishra2023,DeRyck2024}. Composed with~\cref{thm:linearPDEsconvergence},
the geometric decay in~\cref{eqn:linear_convergence_main} drives the training
term below any tolerance; the quadrature term remains.
The composition presumes a stability estimate that we do not establish.

\paragraph{Full-batch gradient descent}
Full-batch updates reflect the problem, not a limitation of the analysis. A
PINN loss over a fixed design (\Cref{def:pde,assump:main})
is a deterministic finite sum without observation noise; on a certified design
with a compatible target it is an interpolation problem with minimum value
zero, and \Cref{thm:linearPDEsconvergence} shows that plain GD reaches any
tolerance. Stochasticity therefore has no statistical role, and the remaining
discretization error is the quadrature term of the design, which no optimizer
changes; row subsampling is a computational device.

\paragraph{Depth}
At depth $H$ (\Cref{def:depth_h_network}) one ingredient holds without any
hypothesis: the tangent Gram matrix is a sum
$\mathbf K=\sum_{l=1}^{H+1}\mathbf K^{[l]}$ of positive semidefinite layer
blocks, so $\lambda_{\min}(\mathbf K)\geq\max_l\lambda_{\min}(\mathbf
K^{[l]})$ and certifying one layer certifies the kernel
(\Cref{lem:layerwise_decomposition}). If, moreover, the rows
satisfy a depth analogue $(\mathsf G^H_\tau)$ of
\Cref{def:local_row_regularity} with a gap $\lambda_0$, the step-size
condition, two of the width conditions, and the spectral and loss conclusions
of \Cref{thm:linearPDEsconvergence} carry over termwise, with a fixed-radius
Euclidean localization in place of the per-neuron one
(\Cref{prop:depth_h_substitution}). Whether $(\mathsf G^H_\tau)$ holds
for PINN rows at depth is open; we state its two missing ingredients as
conjectures: that the analogues of $\Lambda$ and $L_*^2$ are bounded by
$C_0^{(1+r_{\max})H}$, the factor $1+r_{\max}$ arising because a row of order
$r_{\max}$ threads its derivatives through every layer, which order-zero
observations~\citep{Du2019b,AllenZhu2019,Zou2020} avoid
(\Cref{conj:depth_jacobian_constants}), and that one layer block has the
sitewise nullspace of \Cref{prop:limiting_ntk_rank} at every depth
(\Cref{conj:depth_positivity_criterion}). What the two-layer proof uses and
depth lacks is the per-neuron sum $P_i(\theta)=d_1^{-1/2}\sum_mh_i(\theta_m)$
of independent terms.

\paragraph{Nonlinear constraints}
A nonlinear row, such as the convective term $u\,\partial_xu$ at a point, is
a fixed nonlinear function of linear observations of the network.
\Cref{sec:nonlinear_and_depth} shows that, under conditional centering, the tangent Gram matrix at
initialization converges in distribution to a random matrix that is
deterministic for affine rows; under a certificate for the base observations
and centered isotropic output weights, the diagonal entry of a genuinely
nonlinear polynomial row has a nondegenerate random limit, so no
deterministic gap exists to transfer. A larger width exponent removes this
randomness only by sending the initial predictions to zero, after which the
rescaled kernel is that of the rows linearized at zero
(\Cref{sec:nonlinear_scaling}), the regime of~\citet{Sirignano2026}. Whether
GD converges at the standard scaling is open.

\subsection*{Reproducibility statement}
All assumptions are stated in \Cref{assump:main,assump:pde_classes} and
\Cref{def:local_row_regularity}, and the appendix gives complete proofs of
every claim. The experimental designs, tolerances, provenance, and full
tallies are in \Cref{sec:completed_empirical_appendix}. The code,
configurations, tracked hash-validated aggregates, stored certificates, and
locked environment are at
\coderepo; its README gives the
exact commands that run the tests and the smoke audit, reproduce
\cref{fig:theorem_supportive_evidence} and the three-row audit, and rerun the
broad campaign, whose original per-run histories are not included.

%% file: paper_appendix.tex
\appendix
\addtocontents{toc}{\protect\appendixcontentsmark}
\appendixcontents
\input{notation}

\input{related_scopes}

\input{ntk}

\input{appendix_extended_results}

\input{positivity_checks}

\input{vectorvalued}
\input{pd}

\input{optimization_aux}

\input{convergence}

\input{convergence2}
\input{tanh_compatible_experiment}

\input{nonlinear_depth}

\input{theorem_audit_results}

%% file: notation.tex
\section{Notation}
\label{sec:notation}
This appendix consolidates the notation used in the main text and proofs.
For a positive integer $M$, $[M]=\{1,\ldots,M\}$. Input, weight, output,
prediction, and residual vectors, as well as Gram matrices, Jacobians, and
auxiliary coefficient or evaluation matrices, use boldface; the full and
per-neuron parameter vectors $\theta$ and $\theta_m$ are exceptions.
Superscript $\top$ denotes transpose.

\paragraph{Network and parameters}
The input dimension is $d_0$, the width is $d_1$, and the output dimension
is $d_2$.
For neuron $m\in[d_1]$, the hidden weight is
$\mathbf w_m\in\mathbb R^{d_0}$, the hidden bias is $b_m\in\mathbb R$, the
output weight is $\mathbf a_m\in\mathbb R^{d_2}$, and
\[
 \boldsymbol\xi_m=(\mathbf w_m,b_m),\qquad
 \theta_m=(\boldsymbol\xi_m,\mathbf a_m),\qquad
 \theta=(\theta_1,\ldots,\theta_{d_1}).
\]
Thus each unit has $d_\theta=d_0+d_2+1$ trainable coordinates. We parenthesize
iteration indices: $\theta_{(k)}$ is the complete iterate after $k$
gradient-descent steps and $\theta_{m,(k)}$ is its $m$th block.
When declared in $\mathbb R^{d_\theta}$, $z,z'$, and $z_0$ are generic
per-neuron parameter vectors and remain unbold; in activation formulas, $z$
is instead a scalar dummy argument. In one-dimensional single-output examples,
we suppress boldface for the weights $w_m$ and $a_m$.
We denote the initialization law of $\theta_m$ by $\mu$ and its
hidden-parameter marginal by $\mu_h$. The network is
\[
 \mathbf u(\mathbf x;\theta)
 =d_1^{-1/2}\sum_{m=1}^{d_1}
 \sigma(\mathbf w_m^\top\mathbf x+\gamma b_m)\mathbf a_m.
\]
Here $\sigma:\mathbb R\to\mathbb R$ is the scalar activation and
$\gamma\geq0$ is the fixed bias scale
of~\Cref{def:ntk_parameterization}.

\paragraph{Constraints and multi-indices}
The formulation has $N$ scalar constraint rows. Their prediction and target vectors are
$\mathbf v_P(\theta),\mathbf v_T\in\mathbb R^N$, and
$\mathbf r(\theta)=\mathbf v_T-\mathbf v_P(\theta)$ is the residual vector;
$\mathbf r_{(k)}=\mathbf r(\theta_{(k)})$. For a multi-index
$\alpha\in\mathbb N_0^{d_0}$,
$|\alpha|=\sum_j\alpha_j$,
$\alpha!=\prod_j\alpha_j!$,
$\mathbf x^\alpha=\prod_jx_j^{\alpha_j}$, and
$\partial^\alpha=\prod_j\partial_{x_j}^{\alpha_j}$.
For $\boldsymbol\rho\in\mathbb N_0^{d_0+1}$, the analogous parameter-space
quantities are $|\boldsymbol\rho|$, $\boldsymbol\rho!$, and
$\partial_{\boldsymbol\xi}^{\boldsymbol\rho}$.
Likewise,
$\partial_{\mathbf w}^{\alpha}=\prod_j\partial_{w_j}^{\alpha_j}$.
The vector $\mathbf e_q\in\mathbb R^{d_2}$ is the $q$th standard basis
vector, and $p_\alpha(\mathbf x)=\mathbf x^\alpha$.
For a finite index set $I$, $\mathbb R^I$ denotes the associated real
coordinate space. For finite-measure rows, $\nu_{iq}$ and $\omega_{jq}$ denote
the finite signed regular Borel measures of residual row $i$ and boundary row
$j$ in output $q$, with loss scales $s_i,t_j>0$, and
$\bar{\boldsymbol\nu}_i=s_i(\nu_{i1},\ldots,\nu_{id_2})$ is the loss-scaled
measure tuple.

\paragraph{Jacobians, kernels, and matrix order}
The prediction Jacobian
$\mathbf J(\theta)\in\mathbb R^{N\times d_1d_\theta}$ has row
$[\mathbf J(\theta)]_{i,:}=\nabla_\theta P_i(\theta)^\top$. Its tangent Gram
matrix is
$\mathbf K(\theta)=\mathbf J(\theta)\mathbf J(\theta)^\top$.
We abbreviate $\mathbf J_{(k)}=\mathbf J(\theta_{(k)})$ and
$\mathbf K_{(k)}=\mathbf K(\theta_{(k)})$;
$\mathbf K_{(0)}^\infty$ is the deterministic infinite-width limit at
initialization, and $\mathbf K_{\mathrm{out}}^\infty$ is its output-weight
contribution. For symmetric matrices, $\mathbf A\succeq\mathbf B$ (equivalently,
$\mathbf B\preceq\mathbf A$) means that $\mathbf A-\mathbf B$ is positive
semidefinite, and $\mathbf A\succ0$ means that $\mathbf A$ is positive
definite. We use $\ker$, $\rank$,
$\linspan$, $\Range$, and
$\lambda_{\min},\lambda_{\max}$ for nullspace, rank, linear span, column space,
and the extreme eigenvalues. When $\mathbf K$ is positive definite,
$\kappa(\mathbf K)=\lambda_{\max}(\mathbf K)/\lambda_{\min}(\mathbf K)$.
We use $\mathbf I_N$ for the identity of size $N$. The smallest singular
value is $\sigma_{\min}$, and
$\spec(\mathbf A)$ denotes the eigenvalue multiset of a square
matrix $\mathbf A$. For a subspace $V\subseteq\mathbb R^N$,
$\Pi_V$ is the orthogonal projection onto $V$ and $V^\perp$ is its orthogonal
complement.

\paragraph{Per-neuron maps and kernel blocks}
Every loss-scaled linear row $i$, with scale $s_i>0$
(\Cref{def:linear_pde}), has the per-neuron representation
$P_i(\theta)=d_1^{-1/2}\sum_mh_i(\theta_m)$ with
$h_i(\boldsymbol\xi,\mathbf a)=\mathbf a^\top\mathbf f_i(\boldsymbol\xi)$,
where $\mathbf f_i(\boldsymbol\xi)\in\mathbb R^{d_2}$ applies row $i$ to the
ridge functions $\varphi_{\boldsymbol\xi,q}$ of one neuron
(\Cref{def:ntk_parameterization} and the paragraph following
\Cref{tab:pde_coverage}). The single-neuron Jacobian
$\boldsymbol\Psi(z)\in\mathbb R^{N\times d_\theta}$ has entries
$\partial h_i(z)/\partial z_p$ and splits as
$(\boldsymbol\Psi_{\mathrm{hid}},\boldsymbol\Phi_{\mathrm{out}})$ into its
hidden-parameter and output-weight columns, with
$[\boldsymbol\Phi_{\mathrm{out}}(\boldsymbol\xi)]_{i,:}=\mathbf f_i(\boldsymbol\xi)^\top$
(the appendix also writes $\mathbf F=\boldsymbol\Phi_{\mathrm{out}}$);
accordingly
$\mathbf K_{(0)}^\infty=\mathbf K_{\mathrm{out}}^\infty+\mathbf K_{\mathrm{hid}}^\infty$
(\cref{eqn:full_output_hidden_decomposition}). The condition
$(\mathsf G_R)$ of \Cref{def:local_row_regularity} bounds $h_i$, its
gradient, and the gradient's Lipschitz modulus on closed balls of radius $R$
around $\supp\mu$; $\mathcal Z_R=\supp\mu+\overline{\mathbb B}_R(\mathbf0)$
is the corresponding compact per-neuron tube and
$\mathcal T_R(\theta_{(0)})$ the product tube of all neurons around the
initialization. For Banach-space-valued maps, $\|\cdot\|_{\mathrm{op}}$
denotes the operator norm of a Fr\'echet derivative with respect to the
Euclidean norm on the parameter space.

\paragraph{Norms and related operations}
The notation $\|\mathbf v\|_2$ is the Euclidean norm of a vector, whereas
$\|\mathbf A\|_2$ and $\|\mathbf A\|_F$ are the spectral and Frobenius matrix norms.
For functions, we write $\|f\|_{\mathcal X}$;
$\|\nu\|_{\mathrm{TV}}$ denotes the total variation of a signed
measure. Inner products are $\langle\cdot,\cdot\rangle$, with a subscript
identifying the space when needed. Finally, $(t)_+=\max\{t,0\}$, and
$\overline{\mathbb B}_R(z_0)$ is the closed Euclidean ball of radius $R$
centered at $z_0$.

%% file: related_scopes.tex
\section{Comparison with related theorem scopes}
\label{sec:prior_work_comparison}

We compare results whose principal theorems concern an operator-transformed
NTK limit, definiteness or spectrum, or optimization of a linear PINN in a
wide-network or kernel regime. This includes the closest ordinary-NTK
positivity and weak-form conditioning baselines.
\Cref{tab:prior_work_comparison} records each result's stated setting and
conclusion, followed by a constraint family treated here but outside that
theorem's hypotheses. The comparison concerns theorem scope rather than
paper-level generality; an outside-scope example is not a counterexample to
the cited result. Complementary work on weak and nonlocal PINN methods, solution consistency,
generic kernel background, empirical failure modes, and nonlinear-PDE
optimization is summarized after the table.
For pairwise-distinct sites, \Cref{prop:limiting_ntk_rank} under
$\mathsf P_{\mathrm{pd}}$ reduces to the familiar condition that every
constraint have a nonzero evaluated coefficient vector
$\mathbf b_i(\mathbf x_i)$. The condition is necessary under
$\mathsf P_{\mathrm{reg}}$ alone, since $\mathbf b_i(\mathbf x_i)=\mathbf0$
makes $P_i(\theta)=0$ for every $\theta$ and so gives
$\mathbf K_{(0)}^\infty$ a zero row; $\mathsf P_{\mathrm{pd}}$ is what makes
it sufficient. The proof of \Cref{prop:limiting_ntk_rank} is in
\Cref{sec:pd_proof}.

{\scriptsize
\setlength{\tabcolsep}{2.2pt}
\renewcommand{\arraystretch}{1.10}
\begin{longtable}{@{}
>{\raggedright\arraybackslash}p{0.13\linewidth}
>{\raggedright\arraybackslash}p{0.285\linewidth}
>{\raggedright\arraybackslash}p{0.285\linewidth}
>{\raggedright\arraybackslash}p{0.225\linewidth}@{}}
\caption{Comparison of direct theorem-level results. PD and PSD mean positive
definite and positive semidefinite; GF, GD, IGD, and NGD denote gradient flow,
gradient descent, implicit GD, and natural GD; SGD denotes stochastic GD.
The final column identifies
additional constraint structures treated in this paper.}
\label{tab:prior_work_comparison}\\
\toprule
Work and result & Setting and assumptions
& Main conclusion
& Additional scope treated here \\
\midrule
\endfirsthead
\multicolumn{4}{c}{\tablename\ \thetable\ continued}\\
\toprule
Work and result & Setting and assumptions
& Main conclusion
& Additional scope treated here \\
\midrule
\endhead
\midrule
\multicolumn{4}{r}{Continued on the next page}\\
\endfoot
\bottomrule
\endlastfoot

\multicolumn{4}{@{}l}{\textit{Panel A: kernel limits, positivity, rank, and spectrum}}\\
\addlinespace[2pt]

\citet{Wang2022}, limiting PINN NTK
& The block kernel identity holds for general fully connected PINNs, but the
infinite-width theorems treat one hidden layer with standard-Gaussian
initialization, the one-dimensional Poisson model problem with boundary-value
and $u_{xx}$ residual rows, and smooth activations with bounded derivatives;
freezing further assumes uniformly bounded parameters and time-integrated
residual bounds along GF.
& For that model problem, proves the deterministic initialization limit of
the joint block DNTK and its freezing along GF as the width grows; residual
decay is then read off along eigendirections of the PSD limiting kernel, and
the kernel-regression limit assumes an invertible limiting matrix. Exact rank
criteria and finite-width discrete-GD guarantees are outside its scope.
& Two equal rows give a singular joint Gram even when the ordinary input NTK
is PD. \Cref{prop:limiting_ntk_rank} identifies the null direction,
\Cref{cor:active_subspace_convergence} gives GD convergence for equal targets
and convergence to the loss floor for unequal ones, and
\Cref{lem:structural_compatibility_compression} identifies that floor;
see~\Cref{ex:duplicate_active_subspace}. \\

\citet{Carvalho2025}, ordinary-NTK PD
& Sequential infinite-width fully connected network with nondegenerate
Gaussian initialization and distinct raw inputs. With bias, the activation is
continuous, almost everywhere differentiable, and nonpolynomial, with the
integrability that keeps the Gaussian kernel expectations finite; without
bias, sites must also be nonzero and pairwise nonproportional.
& Proves strict PD for ordinary value-observation NTKs under its distinct-input
conditions. Differential, weak, and nonlocal observation rows are outside its
scope.
& Our result treats operator-transformed and co-located rows through their
coefficient rank and uses compact initialization to ensure the required
moments are finite on compact input and parameter sets. \\

\citet{Gao2023}, population PD and finite-width GD
& Biased scalar two-layer cubic-ReLU network; Gaussian hidden and Rademacher
output initialization; both layers trained. The joint heat-residual,
Dirichlet, and initial sites are pairwise distinct, and the boundary weight is
positive.
& Establishes positivity of both population Gram blocks for the specified
heat-equation design and gives GF and GD decay after a finite-width transfer.
& A fourth-order strong row is identically zero for cubic ReLU away from its
kink, whereas \Cref{prop:limiting_ntk_rank} under $\mathsf P_{\mathrm{pd}}$
covers every fixed finite order;
see~\Cref{ex:relu3_fourth_order_obstruction}. \\

\citet{Gan2025}, operator-informed NTK
& Bias-free scalar deep network, sequential widths, one fixed linear
differential operator on a compact convex set, continuous coefficients, and a
$C^{k+2}$ activation for an operator of order $k$, with polynomially bounded
derivatives. The training
theorem assumes a positive limiting Gram and uniform gradient continuity.
& Proves uniform convergence in probability of the transformed kernel and,
under those two training assumptions, uniform closeness of the GF and
limiting-kernel dynamics; it does not characterize rank for arbitrary joint
constraint rows.
& Duplicate compatible rows have zero full gap but a positive one-dimensional
compression. Joint residual--boundary families, weak rows, and nonlocal rows
also lie outside the single-operator theorem. \\

\citet{Jin2026}, qualitative DNTK PD
& Shallow or deep fully connected networks with standard-Gaussian parameters
and positive bias scale; pairwise-distinct physical sites; one nonzero
evaluated scalar differential row per site; smooth nonpolynomial or RePU
activation with order-matched regularity.
& Theorem~2 establishes infinite-width covariance and DNTK positivity under
its distinct-site assumptions. Exact nullspaces for arbitrary repeated-site
and coupled-output rows are outside the stated setting.
& Co-located value/derivative rows can be independent, and coupled-output rows
can be dependent. The sitewise matrices $\mathbf A_{\mathbf x}$ used with
\Cref{prop:limiting_ntk_rank} give both the exact PD criterion and the
nullspace; see~\Cref{ex:shared_point_rank_obstruction,ex:vector_coupling_rank_obstruction}. \\

\citet{Jin2026b}, extreme DNTK eigenvalues
& Bias-free scalar two-layer RePU, Poisson residual plus Dirichlet rows, and
the unweighted infinite DNTK. The lower bound needs $d_0\geq3$, angular
$\delta$-separation of all nonzero sites, and a RePU power restricted to a
range set by that paper's parameters; upper bounds allow $d_0\geq2$.
& Gives explicit lower and upper extreme-eigenvalue bounds by spherical
harmonics under the stated separation assumptions. Loss weights require
congruence scaling; finite-width optimization is outside its scope.
& Repeated sites make the separation parameter vanish.
\Cref{prop:limiting_ntk_rank} still distinguishes an independent
value/derivative pair from a duplicate pair, and
\Cref{cor:active_subspace_convergence} handles the latter, for compatible
targets, on its active subspace. \\

\citet{Moser2026}, transformed-kernel PSD
& A scalar PSD base kernel with the required classical mixed derivatives and
a finite matrix of pointwise linear differential operators whose coefficients
are finite at the evaluation sites.
& Uses reproducing-kernel Hilbert space (RKHS) representers to establish PSD.
Strict PD, exact rank, an NTK limit,
and optimization are outside the theorem; sharp divergence-form interfaces
are explicitly deferred to weak formulations.
& Duplicate operators give an exact structural nullspace, and a discontinuous
diffusion interface gives a bounded weak row; see
\Cref{ex:duplicate_active_subspace,ex:sharp_interface_weak}. \\

\citet{Zhou2024}, DNTK scaling and freezing
& Two-layer pointwise PINNs for continuous, possibly nonlinear differential
operators with width scaling $d_1^{-\varrho}$ as in
\Cref{def:exponent_normalization} (denoted $N^{-s}$ in that paper);
smooth bounded activation derivatives. The freezing theorem also assumes
uniform parameter, operator-derivative, and time-integrated residual bounds.
& Identifies homogeneity-dependent scaling thresholds for initialization
convergence and proves finite-time kernel freezing for $\varrho>1/4$ under the
displayed trajectory bounds. It neither proves PD/rank nor supplies an
optimizer rate.
& Two co-located linear rows can be identical or independent while satisfying
  the same width scaling. Joint independence of their coefficient rows
  determines their rank, which
kernel convergence alone does not. \\

\addlinespace[3pt]
\multicolumn{4}{@{}l}{\textit{Panel B: optimization and weak-form comparators}}\\
\addlinespace[2pt]

\citet{Jiang2026}, population training without a uniform gap
& Scalar second-order linear PDE with the Dirichlet condition enforced
exactly by a lift of the boundary data and a fixed multiplier vanishing on
$\partial\Omega$, so no boundary row is trained; population $L^2$ residual,
optionally with finitely many value data; shallow network with width scaling
$d_1^{-\varrho}$, $\varrho\in(1/2,1)$, as in~\Cref{def:exponent_normalization}
(denoted $N^{-\beta}$ in that paper), and learning rate $d_1^{2\varrho-1}$;
$C_b^4$ nonconstant activation, full-support initialization, and clipped
continuous-time GF.
& Sends width to infinity at each fixed time and then proves asymptotic
residual convergence without a uniform spectral gap; solution convergence
additionally assumes a bounded inverse. It gives neither a rate nor a
uniform-in-time finite-width discrete-GD transfer.
& A finite co-located residual--boundary pair may be singular or PD according
to its evaluated coefficients. Our formula classifies both and gives a
finite-width GD result, without claiming population solution consistency. \\

\citet{Niessen2026}, projected GD
& Scalar homogeneous-Dirichlet Poisson problem on a bounded $C^{2,\beta}$
domain; one hidden layer; fixed paired output weights; only hidden weights
trained; paired truncated-Gaussian initialization; per-neuron projection; and
globally bounded activation derivatives through order four.
& Theorem~6 gives an $O(T^{-1/2})$ time-averaged empirical-loss bound, together
with width and approximation terms, under the stated projected-training
setup. The result concerns time-averaged loss rather than last-iterate
geometric decay.
& Coupled multi-output pointwise residuals with joint boundary rows, trained in
both layers by ordinary unprojected GD. Under $\mathsf P_{\mathrm{pd}}$,
\Cref{prop:limiting_ntk_rank} gives their exact nullspace, and, with
conditional centering, \Cref{cor:classwise_gd_convergence} applies whenever
every co-located coefficient family is independent, as for the system
of~\Cref{ex:vectorvalued}; \Cref{ex:vector_coupling_rank_obstruction} shows
that this independence can fail. \\

\citet{Jin2027}, stochastic wide-PINN training
& Biased scalar two-layer Poisson PINN at fixed residual and Dirichlet sites;
both layers trained; Gaussian/Rademacher initialization; unbiased
with-replacement SGD or modeled stochastic GF. The two population parameter
blocks are each assumed PD, and the step size is small.
& Gives expected loss decay on a no-exit event at sufficient width under the
assumed positive population blocks and stated sampling model. It concerns
stochastic rather than full-batch training and does not treat compatible
singular systems.
& Equal duplicate rows make both block Grams singular but leave a positive
active compression for compatible targets. Our full-batch result applies
after exact compression; see~\Cref{ex:duplicate_active_subspace}. \\

\citet{Xu2027}, implicit GD
& Bias-free scalar two-layer PINN for the heat equation; fixed Rademacher
output weights, with only Gaussian-initialized hidden weights trained; and
pairwise nonparallel training inputs. The analytic nonpolynomial activation
has bounded third derivative and globally Lipschitz derivatives through order
three. In addition, for every $n\geq2$, it assumes that
$\sigma^{(n)}(z)/\varpi(z)$ has a nonzero limit for one common profile
$\varpi$ with $\varpi(z)\to0$ and $z\varpi(\vartheta z)/\varpi(z)\to0$ for
every constant $\vartheta>1$, all limits as $z\to+\infty$ (that paper's
Assumption~2.1). From these conditions that paper proves the limiting Gram
PD (its Lemma~3.2), and each implicit step is assumed to be solved exactly.
& Proves high-probability linear IGD loss decay for any fixed $\eta>0$ at
sufficient width; the sufficient-width condition depends on $\eta$. The
guarantee concerns a special pointwise scalar system and exact implicit
updates, not ordinary GD or a rank classification.
& Co-located independent operators violate the nonparallel-site condition, and
compatible duplicate rows have zero full gap. Our local-rank and
active-subspace results cover both cases for explicit GD. \\

\citet{Xu2026}, natural GD
& Overparameterized scalar two-layer PINNs for the heat-equation model
of~\citet{Gao2023}, with the input augmented by a bias coordinate; fixed
Rademacher output weights, with only Gaussian-initialized hidden weights
trained. For smooth activations under analyticity, derivative-regularity, and
asymptotic-profile conditions like those of~\citet{Xu2027}, the paper proves
a PD limiting Gram at pairwise-distinct sites, which its NGD analysis uses at
sufficient width.
& Establishes improved GD step-size dependence and fast NGD convergence,
including a quadratic local rate for smooth activations. It does not derive
an exact rank/nullspace criterion or cover weak, nonlocal, or compatible
singular systems.
& A duplicated compatible observation makes the natural-gradient Gram
singular but has a positive compressed kernel.
\Cref{cor:active_subspace_convergence} gives explicit-GD convergence without
inverting the full Gram. \\

\citet{Zhao2025}, admissible-operator and random-feature training
& Two-layer tanh models for PINNs and Deep Ritz. The linear result requires an
``admissible'' pointwise operator that preserves feature independence; the
nonlinear results freeze hidden features and use coercivity plus a
\L{}ojasiewicz condition for GF or IGD.
& Proves convergence for its admissible linear class and convergence to
critical points in the nonlinear random-feature setting. The linear
admissibility hypothesis supplies the needed nondegeneracy rather than an
exact rank formula for arbitrary joint rows.
& Repeated-site mixed operators and deliberately dependent rows need not meet
admissibility. The sitewise matrices $\mathbf A_{\mathbf x}$ decide exactly
when they are independent, and \Cref{lem:structural_compatibility_compression}
gives the loss floor, which vanishes exactly for compatible dependencies. \\

\citet{Kouma2026}, variational-NTK conditioning
& Coupled parabolic VPINN with boundary/initial data imposed by a lift; compact
space--time tests; deep vector network. The analysis assumes a frozen,
translation-invariant DNTK with Hermitian-PD Fourier transform, frequency
localization, an invertible symbol, and large convection.
& Theorem~4.1 gives a conditional Schur-complement asymptotic for one frequency
under the frozen, translation-invariant DNTK hypotheses. It does not give an
exact full-matrix rank characterization or a finite-width optimizer theorem.
& Two identical weak tests create a cross-test null vector at every width,
which a per-frequency component-symbol calculation does not classify. Our
weak nullspace and compatible active-subspace results do. \\

\addlinespace[3pt]
\multicolumn{4}{@{}l}{\textit{Panel C: present work}}\\
\addlinespace[2pt]

This work
& Biased two-layer, possibly multi-output networks; a compactly supported
per-neuron law, with conditionally centered output weights for the GD result
and hidden marginal support of nonempty interior for the exact criteria;
analytic nonpolynomial activation with $\gamma>0$ for the exact pointwise and
nonlocal rank/nullspace criteria. Weak rows instead require Banach-valued
analyticity and dense ridge-feature span; nonlocal rows use finite
signed-measure tuples.
& Gives exact limiting-NTK rank and nullspace formulas, a finite-witness lower
bound, and high-probability finite-width explicit-GD convergence with stated
width and step size. A fixed active-subspace version covers compatible exact
dependencies.
& Treats all six constructions below: excessive differential order for a
fixed RePU is shown directly to annihilate the row, outside
$\mathsf P_{\mathrm{reg}}$; dependent or independent same-site rows, coupled
outputs, compatible duplicates, sharp-interface weak rows, and activations
whose Gaussian moments diverge but whose compact-initialization moments remain
finite are classified by the rank formulas. \\

\end{longtable}
}

The table does not include~\citet{Bonfanti2024} and~\citet{Sirignano2026},
whose principal results concern nonlinear PDEs, because for nonlinear
constraints our contribution is negative rather than a comparable positive
result: under conditional centering, second-moment isotropy of the output
weights, and a certificate for the base observations,
\Cref{sec:nonlinear_nondeterminism} shows that at the standard exponent
$\varrho=1/2$ a single genuinely nonlinear polynomial row has a nondegenerate
random limiting kernel entry, so the deterministic limiting kernel on which
every linear-row comparison in this table rests no longer exists. That
qualitative statement is not new: \citet{Bonfanti2024} already find that in
the nonlinear regime the tangent kernel is random at initialization and not
constant during training, and turn to second-order methods; we claim no
priority for it (\Cref{sec:nonlinear_and_depth}).
\Cref{sec:nonlinear_scaling} shows that a larger exponent $\varrho>1/2$
restores a deterministic limit only by sending the initial predictions to
zero, after which the rescaled kernel is that of the rows linearized at the
zero prediction and, along training, depends on the moving predictions.
\citet{Sirignano2026} work in this regime, with width scaling $N^{-\beta}$,
$\beta\in(1/2,1)$, zero initial output, and a deterministic limiting kernel
that depends on the evolving solution, so our negative results do not bear on
their analysis. The table likewise omits
solution consistency and error
control~\citep{Shin2020,Cheng2025,Doumeche2025,Bonito2026}, and empirical
failure or landscape analyses~\citep{Krishnapriyan2021,Rathore2024}. Weak and
nonlocal method papers~\citep{Kharazmi2021,Pang2020} motivate the functional
rows but do not make the theorem-level NTK claims compared above. Functional
kernel perspectives from collocation, operator observations, and measure
embeddings~\citep{Schaback2006,Pfortner2022,Sriperumbudur2011} supply broader
background rather than direct PINN-optimization comparisons.

Within the omitted consistency category, \citet{Cheng2025} give conditions
for consistency of physics-informed kernel GF and wide PINNs, finite-time NTK
convergence, and nonasymptotic loss bounds while assuming a PD base NTK. They
do not derive finite-data DNTK rank or a discrete-GD finite-width transfer.

Property~(LI) of~\citet[Definition~3]{Jin2026} asserts that the limiting
output covariance is nondegenerate on finitely many pairwise-distinct sites,
each carrying several distinct multi-index derivative labels, and they
establish it for $\gamma>0$, standard-Gaussian initialization, and smooth
nonpolynomial or order-matched RePU activations. Every pointwise row is a
fixed linear combination, with coefficient vector $\mathbf b_i(\mathbf x_i)$,
of such labelled observations at its own site, and for two layers their DNTK
dominates this covariance in the Loewner order (their Remark~2).
Property~(LI) therefore already yields, in their Gaussian setting and by a
short linear-algebra step, the sitewise nullspace and rank formulas
of~\Cref{prop:limiting_ntk_rank}. They state neither these formulas nor a
same-site criterion, and their main positivity
theorem~\citep[Theorem~2]{Jin2026} assumes pairwise-distinct physical points;
\Cref{prop:limiting_ntk_rank} states the criterion explicitly, including
coupled outputs, for a compactly supported initialization whose hidden
marginal has support with nonempty interior.

\begin{example}[Cubic ReLU degenerates for a fourth-order strong residual]
\label{ex:relu3_fourth_order_obstruction}
Let $\rho_3(z)=(z_+)^3$ and consider the one-dimensional equation
$u^{(4)}(x)=1$. For the two-layer trial function
\begin{equation*}
 u_\theta(x)=\frac1{\sqrt{d_1}}\sum_{m=1}^{d_1}
 a_m\rho_3(w_mx+b_m),
\end{equation*}
every fixed finite collocation set $\{x_i\}$ avoids the kinks
$w_mx_i+b_m=0$ of $\rho_3$ almost surely whenever each $(w_m,b_m)$ has a
Lebesgue density, as for the Gaussian hidden-layer initialization
of~\citet{Gao2023}. On that event,
each summand either vanishes locally or agrees there with a cubic polynomial, and hence
\begin{equation*}
 \partial_x^4u_\theta(x_i)=0,
 \qquad
 \nabla_\theta\partial_x^4u_\theta(x_i)=\mathbf0.
\end{equation*}
Every such residual row and column of the finite-width and expected
infinite-width tangent kernels is therefore zero, whereas the target $1$ leaves
a strictly positive minimum achievable loss. Moreover, $\rho_3$ is $C^2$ but not
$C^3$, since $\rho_3'''$ jumps from $0$ to $6$ at $z=0$, so even third-derivative
constraints require additional care. Thus the fixed
cubic-ReLU second-order result of~\citet{Gao2023} does not extend to arbitrary
higher-order strong PDEs without changing the activation or formulation.
This example does not contradict~\citet{Jin2026}: their assumptions exclude
$\rho_3$ for fourth derivatives, and their Remark~1 prescribes a matching
rectified-power unit instead.
\end{example}

\begin{example}[A shared PDE--boundary point can be singular or benign]
\label{ex:shared_point_rank_obstruction}
Consider the closed interval $[0,1]$ and collocate at the boundary point
$x_\star=0$ both the PDE operator
$\mathcal A_\star=\partial_x+x\mathcal I$, where $\mathcal I$ is the
identity operator, and the Neumann boundary operator $\mathcal B=\partial_x$.
For positive scales $s_\Omega,s_\partial$, define
$P_\Omega=s_\Omega(\mathcal A_\star u_\theta)(x_\star)$ and
$P_\partial=s_\partial(\mathcal B u_\theta)(x_\star)$. Although
$\mathcal A_\star\neq\mathcal B$ as written operators and both evaluated rows
are nonzero,
\begin{equation*}
 \mathcal A_\star u_\theta(0)=\mathcal Bu_\theta(0)=u_\theta'(0),
 \qquad
 s_\partial\nabla_\theta P_\Omega
 -s_\Omega\nabla_\theta P_\partial=\mathbf0.
\end{equation*}
The joint $2\times2$ Gram block has determinant zero at every width. This
singularity shows why pairwise-distinct-point positivity does not automatically
apply at a repeated physical location under only a nontriviality check on each constraint.
Coincidence is not itself an obstruction: replacing
$\mathcal A_\star$ by $\partial_x+\mathcal I$ gives the
independent local coefficient rows $(1,1)$ and $(0,1)$ in coordinates
$(u,u')$. Under the hypotheses of~\Cref{prop:limiting_ntk_rank}, this pair
yields a PD limiting NTK although the sites coincide.
Linear independence of the local coefficient vectors, rather than coincident
sampling alone, distinguishes the two cases.
\end{example}

\begin{example}[Coupled vector rows require joint rank]
\label{ex:vector_coupling_rank_obstruction}
For $\mathbf u_\theta=(u_{1,\theta},u_{2,\theta})^\top$, fix one physical
point $\mathbf x_\star\in\overline\Omega$ and impose there
\begin{equation*}
 P_1=u_{1,\theta}(\mathbf x_\star),\qquad
 P_2=u_{2,\theta}(\mathbf x_\star),\qquad
 P_3=u_{1,\theta}(\mathbf x_\star)+u_{2,\theta}(\mathbf x_\star).
\end{equation*}
Then $\nabla_\theta P_3=\nabla_\theta P_1+\nabla_\theta P_2$, so the joint
NTK is singular at every width with null direction $(-1,-1,1)^\top$. Thus scalar
positivity for each output coordinate does not by itself imply positivity of
an arbitrary coupled, scalarized multi-output constraint family. Multiple
components alone are not the problem: for example, $u_1+u_2$ and $u_1-u_2$ have
independent coefficient rows. \citet{Jin2026} allow vector outputs with operators
acting componentwise; the additional issue here is coupled constraints at a
repeated site. The local vectors $\mathbf b_i(\mathbf x)$ record both the
component and derivative indices, and their joint linear independence is the
exact criterion under $\mathsf P_{\mathrm{pd}}$; see also~\Cref{ex:vectorvalued}.
\end{example}

\begin{example}[Compatible duplicate rows need only an active-subspace gap]
\label{ex:duplicate_active_subspace}
Let $\ell$ be any nonzero pointwise, weak, or nonlocal constraint covered by
\Cref{prop:limiting_ntk_rank,prop:functional_ntk_rank,prop:nonlocal_ntk_rank},
and duplicate its scaled prediction:
\begin{equation*}
 P_1(\theta)=P_2(\theta)=\ell(u_\theta).
\end{equation*}
If $g(\theta)=\|\nabla_\theta P_1(\theta)\|_2^2$, then the full tangent
kernel is
\begin{equation*}
 \mathbf K(\theta)=g(\theta)
 \begin{bmatrix}1&1\\1&1\end{bmatrix},
 \qquad
 \ker\mathbf K(\theta)=\linspan\{(1,-1)^\top\}
 \ \text{ whenever }g(\theta)>0.
\end{equation*}
Thus every theorem that assumes a strictly positive full Gram matrix excludes
this system. Equal targets $(c,c)^\top$ are nevertheless compatible: the
error remains in $\mathcal U_{\mathrm{dup}}=\linspan\{(1,1)^\top\}$, and the
compressed kernel on $\mathcal U_{\mathrm{dup}}$ is positive whenever
$g(\theta)>0$. The certificate required
by~\Cref{cor:active_subspace_convergence} concerns instead the population
value $\mathbf U^\top\mathbf K_{(0)}^\infty\mathbf U
=2[\mathbf K_{(0)}^\infty]_{11}$ with $\mathbf U=(1,1)^\top/\sqrt2$, which is
positive under the applicable ``pd'' clause because the single nonzero row
$\ell$ is linearly independent; under the ``reg'' clause alone it can vanish
(\Cref{ex:weak_analyticity_necessity}). Given this certificate, conditional
centering $\mathbb E[\mathbf a\mid\boldsymbol\xi]=\mathbf0$, and the
compressed width and step-size conditions,
\Cref{cor:active_subspace_convergence} gives geometric finite-width GD
convergence. For unequal targets $(c_1,c_2)^\top$, the objective satisfies
\begin{equation*}
 \|\mathbf v_P-\mathbf v_T\|_2^2
 =2\Bigl(\ell(u_\theta)-\frac{c_1+c_2}{2}\Bigr)^2+\frac{(c_1-c_2)^2}{2}
 \geq\frac{(c_1-c_2)^2}{2}
 =\|\Pi_{\mathcal N_{\mathrm{str}}}\mathbf v_T\|_2^2,
\end{equation*}
where $\mathcal N_{\mathrm{str}}=\linspan\{(1,-1)^\top\}
=\mathcal U_{\mathrm{dup}}^\perp$ is the structural nullspace
of~\Cref{def:structural_spaces_compatibility}. This is the irreducible term
in~\cref{eqn:structural_loss_compression}; it is attained whenever some
$\theta$ gives $\ell(u_\theta)=(c_1+c_2)/2$, which holds at every width as
soon as $\ell(\varphi_{\boldsymbol\xi,q})\neq0$ for some $\boldsymbol\xi$ and
$q$, since $\ell(u_\theta)$ is linear in the output weights. This single
construction applies, for example, to duplicate collocation rows and to
duplicate weak test functions.
\end{example}

\begin{example}[A sharp interface lies outside the strong-form class but gives a bounded weak row]
\label{ex:sharp_interface_weak}
On $(-1,1)$, let
\begin{equation*}
 \kappa(x)=\begin{cases}1,&x<0,\\2,&x\geq0.\end{cases}
\end{equation*}
The divergence-form operator $-(\kappa u')'$ cannot be expanded at $x=0$
as a strong-form finite sum of classical derivatives with pointwise coefficients unless
an interface interpretation is added. Take the network trial space to be
$\mathcal X=H^1(-1,1)$ and the test space to be $H_0^1(-1,1)$. For
$u\in\mathcal X$ and $v\in H_0^1(-1,1)$, the continuum weak row
\begin{equation*}
 \ell_v(u)=\int_{-1}^{1}\kappa(x)u'(x)v'(x)\,dx
\end{equation*}
is bounded on $\mathcal X$. Dirichlet values are imposed by the two separate,
bounded trace rows $u\mapsto u(-1)$ and $u\mapsto u(1)$. The distinction
between trial and test spaces is essential: an unconstrained tanh ridge
$x\mapsto\tanh(wx+\gamma b)$ generally does not belong to $H_0^1(-1,1)$.
\Cref{cor:continuum_weak_interface_poisson} verifies, rather than assumes, the
Banach-valued analyticity, density, and joint independence required for this
precise formulation. This example lies outside the strong-form
differential-operator theorem
of~\citet{Moser2026}, which explicitly sends sharp interfaces to a weak
formulation.
\end{example}

\begin{example}[Initialization law and moment existence]
\label{ex:gaussian_moment_obstruction}
Consider the analytic nonpolynomial activation $\sigma(z)=\exp(z^4)$ and a
nondegenerate Gaussian preactivation $Z$. Then
\begin{equation*}
 \mathbb E[\sigma(Z)^2]=\mathbb E[\exp(2Z^4)]=\infty.
\end{equation*}
Consequently, the ordinary infinite-width covariance and NTK expectations are
not finite for this activation under Gaussian initialization. This illustrates
why population-kernel statements require moment existence in addition to
activation regularity and nonpolynomiality. Our compact per-neuron
initialization ensures the relevant moments are finite on compact input and
parameter sets; in particular, one nonzero value row has a finite positive
limiting Gram entry under our assumptions.
\end{example}

Five of these six cases illustrate boundaries among theorem settings rather
than counterexamples to the cited results, and
\Cref{ex:gaussian_moment_obstruction} shows only that moment existence is part
of any Gaussian-initialization statement. Together they show why activation
regularity must match differential order, why co-located and coupled rows
require a joint rank test, why full-Gram positivity is unnecessary for
compatible training, why sharp interfaces call for a weak formulation, and
why population-kernel statements need moment existence in addition to
activation regularity.

%% file: ntk.tex
\section{Tangent-kernel limit, finite-width concentration, and local stability}
\label{sec:ntk_proof}
This section states and proves the three kernel properties on which the
positivity and optimization results rest: existence of the deterministic
limit, finite-width concentration, and local stability of the tangent Gram
matrix. \Cref{sec:ntk_statement} states them as~\Cref{thm:linearNTK};
\Cref{sec:linearNTK_proof} isolates the per-neuron regularity condition the
proofs use and verifies it classwise; \Cref{thm:generallinearNTK} then proves
the three items under that condition.

\subsection{Statement}
\label{sec:ntk_statement}

\begin{theorem}[NTK analysis for finite linear constraints]
\label{thm:linearNTK}
Use the network and loss in
\Cref{def:ntk_parameterization,def:pinn_loss}, and suppose the
row-regularity condition in~\Cref{assump:general} holds. Then:
\begin{enumerate}[leftmargin=*]
\item At initialization, the deterministic infinite-width NTK exists and has
the Gram representation
\begin{equation*}
[\mathbf K_{(0)}^\infty]_{ij}
=\mathbb E_{z\sim\mu}
\left[\left\langle\nabla h_i(z),\nabla h_j(z)\right\rangle\right]
=\sum_{p=1}^{d_\theta}
\mathbb E_{z\sim\mu}
\left[\partial_{z_p}h_i(z)\,\partial_{z_p}h_j(z)\right].
\end{equation*}
Thus it is the sum of \(d_\theta=d_0+d_2+1\) positive-semidefinite
contributions, and bounded continuous parameter-derivative functions in an
\(L^2(\mu)\) space define each contribution. If every width-\(d_1\) network uses the first \(d_1\)
vectors of one infinite i.i.d.\ sequence, then
\(\mathbf K_{(0)}\to\mathbf K_{(0)}^\infty\) almost surely; for separately
resampled widths, convergence is in probability. For finite-order pointwise
differential constraints, these parameter-derivative functions are analytic
when \(\sigma\) is real analytic.

\item Suppose the fixed design has a positive-definiteness certificate in the sense
of~\Cref{def:positive_definiteness_certificate}, so
\(\mathbf K_{(0)}^\infty\succ0\), and let
\(0<\lambda_0\leq\lambda_{\min}(\mathbf K_{(0)}^\infty)\) be a certified
lower bound. Define
\begin{equation*}
c_{\mathrm{Ch}}=\frac14+\frac34\log\frac34.
\end{equation*}
For \(0<\delta<1\), if
\begin{equation*}
d_1\geq \frac{R_K}{c_{\mathrm{Ch}}\lambda_0}
\log\!\left(\frac{N}{\delta}\right),
\end{equation*}
then
\(\lambda_{\min}(\mathbf K_{(0)})\geq3\lambda_0/4\) with probability at
 least \(1-\delta\). For the pointwise class under clause
 \(\mathsf P_{\mathrm{pd}}\) of \Cref{assump:pde_classes}, sitewise
 coefficient-vector independence gives
 the required positive-definiteness certificate through
 \Cref{lem:jinwu_sigma_independence,prop:limiting_ntk_rank,%
 sec:pointwise_positivity_certificate}.

\item If \(\theta_{(0)}\in(\supp\mu)^{d_1}\), which holds almost surely
under~\Cref{assump:main}, then for every
\(\theta\in\mathcal T_R(\theta_{(0)})\), direct perturbation
of the prediction Jacobian gives
\begin{equation*}
\lambda_{\min}(\mathbf K(\theta))
\geq\left(
\sqrt{\lambda_{\min}(\mathbf K_{(0)})}
-\frac{L_*}{\sqrt{d_1}}\|\theta-\theta_{(0)}\|_2
\right)_+^2,
\end{equation*}
and \(\lambda_{\max}(\mathbf K(\theta))\leq B_*^2\).
\end{enumerate}
\end{theorem}

The abstract proof is~\Cref{thm:generallinearNTK} below,
and~\Cref{prop:linearNTK} verifies its per-neuron regularity hypotheses for the
stated finite-order strong-form differential constraints. Matrix Chernoff
yields the second conclusion (\Cref{prop:positive_definitenessfinitewidth}),
and singular-value perturbation of the Jacobian yields the third
(\Cref{prop:positive_definiteness_via_parameters}).

\subsection{Abstract representation as a sum of per-neuron contributions}
\label{sec:linearNTK_proof}
We isolate the exact per-neuron sum in~\Cref{thm:linearNTK} and the
regularity conditions on each scalar prediction component used by the kernel and
optimization arguments. We then verify them classwise: for bounded functional
rows in~\Cref{lem:bounded_functional_row_regularity}, with Sobolev instances
in~\Cref{cor:weak_H1_row_regularity,cor:weak_Hk_row_regularity}, for
finite-measure rows in~\Cref{prop:finite_measure_row_regularity}, and for
finite-order strong-form linear differential constraints
in~\Cref{prop:linearNTK}.

\begin{definition}[Abstract row-regularity condition]
    \label{assump:general}
Under~\Cref{assump:main}, fix a target \(\mathbf v_T\) and a
finite collection of prediction components. Assemble them in
\(\mathbf v_P(\theta)\in\mathbb R^N\), and consider the loss
\(\mathcal L(\theta)=\|\mathbf v_P(\theta)-\mathbf v_T\|_2^2\). Write
\(\theta=(\theta_1,\ldots,\theta_{d_1})\), choose \(R>0\), and define the
product neighborhood
\begin{equation}
 \mathcal T_R(\theta_{(0)})
 :=\left\{\theta:
       \max_{1\leq m\leq d_1}
       \|\theta_m-\theta_{m,(0)}\|_2\leq R\right\}.
 \label{eqn:abstract_product_tube}
\end{equation}
For each scalar prediction component \(i\), one loss-scaled linear row
of~\Cref{def:linear_pde}, require
\begin{equation}
 P_i(\theta)=\frac1{\sqrt{d_1}}\sum_{m=1}^{d_1}h_i(\theta_m),
 \qquad
 h_i(\boldsymbol\xi,\mathbf a)
 =\mathbf a^\top\mathbf f_i(\boldsymbol\xi).
 \label{eqn:predictionmap}
\end{equation}
The fixed data encoded by \(\mathbf f_i\) may define any fixed scalar linear
observation satisfying the representation in~\cref{eqn:predictionmap} and the
bounds in~\cref{eqn:abstract_one_neuron_bounds} below. For a strong-form
local differential row, they include its sampled evaluation point (its
collocation point), its evaluated operator coefficients, and its loss scale
\(s_i\). Require deterministic constants
\(A_{0,i},B_i,L_i<\infty\), uniform in $d_1$, such
    that, for every \(z_0\in\supp\mu\) and all
\(z,z'\in\overline{\mathbb B}_R(z_0)\),
\begin{align}
 |h_i(z_0)|&\leq A_{0,i},&
 \|\nabla h_i(z)\|_2&\leq B_i,\nonumber\\
 \|\nabla h_i(z)-\nabla h_i(z')\|_2
   &\leq L_i\|z-z'\|_2.
 \label{eqn:abstract_one_neuron_bounds}
\end{align}
If \(\theta\in\mathcal T_R(\theta_{(0)})\) and every
\(\theta_{m,(0)}\in\supp\mu\), which holds almost surely
under~\Cref{assump:main}, then
\(\theta_m\in\overline{\mathbb B}_R(\theta_{m,(0)})\) for every \(m\), so
the bounds in~\cref{eqn:abstract_one_neuron_bounds} apply to every neuron of
\(\theta\) with center \(z_0=\theta_{m,(0)}\). Every statement in the
appendices that applies these bounds on \(\mathcal T_R(\theta_{(0)})\) is
made on this probability-one event \(\theta_{(0)}\in(\supp\mu)^{d_1}\); where
a statement does not repeat the qualifier, it is understood. Define
\begin{equation}
 A_{0,*}^2=\sum_iA_{0,i}^2,\qquad
 B_*^2=\sum_iB_i^2,\qquad
 L_*^2=\sum_iL_i^2.
 \label{eqn:abstract_aggregate_constants}
\end{equation}
For later use, recall $d_\theta=d_0+d_2+1$ and set
\(\boldsymbol\Psi(z)\in\mathbb R^{N\times d_\theta}\) by
\([\boldsymbol\Psi(z)]_{ip}=\partial h_i(z)/\partial z_p\), $p\in[d_\theta]$.
Differentiating~\cref{eqn:predictionmap} in \(\theta_m\) gives
\(\nabla_{\theta_m}P_i=d_1^{-1/2}\nabla h_i(\theta_m)\), so
\(\mathbf J(\theta)=d_1^{-1/2}[\boldsymbol\Psi(\theta_1)\ \cdots\
\boldsymbol\Psi(\theta_{d_1})]\). The
gradient-Lipschitz bound yields the quadratic Taylor remainder used in the
optimization argument through the integral of the gradient difference; this
derivation does not require an everywhere-defined Hessian. The condition
imposes neither analyticity nor nonpolynomiality.
\end{definition}
For the width-independent bound on the initial training error, we additionally impose
\(\mathbb E_{z\sim\mu}h_i(z)=0\) for every prediction component \(i\).  In
the network parameterization of the main text, the single structural
requirement
\(\mathbb E[\mathbf a\mid\boldsymbol\xi]=\mathbf0\) implies all these
identities; independence between hidden and output weights is unnecessary.
\begin{remark}[Generality of~\Cref{assump:general}]
Finite-order strong-form differential constraints satisfy the condition under
\Cref{assump:main} and order-matched activation smoothness. For a tested weak,
integro-differential, or fractional row, one must first
verify~\cref{eqn:predictionmap,eqn:abstract_one_neuron_bounds} for
that row; linearity and bounded coefficients alone do not imply the
required bounds.
\end{remark}

\begin{lemma}[Automatic regularity for bounded functional rows]
\label{lem:bounded_functional_row_regularity}
Under~\Cref{assump:main}, so that \(\supp\mu\) is compact, let
\(\mathcal X\) be a real Banach space of \(\mathbb R^{d_2}\)-valued functions
embedded continuously in \(L^1_{\mathrm{loc}}\) of its domain, so that an
\(\mathcal X\)-limit that also converges almost everywhere coincides almost
everywhere with its pointwise limit (this identifies Fr\'echet derivatives
with classical parameter derivatives in the corollaries below; the lemma
itself does not use it), let
\(\lambda_i\in\mathcal X^*\), \(i\in[N]\), be fixed bounded linear
functionals, and set
\begin{equation*}
 \mathcal G(z)
 =\mathbf a\,\sigma(\mathbf w^\top\!\cdot+\gamma b)\in\mathcal X,
 \qquad z=(\mathbf w,b,\mathbf a)\in\mathbb R^{d_\theta}.
\end{equation*}
Fix \(0<R<\infty\) and define the compact
parameter tube
\begin{equation*}
 \mathcal Z_R=\supp\mu+\overline{\mathbb B}_R(\mathbf0)
 =\{z:\operatorname{dist}(z,\supp\mu)\leq R\}.
\end{equation*}
Suppose that \(\mathcal G\) is twice continuously Fr\'echet differentiable
as an \(\mathcal X\)-valued map on an open neighborhood of \(\mathcal Z_R\). Define
\begin{align*}
 M_0&=\sup_{z\in\supp\mu}\|\mathcal G(z)\|_{\mathcal X},\\
 M_1&=\sup_{z\in \mathcal Z_R}
       \|D\mathcal G(z)\|_{\mathrm{op}},\\
 M_2&=\sup_{z\in \mathcal Z_R}
       \|D^2\mathcal G(z)\|_{\mathrm{op}}.
\end{align*}
Here \(\|\cdot\|_{\mathrm{op}}\) is taken with respect to the Euclidean norm
on \(\mathbb R^{d_\theta}\); for \(D^2\mathcal G(z)\) it is the bilinear norm
\(\sup_{\|\dot z\|_2=\|\widehat z\|_2=1}
\|D^2\mathcal G(z)[\dot z,\widehat z]\|_{\mathcal X}\).
These constants are finite. For the rows
\(P_i(\theta)=\lambda_i(\mathbf u(\cdot;\theta))\) of types (ii) and (iii)
in~\Cref{def:linear_pde}, since
\(\mathbf u(\cdot;\theta)=d_1^{-1/2}\sum_{m}\mathcal G(\theta_m)\) and
\(\lambda_i\) is linear, the scalar maps
\(h_i=\lambda_i\circ\mathcal G\) give the representation
in~\cref{eqn:predictionmap}, with
\begin{equation*}
 [\mathbf f_i(\boldsymbol\xi)]_q
 =\lambda_i\!\left(
   \mathbf e_q\sigma(\mathbf w^\top\!\cdot+\gamma b)\right),
 \qquad \boldsymbol\xi=(\mathbf w,b),
\end{equation*}
and satisfy~\cref{eqn:abstract_one_neuron_bounds} with the explicit valid
choices
\begin{equation}
 A_{0,i}=\|\lambda_i\|_{\mathcal X^*}M_0,
 \qquad
 B_i=\|\lambda_i\|_{\mathcal X^*}M_1,
 \qquad
 L_i=\|\lambda_i\|_{\mathcal X^*}M_2.
 \label{eqn:bounded_functional_regularity_constants}
\end{equation}
The same conclusion holds if \(\mathcal G\) is continuously Fr\'echet
differentiable on an open neighborhood of \(\mathcal Z_R\) and there is a finite
constant \(M_{1,1}\) such that
\begin{equation*}
 \|D\mathcal G(z)-D\mathcal G(z')\|_{\mathrm{op}}
 \leq M_{1,1}\|z-z'\|_2
 \qquad (z,z'\in \mathcal Z_R).
\end{equation*}
In this \(C^{1,1}\) case, take \(L_i=\|\lambda_i\|_{\mathcal X^*}M_{1,1}\)
in place of the third choice
in~\cref{eqn:bounded_functional_regularity_constants}.
In particular, Banach-valued real analyticity on an open neighborhood of the
tube implies the displayed \(C^2\) hypothesis and hence all three bounds.

As a specialization, let \(\mathcal V_i\) be normed, let
\(a_i:\mathcal X\times\mathcal V_i\to\mathbb R\) be a continuous
bilinear form, fix \(v_i\in\mathcal V_i\), and define the loss-scaled weak
row
\begin{equation*}
 \lambda_i(f)=s_i a_i(f,v_i),\qquad s_i>0.
\end{equation*}
If \(\|a_i\|_{\mathrm{bil}}\) denotes its bilinear operator norm, then
\begin{equation}
 \|\lambda_i\|_{\mathcal X^*}
 \leq s_i\|a_i\|_{\mathrm{bil}}\|v_i\|_{\mathcal V_i}.
 \label{eqn:weak_functional_dual_norm_bound}
\end{equation}
Consequently, substituting the right-hand side
of~\cref{eqn:weak_functional_dual_norm_bound} for
\(\|\lambda_i\|_{\mathcal X^*}\)
in~\cref{eqn:bounded_functional_regularity_constants} gives valid
\(A_{0,i},B_i,L_i\) for every such weak row. Any bounded boundary
functional on the same trial space is covered by the first part of the lemma.
\end{lemma}
\begin{proof}
Compactness of \(\supp\mu\) and finite dimensionality of the per-neuron parameter
space make \(\mathcal Z_R\) compact. Continuity of
\(\mathcal G,D\mathcal G,D^2\mathcal G\) therefore makes
\(M_0,M_1,M_2\) finite. Linearity of \(\lambda_i\) and the coordinate
decomposition
\begin{equation*}
 \mathcal G(\boldsymbol\xi,\mathbf a)
 =\sum_{q=1}^{d_2}(\mathbf a)_q
   \mathbf e_q\sigma(\mathbf w^\top\!\cdot+\gamma b)
\end{equation*}
give the claimed per-neuron representation. The Banach-space chain rule gives
\begin{equation*}
 Dh_i(z)=\lambda_i\circ D\mathcal G(z).
\end{equation*}
Boundedness of \(\lambda_i\) gives
\(|h_i(z_0)|\leq\|\lambda_i\|_{\mathcal X^*}M_0\) on \(\supp\mu\), and since
\(\nabla h_i(z)\) is the Riesz representative of \(Dh_i(z)\),
\(\|\nabla h_i(z)\|_2=\|Dh_i(z)\|_{\mathrm{op}}
\leq\|\lambda_i\|_{\mathcal X^*}M_1\) on \(\mathcal Z_R\).

Fix \(z_0\in\supp\mu\) and
\(z,z'\in\overline{\mathbb B}_R(z_0)\). Their line segment remains in
that ball and hence in \(\mathcal Z_R\). The Banach-space fundamental theorem of
calculus, applied to \(D\mathcal G\), yields
\begin{equation*}
 \|D\mathcal G(z)-D\mathcal G(z')\|_{\mathrm{op}}
 \leq M_2\|z-z'\|_2.
\end{equation*}
Composition with \(\lambda_i\) proves the Lipschitz-gradient estimate and
therefore~\cref{eqn:bounded_functional_regularity_constants}. The
\(C^{1,1}\) variant is the same argument with its defining Lipschitz bound.
Finally, continuity and bilinearity give
\begin{equation*}
 |\lambda_i(f)|
 \leq s_i\|a_i\|_{\mathrm{bil}}
       \|f\|_{\mathcal X}\|v_i\|_{\mathcal V_i},
\end{equation*}
which proves~\cref{eqn:weak_functional_dual_norm_bound} and the weak-row
specialization.
\end{proof}

\begin{corollary}[Automatic regularity for standard weak \(H^1\) rows]
\label{cor:weak_H1_row_regularity}
Let \(\Omega\subset\mathbb R^{d_0}\) be a bounded open set and take
\(\mathcal X=H^1(\Omega;\mathbb R^{d_2})\). Suppose the per-neuron
initialization law is width independent and compactly supported, fix
\(0<R<\infty\), and assume
\(\sigma\in C_{\mathrm{loc}}^3(\mathbb R)\). Then
\begin{equation*}
 z=(\mathbf w,b,\mathbf a)
 \longmapsto
 \mathbf a\,\sigma(\mathbf w^\top\!\cdot+\gamma b)
 \end{equation*}
is a \(C^2\) map into \(\mathcal X\) on an open neighborhood of the compact
tube \(\mathcal Z_R\) of~\Cref{lem:bounded_functional_row_regularity}.
Consequently, every fixed finite family of weak
rows \(\lambda_i(f)=s_i a_i(f,v_i)\), where \(\mathcal V_i\) is normed,
\(a_i:\mathcal X\times\mathcal V_i\to\mathbb R\) is continuous and
bilinear, and \(v_i\in\mathcal V_i\) is fixed, together with any bounded boundary
functionals on \(H^1(\Omega;\mathbb R^{d_2})\), satisfies
\cref{eqn:predictionmap,eqn:abstract_one_neuron_bounds}, with the constants
in~\cref{eqn:bounded_functional_regularity_constants}.

Here a boundary functional is covered only if it is already a bounded element
of \(H^1(\Omega;\mathbb R^{d_2})^*\), for example because an appropriate
tested trace theorem applies. Pointwise Dirichlet evaluation is not
automatically bounded on \(H^1\) when \(d_0\geq2\), and pointwise Neumann
evaluation is not generally bounded on \(H^1\); those rows require a smoother
trial space or a bounded tested trace. This corollary verifies
$(\mathsf G_R)$, equivalently the row bounds in~\Cref{assump:general}. It does
not establish~\Cref{assump:main}, centering,
Banach-valued analyticity or ridge density, joint functional independence, or
a positive-definiteness certificate.
\end{corollary}
\begin{proof}
Write
\begin{equation*}
 t_z(\mathbf x)=\mathbf w^\top\mathbf x+\gamma b,
 \qquad
 \ell_{\dot z}(\mathbf x)=\dot{\mathbf w}^{\top}\mathbf x+
                       \gamma\dot b
\end{equation*}
for a parameter direction
\(\dot z=(\dot{\mathbf w},\dot b,\dot{\mathbf a})\), and use
\(\ell_{\widehat z}\) analogously for
\(\widehat z=(\widehat{\mathbf w},\widehat b,\widehat{\mathbf a})\).
The candidate first and second Fr\'echet derivatives of
\(\mathcal G(z)=\mathbf a\sigma(t_z)\) are
\begin{align}
 D\mathcal G(z)[\dot z]
 &=\dot{\mathbf a}\,\sigma(t_z)
   +\mathbf a\,\sigma'(t_z)\ell_{\dot z},
 \label{eqn:H1_ridge_first_parameter_derivative}\\
 D^2\mathcal G(z)[\dot z,\widehat z]
 &=\dot{\mathbf a}\,\sigma'(t_z)\ell_{\widehat z}
   +\widehat{\mathbf a}\,\sigma'(t_z)\ell_{\dot z}
   +\mathbf a\,\sigma''(t_z)\ell_{\dot z}\ell_{\widehat z}.
 \label{eqn:H1_ridge_second_parameter_derivative}
\end{align}
Their spatial weak derivatives are represented by their classical
derivatives. In particular,
\begin{align*}
 \nabla_{\mathbf x}D\mathcal G(z)[\dot z]
 &=\dot{\mathbf a}\mathbf w^\top\sigma'(t_z)
   +\mathbf a\dot{\mathbf w}^{\top}\sigma'(t_z)
   +\mathbf a\mathbf w^\top\sigma''(t_z)\ell_{\dot z},\\
 \nabla_{\mathbf x}D^2\mathcal G(z)[\dot z,\widehat z]
 &=\dot{\mathbf a}\!\left[
      \mathbf w^\top\sigma''(t_z)\ell_{\widehat z}
      +\widehat{\mathbf w}^{\top}\sigma'(t_z)\right]\\
 &\quad+\widehat{\mathbf a}\!\left[
      \mathbf w^\top\sigma''(t_z)\ell_{\dot z}
      +\dot{\mathbf w}^{\top}\sigma'(t_z)\right]\\
 &\quad+\mathbf a\!\left[
      \mathbf w^\top\sigma'''(t_z)\ell_{\dot z}\ell_{\widehat z}
      +\dot{\mathbf w}^{\top}\sigma''(t_z)\ell_{\widehat z}
      +\widehat{\mathbf w}^{\top}\sigma''(t_z)\ell_{\dot z}
      \right].
\end{align*}

Choose \(\varepsilon>0\) and work on the enlarged compact tube
\(\mathcal Z_{R+\varepsilon}=\supp\mu+\overline{\mathbb B}_{R+\varepsilon}(\mathbf0)\),
whose interior is an open neighborhood of \(\mathcal Z_R\), as
\Cref{lem:bounded_functional_row_regularity} requires. Boundedness of \(\Omega\) and
compactness of this tube uniformly bound \(\mathbf x\), all parameters,
\(t_z(\mathbf x)\), and the affine functions
\(\ell_{\dot z}(\mathbf x)\) for unit parameter directions. Hence
\(\sigma^{(j)}\), \(0\leq j\leq3\), is bounded and uniformly continuous on
one compact interval containing every relevant affine argument. The displayed
formulas and their spatial derivatives are consequently uniformly bounded in
\(L^\infty(\Omega)\), in operator norm over unit directions.

Taylor's theorem in the scalar affine argument, with its remainder uniform on
that compact interval, shows from the same formulas that the difference
quotients defining~\cref{eqn:H1_ridge_first_parameter_derivative,%
eqn:H1_ridge_second_parameter_derivative} converge in
\(W^{1,\infty}(\Omega;\mathbb R^{d_2})\). Uniform continuity of
\(\sigma^{(j)}\), \(j\leq3\), also makes the two displayed derivative maps
continuous in their corresponding operator norms. Thus
\(z\mapsto\mathcal G(z)\) is \(C^2\) into \(W^{1,\infty}\) on an open
neighborhood of \(\mathcal Z_R\). Since \(|\Omega|<\infty\), the embedding
\begin{equation*}
 W^{1,\infty}(\Omega;\mathbb R^{d_2})
 \hookrightarrow H^1(\Omega;\mathbb R^{d_2})
\end{equation*}
is continuous, so the same formulas are its first and second Fr\'echet
derivatives as an \(H^1\)-valued map. Apply
\Cref{lem:bounded_functional_row_regularity} and
\cref{eqn:weak_functional_dual_norm_bound}.
\end{proof}

The same argument applies to Sobolev spaces of any order on domains of any
dimension, and a smoother trial space makes pointwise boundary rows
admissible in every dimension.

\begin{corollary}[Automatic regularity for weak \(H^k\) rows in any dimension]
\label{cor:weak_Hk_row_regularity}
Let \(\Omega\subset\mathbb R^{d_0}\) be a bounded open set,
\(k\in\mathbb N_0\), and \(\mathcal X=H^k(\Omega;\mathbb R^{d_2})\). Suppose
the per-neuron initialization law is width independent and compactly
supported, fix \(0<R<\infty\), and assume
\(\sigma\in C^{k+2}_{\mathrm{loc}}(\mathbb R)\). Then
\(z=(\mathbf w,b,\mathbf a)\mapsto\mathbf a\,\sigma(\mathbf w^\top\!\cdot+\gamma b)\)
is a \(C^2\) map into \(\mathcal X\) on an open neighborhood of the compact
tube \(\mathcal Z_R\), and every fixed finite family of bounded linear
functionals on \(\mathcal X\) satisfies
\cref{eqn:predictionmap,eqn:abstract_one_neuron_bounds}, hence
\((\mathsf G_R)\), with the constants of
\Cref{lem:bounded_functional_row_regularity}. If moreover \(\Omega\) is a
Lipschitz domain and \(k>d_0/2+r\) for
an integer \(r\geq0\), every pointwise evaluation
\(f\mapsto\partial^\alpha f_q(\mathbf x)\) with \(|\alpha|\leq r\) and
\(\mathbf x\in\overline\Omega\) is bounded on \(\mathcal X\), so pointwise
Dirichlet rows (\(r=0\)) and pointwise Neumann or higher-order boundary rows
(\(r\geq1\)) are admissible weak rows in every dimension.
\end{corollary}
\begin{proof}
Write \(t_z(\mathbf x)=\mathbf w^\top\mathbf x+\gamma b\). For a multi-index
\(\beta\) with \(|\beta|\leq k\),
\(\partial^\beta_{\mathbf x}[a_q\sigma(t_z(\mathbf x))]
=a_q\,\mathbf w^\beta\,\sigma^{(|\beta|)}(t_z(\mathbf x))\), which as a
function of \(z\) is a polynomial in \((\mathbf w,\mathbf a)\) times
\(\sigma^{(|\beta|)}(t_z(\mathbf x))\). On the enlarged compact tube
\(\mathcal Z_{R+\varepsilon}\), whose interior is an open neighborhood of
\(\mathcal Z_R\), the arguments
\(t_z(\mathbf x)\), \(\mathbf x\in\overline\Omega\), range over a compact
interval on which \(\sigma^{(|\beta|)}\), \(\sigma^{(|\beta|+1)}\), and
\(\sigma^{(|\beta|+2)}\) are bounded and uniformly continuous, because
\(\sigma\in C^{k+2}_{\mathrm{loc}}\). Hence each
\(z\mapsto\partial^\beta_{\mathbf x}[\mathbf a\,\sigma(t_z(\cdot))]\) is
\(C^2\) into \(C(\overline\Omega;\mathbb R^{d_2})\), with the pointwise
formulas as its derivatives, and since
\(C(\overline\Omega)\hookrightarrow L^2(\Omega)\) continuously on a bounded
domain, the map is \(C^2\) into \(H^k(\Omega;\mathbb R^{d_2})\); this is the
proof of \Cref{cor:weak_H1_row_regularity} with \(k\) spatial derivatives in
place of one. \Cref{lem:bounded_functional_row_regularity} then gives
\((\mathsf G_R)\) for any finite family of bounded functionals. For the last
claim, the Sobolev embedding
\(H^k(\Omega)\hookrightarrow C^r(\overline\Omega)\) holds on bounded Lipschitz
domains whenever \(k>d_0/2+r\)~\citep[Theorem~4.12]{Adams2003}, so pointwise
derivative evaluations of order at most \(r\) are bounded functionals on
\(\mathcal X\).
\end{proof}

\begin{proposition}[Automatic regularity of finite-measure rows]
\label{prop:finite_measure_row_regularity}
Let \(D\subset\mathbb R^{d_0}\) be compact. For \(i\in[N]\) and
\(q\in[d_2]\), let \(\bar\nu_{iq}=s_i\nu_{iq}\) be the loss-scaled finite
signed regular Borel measures of a type (iii) row in~\Cref{def:linear_pde},
so that \(\bar{\boldsymbol\nu}_i=(\bar\nu_{iq})_{q\in[d_2]}\), and
define
\begin{equation*}
 P_i(\theta)=\sum_{q=1}^{d_2}\int_D
 u_q(\mathbf y;\theta)\,d\bar\nu_{iq}(\mathbf y).
\end{equation*}
Suppose the per-neuron initialization law \(\mu\) is width independent and
compactly supported (\Cref{assump:main}), fix \(0<R<\infty\), and assume
\(\sigma\in C_{\mathrm{loc}}^2(\mathbb R)\). Then the induced per-neuron
maps satisfy~\cref{eqn:predictionmap,eqn:abstract_one_neuron_bounds}; in
particular, clause \(\mathsf N_{\mathrm{reg}}\) of~\Cref{assump:pde_classes}
implies \((\mathsf G_R)\) of~\Cref{def:local_row_regularity}.

More precisely, write
\begin{equation*}
 \widetilde{\mathbf y}=(\mathbf y,\gamma),\qquad
 \tau_i=\left(\sum_{q=1}^{d_2}
 \|\bar\nu_{iq}\|_{\mathrm{TV}}^2\right)^{1/2},\qquad
 \mathcal Z_R=\supp\mu+\overline{\mathbb B}_R(\mathbf0),
\end{equation*}
and set
\begin{align*}
 C_D&=\max_{\mathbf y\in D}\|\widetilde{\mathbf y}\|_2,
 &A_R&=\max_{(\boldsymbol\xi,\mathbf a)\in\mathcal Z_R}
       \|\mathbf a\|_2,\\
 T_{D,R}&=\max_{\substack{(\boldsymbol\xi,\mathbf a)\in\mathcal Z_R\\
                       \mathbf y\in D}}
       |\boldsymbol\xi^\top\widetilde{\mathbf y}|,
 &S_{j,R}&=\max_{|t|\leq T_{D,R}}|\sigma^{(j)}(t)|,
 \quad j=0,1,2.
\end{align*}
The following are valid choices:
\begin{align}
 A_{0,i}&=A_RS_{0,R}\tau_i,\nonumber\\
 B_i&=\tau_i\left(S_{0,R}+A_RS_{1,R}C_D\right),\nonumber\\
 L_i&=\tau_i\left(A_RS_{2,R}C_D^2+S_{1,R}C_D\right).
 \label{eqn:finite_measure_regularity_constants}
\end{align}
If \(\mathbb E[\mathbf a\mid\boldsymbol\xi]=\mathbf0\), then also
\(\mathbb E[h_i(\boldsymbol\xi,\mathbf a)]=0\) for every row.
\end{proposition}
\begin{proof}
Define \(\mathbf f_i:\mathbb R^{d_0+1}\to\mathbb R^{d_2}\) by
\begin{equation*}
 [\mathbf f_i(\boldsymbol\xi)]_q
 =\int_D\sigma(\boldsymbol\xi^\top\widetilde{\mathbf y})
 \,d\bar\nu_{iq}(\mathbf y).
\end{equation*}
Then
\begin{equation*}
 P_i(\theta)=\frac1{\sqrt{d_1}}\sum_{m=1}^{d_1}h_i(\theta_m),
 \qquad
 h_i(\boldsymbol\xi,\mathbf a)=\mathbf a^\top\mathbf f_i(\boldsymbol\xi).
\end{equation*}
Compactness of \(D\), local smoothness of \(\sigma\), and finite total
variation justify differentiation under each integral. For
\(\mathbf v,\mathbf v'\in\mathbb R^{d_0+1}\),
\begin{align*}
 D[\mathbf f_i]_q(\boldsymbol\xi)[\mathbf v]
 &=\int_D\sigma'(\boldsymbol\xi^\top\widetilde{\mathbf y})
   \mathbf v^\top\widetilde{\mathbf y}\,d\bar\nu_{iq}(\mathbf y),\\
 D^2[\mathbf f_i]_q(\boldsymbol\xi)[\mathbf v,\mathbf v']
 &=\int_D\sigma''(\boldsymbol\xi^\top\widetilde{\mathbf y})
   (\mathbf v^\top\widetilde{\mathbf y})
   (\mathbf v'^\top\widetilde{\mathbf y})
   \,d\bar\nu_{iq}(\mathbf y).
\end{align*}
Consequently, throughout \(\mathcal Z_R\),
\begin{equation*}
 \|\mathbf f_i\|_2\leq S_{0,R}\tau_i,\qquad
 \|D\mathbf f_i\|_{\mathrm{op}}\leq S_{1,R}C_D\tau_i,
 \qquad
 \|D^2\mathbf f_i\|_{\mathrm{op}}\leq S_{2,R}C_D^2\tau_i.
\end{equation*}
The gradient and Hessian of \(h_i\) have the block forms
\begin{equation*}
 \nabla h_i=
 \begin{pmatrix}(D\mathbf f_i)^\top\mathbf a\\ \mathbf f_i\end{pmatrix},
 \qquad
 \nabla^2h_i=
 \begin{pmatrix}
  \sum_q a_qD^2[\mathbf f_i]_q&(D\mathbf f_i)^\top\\
  D\mathbf f_i&0
 \end{pmatrix}.
\end{equation*}
Bounding \(\|D\mathbf f_i\|_{\mathrm{op}}\) by its Frobenius norm, using
\(\|\nabla h_i\|_2\leq\|(D\mathbf f_i)^\top\mathbf a\|_2+\|\mathbf f_i\|_2\),
\(\|\sum_qa_qD^2[\mathbf f_i]_q\|_{\mathrm{op}}
\leq\sum_q|a_q|\,S_{2,R}C_D^2\|\bar\nu_{iq}\|_{\mathrm{TV}}
\leq A_RS_{2,R}C_D^2\tau_i\) (Cauchy--Schwarz over \(q\)), and
\(
 \left\|\left(\begin{smallmatrix}A&B^\top\\B&0\end{smallmatrix}\right)\right\|_{\mathrm{op}}
 \leq\|A\|_{\mathrm{op}}+\|B\|_{\mathrm{op}}
\)
gives~\cref{eqn:finite_measure_regularity_constants}. The mean-value
estimate on each convex radius-\(R\) ball then gives the required
Lipschitz-gradient bound. Finally,
\(
 \mathbb E[h_i\mid\boldsymbol\xi]
 =\mathbb E[\mathbf a\mid\boldsymbol\xi]^\top\mathbf f_i(\boldsymbol\xi)=0
\), proving the centering assertion.
\end{proof}

We next check the componentwise conditions for the finite-order
strong-form differential operators in~\Cref{def:linear_pde}. The general results then apply
directly to any other linear row for which the same requirements are
verified.
\begin{proposition}[Row regularity of finite-order strong-form linear constraints]
    \label{prop:linearNTK}
For the network in~\Cref{def:ntk_parameterization}, suppose
\Cref{assump:main} holds. Let
\(r_{\max}\) be the largest differential order among the finite collection of
strong-form differential constraints in~\Cref{def:linear_pde}. If
\(\sigma\in C_{\mathrm{loc}}^{r_{\max}+2}(\mathbb R)\), or more generally
\(\sigma\in C_{\mathrm{loc}}^{r_{\max}+1,1}(\mathbb R)\), then, for every
fixed \(0<R<\infty\), this family satisfies
\cref{eqn:predictionmap,eqn:abstract_one_neuron_bounds} with radius \(R\) and
width-independent constants. In particular,
\cref{eqn:gd_row_condition} is then automatic for such constraints. If
\(\sigma\) is real analytic, each \(h_i\), hence every entry of
\(\boldsymbol\Psi\), is real analytic in \(z=(\boldsymbol\xi,\mathbf a)\).
\end{proposition}
\begin{proof}
    \phantomsection
    \label{proof:linearNTK}
    For row $i$, set
    $t_m(\mathbf x_i)=\mathbf w_m^\top\mathbf x_i+\gamma b_m$. Applying
    each mixed derivative in~\cref{eqn:scalar_constraint_main} to one neuron
    gives
    \begin{equation}
      P_i(\theta)=\frac1{\sqrt{d_1}}\sum_{m=1}^{d_1}h_i(\theta_m),
      \qquad
      h_i(\boldsymbol\xi_m,\mathbf a_m)
      =s_i\sum_{q=1}^{d_2}\sum_{\alpha\in I_{iq}}
      a_{iq\alpha}(\mathbf x_i)(\mathbf a_m)_q
      \sigma^{(|\alpha|)}(t_m(\mathbf x_i))\mathbf w_m^\alpha.
      \label{eqn:classical_one_neuron_contribution}
    \end{equation}
    where \(a_{iq\alpha}(\mathbf x_i)\) are the operator coefficients
    of~\cref{eqn:scalar_constraint_main} and \((\mathbf a_m)_q\) the output
    weights. Thus, the representation in~\cref{eqn:predictionmap} holds with
    \begin{equation*}
      [\mathbf f_i(\boldsymbol\xi_m)]_q
      =s_i\sum_{\alpha\in I_{iq}}a_{iq\alpha}(\mathbf x_i)
      \sigma^{(|\alpha|)}(t_m(\mathbf x_i))\mathbf w_m^\alpha.
    \end{equation*}
    Each $h_i$ is a finite sum of products of hidden-weight monomials,
    output-weight coordinates, and derivatives of $\sigma$ of order at most
    $r_{\max}$. Hence $\sigma\in C_{\mathrm{loc}}^{r_{\max}+2}$ gives a
    continuously differentiable $h_i$ with locally Lipschitz gradient;
    the weaker condition
    $\sigma\in C_{\mathrm{loc}}^{r_{\max}+1,1}$ yields the same conclusion
    directly. The tube $\mathcal Z_R=\supp\mu+\overline{\mathbb B}_R(\mathbf0)$
    is compact and width independent, so $h_i$ and $\nabla h_i$ are uniformly
    bounded there, and, each ball $\overline{\mathbb B}_R(z_0)$ being convex,
    the mean-value inequality turns the uniform local Lipschitz constant of
    $\nabla h_i$ on $\mathcal Z_R$ into a Lipschitz constant on the ball. This
    proves~\cref{eqn:abstract_one_neuron_bounds}; explicit constants follow.
    Write $\widetilde{\mathbf x}_i=(\mathbf x_i,\gamma)$,
    $X_i=\|\widetilde{\mathbf x}_i\|_2$,
    $W_R=\max_{\mathcal Z_R}\max\{1,\|\mathbf w\|_2\}$,
    $A_R=\max_{\mathcal Z_R}\|\mathbf a\|_2$,
    $T_{\mathcal S,R}=\max_{i}\max_{\mathcal Z_R}
    |\boldsymbol\xi^\top\widetilde{\mathbf x}_i|$, and
    $S_j=\max_{|t|\leq T_{\mathcal S,R}}|\sigma^{(j)}(t)|$ for
    $j\leq r_{\max}+2$; in the $C^{r_{\max}+1,1}_{\mathrm{loc}}$ case let
    $S_{r_{\max}+2}$ be the Lipschitz constant of $\sigma^{(r_{\max}+1)}$ on
    $[-T_{\mathcal S,R},T_{\mathcal S,R}]$. For
    $g_\alpha(\boldsymbol\xi)=\mathbf w^\alpha
    \sigma^{(|\alpha|)}(\boldsymbol\xi^\top\widetilde{\mathbf x}_i)$, the
    bounds $\|\nabla_{\mathbf w}\mathbf w^\alpha\|_2\leq|\alpha|W_R^{|\alpha|-1}$
    and $\|\nabla_{\mathbf w}^2\mathbf w^\alpha\|_{\mathrm{op}}
    \leq|\alpha|(|\alpha|-1)W_R^{|\alpha|-2}$ on $\mathcal Z_R$ and the
    product rule give, on $\mathcal Z_R$,
    \begin{align*}
     |g_\alpha|&\leq G_0(\alpha):=W_R^{|\alpha|}S_{|\alpha|},\\
     \|\nabla g_\alpha\|_2&\leq G_1(\alpha)
      :=W_R^{|\alpha|}\bigl(|\alpha|S_{|\alpha|}+X_iS_{|\alpha|+1}\bigr),\\
     \operatorname{Lip}(\nabla g_\alpha)&\leq G_2(\alpha)
      :=W_R^{|\alpha|}\bigl(|\alpha|(|\alpha|-1)S_{|\alpha|}
      +2|\alpha|X_iS_{|\alpha|+1}+X_i^2S_{|\alpha|+2}\bigr).
    \end{align*}
    With $F_{k,iq}=s_i\sum_{\alpha\in I_{iq}}|a_{iq\alpha}(\mathbf x_i)|\,
    G_k(\alpha)$ and $\mathbf F_{k,i}=(F_{k,iq})_{q\in[d_2]}$, the block
    computation in the proof of~\Cref{prop:finite_measure_row_regularity}
    yields the valid choices
    \begin{equation*}
     A_{0,i}=A_R\|\mathbf F_{0,i}\|_2,\qquad
     B_i=\|\mathbf F_{0,i}\|_2+A_R\|\mathbf F_{1,i}\|_2,\qquad
     L_i=\|\mathbf F_{1,i}\|_2+A_R\|\mathbf F_{2,i}\|_2,
    \end{equation*}
    all proportional to $s_i$ and independent of $d_1$. Finiteness of the
    sampled family makes the aggregate constants
    in~\cref{eqn:abstract_aggregate_constants} finite. If $\sigma$ is real
    analytic,~\cref{eqn:classical_one_neuron_contribution} is real analytic in
    the parameters of one neuron, and so is every entry of
    $\boldsymbol\Psi$.
\end{proof}

\begin{remark}[Weak and nonlocal rows]
Let \(\mathcal X\) be a real Banach space of \(\mathbb R^{d_2}\)-valued
functions embedded continuously in \(L^1_{\mathrm{loc}}\) as in clause
\(\mathsf W_{\mathrm{reg}}\), and let a loss-scaled bounded linear functional
\(\lambda_i\in\mathcal X^*\) induce a tested scalar constraint. Suppose the
complete one-neuron map
\(z\mapsto\mathbf a\,\sigma(\mathbf w^\top\cdot+\gamma b)\) is \(C^2\), or
more generally \(C^{1,1}\), as an \(\mathcal X\)-valued map on an open
neighborhood of the compact parameter tube \(\mathcal Z_R\). Then
\Cref{lem:bounded_functional_row_regularity} verifies
\cref{eqn:predictionmap,eqn:abstract_one_neuron_bounds}, with the explicit
constants in~\cref{eqn:bounded_functional_regularity_constants} (with
\(L_i=\|\lambda_i\|_{\mathcal X^*}M_{1,1}\) in the \(C^{1,1}\) case). In
particular, a continuous weak bilinear form with a fixed test function
satisfies the dual-norm estimate in
\cref{eqn:weak_functional_dual_norm_bound}; mixed weak-residual and bounded
boundary rows are covered when they act on one common trial space.
Finite-measure rows are the case \(\mathcal X=C(D;\mathbb R^{d_2})\) with
\(\|\lambda_i\|_{\mathcal X^*}\leq\sum_q\|\bar\nu_{iq}\|_{\mathrm{TV}}\);
\Cref{prop:finite_measure_row_regularity} gives the sharper
constants~\cref{eqn:finite_measure_regularity_constants} for them directly
under \(\sigma\in C^2_{\mathrm{loc}}\). Fixed-node quadrature rules without
derivative evaluations are atomic finite-measure rows and fall under that
proposition; rules containing derivative evaluations are a special case here
whenever every evaluation used by the rule is bounded on the chosen trial
space. Singular
integral, fractional, and continuum weak constraints require this regularity
check; coefficient boundedness alone does not suffice.
\end{remark}

\subsection{Abstract NTK analysis}
\begin{theorem}[Abstract NTK analysis]
    \label{thm:generallinearNTK}
    If the row-regularity condition in~\Cref{assump:general} holds, the
    deterministic-limit, finite-width concentration, and local stability
    statements in~\Cref{thm:linearNTK}
    hold, with bounded continuous parameter-derivative functions.
    For strong-form differential constraints of any fixed order, these
    functions are real analytic in the per-neuron parameters when \(\sigma\)
    is, by~\Cref{prop:linearNTK}.
    Weak or nonlocal cases require separate verification of this property.
\end{theorem}
We prove the three statements in~\Cref{thm:linearNTK} individually under this
condition.
\begin{proof}[Proof of item 1 of \Cref{thm:linearNTK}]
        \phantomsection
        \label{proof:ntk_deterministic}
Order the coordinates of \(z=\theta_m\), and define the parameter-derivative functions
\begin{equation*}
 \psi_{i,p}(z)=\partial_{z_p}h_i(z),\qquad p\in[d_\theta].
\end{equation*}
Take one infinite i.i.d.\ sequence of per-neuron parameter vectors and construct
the width-\(d_1\) network from its first \(d_1\) vectors. Embedding these models in
a common probability space gives almost-sure convergence as
\(d_1\to\infty\); fresh independent resampling at each size gives
convergence in probability. The prediction components
and constraint design remain fixed, or one samples them independently of the
parameter sequence; in the latter case, the following limit is conditional on
the realized design. The sum over neurons in~\cref{eqn:predictionmap} gives,
exactly,
\begin{equation*}
 [\mathbf K_{(0)}]_{ij}
 =\frac1{d_1}\sum_{m=1}^{d_1}\sum_{p=1}^{d_\theta}
 \psi_{i,p}(\theta_{m,(0)})\psi_{j,p}(\theta_{m,(0)}).
\end{equation*}
For fixed \(i,j,p\), these terms are i.i.d.\ by~\Cref{assump:main} and
bounded by~\cref{eqn:abstract_one_neuron_bounds}. The strong law therefore gives the
almost-sure deterministic limit
\begin{equation*}
 [\mathbf K_{(0)}^\infty]_{ij}
 =\sum_{p=1}^{d_\theta}\mathbb E_{z\sim\mu}
   [\psi_{i,p}(z)\psi_{j,p}(z)].
\end{equation*}
Equivalently,
\(\mathbf K_{(0)}^\infty=\mathbb E_{z\sim\mu}[\boldsymbol\Psi(z)\boldsymbol\Psi(z)^\top]\),
the population NTK of the main text. Among the \(d_\theta\) coordinates of
\(z\), \(d_2\) correspond to output weights and \(d_0+1\) to hidden affine
parameters. Each trainable coordinate \(p\in[d_\theta]\) contributes the
positive-semidefinite Gram matrix
\([\langle\Phi_i^{(p)},\Phi_j^{(p)}\rangle_{L^2(\mu)}]_{i,j\in[N]}\) of the
functions \(\Phi_i^{(p)}:=\psi_{i,p}\in L^2(\mu)\), so
\(\mathbf K_{(0)}^\infty\) is the positive-semidefinite sum of these
\(d_\theta\) matrices. These functions are bounded by \(B_i\) and Lipschitz
on every ball \(\overline{\mathbb B}_R(z_0)\), \(z_0\in\supp\mu\),
by~\Cref{assump:general}, hence continuous on \(\supp\mu\); in particular
\(\psi_{i,p}\in L^2(\mu)\). For the strong-form differential rows
of~\Cref{prop:linearNTK} with real analytic \(\sigma\), they are analytic in
the parameters of a single neuron.
\end{proof}

\begin{proposition}[Finite-width positive definiteness at initialization]
    \label{prop:positive_definitenessfinitewidth}
    Suppose the condition in~\Cref{assump:general} holds and the fixed design has a
    positive-definiteness certificate as
    in~\Cref{def:positive_definiteness_certificate}, so
    \(\mathbf K_{(0)}^\infty\succ0\), and let
    \(0<\lambda_0\leq\lambda_{\min}(\mathbf K_{(0)}^\infty)\) be a certified
    lower bound, and fix $0<\delta<1$. Set
    \begin{equation*}
    R_K:=\esssup_{z\sim\mu}
       \|\boldsymbol\Psi(z)\|_2^2\leq B_*^2
    \quad\text{(by~\cref{eqn:abstract_one_neuron_bounds})},
    \qquad
    c_{\mathrm{Ch}}:=\frac14+\frac34\log\frac34.
    \end{equation*}
    If
    \begin{equation*}
    d_1\geq \frac{R_K}{c_{\mathrm{Ch}}\lambda_0}
       \log\!\left(\frac{N}{\delta}\right),
    \end{equation*}
    then $\lambda_{\min}(\mathbf K_{(0)})\geq3\lambda_0/4$ with probability at least $1-\delta$.
\end{proposition}

\begin{proof}[Proof of \Cref{prop:positive_definitenessfinitewidth} (item 2 of \Cref{thm:linearNTK})]
    \phantomsection
    \label{proof:positive_definitenessfinitewidth}
Let $\mathbf J_{(0),m}\in\mathbb R^{N\times d_\theta}$ denote the
initialization-Jacobian block for neuron $m$, and define
\begin{equation*}
\boldsymbol\Psi_m=\sqrt{d_1}\,\mathbf J_{(0),m}
=\boldsymbol\Psi(\theta_{m,(0)}),
\qquad
\mathbf G_m=\boldsymbol\Psi_m\boldsymbol\Psi_m^\top.
\end{equation*}
The per-neuron parameter vectors are i.i.d., each $\mathbf G_m$ is positive semidefinite, and
\begin{equation*}
\mathbf K_{(0)}=\frac1{d_1}\sum_{m=1}^{d_1}\mathbf G_m,
\qquad
\mathbb E\mathbf G_m=\mathbf K_{(0)}^\infty.
\end{equation*}
Because \(\lambda_{\max}(\mathbf G_m)=\|\boldsymbol\Psi_m\|_2^2\) and
\(\theta_{m,(0)}\in\supp\mu\) almost surely, the bounds
in~\cref{eqn:abstract_one_neuron_bounds} give the almost-sure estimate
\begin{equation*}
 \lambda_{\max}(\mathbf G_m)
 \leq R_K
 =\esssup_{z\sim\mu}\|\boldsymbol\Psi(z)\|_2^2
 \leq\esssup_{z\sim\mu}
       \|\boldsymbol\Psi(z)\|_F^2
 \leq B_*^2.
\end{equation*}

Apply the lower-tail matrix Chernoff
inequality~\citep[Corollary~5.2]{Tropp2012} to the independent summands
\(\mathbf X_m=\mathbf G_m/d_1\), which satisfy
\(\lambda_{\max}(\mathbf X_m)\leq R_K/d_1\) almost surely and
\(\mu_{\min}:=\lambda_{\min}(\sum_m\mathbb E\mathbf X_m)
=\lambda_{\min}(\mathbf K_{(0)}^\infty)\geq\lambda_0\). Since
\(\{\lambda_{\min}(\mathbf K_{(0)})\leq(1-\varepsilon)\lambda_0\}\subseteq
\{\lambda_{\min}(\mathbf K_{(0)})\leq(1-\varepsilon)\mu_{\min}\}\) and the
base \(e^{-\varepsilon}/(1-\varepsilon)^{1-\varepsilon}\) lies in \((0,1]\),
replacing \(\mu_{\min}\) by \(\lambda_0\) in the exponent only weakens the
bound, and we obtain, for $0\leq\varepsilon<1$,
\begin{equation*}
\Pr\!\left\{\lambda_{\min}(\mathbf K_{(0)})
\leq(1-\varepsilon)\lambda_0\right\}
\leq N\left[
 \frac{e^{-\varepsilon}}{(1-\varepsilon)^{1-\varepsilon}}
 \right]^{d_1\lambda_0/R_K}.
\end{equation*}
Setting $\varepsilon=1/4$ gives the exponent $c_{\mathrm{Ch}}>0$
(so $c_{\mathrm{Ch}}^{-1}\approx29.207$) and yields
\begin{equation*}
\Pr\!\left\{\lambda_{\min}(\mathbf K_{(0)})
\leq\frac34\lambda_0\right\}
\leq N\exp\!\left(-c_{\mathrm{Ch}}
                      \frac{d_1\lambda_0}{R_K}\right).
\end{equation*}
The stated width makes the right-hand side at most $\delta$, proving the claim.
\end{proof}

\begin{remark}[Chernoff bound in limiting-kernel geometry]
The same argument can retain the population geometry. Define the
largest eigenvalue of a one-neuron Gram term in limiting-NTK coordinates:
\begin{equation*}
 \tau_K:=\esssup_{z\sim\mu}
 \lambda_{\max}\!\left(
   (\mathbf K_{(0)}^\infty)^{-1/2}
   \boldsymbol\Psi(z)\boldsymbol\Psi(z)^\top
   (\mathbf K_{(0)}^\infty)^{-1/2}\right).
\end{equation*}
The resulting whitened matrices have expectation \(\mathbf I_N\) and are bounded above
by \(\tau_K\mathbf I_N\). Matrix Chernoff therefore yields
\begin{equation*}
 \Pr\!\left\{\mathbf K_{(0)}\not\succeq
                   \frac34\mathbf K_{(0)}^\infty\right\}
 \leq N\exp\!\left(-c_{\mathrm{Ch}}d_1/\tau_K\right).
\end{equation*}
Indeed, the inequality first gives the good event for
\((\mathbf K_{(0)}^\infty)^{-1/2}\mathbf K_{(0)}
 (\mathbf K_{(0)}^\infty)^{-1/2}\), and congruence by
\((\mathbf K_{(0)}^\infty)^{1/2}\) gives the displayed matrix-order inequality.
Thus
\(d_1\geq c_{\mathrm{Ch}}^{-1}\tau_K\log(N/\delta)\) is an alternative
sufficient width for the event
\(\mathbf K_{(0)}\succeq\frac34\mathbf K_{(0)}^\infty\). Since
\(\tau_K\leq R_K/\lambda_{\min}(\mathbf K_{(0)}^\infty)\), it is never larger
than the width in~\Cref{prop:positive_definitenessfinitewidth} with
\(\lambda_0=\lambda_{\min}(\mathbf K_{(0)}^\infty)\), but it requires the
whitened quantity \(\tau_K\) rather than only \(R_K\) and a certified
\(\lambda_0\).
\end{remark}

\begin{remark}[Comparison with entrywise concentration]
An entrywise Hoeffding alternative with $B_K$ an almost-sure upper bound on
the range of each scalar summand requires
$d_1=\Omega(B_K^2N^2\lambda_0^{-2}\log(N/\delta))$. It takes a union over
the $N^2$ entries and then uses the Frobenius norm. Matrix
Chernoff exploits positivity and controls the required lower eigenvalue directly.
Its sufficient width is
$d_1=\Omega(R_K\lambda_0^{-1}\log(N/\delta))$. With a fixed number of loss
groups and each row divided by the square root of its group's sample count,
$B_K=O(1/N)$ and $R_K=O(1)$, so the two widths are
$\Omega(\lambda_0^{-2}\log(N/\delta))$ and
$\Omega(\lambda_0^{-1}\log(N/\delta))$; without normalization $B_K=O(1)$ and
$R_K=O(N)$ generically, and they are
$\Omega(N^2\lambda_0^{-2}\log(N/\delta))$ and
$\Omega(N\lambda_0^{-1}\log(N/\delta))$. In both regimes the Chernoff width
is never larger, since, with $B_K=\max_iB_i^2$,
$\lambda_0\leq\lambda_{\min}(\mathbf K_{(0)}^\infty)\leq N^{-1}\sum_iB_i^2\leq B_K$
and $R_K\leq B_*^2\leq NB_K$.
\end{remark}

\begin{proposition}[Positive definiteness during training]
    \label{prop:positive_definiteness_via_parameters}
    Suppose the condition in~\Cref{assump:general} holds, let
    $\theta_{(0)}\in(\supp\mu)^{d_1}$, which holds almost surely
    under~\Cref{assump:main}, and let
    $\theta\in\mathcal T_R(\theta_{(0)})$. Then
    $\lambda_{\max}(\mathbf K(\theta))\leq B_*^2$ and
    \begin{equation}
    \lambda_{\min}(\mathbf K(\theta))
    \geq\left(
      \sqrt{\lambda_{\min}(\mathbf K_{(0)})}
      -\frac{L_*}{\sqrt{d_1}}
       \|\theta-\theta_{(0)}\|_2
    \right)_+^2;
    \label{eqn:direct_jacobian_stability_bound}
    \end{equation}
    for an iterate $\theta=\theta_{(k)}$ this bounds
    $\lambda_{\min}(\mathbf K_{(k)})$. In particular, if $0<\lambda_{\mathrm{thr}}<\lambda_{\min}(\mathbf K_{(0)})$ and
    \begin{equation}
    \frac{L_*}{\sqrt{d_1}}
      \|\theta_{(k)}-\theta_{(0)}\|_2
    \leq
      \sqrt{\lambda_{\min}(\mathbf K_{(0)})}-\sqrt{\lambda_{\mathrm{thr}}},
    \label{eqn:direct_jacobian_stability_condition}
    \end{equation}
    then $\lambda_{\min}(\mathbf K_{(k)})\geq\lambda_{\mathrm{thr}}$.
\end{proposition}
\begin{proof}[Proof of \Cref{prop:positive_definiteness_via_parameters} (item 3 of \Cref{thm:linearNTK})]
    \phantomsection
    \label{proof:positive_definiteness_via_parameters}
For any real matrix $\mathbf A$ with $N$ rows, write
\(\sigma_{\min}(\mathbf A):=
\sqrt{\lambda_{\min}(\mathbf A\mathbf A^\top)}\); this convention includes
the necessary zero singular values when $\mathbf A$ has fewer than $N$
columns.
Recall that
\(\boldsymbol\Psi(z)\in\mathbb R^{N\times d_\theta}\) satisfies
\([\boldsymbol\Psi(z)]_{ip}=\partial h_i(z)/\partial z_p\). Summing the row
bounds in~\cref{eqn:abstract_one_neuron_bounds} over $i\in[N]$ and
using~\cref{eqn:abstract_aggregate_constants} together with
$\|\cdot\|_2\leq\|\cdot\|_F$ gives, for every $z_0\in\supp\mu$ and all
$z,z'\in\overline{\mathbb B}_R(z_0)$,
\begin{equation*}
 \|\boldsymbol\Psi(z)\|_2\leq B_*,\qquad
 \|\boldsymbol\Psi(z)-\boldsymbol\Psi(z')\|_F^2
 =\sum_{i=1}^N\|\nabla h_i(z)-\nabla h_i(z')\|_2^2
 \leq L_*^2\|z-z'\|_2^2.
\end{equation*}
Fix $\theta\in\mathcal T_R(\theta_{(0)})$. By~\cref{eqn:abstract_product_tube},
$\theta_m\in\overline{\mathbb B}_R(\theta_{m,(0)})$ for every $m$, and
$\theta_{m,(0)}\in\supp\mu$ by hypothesis, so the displayed bounds apply to
each pair $(z,z')=(\theta_m,\theta_{m,(0)})$ with $z_0=\theta_{m,(0)}$.
Write the full prediction Jacobian as a horizontal concatenation over neurons:
\begin{equation*}
 \mathbf J(\theta)=\frac1{\sqrt{d_1}}
 [\boldsymbol\Psi(\theta_1)\ \cdots\ \boldsymbol\Psi(\theta_{d_1})].
\end{equation*}
Then
\begin{align}
 \|\mathbf J(\theta)-\mathbf J(\theta_{(0)})\|_2^2
 &\leq \|\mathbf J(\theta)-\mathbf J(\theta_{(0)})\|_F^2
 =\frac1{d_1}\sum_{m=1}^{d_1}
  \|\boldsymbol\Psi(\theta_m)-\boldsymbol\Psi(\theta_{m,(0)})\|_F^2\nonumber\\
 &\leq \frac{L_*^2}{d_1}\sum_{m=1}^{d_1}\|\theta_m-\theta_{m,(0)}\|_2^2
 =\frac{L_*^2}{d_1}\|\theta-\theta_{(0)}\|_2^2,
 \label{eqn:direct_jacobian_perturbation}
\end{align}
so that
\(\|\mathbf J(\theta)-\mathbf J(\theta_{(0)})\|_2
\leq L_*\|\theta-\theta_{(0)}\|_2/\sqrt{d_1}\).
Also, because \(\mathbf K(\theta)=\mathbf J(\theta)\mathbf J(\theta)^\top\), for
every \(\theta\in\mathcal T_R(\theta_{(0)})\),
\begin{equation*}
 \lambda_{\max}(\mathbf K(\theta))
 \leq\frac1{d_1}\sum_{m=1}^{d_1}\|\boldsymbol\Psi(\theta_m)\|_2^2
 \leq B_*^2.
\end{equation*}
Weyl's inequality for singular values,
\(|\sigma_N(\mathbf A)-\sigma_N(\mathbf B)|\leq\|\mathbf A-\mathbf B\|_2\),
and~\cref{eqn:direct_jacobian_perturbation} yield
\begin{align*}
 \sigma_{\min}(\mathbf J(\theta))
 &\geq \sigma_{\min}(\mathbf J(\theta_{(0)}))
       -\|\mathbf J(\theta)-\mathbf J(\theta_{(0)})\|_2\\
 &\geq \sqrt{\lambda_{\min}(\mathbf K_{(0)})}
       -\frac{L_*}{\sqrt{d_1}}
        \|\theta-\theta_{(0)}\|_2.
\end{align*}
Taking the positive part and using
\(\lambda_{\min}(\mathbf J\mathbf J^\top)=\sigma_{\min}(\mathbf J)^2\)
proves~\cref{eqn:direct_jacobian_stability_bound}. Under the condition
in~\cref{eqn:direct_jacobian_stability_condition}, the right-hand side
of~\cref{eqn:direct_jacobian_stability_bound} at \(\theta=\theta_{(k)}\) is
at least \(\lambda_{\mathrm{thr}}\), which establishes the final assertion.
\end{proof}

%% file: appendix_extended_results.tex
\section{Extended positivity results}
\label{sec:extended_positivity}

\subsection{Independence lemma, finite-witness bound, and an explicit example}

The local rank characterization in~\Cref{prop:limiting_ntk_rank} uses the
following specialization of~\citet[Theorem~4]{Jin2026}.

\begin{lemma}[Independence of activation-derivative polynomial functions]
\label{lem:jinwu_sigma_independence}
Let \(\sigma\in C^\infty(\mathbb R)\) be nonpolynomial, and let
\(\mathbf x_1,\ldots,\mathbf x_M\in\mathbb R^{d_0}\) be pairwise distinct.
For each \(r\in[M]\), let \(J_r\subset\mathbb N_0^{d_0}\) be finite and
nonempty and let \((\kappa_{r,\alpha})_{\alpha\in J_r}\) be nonzero
coefficients. Define
\begin{equation}
Q_r(\mathbf w,v)
=\sum_{\alpha\in J_r}\kappa_{r,\alpha}
\sigma^{(|\alpha|)}(\mathbf w^\top\mathbf x_r+v)\mathbf w^\alpha.
\label{eqn:jinwu_specialized_polynomial}
\end{equation}
Then \(Q_1,\ldots,Q_M\) are linearly independent as functions on
\(\mathbb R^{d_0+1}\).
\end{lemma}
\begin{proof}
Because \(J_r\) is a set, the monomials \(\mathbf w^\alpha\),
\(\alpha\in J_r\), are pairwise distinct,
so~\cref{eqn:jinwu_specialized_polynomial} contains no like terms; every
coefficient \(\kappa_{r,\alpha}\) is nonzero; and the derivative order
\(|\alpha|\) of every summand equals the degree of its monomial
\(\mathbf w^\alpha\). Hence every \(Q_r\) is orderly and nondegenerate, that
is, in the standard form of~\citet[Definition~2]{Jin2026}. Theorem~4
of~\citet{Jin2026} states that if a linear combination, with all coefficients
nonzero, of such standard-form expressions at pairwise distinct sites vanishes
identically on \(\mathbb R^{d_0+1}\), then \(\sigma\) is a polynomial. Suppose
\(\sum_{r=1}^Mc_rQ_r\equiv0\) with not all \(c_r\) zero. Omitting the indices
with \(c_r=0\) leaves a nonempty relation satisfying these hypotheses, which
would force \(\sigma\) to be a polynomial, a contradiction.
\end{proof}

The exact rank results are qualitative. The next proposition gives a generic
finite-witness bound from finitely many values of the output feature matrix
$\mathbf F=\boldsymbol\Phi_{\mathrm{out}}$ defined below and its
variation on sets of positive initialization probability. Unlike the
Poisson-specific lower bound of~\citet{Jin2026b}, which is explicit in the
angular separation of the sites and holds for every separated site set under
that paper's dimension and RePU-order restrictions, it is an estimate for one
fixed design whose constants come from witness data for that design.

\begin{proposition}[Finite-witness lower bound for the smallest eigenvalue]
\label{prop:quantitative_dntk_bound}
Use the two-layer network in~\Cref{def:ntk_parameterization}, with its output
weights trainable, and let $\lambda_1,\ldots,\lambda_N$ be any fixed linear
observations on a common function space containing the coordinate ridge
features. Set $P_i(\theta)=\lambda_i(\mathbf u(\cdot;\theta))$ and write
\(\mathbf F(\boldsymbol\xi)=\boldsymbol\Phi_{\mathrm{out}}(\boldsymbol\xi)
\in\mathbb R^{N\times d_2}\) for the output feature matrix of the main text,
\begin{equation*}
F_{iq}(\boldsymbol\xi)
=\lambda_i\!\left(\mathbf e_q\sigma_{\mathbf w,b}\right),
\qquad \boldsymbol\xi=(\mathbf w,b),\qquad
\sigma_{\mathbf w,b}(\mathbf x)=\sigma(\mathbf w^\top\mathbf x+\gamma b).
\end{equation*}
Assume $F_{iq}\in L^2(\mu_h)$ for every $i,q$ and that the full limiting NTK
over all trainable parameters exists. The output-weight contribution is then
\(\mathbf K_{\mathrm{out}}^\infty
=\mathbb E_{\boldsymbol\xi\sim\mu_h}
[\mathbf F(\boldsymbol\xi)\mathbf F(\boldsymbol\xi)^\top]\). Suppose there are
pairwise disjoint measurable sets \(E_1,\ldots,E_L\), parameter values
\(\boldsymbol\xi_\ell\), and numbers \(\beta>0\),
\(\varepsilon\geq0\)
such that
\begin{equation*}
\mu_h(E_\ell)\geq\beta,
\qquad
\sup_{\boldsymbol\xi\in E_\ell}
\|\mathbf F(\boldsymbol\xi)-\mathbf F(\boldsymbol\xi_\ell)\|_2
\leq\varepsilon.
\end{equation*}
Let
\(\mathbf V=\left(\mathbf F(\boldsymbol\xi_1)^\top,\ldots,
\mathbf F(\boldsymbol\xi_L)^\top\right)^\top
\in\mathbb R^{Ld_2\times N}\), and assume
\(c_{\mathrm{sep}}:=\sigma_{\min}(\mathbf V)
=\sqrt{\lambda_{\min}(\mathbf V^\top\mathbf V)}>0\). Then
\begin{equation}
\lambda_{\min}(\mathbf K_{(0)}^\infty)
\geq\lambda_{\min}(\mathbf K_{\mathrm{out}}^\infty)
\geq \frac{\beta c_{\mathrm{sep}}^2}{2}-\varepsilon^2.
\label{eqn:quantitative_dntk_bound}
\end{equation}
In particular, \(\varepsilon\leq c_{\mathrm{sep}}\sqrt\beta/2\) implies
\(\lambda_{\min}(\mathbf K_{(0)}^\infty)\geq\beta c_{\mathrm{sep}}^2/4\).

For the pointwise constraints of~\Cref{def:linear_pde}, this statement
specializes to
\begin{equation*}
 F_{iq}(\boldsymbol\xi)
 =s_i(\mathcal A_{iq}\sigma_{\mathbf w,b})(\mathbf x_i).
\end{equation*}
For the weak and nonlocal rows considered below, respectively, it specializes
to
\begin{equation*}
 F_{iq}(\boldsymbol\xi)
 =s_i a_i(\mathbf e_q\sigma_{\mathbf w,b},v_i),
 \qquad
 F_{iq}(\boldsymbol\xi)
 =s_i\int_D\sigma_{\mathbf w,b}(\mathbf y)\,d\nu_{iq}(\mathbf y).
\end{equation*}
Thus the same finite-witness lower bound applies to a combined
residual--boundary feature matrix in the pointwise, weak, or nonlocal setting. No
analyticity or density hypothesis is needed for this quantitative implication
once the displayed features and limiting kernels exist.
\end{proposition}
See~\hyperref[proof:quantitative_dntk_bound]{the positivity appendix} for the
proof.

\begin{example}[A fully numerical finite-width GD certificate]
\label{ex:fully_numerical_gd_certificate}
This example instantiates every constant in~\Cref{thm:linearPDEsconvergence}
for one continuum nonlocal observation. It is a deliberately conservative
certificate, not a claim that the resulting width is necessary or practically
optimal. Let $d_0=d_2=1$, $\Omega=(-1,1)$, $D=\overline\Omega$, $\gamma=1$,
$\sigma=\tanh$, $s=1$, and
\begin{equation*}
 \lambda(f)=\frac12\int_{-1}^{1}f(x)\,dx,
 \qquad v_T=0.
\end{equation*}
Thus $N=1$ and, for $z=(w,b,a)$, the per-neuron prediction is
\begin{equation*}
 h(z)=aF(w,b),
 \qquad
 F(w,b)=\frac12\int_{-1}^{1}\tanh(wx+b)\,dx.
\end{equation*}
Initialize the neuron parameters independently across neurons, with mutually
independent coordinates
\begin{equation*}
 w\sim\operatorname{Unif}[-0.1,0.1],\qquad
 b\sim\operatorname{Unif}[1,2],\qquad
 a\sim\operatorname{Unif}[-0.1,0.1].
\end{equation*}
The law is width independent and compactly supported, its hidden marginal has
support with nonempty interior, and
$\mathbb E[a\mid w,b]=0$. Set $R=0.1$ and $\delta=0.05$.

We now verify the constants. At initialization, $wx+b\geq0.9$ for every
$x\in[-1,1]$. Since $\tanh$ is increasing,
\begin{equation}
\begin{aligned}
 \lambda_{\min}(\mathbf K_{(0)}^\infty)
 &=[\mathbf K_{(0)}^\infty]_{11}
  =\mathbb E\|\nabla h(z)\|_2^2\\
 &\geq\mathbb E\bigl[(\partial_a h(z))^2\bigr]
  =\mathbb E[F(w,b)^2]
  \geq\tanh(0.9)^2=:\lambda_0>0.51308.
\end{aligned}
\label{eqn:numerical_certificate_gap}
\end{equation}
Writing subscripts for derivatives of $F$, differentiation under the integral
gives
\begin{equation*}
 |F|\leq1,\qquad |F_w|\leq\frac12,\qquad |F_b|\leq1.
\end{equation*}
Because
$\nabla h=(aF_w,aF_b,F)$, the initialization bound $|a|\leq0.1$ yields
\begin{equation}
 R_K\leq1+0.1^2\left(\frac14+1\right)=1.0125.
 \label{eqn:numerical_certificate_RK}
\end{equation}
Every point in a radius-$R$ per-neuron neighborhood satisfies $|a|\leq0.2$.
The same derivative bounds therefore give the valid local choice
\begin{equation}
 B_*^2=1.05.
 \label{eqn:numerical_certificate_B}
\end{equation}

For completeness, the gradient-Lipschitz bound follows from the Hessian rather
than from an unspecified smoothness constant. The sharp global bound
\begin{equation*}
 \|\tanh''\|_\infty=\frac{4}{3\sqrt3}=:c_2
\end{equation*}
implies
\begin{equation*}
 |F_{ww}|\leq\frac{c_2}{3},\qquad
 |F_{wb}|\leq\frac{c_2}{2},\qquad
 |F_{bb}|\leq c_2.
\end{equation*}
Since
\begin{equation*}
 \nabla^2h=
 \begin{pmatrix}
  aF_{ww}&aF_{wb}&F_w\\
  aF_{wb}&aF_{bb}&F_b\\
  F_w&F_b&0
 \end{pmatrix},
\end{equation*}
its operator norm is at most its Frobenius norm, and throughout the local
neighborhood
\begin{align}
 \|\nabla^2h\|_2^2
 &\leq0.2^2c_2^2
 \left(\frac19+\frac12+1\right)
 +2\left(\frac14+1\right)\notag\\
 &=\frac{30839}{12150}<2.5382.
 \label{eqn:numerical_certificate_L}
\end{align}
Hence $L_*^2=2.5382$ is valid. Finally, $A_{0,*}=0.1$ and independence of
$a$ and $(w,b)$ give
\begin{equation*}
 H_0^2=\mathbb E[h(z)^2]
 \leq\mathbb E[a^2]=\frac{0.1^2}{3}.
\end{equation*}
Consequently, the initialization-residual quantity at $\delta=0.05$ obeys
\begin{equation}
 \overline r_{0.05}
 \leq\frac{0.1}{\sqrt3}
 +0.1\sqrt{2\log40}
 =:\widehat r<0.32936.
 \label{eqn:numerical_certificate_rbar}
\end{equation}
Here $\widehat r$ denotes the exact expression on the left of the last
inequality; $0.32936$ is its display rounding, and the bounds below are
evaluated with the exact value (with the rounded value the second bound
would read $1356.02$ and the integer width below would become $1357$).

Substitution of
\begin{equation*}
 \lambda_0=\tanh(0.9)^2,\quad
 R_K\leq1.0125,\quad \Lambda=B_*^2=1.05,\quad
 L_*^2=2.5382,\quad \overline r_{0.05}\leq\widehat r
\end{equation*}
into the three width terms
in~\cref{eqn:linear_pde_convergence_conditions} gives, using outward-rounded
interval evaluation and
$c_{\mathrm{Ch}}=\frac14+\frac34\log\frac34$,
\begin{align*}
 \frac{1.0125\log40}{c_{\mathrm{Ch}}\lambda_0}
 &<212.613,\\
 (320+128\sqrt6)
 \frac{2.5382(1.05)\widehat r^2}{\lambda_0^3}
 &<1355.980,\\
 \frac{16(1.05)\widehat r^2}
 {0.1^2\lambda_0^2}
 &<692.252,
\end{align*}
respectively. Therefore the integer width and step-size choices
\begin{equation}
 d_1\geq1356,\qquad 0<\eta\leq\frac{10}{21}
 \label{eqn:numerical_certificate_width_step}
\end{equation}
satisfy the theorem. With probability at least $0.95$, simultaneously for
every $k\geq0$,
\begin{align*}
 \max_{m\in[d_1]}\|\theta_{m,(k)}-\theta_{m,(0)}\|_2&\leq0.1,\\
 \mathbf K_{(k)}&\succeq\frac{\tanh(0.9)^2}{2}\mathbf I_1,\\
 \mathcal L(\theta_{(k)})
 &\leq\bigl(1-\eta\tanh(0.9)^2\bigr)^k\mathcal L(\theta_{(0)}).
\end{align*}
This calculation is for the exact integral functional, not a quadrature
surrogate. Its purpose is to show that the theorem's sufficient conditions can
be evaluated end to end; the constants deliberately use elementary global
derivative bounds and are not optimized.
\end{example}

\subsection{General Gram and direct-sum criteria}

These criteria apply when the pointwise rank criterion of
\Cref{prop:limiting_ntk_rank} is unavailable,
for example for a bias-free architecture or a constraint outside
the pointwise finite-order setting.

\begin{lemma}[Positive definiteness from independent continuous parameter-derivative functions]
\label{lem:positive_definiteness}
For $N$ sampled scalar rows, let \(\Phi_i\in L^2(\mu)\)
be real-valued parameter-derivative functions that are continuous on
\(\supp(\mu)\), and define
\([\mathbf K]_{ij}=\langle\Phi_i,\Phi_j\rangle_{L^2(\mu)}\).
Then \(\mathbf K\) is positive definite if and only if
\(\{\Phi_i\}_{i=1}^N\) is linearly independent in \(L^2(\mu)\), equivalently
on \(\supp(\mu)\).
\end{lemma}
See~\Cref{sec:pd_proof} for the
proof.

\begin{corollary}[Exact direct-sum criterion for the NTK]
\label{cor:positive_definiteness}
Given the limiting representation in the first item of~\Cref{thm:linearNTK}, let
\(\Phi_i^{(p)}\) denote the feature associated with constraint \(i\) in the
NTK contribution from coordinate \(p\), and
define its coordinate Gram matrix by
\begin{equation*}
 [\mathbf K_{(0)}^{\infty,(p)}]_{ij}
 =\langle\Phi_i^{(p)},\Phi_j^{(p)}\rangle_{L^2(\mu)}.
\end{equation*}
Then \(\mathbf K_{(0)}^\infty\) is positive definite if and only if the direct sums
of parameter-derivative functions
\begin{equation*}
\left(\Phi_i^{(1)},\ldots,\Phi_i^{(d_\theta)}\right),
\qquad i\in[N],
\end{equation*}
are linearly independent in \(L^2(\mu)^{d_\theta}\). Moreover,
\begin{equation*}
\ker\mathbf K_{(0)}^\infty
=\bigcap_{p=1}^{d_\theta}\ker\mathbf K_{(0)}^{\infty,(p)},
\qquad
\mathbf K_{(0)}^\infty\succ0
\quad\Longleftrightarrow\quad
\bigcap_{p=1}^{d_\theta}\ker\mathbf K_{(0)}^{\infty,(p)}=\{0\}.
\end{equation*}
In particular, a positive-definite contribution from one coordinate
is sufficient but, in general, not necessary.
\end{corollary}
\begin{proof}
The claim follows by applying~\Cref{lem:direct_sum_feature_criterion} to the
summands in~\Cref{thm:linearNTK}.
\end{proof}

\subsection{Finite-dimensional rank conditions for analytic functionals}

Under the nonempty-interior support hypothesis below, analyticity makes the
preceding Gram condition equivalent to full row rank for both a finite
Taylor-coefficient matrix and one evaluated at finitely many parameter values.
Without an a priori degree bound, failure at one fixed truncation does not imply
that the NTK is singular.

\begin{proposition}[Taylor-coefficient and evaluation rank characterization]
\label{prop:analytic_functional_rank}
For the two-layer network in~\Cref{def:ntk_parameterization}, with its output
weights trainable, let
$P_i(\theta)=\lambda_i(\mathbf u(\cdot;\theta))$, where the $\lambda_i$ are
linear on a common space containing all network realizations and all
single-neuron functions displayed below. Define
\begin{equation*}
 F_{iq}(\boldsymbol\xi)
 =\lambda_i\!\left(
 \mathbf e_q\sigma(\mathbf w^\top\!\cdot+\gamma b)\right),
 \qquad \boldsymbol\xi=(\mathbf w,b).
\end{equation*}
Suppose every $F_{iq}$ is real analytic on $\mathbb R^{d_0+1}$, the hidden
law $\mu_h$ has compact support with nonempty interior, and the full limiting
NTK exists. Fix $\boldsymbol\xi_\star\in\mathbb R^{d_0+1}$. For
$K\in\mathbb N_0$, let
$\mathcal I_K=\{\boldsymbol\rho\in\mathbb N_0^{d_0+1}:
|\boldsymbol\rho|\leq K\}$, and define
\begin{equation*}
 \mathbf C_K(\boldsymbol\xi_\star)
 =\left[
 \frac{1}{\boldsymbol\rho!}
 \partial_{\boldsymbol\xi}^{\boldsymbol\rho}
 F_{iq}(\boldsymbol\xi_\star)
 \right]_{\substack{i\in[N]\\(q,\boldsymbol\rho)\in[d_2]\times\mathcal I_K}}.
\end{equation*}
Then
\begin{equation}
 \ker\mathbf K_{\mathrm{out}}^\infty
 =\ker\mathbf K_{(0)}^\infty
 =\bigcap_{K\geq0}\ker\mathbf C_K(\boldsymbol\xi_\star)^\top.
 \label{eqn:analytic_coefficient_nullspace}
\end{equation}
Consequently, the following are equivalent:
\begin{enumerate}[label=(\roman*),nosep]
\item $\mathbf K_{(0)}^\infty\succ0$;
\item $\rank\mathbf C_K(\boldsymbol\xi_\star)=N$ for some finite $K$;
\item for some finite parameter set
$\Xi=\{\boldsymbol\xi^{(1)},\ldots,\boldsymbol\xi^{(T)}\}$, the following
evaluation matrix has row rank $N$:
\begin{equation*}
 \mathbf E_\Xi=\left[F_{iq}(\boldsymbol\xi^{(t)})\right]_{i,(t,q)}.
\end{equation*}
\end{enumerate}
Moreover, $\Xi$ may be chosen within any prescribed nonempty open subset of
$\mathbb R^{d_0+1}$.
\end{proposition}
\begin{proof}
For $\mathbf c\in\mathbb R^N$, the output-weight Gram representation gives
\begin{equation*}
 \mathbf c^\top\mathbf K_{\mathrm{out}}^\infty\mathbf c
 =\sum_{q=1}^{d_2}\int
 \left|\sum_{i=1}^Nc_iF_{iq}(\boldsymbol\xi)\right|^2
 \,d\mu_h(\boldsymbol\xi).
\end{equation*}
If the quadratic form vanishes, continuity forces each such sum to be null on
$\supp\mu_h$.
Because this set contains an open subset, the identity theorem makes each
combination identically zero. This conclusion is equivalent to the vanishing of all Taylor
coefficients at $\boldsymbol\xi_\star$, proving the output-weight equality
in~\cref{eqn:analytic_coefficient_nullspace}.

The same relation among these features also gives, at every
finite width and parameter value,
\begin{equation*}
 \sum_{i=1}^N c_iP_i(\theta)
 =\frac1{\sqrt{d_1}}\sum_{m=1}^{d_1}\sum_{q=1}^{d_2}(\mathbf a_m)_q
 \sum_{i=1}^N c_iF_{iq}(\boldsymbol\xi_m)=0.
\end{equation*}
Differentiation shows that $\mathbf c$ annihilates every gradient block.
Conversely, the output-weight kernel is a positive-semidefinite
summand of the full limiting kernel. The two implications prove the
identity in~\cref{eqn:analytic_coefficient_nullspace}.

The columns of the matrices $\mathbf C_K(\boldsymbol\xi_\star)$, $K\geq0$,
span $\mathbb R^N$ exactly when
$\bigcap_K\ker\mathbf C_K(\boldsymbol\xi_\star)^\top=\{0\}$. Since
$\rank\mathbf C_K(\boldsymbol\xi_\star)$ is nondecreasing in $K$ and bounded
by $N$, it stabilizes, so full rank is reached at a finite $K$. Applying the
same left-nullspace
argument to the evaluation vectors proves assertion (iii). If their span were deficient on
a nonempty open set, a nonzero such combination would vanish there and
hence everywhere, contradicting condition (i).
\end{proof}

\begin{corollary}[Rank conditions from monomials and moments]
\label{cor:monomial_moment_rank}
In the setting of~\Cref{prop:analytic_functional_rank}, suppose that
$\sigma$ is real analytic and nonpolynomial and $\gamma>0$ (two conditions
not contained in clause $\mathsf W_{\mathrm{pd}}$), and that the maps
$\boldsymbol\xi\mapsto
\mathbf e_q\sigma(\mathbf w^\top\!\cdot+\gamma b)$ are analytic in a common
Banach trial space $\mathcal X$ on which every $\lambda_i$ is bounded and
whose norm convergence implies pointwise convergence along a subsequence,
almost everywhere on $\Omega$ for the $L^1_{\mathrm{loc}}$-embedded spaces of
clause $\mathsf W_{\mathrm{reg}}$ and everywhere on $D$ for
$C(D;\mathbb R^{d_2})$ on a compact $D$. For
$p_\alpha(\mathbf x)=\mathbf x^\alpha$, choose,
for each $\alpha$, an $r_\alpha\in\mathbb N_0$ such that
$\sigma^{(|\alpha|+r_\alpha)}(0)\neq0$. Banach-valued differentiation gives
\begin{equation*}
 \left.
 \partial_{\mathbf w}^{\alpha}\partial_b^{r_\alpha}
 \left[\mathbf e_q\sigma(\mathbf w^\top\!\cdot+\gamma b)\right]
 \right|_{(\mathbf w,b)=(\mathbf0,0)}
 =\gamma^{r_\alpha}\sigma^{(|\alpha|+r_\alpha)}(0)
 \mathbf e_q p_\alpha\in\mathcal X.
\end{equation*}
Thus $\mathbf e_q p_\alpha\in\mathcal X$, and we may define
\begin{equation*}
 \mathbf M_K=\left[\lambda_i(\mathbf e_q p_\alpha)\right]_
 {\substack{i\in[N]\\q\in[d_2],\ |\alpha|\leq K}}.
\end{equation*}
Then
\begin{equation}
 \ker\mathbf K_{(0)}^\infty
 =\bigcap_{K\geq0}\ker\mathbf M_K^\top,
 \qquad
 \mathbf K_{(0)}^\infty\succ0
 \quad\Longleftrightarrow\quad
 \rank\mathbf M_K=N\ \text{for some finite }K.
 \label{eqn:monomial_rank_equivalence}
\end{equation}
For weak functionals $\lambda_i(f)=s_i a_i(f,v_i)$ and integral observations,
respectively, the entries become
\begin{align}
 [\mathbf M_K^{\mathrm{weak}}]_{i,(q,\alpha)}
 &=s_i a_i(\mathbf e_q p_\alpha,v_i),
 \label{eqn:weak_monomial_matrix}\\
 [\mathbf M_K^{\mathrm{nl}}]_{i,(q,\alpha)}
 &=s_i\int_D p_\alpha(\mathbf y)\,d\nu_{iq}(\mathbf y).
 \label{eqn:nonlocal_moment_matrix}
\end{align}
On compact $D$, the loss-scaled measure tuples
$\bar{\boldsymbol\nu}_1,\ldots,\bar{\boldsymbol\nu}_N$ are linearly independent if
and only if $\mathbf M_K^{\mathrm{nl}}$ is surjective onto $\mathbb R^N$ for some finite $K$. If instead
every $\nu_{iq}$ is a combination of Dirac masses at a common finite set of
distinct nodes $\mathbf y_1,\ldots,\mathbf y_A$,
$\nu_{iq}=\sum_{a=1}^A m_{iqa}\delta_{\mathbf y_a}$, then independence of the
loss-scaled measure tuples is equivalent to
row rank $N$ of the Dirac-coefficient matrix
\begin{equation}
 \mathbf C^{\mathrm{Dirac}}=\left[s_im_{iqa}\right]_{i,(q,a)}.
 \label{eqn:nonlocal_dirac_coefficient_matrix}
\end{equation}
Include all residual and boundary rows jointly in these matrices.
\end{corollary}
\begin{proof}
At $\boldsymbol\xi_\star=(\mathbf0,b_\star)$, boundedness of $\lambda_i$
and Banach-valued analyticity allow differentiation through the functional.
For every $r\in\mathbb N_0$,
\begin{equation}
 \partial_{\mathbf w}^{\alpha}\partial_b^rF_{iq}(\mathbf0,b_\star)
 =\gamma^r\sigma^{(|\alpha|+r)}(\gamma b_\star)
 \lambda_i(\mathbf e_q p_\alpha).
 \label{eqn:functional_feature_taylor_moment_identity}
\end{equation}
At every real center, a nonpolynomial analytic function has nonzero
derivatives of arbitrarily high order. Hence, for each $\alpha$, some
$r\in\mathbb N_0$ makes the prefactor
in~\cref{eqn:functional_feature_taylor_moment_identity} not vanish. The
Taylor-coefficient vectors and
$[\lambda_i(\mathbf e_q p_\alpha)]_{i=1}^N$ therefore span the same subspace of
$\mathbb R^N$. Equivalently, their common left nullspaces coincide.
\Cref{prop:analytic_functional_rank} then
proves~\cref{eqn:monomial_rank_equivalence} and the two displayed
specializations.

For compact $D$, the Stone--Weierstrass theorem makes polynomial restrictions
dense in $C(D)$, so a finite signed regular Borel measure that annihilates
all polynomials annihilates $C(D)$ and vanishes by Riesz uniqueness.
Thus the infinite moment matrix has row rank $N$ exactly when the loss-scaled
tuples $\bar{\boldsymbol\nu}_1,\ldots,\bar{\boldsymbol\nu}_N$ admit no nonzero
linear relation; finite dimensionality makes the rank
visible at some finite $K$. For the discrete measures above, distinct Dirac
masses are linearly independent,
giving~\cref{eqn:nonlocal_dirac_coefficient_matrix}.
\end{proof}

Cauchy estimates for $\sigma$ on a compact interval and the same
differentiation identity, with density of polynomials in $H^k$, verify,
respectively, the analyticity and density
clauses of $\mathsf W_{\mathrm{pd}}$ on Sobolev trial spaces in every
dimension, so the weak class is not restricted to one spatial dimension.

\begin{lemma}[Ridge analyticity and density in Sobolev spaces, any dimension]
\label{lem:ridge_density_sobolev}
Let \(\Omega\subset\mathbb R^{d_0}\) be a bounded open set with the segment
property (for instance a bounded Lipschitz domain), \(k\in\mathbb N_0\), and
\(\mathcal X=H^k(\Omega;\mathbb R^{d_2})\). Let \(\sigma\) be real analytic
and nonpolynomial and \(\gamma>0\). Then every map
\(\boldsymbol\xi\mapsto\varphi_{\boldsymbol\xi,q}
=\mathbf e_q\sigma(\mathbf w^\top\!\cdot+\gamma b)\) is real analytic from
\(\mathbb R^{d_0+1}\) into \(\mathcal X\), and
\begin{equation*}
 \overline{\linspan\{\varphi_{\boldsymbol\xi,q}:
 \boldsymbol\xi\in\mathbb R^{d_0+1},\ q\in[d_2]\}}^{\mathcal X}
 =\mathcal X .
\end{equation*}
Consequently the analyticity and density conditions displayed in clause
\(\mathsf W_{\mathrm{pd}}\) of~\Cref{assump:pde_classes} hold for this trial
space in every dimension \(d_0\); the remaining condition of
\(\mathsf W_{\mathrm{pd}}\), \(\operatorname{int}(\supp\mu_h)\neq\varnothing\),
concerns \(\mu_h\) alone. Clause
\(\mathsf W_{\mathrm{reg}}\) is then verified for a given finite family of
rows by~\Cref{cor:weak_Hk_row_regularity} as soon as every row is bounded on
\(\mathcal X\).
\end{lemma}
\begin{proof}
\emph{Analyticity.} Fix \(\boldsymbol\xi_0=(\mathbf w_0,b_0)\). The arguments
\(t_0(\mathbf x)=\mathbf w_0^\top\mathbf x+\gamma b_0\), \(\mathbf x\in\overline\Omega\),
lie in a compact interval \(I\). A real analytic \(\sigma\) extends
holomorphically to a complex neighborhood of \(I\), so the Cauchy estimates
give \(\sup_{t\in I}|\sigma^{(n)}(t)|\leq C\,n!\,\rho^{-n}\) for some
\(C,\rho>0\) and all \(n\). For
\(\boldsymbol\xi=\boldsymbol\xi_0+\boldsymbol\delta\) with
\(\boldsymbol\delta=(\boldsymbol\delta_{\mathbf w},\delta_b)\), Taylor's
theorem in the scalar argument gives
\begin{equation*}
 \sigma(\mathbf w^\top\mathbf x+\gamma b)
 =\sum_{n\geq0}\frac1{n!}\,\sigma^{(n)}(t_0(\mathbf x))
 \bigl(\boldsymbol\delta_{\mathbf w}^\top\mathbf x+\gamma\delta_b\bigr)^n .
\end{equation*}
The \(n\)-th term is a homogeneous polynomial of degree \(n\) in
\(\boldsymbol\delta\) with coefficients in \(C^k(\overline\Omega)\):
differentiating it \(j\leq k\) times in \(\mathbf x\) produces
\(\sigma^{(n+i)}(t_0(\mathbf x))\), \(i\leq j\), times polynomials in
\(\boldsymbol\delta_{\mathbf w}\) and in
\(\boldsymbol\delta_{\mathbf w}^\top\mathbf x+\gamma\delta_b\) of total degree
\(n\), so by the Cauchy estimates its \(C^k(\overline\Omega)\) norm is at
most \(C'(n+k)^k(\Lambda_\Omega\|\boldsymbol\delta\|_2/\rho)^n\) with
\(\Lambda_\Omega=\sup_{\mathbf x\in\overline\Omega}\|(\mathbf x,\gamma)\|_2\).
The series therefore converges absolutely in \(C^k(\overline\Omega)\), hence
in \(H^k(\Omega)\) on the bounded set \(\Omega\), for
\(\|\boldsymbol\delta\|_2<\rho/\Lambda_\Omega\), which is real analyticity
into \(\mathcal X\).

\emph{Density.} Let \(V\) be the closed linear span. Derivatives of a
Banach-space-valued analytic map at a point are limits of finite linear
combinations of its values, hence lie in \(V\). By the identity
in~\Cref{cor:monomial_moment_rank},
\(\partial_{\mathbf w}^{\alpha}\partial_b^{r_\alpha}\varphi_{\boldsymbol\xi,q}
|_{\boldsymbol\xi=\mathbf0}
=\gamma^{r_\alpha}\sigma^{(|\alpha|+r_\alpha)}(0)\,\mathbf e_q\mathbf x^\alpha\),
and because \(\sigma\) is nonpolynomial, for each \(\alpha\) some
\(r_\alpha\) has \(\sigma^{(|\alpha|+r_\alpha)}(0)\neq0\): otherwise the
Taylor series of \(\sigma\) at \(0\) would terminate and \(\sigma\) would be
a polynomial by analyticity. Thus \(V\) contains every vector monomial
\(\mathbf e_q\mathbf x^\alpha\), hence all \(\mathbb R^{d_2}\)-valued
polynomials. Polynomials are dense in \(H^k(\Omega)\): restrictions to
\(\Omega\) of functions in \(C_c^\infty(\mathbb R^{d_0})\) are dense in
\(H^k(\Omega)\) for domains with the segment
property~\citep[Theorem~3.22]{Adams2003}, and any such function is
approximated in \(C^k\) on a compact box containing \(\overline\Omega\) by
polynomials (Weierstrass approximation of the function and its derivatives,
for instance by Bernstein polynomials), hence in \(H^k(\Omega)\). So
\(V=\mathcal X\).
\end{proof}

\begin{remark}[Quadrature and numerical rank checks]
For a quadrature-based weak loss, combine repetitions and enumerate the
resulting data as
$(q_a,\mathbf x_a,\alpha_a)\in[d_2]\times\overline\Omega\times
\mathbb N_0^{d_0}$, $a\in[A]$. Define
\begin{equation}
 \mathcal Q_a(f)=\partial^{\alpha_a}f_{q_a}(\mathbf x_a),
 \qquad
 \mathbf B=[B_{ia}]\in\mathbb R^{N\times A},\qquad
 \lambda_i=\sum_{a=1}^A B_{ia}\mathcal Q_a.
 \label{eqn:quadrature_derivative_functionals}
\end{equation}
If $\{\mathcal Q_a\}_{a=1}^A$ is linearly independent in $\mathcal X^*$,
then
\begin{equation}
 \begin{aligned}
 \sum_{i=1}^N c_i\lambda_i=0\quad(\mathbf c\in\mathbb R^N)
 &\quad\Longleftrightarrow\quad \mathbf c^\top\mathbf B=0,\\
 \{\lambda_i\}_{i=1}^N\text{ is linearly independent}
 &\quad\Longleftrightarrow\quad \rank\mathbf B=N.
 \end{aligned}
 \label{eqn:quadrature_coefficient_rank}
\end{equation}
The stated equivalence relies on this property in $\mathcal X^*$; distinct
triples alone need not guarantee it for an arbitrary function space. For example, an architecture that
enforces exact boundary conditions can create linear relations among the
restrictions of the $\mathcal Q_a$ to the homogeneous-boundary tangent
subspace $\mathcal X_0$ of~\Cref{rem:derivative_hard_boundary}.
A symbolic nonzero $N\times N$ minor, or a rigorous interval bound containing
its determinant that excludes zero, proves full row rank; a floating-point
singular value is only a numerical diagnostic.
Rescaled orthogonal polynomials are preferable to raw monomials in computations.
Neither~\cref{eqn:monomial_rank_equivalence}
nor~\cref{eqn:quadrature_coefficient_rank} provides an a priori truncation degree or a mesh-uniform positive lower bound
on the smallest eigenvalue, and
$\rank\mathbf B=N$ concerns the discretized functionals rather than the
continuum weak problem. A quantitative bound instead needs the finite-witness
data of~\Cref{prop:quantitative_dntk_bound}: pairwise disjoint sets of
$\mu_h$-measure at least $\beta>0$ on which the output feature matrix varies
by at most $\varepsilon$, and $c_{\mathrm{sep}}>0$. The resulting
estimate~\cref{eqn:quantitative_dntk_bound} depends on the design and is not
mesh-uniform either.
\end{remark}

\subsection{Weak residuals and boundary constraints}

Standard VPINN and hp-VPINN formulations enforce finitely many fixed test
functionals~\citep{Kharazmi2021}, whereas weak adversarial networks learn the
tests during optimization~\citep{Zang2020}. The theorem below concerns a fixed
finite family. The output-weight block is a Gram matrix of functional ridge
features; analyticity and density convert its kernel into exact relations
among the observations. This viewpoint parallels generalized kernel collocation and bounded
operator observations in physics-informed Gaussian
processes~\citep{Schaback2006,Pfortner2022}. Recent variational-NTK analysis
for coupled parabolic systems instead studies screening and conditioning for a
specific matrix-valued weak formulation~\citep{Kouma2026}.

\begin{theorem}[NTK existence and exact nullspace for weak-form constraints]
\label{thm:weak_constraint_positivity}
Let \(\mathcal X\) be a Banach space over $\mathbb R$ of
\(\mathbb R^{d_2}\)-valued functions on \(\overline\Omega\), continuously
embedded in \(L^1_{\mathrm{loc}}\) as in clause \(\mathsf W_{\mathrm{reg}}\). For
\(i\in[N]\), let \(\mathcal V_i\) be normed over $\mathbb R$, let
\(a_i:\mathcal X\times\mathcal V_i\to\mathbb R\) be a continuous bilinear
form, fix \(v_i\in\mathcal V_i\), and define the weak prediction functional
and its loss-scaled version by
\begin{equation*}
\ell_i(f)=a_i(f,v_i),
\qquad
\lambda_i=s_i\ell_i\in\mathcal X^*,
\qquad s_i>0.
\end{equation*}
Use \(P_i(\theta)=\lambda_i(\mathbf u(\cdot;\theta))\) with the network
in~\Cref{def:ntk_parameterization}. Assume~\Cref{assump:main} and the weak
clauses of~\Cref{assump:pde_classes} for this $\mathcal X$, the combined
family $\{\lambda_i\}_{i=1}^N$, and the ridge features
$\varphi_{\boldsymbol\xi,q}$ of~\Cref{def:ntk_parameterization}.
Define the space of exact relations among the weak constraints by
\begin{equation}
\mathcal N_{\mathrm{weak}}
=\left\{\mathbf c\in\mathbb R^N:
\sum_{i=1}^N c_i\lambda_i=0\ \text{in }\mathcal X^*\right\}.
\label{eqn:weak_functional_nullspace}
\end{equation}
Then \(\mathbf K_{(0)}^\infty\) exists and
\(\mathbf K_{(0)}\to\mathbf K_{(0)}^\infty\) almost surely along nested widths
(in probability under fresh resampling). Moreover,
\begin{equation}
\ker\mathbf K_{\mathrm{out}}^\infty
=\ker\mathbf K_{(0)}^\infty
=\mathcal N_{\mathrm{weak}},
\qquad
\rank\mathbf K_{(0)}^\infty
=\dim\linspan\{\lambda_1,\ldots,\lambda_N\}.
\label{eqn:weak_exact_ntk_nullspace}
\end{equation}
Consequently, this NTK is positive definite if and only if the
loss-scaled weak functionals are linearly independent in \(\mathcal X^*\).
In particular, suppose \(\mathcal V_i=\mathcal X\), all \(a_i=a\), and
\(a:\mathcal X\times\mathcal X\to\mathbb R\) is coercive:
\(a(v,v)\geq c_a\|v\|_{\mathcal X}^2\) for some \(c_a>0\).
Then linearly independent test functions \(v_1,\ldots,v_N\) imply
\(\mathbf K_{(0)}^\infty\succ0\).
\end{theorem}

\begin{corollary}[Joint weak-residual and boundary criterion]
\label{cor:weak_boundary_joint_positivity}
Assume~\Cref{assump:main} and the weak clauses of
\Cref{assump:pde_classes} for the combined prediction vector. Let
\(\lambda_i^\Omega\in\mathcal X^*\), \(i\in[N_\Omega]\), be loss-scaled weak
PDE-residual functionals, and let
\(\beta_j^\partial\in\mathcal X^*\), \(j\in[N_\partial]\), be loss-scaled
boundary functionals on the same trial space. Set $N=N_\Omega+N_\partial$.
A boundary functional may be a
tested boundary-restriction (trace) functional or a strong-form pointwise
boundary observation, provided that it is bounded on \(\mathcal X\). Order
the residual rows before the others. Use the combined predictions
\(P_i^\Omega(\theta)=\lambda_i^\Omega(\mathbf u(\cdot;\theta))\) and
\(P_j^\partial(\theta)=\beta_j^\partial(\mathbf u(\cdot;\theta))\), and define
\begin{equation}
\mathcal N_{\Omega,\partial}^{\mathrm{weak}}
=\left\{(\mathbf c,\mathbf d)\in
\mathbb R^{N_\Omega}\times\mathbb R^{N_\partial}:
\sum_{i=1}^{N_\Omega}c_i\lambda_i^\Omega
+\sum_{j=1}^{N_\partial}d_j\beta_j^\partial=0
\ \text{in }\mathcal X^*\right\}.
\label{eqn:weak_boundary_joint_nullspace}
\end{equation}
We identify $(\mathbf c,\mathbf d)$ with its concatenation in
$\mathbb R^N$ when comparing the set with a kernel.
Then the mixed output-weight and full limiting NTKs obey
\begin{equation*}
\ker\mathbf K_{\mathrm{out}}^\infty
=\ker\mathbf K_{(0)}^\infty
=\mathcal N_{\Omega,\partial}^{\mathrm{weak}},
\qquad
\rank\mathbf K_{(0)}^\infty
=\dim\linspan\!\left(
\{\lambda_i^\Omega\}_{i=1}^{N_\Omega}
\cup\{\beta_j^\partial\}_{j=1}^{N_\partial}\right).
\end{equation*}
Consequently, this matrix is positive definite if and only if the
residual and boundary functionals are \emph{jointly} linearly independent.
Checking the two families individually is not sufficient in general.
\end{corollary}
\begin{proof}
Apply~\Cref{thm:weak_constraint_positivity} to the concatenated family
\(\{\lambda_i^\Omega\}_i\cup\{\beta_j^\partial\}_j\). Any functional
\(\beta\in\mathcal X^*\) not already written as \(s\,a(\cdot,v)\), in
particular each boundary functional, fits the theorem's bilinear notation by
taking \(\mathcal V=\mathbb R\), \(a(f,t)=t\beta(f)\), \(v=1\), and loss
scale one. The
nullspace in~\cref{eqn:weak_functional_nullspace} is then exactly the set
in~\cref{eqn:weak_boundary_joint_nullspace}.
\end{proof}

For finite-width optimization, clause $\mathsf W_{\mathrm{reg}}$ of
\Cref{assump:pde_classes} and
\Cref{lem:bounded_functional_row_regularity} supply the bounded and
Lipschitz-Jacobian constants required by
\Cref{thm:linearPDEsconvergence}. In the standard bounded-domain
$H^1$ setting, \Cref{cor:weak_H1_row_regularity} verifies this premise directly
for $\sigma\in C_{\mathrm{loc}}^3$, and \Cref{cor:weak_Hk_row_regularity}
(for $\sigma\in C_{\mathrm{loc}}^{k+2}$) together with
\Cref{lem:ridge_density_sobolev} (for real analytic nonpolynomial $\sigma$
and $\gamma>0$) verifies the regularity and density clauses on
$H^k(\Omega;\mathbb R^{d_2})$ over bounded Lipschitz domains in every
dimension. Energy spaces outside this Sobolev family still require the
trial-space regularity, analyticity, and density checks in the weak clauses
of~\Cref{assump:pde_classes}; the arguments
of~\Cref{cor:weak_Hk_row_regularity,lem:ridge_density_sobolev} provide a
reusable proof template, but no blanket claim for arbitrary energy spaces is
made here.

\begin{corollary}[Continuum weak Poisson problem with a discontinuous coefficient]
\label{cor:continuum_weak_interface_poisson}
Let $\Omega=(-1,1)$, let $\kappa\in L^\infty(\Omega)$ satisfy
$0<\kappa_-\leq\kappa(x)\leq\kappa_+<\infty$ almost everywhere, and take
the common scalar trial space $\mathcal X=H^1(\Omega)$. Fix linearly
independent tests $v_1,\ldots,v_{N_\Omega}\in H_0^1(\Omega)$ and positive
row scales $s_i,t_-,t_+$. Define the continuum weak-residual and soft
Dirichlet prediction functionals by
\begin{equation}
 \lambda_i^\Omega(f)
 =s_i\int_{-1}^{1}\kappa(x)f'(x)v_i'(x)\,dx,
 \qquad
 \beta_-^\partial(f)=t_-f(-1),
 \qquad
 \beta_+^\partial(f)=t_+f(1).
 \label{eqn:continuum_weak_interface_rows}
\end{equation}
Use the biased scalar two-layer network in~\Cref{def:ntk_parameterization}
with $\sigma=\tanh$ and $\gamma>0$, and assume~\Cref{assump:main} with
$\operatorname{int}(\supp\mu_h)\neq\varnothing$. Then the full
limiting NTK for the joint family
$\{\lambda_i^\Omega\}_{i=1}^{N_\Omega}\cup
\{\beta_-^\partial,\beta_+^\partial\}$ exists and is positive definite; the
independence proof below supplies its positive-definiteness certificate, and
the monomial procedure in~\Cref{sec:weak_positivity_certificate} gives an
alternative finite certificate for any given test family.
For every fixed localization radius $R>0$ the induced per-neuron prediction
maps satisfy $(\mathsf G_R)$ by~\Cref{cor:weak_H1_row_regularity}. Hence,
under the centering condition
$\mathbb E[a_m\mid w_m,b_m]=0$, for any fixed targets and
$0<\delta<1$, the explicit width and step-size conditions
in~\cref{eqn:linear_pde_convergence_conditions} imply, with probability at
least $1-\delta$, the all-iteration localization, kernel-gap preservation,
and geometric empirical-loss decay in~\cref{eqn:linear_convergence_main}.
In particular, the conclusion applies to the discontinuous
coefficient $\kappa=\mathbf1_{(-1,0)}+2\mathbf1_{[0,1)}$. For the equation
$-(\kappa u')'=g$ with $g\in L^2(\Omega)$ and prescribed endpoint values
$u(-1)=g_-$ and $u(1)=g_+$, the loss-scaled targets paired with
the rows in~\cref{eqn:continuum_weak_interface_rows} are
$s_i\int_{-1}^{1}g(x)v_i(x)\,dx$, $t_-g_-$, and $t_+g_+$, respectively.
\end{corollary}
\begin{proof}
The bilinear form
\begin{equation*}
 a(f,v)=\int_{-1}^{1}\kappa(x)f'(x)v'(x)\,dx,
 \qquad (f,v)\in H^1(\Omega)\times H_0^1(\Omega),
\end{equation*}
is continuous by Cauchy--Schwarz, and endpoint evaluation is bounded on
$H^1(\Omega)$ by the one-dimensional trace theorem. Thus all rows
in~\cref{eqn:continuum_weak_interface_rows} belong to $\mathcal X^*$.

For each $(w,b)\in\mathbb R^2$, the ridge $x\mapsto\tanh(wx+\gamma b)$
belongs to $H^1(\Omega)$. Moreover,
\begin{equation}
 (w,b)\longmapsto\tanh(w\,\cdot+\gamma b)
 \label{eqn:tanh_H1_analytic_map}
\end{equation}
is real analytic as an $H^1(\Omega)$-valued map. Indeed, at any real
$(w_0,b_0)$, the holomorphic extension of $\tanh$ has no pole on the compact
segment $w_0[-1,1]+\gamma b_0$; a sufficiently small complex neighborhood
of $(w_0,b_0)$ therefore gives a holomorphic map into
$C^1([-1,1];\mathbb C)$, and the embedding $C^1([-1,1])\hookrightarrow
H^1(\Omega)$ is continuous.

The span of the ridge functions $\{\tanh(w\,\cdot+\gamma b):(w,b)\in\mathbb R^2\}$
is dense in $H^1(\Omega)$. To see this without invoking a
Sobolev universal-approximation theorem, fix $n\in\mathbb N_0$. Since
$\tanh$ is analytic and nonpolynomial, $\tanh^{(n)}$ is not identically
zero; choose $b_n$ with $\tanh^{(n)}(\gamma b_n)\neq0$. Banach-valued
differentiation of~\cref{eqn:tanh_H1_analytic_map} gives
\begin{equation*}
 \left.\partial_w^n\tanh(wx+\gamma b_n)\right|_{w=0}
 =\tanh^{(n)}(\gamma b_n)x^n.
\end{equation*}
This derivative is an $H^1$ limit of finite linear combinations of ridge
features, so every monomial, and hence every polynomial, lies in the closure
of their span. Polynomials are dense in $H^1(-1,1)$: approximate the weak
derivative of an $H^1$ function in $L^2$ by a polynomial and integrate that
polynomial, matching one endpoint value. Thus the Banach-valued analyticity
and ridge-density requirements in clause $\mathsf W_{\mathrm{pd}}$ of
\Cref{assump:pde_classes} hold; its hidden-support requirement is assumed in
the statement.
For any fixed $R$, the closed radius-$R$ neighborhood of the compact
initialization support is compact. The complete one-neuron map
$(w,b,a)\mapsto a\tanh(w\,\cdot+\gamma b)$ is $C^\infty$ into
$H^1(\Omega)$, so boundedness of the finitely many rows makes every induced
scalar prediction map $C^{1,1}$ there. Hence clause
$\mathsf W_{\mathrm{reg}}$ also holds, and
\Cref{prop:functional_ntk_rank} gives existence of the population NTK. Once
joint independence is established below, conditional centering and
\Cref{cor:functional_gd_convergence} give the stated GD conclusion.

It remains to verify joint independence. Suppose, for every $f\in H^1(\Omega)$,
\begin{equation*}
 \sum_{i=1}^{N_\Omega}c_i\lambda_i^\Omega(f)
 +d_-\beta_-^\partial(f)+d_+\beta_+^\partial(f)=0,
\end{equation*}
and set $v=\sum_i c_i s_i v_i\in H_0^1(\Omega)$. Restricting to
$f\in H_0^1(\Omega)$ and then choosing $f=v$ yields
$\int_{-1}^{1}\kappa|v'|^2=0$. Coercivity and the zero trace imply $v=0$;
linear independence of the $v_i$ and positivity of the $s_i$ give $c_i=0$
for every $i$. The remaining relation is
$d_-t_-f(-1)+d_+t_+f(1)=0$. Testing with the two affine endpoint cardinal
functions gives $d_-=d_+=0$. The joint criterion
in~\Cref{cor:weak_boundary_joint_positivity} now supplies the claimed
positive-definiteness certificate.
\end{proof}

The same route runs in higher dimension, where boundary data enter either as
tested traces on $H^1$ or pointwise on a smoother trial space.

\begin{corollary}[Two-dimensional weak Poisson problem with tested or pointwise boundary data]
\label{cor:weak_poisson_2d}
Let $\Omega=(-1,1)^2$, let $\kappa\in L^\infty(\Omega)$ satisfy
$0<\kappa_-\leq\kappa\leq\kappa_+<\infty$ almost everywhere, and take the
scalar trial space $\mathcal X=H^1(\Omega)$. Fix linearly independent
polynomial tests $v_1,\ldots,v_{N_\Omega}\in H_0^1(\Omega)$, for instance
the products $(1-x_1^2)(1-x_2^2)\mathbf x^\alpha$ with distinct
$\alpha\in\mathbb N_0^2$,
linearly independent boundary test
functions $\psi_1,\ldots,\psi_{N_\partial}\in L^2(\partial\Omega)$, and
positive row scales $s_i,t_j$. Define
\begin{equation*}
 \lambda_i^\Omega(f)=s_i\int_\Omega\kappa\,\nabla f\cdot\nabla v_i\,d\mathbf x,
 \qquad
 \beta_j^\partial(f)=t_j\int_{\partial\Omega}(\operatorname{Tr}f)\,\psi_j\,dS.
\end{equation*}
Use the biased scalar two-layer network of~\Cref{def:ntk_parameterization}
with a real analytic nonpolynomial $\sigma$ and $\gamma>0$, and
assume~\Cref{assump:main} with $\operatorname{int}(\supp\mu_h)\neq\varnothing$. Then the full limiting NTK of the joint family
$\{\lambda_i^\Omega\}_{i=1}^{N_\Omega}\cup\{\beta_j^\partial\}_{j=1}^{N_\partial}$
exists and is positive definite, and for every $R>0$ the rows satisfy
$(\mathsf G_R)$; if moreover
$\mathbb E[\mathbf a\mid\boldsymbol\xi]=\mathbf0$, then
\Cref{thm:linearPDEsconvergence} applies with
$\lambda_0=\lambda_{\min}(\mathbf K_{(0)}^\infty)$ and its width and
step-size conditions. The same conclusions hold with the boundary rows replaced by
pointwise Dirichlet rows $\beta_j^\partial(f)=t_jf(\mathbf x_j)$ at distinct
points $\mathbf x_j\in\partial\Omega$, provided the trial space is
$\mathcal X=H^2(\Omega)$.
\end{corollary}
\begin{proof}
$\Omega$ is a bounded Lipschitz domain. The bilinear form
$a(f,v)=\int_\Omega\kappa\nabla f\cdot\nabla v$ is continuous on
$H^1(\Omega)\times H_0^1(\Omega)$ by Cauchy--Schwarz, and the trace
$\operatorname{Tr}:H^1(\Omega)\to L^2(\partial\Omega)$ is bounded, so every
row lies in $\mathcal X^*$. \Cref{lem:ridge_density_sobolev} with $k=1$
gives the analyticity and density clauses of $\mathsf W_{\mathrm{pd}}$, and
\Cref{cor:weak_Hk_row_regularity} with $k=1$ gives the $C^2$ regularity of
$\mathsf W_{\mathrm{reg}}$ and hence $(\mathsf G_R)$; the interior-support
clause is assumed. For joint independence, suppose
$\sum_ic_i\lambda_i^\Omega+\sum_jd_j\beta_j^\partial=0$ on $\mathcal X$.
Every $f\in H_0^1(\Omega)$ has zero trace, so
$\sum_ic_is_i\,a(f,v_i)=0$ for all $f\in H_0^1(\Omega)$; taking
$f=v:=\sum_ic_is_iv_i\in H_0^1(\Omega)$ gives $a(v,v)=0$, and coercivity of
$a$ on $H_0^1(\Omega)$ (from $\kappa\geq\kappa_-$ and the Poincar\'e
inequality) forces $v=0$, hence $\mathbf c=\mathbf0$ because the $v_i$ are
independent and $s_i>0$. Then $\sum_jd_jt_j\int_{\partial\Omega}
(\operatorname{Tr}f)\psi_j\,dS=0$ for all $f\in H^1(\Omega)$; since the
trace map has range $H^{1/2}(\partial\Omega)$, which is dense in
$L^2(\partial\Omega)$, the function $\sum_jd_jt_j\psi_j$ is orthogonal to a
dense subspace and vanishes in $L^2(\partial\Omega)$; independence of the
$\psi_j$ and $t_j>0$ give $\mathbf d=\mathbf0$. Thus the set
$\mathcal N_{\Omega,\partial}^{\mathrm{weak}}$
in~\cref{eqn:weak_boundary_joint_nullspace} is trivial,
\Cref{cor:weak_boundary_joint_positivity} gives
$\mathbf K_{(0)}^\infty\succ0$, and \Cref{cor:classwise_gd_convergence}
applies. For pointwise boundary rows on $H^2(\Omega)$:
\Cref{cor:weak_Hk_row_regularity} with $k=2$ (admissible since a real
analytic $\sigma$ lies in $C^4_{\mathrm{loc}}$) gives the $C^2$ regularity of
$\mathsf W_{\mathrm{reg}}$ into $H^2(\Omega)$, hence $(\mathsf G_R)$, and,
since $k=2>d_0/2$ on the Lipschitz domain $\Omega$, bounded point evaluation;
\Cref{lem:ridge_density_sobolev} with $k=2$ gives the analyticity and density
clauses of $\mathsf W_{\mathrm{pd}}$ in $H^2(\Omega)$; and the residual rows
remain bounded on $H^2(\Omega)\hookrightarrow H^1(\Omega)$.
In the independence argument, test first on $f\in H^2(\Omega)\cap H_0^1(\Omega)$,
which contains the polynomial tests $v_i$ and on which every pointwise
boundary row vanishes, since such $f$ is continuous on $\overline\Omega$ and
its trace is its restriction to $\partial\Omega$; this gives
$\mathbf c=\mathbf0$ as before. Then $\sum_jd_jt_jf(\mathbf x_j)=0$ for all
$f\in H^2(\Omega)$, and
a polynomial taking prescribed values at the distinct points $\mathbf x_j$
gives $\mathbf d=\mathbf0$.
\end{proof}

For the nonlocal class, under $\mathsf N_{\mathrm{pd}}$, a two-dimensional
instance is immediate: averages
over pairwise disjoint closed subsquares of $\overline\Omega$ together with
Dirac data at distinct points outside those subsquares are measure tuples
with pairwise disjoint supports, hence linearly independent, and
\Cref{prop:nonlocal_ntk_rank} applies on $C(\overline\Omega)$; Dirac points
inside a subsquare are covered instead by the mutual-singularity argument in
the proof of \Cref{cor:nonlocal_boundary_joint_positivity}.

The boundary functionals $\beta_j^\partial$ admitted by
\Cref{cor:weak_boundary_joint_positivity} are arbitrary bounded functionals
on $\mathcal X$; the two forms used in this paper are the following. For a scale
$t_j>0$, a specified boundary operator $\mathcal B$,
and a fixed boundary test function $\psi_j$ such that
$f\mapsto\langle\mathcal Bf,\psi_j\rangle_{\partial\Omega}$ is bounded on
$\mathcal X$, a tested weak boundary functional has the form
\(\beta_j^\partial(f)=t_j\langle\mathcal Bf,\psi_j\rangle_{\partial\Omega}\).
A strong-form pointwise row may instead be
\begin{equation*}
\beta_j^\partial(f)=t_j\sum_{q=1}^{d_2}
\sum_{\alpha\in J_{jq}}b_{jq\alpha}
\partial^\alpha f_q(\mathbf x_j),
\qquad \mathbf x_j\in\partial\Omega.
\end{equation*}
Here every $J_{jq}\subset\mathbb N_0^{d_0}$ is finite and
$b_{jq\alpha}\in\mathbb R$.
The criterion covers both rows exactly when they are bounded on the common
space \(\mathcal X\). The independence to be checked is that of the induced
functionals in $\mathcal X^*$, not merely of the written test functions $v_i$
and $\psi_j$. For a quadrature-based variational loss,
write its distinct derivative evaluations and coefficient matrix
as in~\cref{eqn:quadrature_derivative_functionals}. If the retained
$\mathcal Q_a$ are linearly independent in $\mathcal X^*$,
then~\cref{eqn:quadrature_coefficient_rank} shows that full row rank of
$\mathbf B$ is equivalent to independence of the discretized rows. A separate
consistency argument is still needed to relate that loss to the continuum
weak problem.

\subsection{Nonlocal integral constraints and boundary measures}
\label{sec:nonlocal_boundary_measures}

Fractional and nonlocal PINN formulations supply prominent examples of
integral or integro-differential residuals~\citep{Pang2019,Pang2020}. For the
bounded integral observations treated here, uniform universal approximation
turns the functional criterion into an exact statement about the defining
measures. The resulting signed-measure independence criterion is related to
kernel mean-embedding characterizations~\citep{Sriperumbudur2011}, but it is
applied here to the output-weight block of the limiting tangent kernel at
initialization, whose kernel function is
$k(\mathbf x,\mathbf y)=\mathbb E_{\boldsymbol\xi\sim\mu_h}
[\sigma(\mathbf w^\top\mathbf x+\gamma b)\,\sigma(\mathbf w^\top\mathbf y+\gamma b)]$.
The scope is specifically finite signed regular Borel measures acting on
$C(D)$; the theorem below does not, by itself, cover principal-value or
hypersingular kernels, distributional derivatives, or continuum Caputo,
Riemann--Liouville, or Riesz fractional derivatives. A quadrature formula for
one of those operators that uses only point values of the unknown, such as a
Gr\"unwald--Letnikov or L1-type rule, produces a finite signed atomic measure
and is therefore covered as a \emph{discrete} constraint; a rule that samples
derivatives of the unknown is instead a weak-form row on a trial space on
which those evaluations are bounded. That conclusion supplies neither
quadrature consistency nor a continuum fractional-operator result. A
continuum operator outside the finite-measure class must instead be placed in
a Banach trial space on which its tested rows are bounded. There the $C^2$
one-neuron condition of clause $\mathsf W_{\mathrm{reg}}$ and, for the exact
nullspace, the interior-support, analyticity, and density clauses of
$\mathsf W_{\mathrm{pd}}$ together with joint independence of the rows in
$\mathcal X^*$ (or another positive-definiteness certificate) must be
verified; the $C^2$ condition then yields the optimization row regularity
$(\mathsf G_R)$, with explicit constants,
through~\Cref{lem:bounded_functional_row_regularity}.

\begin{theorem}[NTK existence and exact nullspace for nonlocal integral constraints]
\label{thm:nonlocal_constraint_positivity}
Let \(D\subseteq\overline\Omega\subset\mathbb R^{d_0}\) be compact and set
$\mathcal X=C(D;\mathbb R^{d_2})$. For every
\(i\in[N]\) and \(q\in[d_2]\), let \(\nu_{iq}\) be a finite signed regular
Borel measure on \(D\), and define
\begin{equation*}
\ell_i(f)=\sum_{q=1}^{d_2}\int_D f_q(\mathbf y)\,d\nu_{iq}(\mathbf y),
\qquad
\lambda_i=s_i\ell_i\in\mathcal X^*,\qquad
P_i(\theta)=\lambda_i(\mathbf u(\cdot;\theta)),
\qquad s_i>0.
\end{equation*}
Assume~\Cref{assump:main} and the nonlocal clauses of
\Cref{assump:pde_classes}. Set
\begin{equation*}
\mathcal N_{\mathrm{nl}}
=\left\{\mathbf c\in\mathbb R^N:
\sum_{i=1}^N c_i s_i\nu_{iq}=0
\ \text{as a signed measure for every }q\in[d_2]\right\}.
\end{equation*}
Then \(\mathbf K_{(0)}^\infty\) exists and
\(\mathbf K_{(0)}\to\mathbf K_{(0)}^\infty\) almost surely along nested widths
(in probability under fresh resampling). Moreover,
\begin{equation}
\ker\mathbf K_{\mathrm{out}}^\infty
=\ker\mathbf K_{(0)}^\infty
=\mathcal N_{\mathrm{nl}},
\qquad
\rank\mathbf K_{(0)}^\infty
=\dim\linspan
\left\{\bar{\boldsymbol\nu}_i:i\in[N]\right\}.
\label{eqn:nonlocal_exact_ntk_nullspace}
\end{equation}
Consequently, \(\mathbf K_{(0)}^\infty\succ0\) if and only if the loss-scaled
tuples \(\bar{\boldsymbol\nu}_i=s_i(\nu_{i1},\ldots,\nu_{id_2})\),
\(i\in[N]\), are linearly independent in
\(\mathcal M(D)^{d_2}\), where \(\mathcal M(D)=C(D)^*\) is the space of finite
signed regular Borel measures on \(D\).
\end{theorem}

\begin{corollary}[Nonlocal residuals with nonlocal or Dirichlet boundary data]
\label{cor:nonlocal_boundary_joint_positivity}
Assume~\Cref{assump:main} and the nonlocal clauses of
\Cref{assump:pde_classes} for the combined prediction map.
For \(i\in[N_\Omega]\), let
\(\bar{\boldsymbol\nu}_i=s_i(\nu_{i1},\ldots,\nu_{id_2})\) define a nonlocal
PDE-residual row. For \(j\in[N_\partial]\), let
\(\bar{\boldsymbol\omega}_j=t_j(\omega_{j1},\ldots,\omega_{jd_2})\), \(t_j>0\), be a
vector of finite signed regular Borel measures on \(D\), typically supported
on \(D\cap\partial\Omega\), defining another constraint. Set
$N=N_\Omega+N_\partial$ and order the residual rows before the boundary rows.
The corresponding predictions are
\[
P_i^\Omega(\theta)=s_i\sum_{q=1}^{d_2}\int_Du_q(\mathbf y;\theta)\,d\nu_{iq}(\mathbf y),
\qquad
P_j^\partial(\theta)=t_j\sum_{q=1}^{d_2}\int_Du_q(\mathbf y;\theta)\,d\omega_{jq}(\mathbf y).
\]
Set
\begin{equation*}
\begin{aligned}
\mathcal N_{\Omega,\partial}^{\mathrm{nl}}
=\left\{(\mathbf c,\mathbf d)\right.&\in
\mathbb R^{N_\Omega}\times\mathbb R^{N_\partial}:\\
&\sum_{i=1}^{N_\Omega}c_i s_i\nu_{iq}
+\sum_{j=1}^{N_\partial}d_j t_j\omega_{jq}=0\\[-0.2em]
&\left.\text{as a signed measure for every }q\in[d_2]\right\}.
\end{aligned}
\end{equation*}
We identify $(\mathbf c,\mathbf d)$ with its concatenation in
$\mathbb R^N$.
Then
\begin{equation*}
\ker\mathbf K_{\mathrm{out}}^\infty
=\ker\mathbf K_{(0)}^\infty
=\mathcal N_{\Omega,\partial}^{\mathrm{nl}},
\qquad
\rank\mathbf K_{(0)}^\infty
=\dim\linspan\!\left(
\{\bar{\boldsymbol\nu}_i\}_{i=1}^{N_\Omega}
\cup\{\bar{\boldsymbol\omega}_j\}_{j=1}^{N_\partial}\right).
\end{equation*}
Thus the mixed limiting NTK is positive definite exactly when the combined
family of residual and boundary measure tuples is linearly independent.
Taking \(\omega_{jq}=b_{jq}\delta_{\mathbf x_j}\) with \(\mathbf x_j\in D\) and
\(b_{jq}\in\mathbb R\) gives the order-zero pointwise row
\(t_j\sum_{q=1}^{d_2}b_{jq}f_q(\mathbf x_j)\). When
\(\mathbf x_j\in D\cap\partial\Omega\), \(b_{jq_0}=1\), and \(b_{jq}=0\) for
\(q\neq q_0\), this is a Dirichlet row for component \(q_0\); when
\(\mathbf x_j\in D\cap\Omega\), the same functional is interior point data.

Suppose additionally that a positive regular Borel measure \(\rho\) on \(D\)
satisfies \(\nu_{iq}\ll\rho\) (absolute continuity with respect to \(\rho\))
and \(\omega_{jq}\perp\rho\) (mutual singularity with \(\rho\)) for every
index.
Then the combined family is linearly independent if and only if the residual
family \(\{\bar{\boldsymbol\nu}_i\}_i\) and boundary family
\(\{\bar{\boldsymbol\omega}_j\}_j\) are each linearly independent. Hence separate
checks suffice under this mutual-singularity condition.
\end{corollary}
\begin{proof}
Apply~\Cref{thm:nonlocal_constraint_positivity} to the concatenated collection of
measure tuples. For the final assertion, in each output coordinate \(q\) the
residual part \(\zeta_q^\Omega=\sum_ic_is_i\nu_{iq}\) of a null combination is
absolutely continuous with respect to \(\rho\) and the boundary part
\(\zeta_q^\partial=\sum_jd_jt_j\omega_{jq}\) is singular with respect to
\(\rho\). Since \(\zeta_q^\Omega=-\zeta_q^\partial\) is both, it vanishes: a
Borel set \(E\) with \(\rho(E)=0\) carries all of \(|\zeta_q^\partial|\), and
absolute continuity gives \(|\zeta_q^\Omega|(E)=0\). Hence both parts vanish
separately, after which within-family independence makes every coefficient
zero. The converse is immediate.
\end{proof}

Under~\Cref{assump:main} and the nonlocal clauses of
\Cref{assump:pde_classes},
\Cref{prop:finite_measure_row_regularity} supplies explicit
total-variation-based constants for every row in
\Cref{cor:nonlocal_boundary_joint_positivity}. Hence joint measure-tuple
independence and conditional output-weight centering imply the finite-width GD
conclusion in~\Cref{cor:nonlocal_gd_convergence}. This implication is
qualitative until a numerical lower bound on
$\lambda_{\min}(\mathbf K_{(0)}^\infty)$ is also certified.

Let $\rho_i$ be a positive regular Borel measure on $D$, not necessarily the
reference measure of~\Cref{cor:nonlocal_boundary_joint_positivity}, and fix a
site $\mathbf x_i\in D$. Define an integral-operator observation by
\begin{equation*}
\ell_i(f)=\sum_{q=1}^{d_2}\int_D
\mathfrak k_{iq}(\mathbf x_i,\mathbf y)f_q(\mathbf y)\,d\rho_i(\mathbf y).
\end{equation*}
The criterion applies whenever each kernel section
$\mathbf y\mapsto\mathfrak k_{iq}(\mathbf x_i,\mathbf y)$ is
\(\rho_i\)-integrable, by taking
\(d\nu_{iq}(\mathbf y)=\mathfrak k_{iq}(\mathbf x_i,\mathbf y)\,d\rho_i(\mathbf y)\).
A nonlocal-diffusion row with a bounded kernel and \(\rho_i(D)<\infty\), in the
scalar case
\(\ell_i(f)=\int_D\mathfrak k_i(\mathbf x_i,\mathbf y)\,
(f(\mathbf x_i)-f(\mathbf y))\,d\rho_i(\mathbf y)\), is the finite signed
measure
\(\nu_i=\bigl(\int_D\mathfrak k_i(\mathbf x_i,\cdot)\,d\rho_i\bigr)
\delta_{\mathbf x_i}-\mathfrak k_i(\mathbf x_i,\cdot)\,\rho_i\) and is covered
in the same way. Its mass at \(\mathbf x_i\) is
\(m_i=\int_{D\setminus\{\mathbf x_i\}}\mathfrak k_i(\mathbf x_i,\cdot)\,d\rho_i\);
when \(m_i\neq0\), \(\nu_i\ll\rho\) forces \(\rho(\{\mathbf x_i\})>0\), so a
reference measure without an atom at \(\mathbf x_i\) cannot be used. The
separate checks remain available with
\(\rho=\sum_i(\rho_i+\delta_{\mathbf x_i})\) provided every boundary measure
is singular with respect to this \(\rho\), for example Dirac boundary data at
points that carry no mass of any \(\rho_i\) and are not residual sites;
otherwise the joint check
of~\Cref{cor:nonlocal_boundary_joint_positivity} is required.

\begin{remark}[Derivative and hard boundary conditions]
\label{rem:derivative_hard_boundary}
A Dirichlet evaluation is integration against a Dirac mass and therefore
falls under~\Cref{cor:nonlocal_boundary_joint_positivity}. In contrast, a
Neumann or Robin normal-derivative term is a distributional derivative, not a
finite measure acting on \(C(D)\). Combined with finite-measure residual
rows, such a row falls instead under~\Cref{cor:weak_boundary_joint_positivity}
on a common space \(\mathcal X\) of functions on \(\overline\Omega\), for instance
\(C^1(\overline\Omega;\mathbb R^{d_2})\) for bounded \(\Omega\), with measures on
\(D\subseteq\overline\Omega\) extended by zero, on
which both the measure rows and the boundary derivative are bounded; combined
with pointwise residual rows it is covered directly by the pointwise clauses
of~\Cref{assump:pde_classes}. The single-neuron
features must also belong to $\mathcal X$ and satisfy the stated Banach-valued
analyticity and density hypotheses. The weak formulation may
alternatively incorporate tested Neumann or Robin data.

If the architecture enforces an essential boundary condition \emph{hard}
rather than penalizing boundary-loss rows, one may write the constrained
model as $\mathbf g+\mathcal E\mathbf u(\cdot;\theta)$, where the network
$\mathbf u(\cdot;\theta)$ of~\Cref{def:ntk_parameterization} is regarded as an
element of a Banach space $\widetilde{\mathcal X}$, and the predictions are the
residual rows
$P_i^\Omega(\theta)=\lambda_i^\Omega\bigl(\mathbf g+\mathcal E\mathbf u(\cdot;\theta)\bigr)$,
$i\in[N_\Omega]$, with loss-scaled $\lambda_i^\Omega\in\mathcal X^*$. Here
$\mathbf g\in\mathcal X$ is a fixed lifting satisfying the inhomogeneous
data. A bounded linear operator
$\mathcal E:\widetilde{\mathcal X}\to\mathcal X_0$ maps into a closed
homogeneous-boundary tangent subspace $\mathcal X_0\subseteq\mathcal X$ with
the inherited norm. No additional penalty rows are present. The
tangent features become $\mathcal E\varphi_{\boldsymbol\xi,q}$, and the exact
criterion concerns the restrictions of the residual functionals to
$\mathcal X_0$:
\begin{equation*}
\ker\mathbf K_{\mathrm{out}}^\infty
=\ker\mathbf K_{(0)}^\infty
=\left\{\mathbf c\in\mathbb R^{N_\Omega}:
\left.\sum_{i=1}^{N_\Omega}c_i\lambda_i^\Omega\right|_{\mathcal X_0}=0
\ \text{in }\mathcal X_0^*\right\},
\end{equation*}
provided that \Cref{assump:main} holds with
$\operatorname{int}(\supp\mu_h)\neq\varnothing$, that the raw features
$\varphi_{\boldsymbol\xi,q}$ belong to $\widetilde{\mathcal X}$ and depend
real-analytically on their parameters in that space (with boundedness of
$\mathcal E$ this gives the $C^2$ regularity of clause
$\mathsf W_{\mathrm{reg}}$ for the adapted one-neuron map, hence
$(\mathsf G_R)$ by \Cref{lem:bounded_functional_row_regularity}), and that the
adapted family $\mathcal E\varphi_{\boldsymbol\xi,q}$ has dense span in
$\mathcal X_0$.
The fixed lifting changes the target but
not the NTK. The density hypothesis concerns the adapted family itself:
unconstrained single-neuron features generally do
not belong to a zero-boundary-value trial space, so one cannot simply reuse a
density result for the unconstrained features in a hard-boundary construction.
\end{remark}

%% file: positivity_checks.tex
\section{Alternative positivity checks and singular examples}

\subsection{Sharpness of the classwise hypotheses}
\label{sec:hypothesis_sharpness}

The clauses of~\Cref{assump:main,assump:pde_classes} are sufficient
conditions, and it is fair to ask whether they are decorative. The six
constructions below answer that question. Each deletes or weakens exactly one
clause, retains every remaining clause of the same assumption, and verifies
the retained clauses explicitly; each then exhibits a system whose rows are
linearly independent yet whose limiting NTK is singular, or, in the last case,
does not exist. None of them contradicts any result as stated: every
construction violates a hypothesis that the corresponding theorem assumes, so
that theorem simply does not apply to it. For each example we record which
half of the conclusion survives the deletion and which does not.

The first two examples work by shrinking the closed span of the features the
network can actually reach, so that a nonzero combination of independent rows
annihilates all of them.

\begin{example}[The ridge-density clause of $\mathsf W_{\mathrm{pd}}$ is load-bearing]
\label{ex:ridge_density_obstruction}
Delete from clause $\mathsf W_{\mathrm{pd}}$ of~\Cref{assump:pde_classes} only
the ridge-density requirement
\begin{equation*}
 \overline{\linspan\{\varphi_{\boldsymbol\xi,q}:
 \boldsymbol\xi\in\mathbb R^{d_0+1},\ q\in[d_2]\}}^{\mathcal X}
 =\mathcal X,
\end{equation*}
and keep every other hypothesis
of~\Cref{prop:functional_ntk_rank,thm:weak_constraint_positivity}. Let
$\Omega=(0,1)$, so $\overline\Omega=[0,1]$, let $d_0=d_2=1$, and let
$\mathcal X=C([0,1])$ with the supremum norm. Use the network
of~\Cref{def:ntk_parameterization} with the linear activation $\sigma(z)=z$,
bias scale $\gamma=1$, and i.i.d.\ per-neuron initialization
$\theta_m=(w_m,b_m,a_m)\sim\Unif([-1,1]^3)$. Take the $N=3$ unit-scaled
evaluation rows
\begin{equation*}
 \lambda_i(f)=f(x_i),
 \qquad
 (x_1,x_2,x_3)=(0,1,\tfrac12),
\end{equation*}
which lie in $\mathcal X^*$ and are therefore admissible
in~\Cref{thm:weak_constraint_positivity} through the reduction
$\mathcal V_i=\mathbb R$, $v_i=s_i=1$, $a_i(f,t)=t\lambda_i(f)$ used in the
proof of~\Cref{cor:weak_boundary_joint_positivity}. Positive loss scales $s_i$ would
only replace the null vector $\mathbf c$ below by $(c_i/s_i)_{i=1}^3$.

Every retained clause holds. The embedding
$\mathcal X\hookrightarrow L^1(0,1)$ is continuous, each
$\lambda_i\in\mathcal X^*$, and the one-neuron map
$\mathcal G(\boldsymbol\xi,\mathbf a)=a(w\,\cdot+\gamma b)$ is a quadratic
polynomial map $\mathbb R^3\to\mathcal X$, hence $C^\infty$ and real analytic
on all of $\mathbb R^3$; clause $\mathsf W_{\mathrm{reg}}$ and the
feature-analyticity part of $\mathsf W_{\mathrm{pd}}$ therefore hold. The law
$\mu$ is width independent and compactly supported, so~\Cref{assump:main}
holds; $\operatorname{int}(\supp\mu_h)=(-1,1)^2\neq\varnothing$, so the
support part of $\mathsf W_{\mathrm{pd}}$ holds; and
$\mathbb E[a_m\mid w_m,b_m]=0$, so the conditional output-weight centering
that~\Cref{assump:pde_classes} imposes for gradient descent holds as well.
Only density fails, and it fails by a definite margin: with $d_2=1$ the single
coordinate feature is $\varphi_{(w,b)}=w\,\cdot+\gamma b$, so
$\linspan\{\varphi_{\boldsymbol\xi}:\boldsymbol\xi\in\mathbb R^2\}
=\linspan\{1,x\}=:\mathcal P_1$, a two-dimensional and therefore closed proper
subspace of $\mathcal X$.

The three rows are linearly independent in $\mathcal X^*$: testing
$\sum_ic_i\lambda_i=0$ against continuous hat functions $f_j$ with $f_j(x_j)=1$
and $f_j(x_k)=0$ for $k\neq j$ gives $c_j=0$ for each $j$. Independence
notwithstanding, $\mathbf c=(1,1,-2)^\top$ and
$\lambda_{\mathbf c}=\lambda_1+\lambda_2-2\lambda_3$ satisfy
\begin{equation*}
 \lambda_{\mathbf c}(\alpha+\beta x)
 =\alpha+(\alpha+\beta)-2\left(\alpha+\tfrac{\beta}{2}\right)=0
 \qquad\text{for all }\alpha,\beta\in\mathbb R,
\end{equation*}
so the nonzero functional $\lambda_{\mathbf c}$ annihilates $\mathcal P_1$ and
hence every ridge feature. The same vector quantifies the failure of density:
$\lambda_{\mathbf c}(x^2)=\tfrac12$ and
$\|\lambda_{\mathbf c}\|_{\mathcal X^*}=4$ give
$\inf_{p\in\mathcal P_1}\|x^2-p\|_{\mathcal X}\geq\tfrac18$, which is exactly
the Chebyshev distance attained by $x-\tfrac18$.

Because $\mathbf u(\cdot;\theta)\in\mathcal P_1$ at every width and every
parameter value,
\begin{equation*}
 \sum_{i=1}^3c_iP_i(\theta)
 =\lambda_{\mathbf c}(\mathbf u(\cdot;\theta))=0,
 \qquad\text{hence}\qquad
 \mathbf J(\theta)^\top\mathbf c=\mathbf0,
 \quad
 \mathbf K(\theta)\mathbf c=\mathbf0 .
\end{equation*}
That is the one-line reason. Explicitly, with
$h_i(w,b,a)=a(wx_i+\gamma b)$ in~\cref{eqn:predictionmap}, the Gram
representation in~\Cref{thm:linearNTK} gives
\begin{equation*}
 [\mathbf K_{(0)}^\infty]_{ij}
 =\mathbb E\!\left[a^2\right]\!\left(x_ix_j+\gamma^2\right)
 +\mathbb E\!\left[(wx_i+\gamma b)(wx_j+\gamma b)\right]
 =\tfrac23\left(1+x_ix_j\right),
\end{equation*}
using $\mathbb E[a^2]=\mathbb E[w^2]=\mathbb E[b^2]=\tfrac13$ and
$\mathbb E[wb]=0$. Writing $\mathbf V_{i,:}=(1,x_i)$, this says
\begin{equation*}
 \mathbf K_{(0)}^\infty=\tfrac23\mathbf V\mathbf V^\top,
 \qquad
 \mathbf K_{\mathrm{out}}^\infty=\tfrac13\mathbf V\mathbf V^\top,
\end{equation*}
so both matrices have rank $2$ and kernel $\linspan\{(1,1,-2)^\top\}$.

The weakened statement therefore fails. The relation space
in~\cref{eqn:weak_functional_nullspace} is
$\mathcal N_{\mathrm{weak}}=\{\mathbf0\}$, whereas
$\ker\mathbf K_{\mathrm{out}}^\infty=\ker\mathbf K_{(0)}^\infty\neq\{\mathbf0\}$
and $\rank\mathbf K_{(0)}^\infty=2$, which is smaller than
$\dim\linspan\{\lambda_1,\lambda_2,\lambda_3\}=3$. Both identities
in~\cref{eqn:weak_exact_ntk_nullspace} fail; equivalently, the rows are
linearly independent while $\mathbf K_{(0)}^\infty\not\succ0$. The clause as
actually stated excludes the construction, since $\mathcal P_1$ is not dense in
$\mathcal X$, so~\Cref{prop:functional_ntk_rank,thm:weak_constraint_positivity}
are untouched.

Three scoping remarks. First, only one inclusion is lost. Density is consumed
exactly once, in the step
$\ker\mathbf K_{\mathrm{out}}^\infty\subseteq\mathcal N_{\mathrm{weak}}$. The
opposite inclusion
$\mathcal N_{\mathrm{weak}}\subseteq\ker\mathbf K_{(0)}^\infty$ follows by
differentiating $\sum_ic_iP_i(\theta)=0$ and, for
$\mathbf K_{\mathrm{out}}^\infty$, is immediate
from~\cref{eqn:weak_output_quadratic}; it survives the deletion, as do the
resulting bound
$\rank\mathbf K_{(0)}^\infty\leq\dim\linspan\{\lambda_i\}_{i=1}^N$ and the
existence and convergence conclusions, which use only clause
$\mathsf W_{\mathrm{reg}}$. Second, the rows are not at fault. Replacing
$\sigma$ by $\tanh$ with $\gamma>0$ restores density by the ridge-monomial
argument in the proof of~\Cref{cor:continuum_weak_interface_poisson} together
with the Weierstrass theorem, and the same three rows then give
$\mathbf K_{(0)}^\infty\succ0$ by~\Cref{prop:functional_ntk_rank}, as they also
do by~\Cref{prop:nonlocal_ntk_rank}, since the Dirac measures at
$x_1,x_2,x_3$ are linearly independent. Third, the mechanism is not tied to an
affine activation. For any polynomial $\sigma$ of degree $p$ on this
one-dimensional domain the closed ridge span lies in the polynomials of degree
at most $p$, so any $p+2$ distinct evaluation rows are independent in
$\mathcal X^*$ yet dependent on that span. For $\sigma=\tanh$ with the
bias-free scale $\gamma=0$ permitted by~\Cref{def:ntk_parameterization} every
ridge vanishes at $x=0$, so the closed span lies in
$\{f\in\mathcal X:f(0)=0\}$ and the single row $\lambda_1$, which is nonzero
and hence trivially independent, already yields $\mathbf K_{(0)}^\infty=0$
for an analytic nonpolynomial activation. Clauses $\mathsf P_{\mathrm{pd}}$ and
$\mathsf N_{\mathrm{pd}}$ exclude such configurations by explicitly requiring a
nonpolynomial $\sigma$ and $\gamma>0$; clause $\mathsf W_{\mathrm{pd}}$ states
neither requirement, and its density clause is precisely what does that work on
a general trial space.

What the example does not show: it establishes only that the clause cannot be
deleted outright, not that density is the weakest usable replacement, and it is
purely qualitative, since a feature family whose span is dense but poorly
conditioned still yields a positive definite kernel whose smallest eigenvalue
requires~\Cref{prop:quantitative_dntk_bound}. It also says nothing about the
pointwise or nonlocal classes. Like~\Cref{ex:compatible_singular_ntk}, which
restricts the model space by a polynomial activation for strong-form rows, the
construction is degenerate by design; the point here is the identification of
the single weak-class clause that rules such degeneracies out.
\end{example}

\begin{example}[Deleting $\gamma>0$ from $\mathsf N_{\mathrm{pd}}$: an even activation collapses two distinct Dirac rows]
\label{ex:gamma_zero_parity_obstruction}
Take $d_0=d_2=1$, $\Omega=(-1,1)$, and $D=\overline\Omega=[-1,1]$, so that
$\mathcal X=C([-1,1])$. Use~\Cref{def:ntk_parameterization} with the bias-free
scale $\gamma=0$ and the activation $\sigma=\cosh$,
\begin{equation*}
 u_\theta(x)=\frac1{\sqrt{d_1}}\sum_{m=1}^{d_1}a_m\cosh(w_mx),
\end{equation*}
and initialize every coordinate of $(w_m,b_m,a_m)$ i.i.d.\ from
$\Unif[-1,1]$. As in~\Cref{ex:compatible_singular_ntk}, the formal hidden
biases $b_m$ do not enter the prediction when $\gamma=0$, so their Jacobian
columns vanish. Impose the two nonlocal rows
of~\Cref{thm:nonlocal_constraint_positivity} with $s_1=s_2=1$ and
\begin{equation*}
 \nu_1=\delta_{1},\qquad \nu_2=\delta_{-1},
 \qquad
 P_1(\theta)=u_\theta(1),\qquad P_2(\theta)=u_\theta(-1).
\end{equation*}
Every clause of $\mathsf N_{\mathrm{reg}}$ and $\mathsf N_{\mathrm{pd}}$
in~\Cref{assump:pde_classes} holds except one: the rows are finite signed
regular Borel measures on the compact set $D$, the law $\mu$
of~\Cref{assump:main} is compactly supported,
$\cosh\in C_{\mathrm{loc}}^2(\mathbb R)$ is real analytic and nonpolynomial,
and $\operatorname{int}(\supp\mu_h)=(-1,1)^2\neq\varnothing$. Only the clause
$\gamma>0$ is deleted, and \Cref{def:ntk_parameterization} admits $\gamma=0$.

The deleted clause is exactly what the conclusion needs. Testing the identity
$c_1\delta_1+c_2\delta_{-1}=0$ on the Borel sets $\{1\}$ and $\{-1\}$ gives
$\mathcal N_{\mathrm{nl}}=\{\mathbf0\}$ and
$\dim\linspan\{\bar{\boldsymbol\nu}_1,\bar{\boldsymbol\nu}_2\}=2$, so the
weakened forms of~\cref{eqn:nonlocal_exact_ntk_nullspace}
and~\cref{eqn:nonlocal_rank_main} would assert
$\rank\mathbf K_{(0)}^\infty=2$. It is instead $1$. Because $\cosh$ is even
and $\gamma=0$, every ridge $y\mapsto\cosh(wy)$ is even, hence
$u_\theta(-x)=u_\theta(x)$ for every $\theta$ and every width, and
$P_1-P_2\equiv0$ as a function of $\theta$. Equivalently, the per-neuron
Jacobian $\boldsymbol\Psi$ defined before~\cref{eqn:full_output_hidden_decomposition},
evaluated at $z=(w,b,a)$ with columns ordered as
$(\partial_w,\partial_b,\partial_a)$, has two identical rows,
\begin{equation*}
 \boldsymbol\Psi(z)
 =\begin{bmatrix}
 a\sinh w&0&\cosh w\\
 a\sinh w&0&\cosh w
 \end{bmatrix},
\end{equation*}
since $\cosh(-w)=\cosh w$ and $(-1)\sinh(-w)=\sinh w$. Taking the expectation
in $\mathbf K_{(0)}^\infty=\mathbb E_{z\sim\mu}[\boldsymbol\Psi(z)
\boldsymbol\Psi(z)^\top]$ and using
$\mathbb E[\cosh^2w]=\tfrac14(2+\sinh2)$,
$\mathbb E[\sinh^2w]=\tfrac14(\sinh2-2)$, and $\mathbb E[a^2]=\tfrac13$ gives
\begin{equation}
 \mathbf K_{(0)}^\infty
 =\frac{1+\sinh 2}{3}
 \begin{bmatrix}1&1\\1&1\end{bmatrix},
 \qquad
 \ker\mathbf K_{(0)}^\infty=\linspan\{(1,-1)^\top\},
 \qquad
 \rank\mathbf K_{(0)}^\infty=1.
 \label{eqn:gamma_zero_singular_ntk}
\end{equation}
Both summands of~\cref{eqn:full_output_hidden_decomposition} are separately
singular here, with every entry of $\mathbf K_{\mathrm{out}}^\infty$ equal to
$\tfrac14(2+\sinh2)$ and every entry of $\mathbf K_{\mathrm{hid}}^\infty$ equal
to $\tfrac1{12}(\sinh2-2)$.

The failure is localized to a single step of the proof
of~\Cref{thm:nonlocal_constraint_positivity}, where $\gamma>0$ makes the
shifts $\gamma b$ exhaust $\mathbb R$ and thereby licenses the uniform
universal-approximation theorem for a nonpolynomial activation \emph{with a
bias}~\citep{Leshno1993}. When $\gamma=0$ the available features are
$\{y\mapsto\cosh(wy):w\in\mathbb R\}$, whose closed span in $C([-1,1])$
consists of even functions only, so it is a proper closed subspace and the
Riesz--Markov step is unavailable. Concretely,
$\zeta=\delta_1-\delta_{-1}$ satisfies
$\int_D\cosh(wy)\,d\zeta(y)=\cosh(w)-\cosh(-w)=0$ for every $w$ while
$\zeta\neq0$, so~\cref{eqn:nonlocal_all_ridges_annihilated} no longer forces
the corresponding signed measure to vanish. The degeneracy is exactly one
dimensional: $c_1\delta_1+c_2\delta_{-1}$ annihilates every even continuous
function precisely when $c_1+c_2=0$,
matching~\cref{eqn:gamma_zero_singular_ntk}.

Restoring $\gamma>0$ repairs the same two rows. With $\gamma=1$ and the same
law, the output-weight features
$[\boldsymbol\Phi_{\mathrm{out}}(\boldsymbol\xi)]_{i,:}
=\mathbf f_i(\boldsymbol\xi)^\top$
of~\cref{eqn:full_output_hidden_decomposition} are
$\mathbf f_1(w,b)=\cosh(w+b)$ and $\mathbf f_2(w,b)=\cosh(w-b)$, so
$\mathbb E[\mathbf f_1\mathbf f_2]
=\tfrac12\mathbb E[\cosh2w+\cosh2b]=\tfrac12\sinh2$ and
$\mathbb E[\mathbf f_1^2]=\mathbb E[\mathbf f_2^2]
=\tfrac12+\tfrac18\sinh^22$, whence
\begin{equation*}
 \mathbf K_{\mathrm{out}}^\infty
 =\frac18\begin{bmatrix}
 4+\sinh^22&4\sinh 2\\
 4\sinh 2&4+\sinh^22
 \end{bmatrix},
 \qquad
 \spec(\mathbf K_{\mathrm{out}}^\infty)
 =\left\{\frac{(\sinh 2+2)^2}{8},\frac{(\sinh 2-2)^2}{8}\right\},
\end{equation*}
approximately $\{3.958,0.331\}$ with eigenvectors $(1,\pm1)^\top$. Both
eigenvalues are positive because $\sinh2\neq2$, so
$\mathbf K_{\mathrm{out}}^\infty\succ0$ and therefore
$\mathbf K_{(0)}^\infty\succ0$ by~\cref{eqn:block_pd_certificate_bound}: an
exact certificate in the sense
of~\Cref{def:positive_definiteness_certificate}, agreeing
with~\Cref{thm:nonlocal_constraint_positivity}. The mechanism is general. For
any $\gamma>0$ and any $\mu_h$ whose support has nonempty interior,
$\mathbf c^\top\mathbf K_{\mathrm{out}}^\infty\mathbf c=0$ forces the
real-analytic function
$g_{\mathbf c}(w,b)=(c_1+c_2)\cosh w\cosh(\gamma b)
+(c_1-c_2)\sinh w\sinh(\gamma b)$
to vanish on $\operatorname{int}(\supp\mu_h)$, hence on all of $\mathbb R^2$
by the identity theorem; setting $b=0$ gives $c_1+c_2=0$ because
$\cosh\geq1$, and then $(w,b)=(1,1/\gamma)$ gives $c_1=c_2$. The bias is what
breaks the parity.

Two limitations are worth stating. First, the construction leaves the
existence half of~\Cref{thm:nonlocal_constraint_positivity} intact:
\Cref{prop:finite_measure_row_regularity} imposes no condition on $\gamma$, so
$\mathbf K_{(0)}^\infty$ still exists and
$\mathbf K_{(0)}\to\mathbf K_{(0)}^\infty$ at $\gamma=0$. Only the nullspace
and rank identity in~\cref{eqn:nonlocal_exact_ntk_nullspace} fails. Second,
the example isolates the hypothesis rather than the architecture: with
$\gamma=0$ and the also real-analytic nonpolynomial $\sigma=\exp$, the
identity $\int_De^{wy}\,d\zeta(y)=0$ for all $w\in\mathbb R$ continues
analytically to the Fourier transform of $\zeta$ and still forces $\zeta=0$,
so the conclusion survives for that instance. What the construction shows is
that the weakened hypothesis set admits false instances, not that every
$\gamma=0$ model is degenerate.

The same pair also settles the pointwise class. Since $\delta_{\pm1}$ are
order-zero rows of~\Cref{def:linear_pde} at the distinct sites
$\mathbf x=\pm1\in\overline\Omega$ with local coefficient vectors
$\mathbf b_1(1)=\mathbf b_2(-1)=(1)$, the weakened form
of~\cref{eqn:exact_dntk_rank} would likewise predict
$\rank\mathbf K_{(0)}^\infty=1+1=2$, and $\cosh$ satisfies
$\mathsf P_{\mathrm{reg}}$ for $r_{\max}=0$. So $\gamma>0$ is load-bearing in
$\mathsf P_{\mathrm{pd}}$ for~\Cref{prop:limiting_ntk_rank} exactly as it is
in $\mathsf N_{\mathrm{pd}}$
for~\Cref{prop:nonlocal_ntk_rank,thm:nonlocal_constraint_positivity}. The
example is distinct from~\Cref{ex:compatible_singular_ntk}, which also sets
$\gamma=0$ but additionally violates the nonpolynomiality clause; here
analyticity and nonpolynomiality are retained and only $\gamma>0$ is removed.
\end{example}

In the next three examples the global feature family is rich enough, and what
fails is the passage from $\supp\mu_h$ to all of $\mathbb R^{d_0+1}$ on which
the density clause is stated. The first two remove analyticity of the feature
map; the third removes the interior hypothesis on $\supp\mu_h$.

\begin{example}[Smoothness without analyticity collapses the weak feature span]
\label{ex:weak_analyticity_necessity}
Delete from clause $\mathsf W_{\mathrm{pd}}$ of~\Cref{assump:pde_classes} only
the requirement that every $\boldsymbol\xi\mapsto\varphi_{\boldsymbol\xi,q}$ be
real analytic into $\mathcal X$, keeping
$\operatorname{int}(\supp\mu_h)\neq\varnothing$, the ridge-density clause, and
all of $\mathsf W_{\mathrm{reg}}$. Both identities
in~\cref{eqn:weak_exact_ntk_nullspace} then fail, and with them the
positive-definiteness criteria
of~\Cref{prop:functional_ntk_rank,cor:weak_boundary_joint_positivity}.

Take the row family of~\Cref{cor:continuum_weak_interface_poisson} with
$d_0=d_2=1$, $\Omega=(-1,1)$, $\mathcal X=H^1(\Omega)$, $\kappa\equiv1$,
$\gamma=1$, unit row scales, the single test function
$v_1(x)=1-x^2\in H_0^1(\Omega)$, and the network
of~\Cref{def:ntk_parameterization}; only the activation will be changed. The
$N=3$ rows
\begin{equation*}
 \lambda_1^\Omega(f)=\int_{-1}^{1}f'(x)v_1'(x)\,dx,
 \qquad
 \beta_-^\partial(f)=f(-1),
 \qquad
 \beta_+^\partial(f)=f(1)
\end{equation*}
are bounded on $\mathcal X$ by Cauchy--Schwarz and the one-dimensional trace
theorem, and they are linearly independent in $\mathcal X^*$: testing a
vanishing combination against $v_1$, then against $x\mapsto(1-x)/2$, then
against the constant function $1$ forces its three coefficients to vanish in
that order. Hence $\mathcal N_{\mathrm{weak}}=\{\mathbf0\}$
in~\cref{eqn:weak_functional_nullspace}, and the weakened statement would
assert $\mathbf K_{(0)}^\infty\succ0$ with rank $3$. For $\sigma=\tanh$ this is
exactly what~\Cref{cor:continuum_weak_interface_poisson} proves.

Replace $\tanh$ by
\begin{equation*}
 \sigma_\star(z)=z+\chi\!\left(\tfrac12-z\right),
 \qquad
 \chi(t)=e^{-1/t}\mathbf 1_{\{t>0\}},
\end{equation*}
and let $\mu=\Unif([1,2]\times[3,4])\otimes\Unif([-1,1])$. Then
$\sigma_\star\in C^\infty(\mathbb R)$, it agrees with the identity on
$[\tfrac12,\infty)$, and it is neither polynomial nor real analytic, because
$\sigma_\star(0)=e^{-2}\neq0$ while any polynomial, or any real-analytic
function, that agrees with $z$ on a half-line equals $z$ everywhere. The law
$\mu$ is width independent, compactly supported, and sampled i.i.d., with
$\operatorname{int}(\supp\mu_h)=(1,2)\times(3,4)\neq\varnothing$ and
$\mathbb E[a_m\mid\boldsymbol\xi_m]=0$, so~\Cref{assump:main} holds.

On the open cone $Q=\{\boldsymbol\xi=(w,b):\gamma b-|w|>\tfrac12\}$ every
preactivation seen on the domain lies in the affine piece of $\sigma_\star$:
for $x\in\overline\Omega$,
\begin{equation*}
 wx+\gamma b\geq\gamma b-|w|>\tfrac12,
 \qquad\text{so}\qquad
 \varphi_{\boldsymbol\xi}=w\,p_1+\gamma b\,p_0
 \quad\text{exactly on }\overline\Omega,
\end{equation*}
where $p_0=1$ and $p_1=x$. Moreover $\supp\mu_h\subset Q$, because
$b-|w|\geq3-2=1$ there; any $\gamma>0$ works after rescaling the bias support
to $[3/\gamma,4/\gamma]$. Thus
$\linspan\{\varphi_{\boldsymbol\xi}:\boldsymbol\xi\in Q\}
=\linspan\{p_0,p_1\}$ is two-dimensional, although the full ridge family over
$\mathbb R^2$ is dense. Writing
$F_i(\boldsymbol\xi)=\lambda_i(\varphi_{\boldsymbol\xi})$ for the three rows in
the order above, and using that $v_1\in H_0^1(\Omega)$ makes
$\int_{-1}^{1}v_1'=0$, so that $\lambda_1^\Omega$ annihilates every affine
function, we obtain on $Q$
\begin{equation*}
 F_1\equiv0,
 \qquad
 F_2=\gamma b-w,
 \qquad
 F_3=\gamma b+w,
 \qquad
 \nabla_{\boldsymbol\xi}F_2=(-1,\gamma),
 \qquad
 \nabla_{\boldsymbol\xi}F_3=(1,\gamma).
\end{equation*}
The Gram representation in~\Cref{thm:linearNTK}, applied to the per-neuron maps
$h_i(z)=aF_i(\boldsymbol\xi)$ supplied
by~\Cref{lem:bounded_functional_row_regularity} for $z=(\boldsymbol\xi,a)$,
gives for every $\mathbf c\in\mathbb R^3$
\begin{equation*}
 \mathbf c^\top\mathbf K_{(0)}^\infty\mathbf c
 =\mathbb E_{\mu_h}\!\left[
 \big((c_3-c_2)w+(c_2+c_3)\gamma b\big)^2\right]
 +\tfrac13\left\|\big(c_3-c_2,\
 \gamma(c_2+c_3)\big)\right\|_2^2,
\end{equation*}
since $\mathbb E[a^2]=\tfrac13$ and $a$ is independent of $\boldsymbol\xi$. The
second term vanishes precisely when $c_3-c_2=0$ and $c_2+c_3=0$, that is, when
$c_2=c_3=0$; the first term then vanishes as well, and $c_1$ is unconstrained,
so
\begin{equation*}
 \ker\mathbf K_{(0)}^\infty
 =\linspan\{(1,0,0)^\top\}
 \neq\{\mathbf0\}=\mathcal N_{\mathrm{weak}},
 \qquad
 \rank\mathbf K_{(0)}^\infty=2<3
 =\dim\linspan\{\lambda_1^\Omega,\beta_-^\partial,\beta_+^\partial\}.
\end{equation*}
The annihilating functional is explicit: $\lambda_1^\Omega$ is nonzero on
$\mathcal X$, since $\lambda_1^\Omega(v_1)=\int_{-1}^{1}4x^2\,dx=\tfrac83$, yet
it kills every $\varphi_{\boldsymbol\xi}$ with $\boldsymbol\xi\in Q$. The
degeneracy is not an infinite-width artifact: $F_1\equiv0$ on the open set $Q$
and $\boldsymbol\xi_m\in\supp\mu_h\subset Q$ almost surely, so
$\nabla_\theta P_1(\theta_{(0)})=\mathbf0$ almost surely and the first row and
column of $\mathbf K_{(0)}$ vanish at every width, while the target
$\int_{-1}^{1}gv_1$ of that row is generically nonzero.

Analyticity is the only clause that fails. Clause $\mathsf W_{\mathrm{reg}}$
holds in full, because the rows lie in $\mathcal X^*$ and
$\sigma_\star\in C_{\mathrm{loc}}^3(\mathbb R)$, so
\Cref{cor:weak_H1_row_regularity} makes $\mathcal G$ a $C^2$ map into
$H^1(\Omega)$ on an open neighborhood of every fixed parameter tube; the
existence and convergence half of~\Cref{prop:functional_ntk_rank} is therefore
intact and only its positive-definiteness half fails. The hidden support has
nonempty interior. The ridge-density clause also holds: for each
$n\in\mathbb N_0$, nonpolynomiality gives $b_n$ with
$\sigma_\star^{(n)}(\gamma b_n)\neq0$, and
$\partial_w^n\sigma_\star(wx+\gamma b_n)|_{w=0}
=\sigma_\star^{(n)}(\gamma b_n)x^n$ is an $H^1$ limit of finite combinations of
ridge features, so the monomials, and hence the $H^1(-1,1)$-dense polynomials,
lie in the closed span, exactly as in the proof
of~\Cref{cor:continuum_weak_interface_poisson}. Only Banach-valued analyticity
fails, and visibly so: $\varphi_{(0,b)}=\sigma_\star(\gamma b)p_0$ is not a
real-analytic function of $b$.

The failure occurs at exactly the step the proof needs. In the proof
of~\Cref{thm:weak_constraint_positivity}, analyticity is used once: a
combination of the $F_{iq}$ that vanishes $\mu_h$-almost surely vanishes on
$\operatorname{int}(\supp\mu_h)$ by continuity, and the identity theorem
propagates this to all of $\mathbb R^{d_0+1}$, where the density clause turns
it into $\sum_ic_i\lambda_i=0$ in $\mathcal X^*$; the same step appears
in~\Cref{prop:analytic_functional_rank}. Here the propagation is the only
thing that fails: the relation $F_1\equiv0$ holds on $Q$ and stops there, while
the density clause quantifies over all of $\mathbb R^{d_0+1}$, so the two
clauses never meet. The reverse inclusion
$\mathcal N_{\mathrm{weak}}\subseteq\ker\mathbf K_{(0)}^\infty$ uses only
linearity and still holds. The example also shows that the clause must be read
globally in $\boldsymbol\xi$: here
$\boldsymbol\xi\mapsto\varphi_{\boldsymbol\xi}$ is real analytic on a
neighborhood of $\supp\mu_h$, indeed affine there, and that is not enough.

Two remarks delimit the claim. First, $\sigma=\ReLU$ with the same $\mu$
produces the identical collapse, because $\gamma b-|w|\geq1>0$ on $\supp\mu_h$
places the kink $x=-\gamma b/w\leq-\tfrac32$ outside $\overline\Omega$;
however, $\boldsymbol\xi\mapsto\varphi_{\boldsymbol\xi}$ is then not
differentiable into $H^1(\Omega)$ once the kink enters $\Omega$, so $\ReLU$
also violates $\mathsf W_{\mathrm{reg}}$ on parameter tubes of radius at least
$1/\sqrt2$ and is not a single-clause counterexample. The smooth
$\sigma_\star$ shows that neither smoothness nor nonpolynomiality substitutes
for analyticity. Second, the example does not claim that non-analytic
activations give degenerate kernels in general: the collapse is a property of
the pair $(\sigma,\mu_h)$, and a hidden support over which the affine piece of
$\sigma_\star$ fails to cover $\overline\Omega$ produces a far richer feature
family. Nor is analyticity the weakest possible clause; density of
$\linspan\{\varphi_{\boldsymbol\xi,q}:\boldsymbol\xi\in\supp\mu_h\}$ would
suffice, and analyticity is the checkable condition that transports the stated
density clause from $\mathbb R^{d_0+1}$ to $\supp\mu_h$.

The same device applies to the other two classes with $\sigma_\star$, whose
smoothness already gives $\mathsf P_{\mathrm{reg}}$ and
$\mathsf N_{\mathrm{reg}}$. On $Q$ the strong-form row $u\mapsto u''(0)$ and
the nonlocal row $f\mapsto\int f\,d\nu$ with
$\nu=\delta_{-1}-2\delta_0+\delta_1$ both annihilate $\linspan\{p_0,p_1\}$ and
give an identically zero limiting NTK, although each row is nonzero and its
local coefficient vector, respectively measure tuple, is nonzero. The
analyticity requirements in $\mathsf P_{\mathrm{pd}}$ and
$\mathsf N_{\mathrm{pd}}$, used
by~\Cref{prop:limiting_ntk_rank,prop:nonlocal_ntk_rank}, are load-bearing for
the same reason. Unlike~\Cref{ex:relu3_fourth_order_obstruction}, where the
obstruction is a mismatch between activation smoothness and differential order,
the activation here may be taken as smooth as desired and the obstruction is
purely the loss of analytic continuation.
\end{example}

\begin{example}[A compactly supported $C^\infty$ activation breaks the weak criterion]
\label{ex:compact_support_analyticity_obstruction}
This example weakens exactly one clause of the hypotheses
of~\Cref{thm:weak_constraint_positivity}: in clause
$\mathsf W_{\mathrm{pd}}$ of~\Cref{assump:pde_classes}, replace ``every
$\boldsymbol\xi\mapsto\varphi_{\boldsymbol\xi,q}$ is real analytic into
$\mathcal X$'' by ``$C^\infty$ into $\mathcal X$'', and keep every other
requirement verbatim, including~\Cref{assump:main}, clause
$\mathsf W_{\mathrm{reg}}$, $\operatorname{int}(\supp\mu_h)\neq\varnothing$,
and density of the ridge span over all
$\boldsymbol\xi\in\mathbb R^{d_0+1}$. The weakened statement is false, and
with it the weakened forms of~\Cref{prop:functional_ntk_rank}
and~\Cref{cor:weak_boundary_joint_positivity}. It is the same clause as
in~\Cref{ex:weak_analyticity_necessity}, broken by a different mechanism:
there the features are locally affine on the region the row observes, here
they are locally zero.

Take $d_0=d_2=1$, $\Omega=(-1,1)$, $\mathcal X=H^1(-1,1)$, and $\gamma=1$
in~\Cref{def:ntk_parameterization}, with the standard bump activation
\begin{equation*}
 \sigma(z)=
 \begin{cases}
  \exp\!\big((z^2-1)^{-1}\big),&|z|<1,\\[2pt]
  0,&|z|\geq1,
 \end{cases}
 \qquad
 \sigma\in C_c^\infty(\mathbb R),
 \qquad
 \supp\sigma=[-1,1].
\end{equation*}
Initialize $(w_m,b_m)\sim\Unif\!\left([1,2]\times[1,\tfrac32]\right)$ and
$a_m\sim\Unif\{\pm1\}$ independently, so $\mu$ is compactly supported and
width independent, $\mathbb E[a_m\mid w_m,b_m]=0$, and
$\operatorname{int}(\supp\mu_h)=(1,2)\times(1,\tfrac32)\neq\varnothing$.
Impose the single ($N=1$) weak residual row generated by the $H^1$ inner
product, that is, the weak form of $-u''+u$ tested against a fixed
$v_1\in C_c^\infty\!\left((\tfrac12,1)\right)$ with $v_1\geq0$ and
$v_1\not\equiv0$:
\begin{equation*}
 \mathcal V_1=\mathcal X,
 \qquad
 a(f,v)=\int_{-1}^{1}\big(f'v'+fv\big)\,dx=\langle f,v\rangle_{H^1},
 \qquad
 \lambda_1=a(\cdot,v_1),\qquad s_1=1.
\end{equation*}
Write $\varphi_{\boldsymbol\xi}(x)=\sigma(wx+b)$ and
$F_1(\boldsymbol\xi)=\lambda_1(\varphi_{\boldsymbol\xi})$ for
$\boldsymbol\xi=(w,b)$.

The weakened hypotheses hold. Since
$\sigma\in C^\infty\subset C_{\mathrm{loc}}^3$,
\Cref{cor:weak_H1_row_regularity} gives the required $C^2$ map into
$\mathcal X$ on a neighborhood of every radius-$R$ parameter tube, and $a$ is
continuous and bilinear with $\lambda_1\in\mathcal X^*$,
$\|\lambda_1\|_{\mathcal X^*}=\|v_1\|_{H^1}$; clause
$\mathsf W_{\mathrm{reg}}$ therefore holds. All derivatives of $\sigma$ are
bounded and uniformly continuous, so
$\boldsymbol\xi\mapsto\varphi_{\boldsymbol\xi}$ is $C^\infty$ into
$C^1([-1,1])\hookrightarrow H^1(-1,1)$. The ridge span is dense: by the
Banach-valued differentiation argument in the proof
of~\Cref{cor:continuum_weak_interface_poisson},
$\partial_w^n\sigma(wx+b)\big|_{w=0}=\sigma^{(n)}(b)\,x^n$ is an $H^1$ limit of
finite linear combinations of features $\varphi_{(w,b)}$ with $b$ fixed, and
$\sigma^{(n)}\not\equiv0$ for every $n$, since otherwise $\sigma$ would be a
polynomial vanishing on $[1,\infty)$, hence identically zero, contradicting
$\sigma(0)=e^{-1}$. Choosing $b_n$ with $\sigma^{(n)}(b_n)\neq0$ puts every
monomial, and hence every polynomial, in the closed span, and polynomials are
dense in $H^1(-1,1)$. Finally $a$ is coercive with $c_a=1$ and $\{v_1\}$ is
linearly independent, so~\Cref{thm:weak_constraint_positivity} in its weakened
form, and likewise its coercive special case, would assert
$\mathbf K_{(0)}^\infty\succ0$.

The conclusion fails. Let
$U=\{(w,b):w>0,\ \tfrac w2+b>1\}$, an open set containing
$\supp\mu_h=[1,2]\times[1,\tfrac32]$, where $\tfrac w2+b\geq\tfrac32$. For
$\boldsymbol\xi\in U$ and $x\geq\tfrac12$ we have $wx+b\geq\tfrac w2+b>1$, and
$\{x:wx+b>1\}=\big((1-b)/w,\infty\big)$ with $(1-b)/w<\tfrac12$; hence
$\varphi_{\boldsymbol\xi}$ and all of its $x$-derivatives vanish on a
neighborhood of $[\tfrac12,1]$ in $\overline\Omega$, which contains
$\supp v_1$. Therefore
\begin{equation*}
 F_1(\boldsymbol\xi)
 =\int_{1/2}^{1}\big(\varphi_{\boldsymbol\xi}'v_1'
 +\varphi_{\boldsymbol\xi}v_1\big)\,dx=0
 \qquad\text{for every }\boldsymbol\xi\in U,
\end{equation*}
and, $U$ being open, every partial derivative of $F_1$ vanishes on $U$ as
well. This is consistent with the density just established: $F_1$ is nonzero
off $U$, for instance $F_1(0,0)=e^{-1}\int_{1/2}^{1}v_1>0$. Because
$P_1(\theta)=d_1^{-1/2}\sum_m a_mF_1(\boldsymbol\xi_m)$, both
$\partial P_1/\partial a_m=d_1^{-1/2}F_1(\boldsymbol\xi_m)$ and
$\partial P_1/\partial\boldsymbol\xi_m
=d_1^{-1/2}a_m\nabla F_1(\boldsymbol\xi_m)$ vanish, so
$\nabla_\theta P_1(\theta)=\mathbf0$ at every parameter vector whose neuron
blocks lie in $\supp\mu$. Hence $\mathbf K_{(0)}=\mathbf0$ almost surely at
every width and $\mathbf K_{(0)}^\infty=\mathbf0$. Yet
$\lambda_1(v_1)=\|v_1\|_{H^1}^2>0$, so $\lambda_1\neq0$ and
$\mathcal N_{\mathrm{weak}}=\{0\}$. Therefore
\begin{equation*}
 \ker\mathbf K_{(0)}^\infty=\mathbb R\neq\{0\}=\mathcal N_{\mathrm{weak}},
 \qquad
 \rank\mathbf K_{(0)}^\infty=0<1=\dim\linspan\{\lambda_1\},
\end{equation*}
contradicting~\cref{eqn:weak_exact_ntk_nullspace}. Since
$\nabla_\theta\mathcal L(\theta_{(0)})=\mathbf0$, gradient descent is
stationary at every width and step size, and by~\Cref{def:pinn_loss} with
$s_1=1$ the loss is frozen at $\mathcal L(\theta_{(0)})=y_1^2$.

The kernel need not collapse to zero. Append the boundary row
$\lambda_2=\delta_{-1}$, that is, $\lambda_2(f)=f(-1)$, which is bounded on
$H^1(-1,1)$ by the one-dimensional trace theorem and enters
\Cref{thm:weak_constraint_positivity} through the device
$\mathcal V_2=\mathbb R$, $a_2(f,t)=tf(-1)$, $v_2=s_2=1$ used in the proof
of~\Cref{cor:weak_boundary_joint_positivity}. Then
$F_2(\boldsymbol\xi)=\sigma(b-w)$, and $b-w\in[-1,\tfrac12]$ on $\supp\mu_h$,
so $F_2>0$ $\mu_h$-almost everywhere, with $F_2(1,1)=\sigma(0)=e^{-1}$;
consequently $[\mathbf K_{\mathrm{out}}^\infty]_{22}=\int F_2^2\,d\mu_h>0$, and
the full kernel dominates the output-weight block
by~\cref{eqn:block_pd_certificate_bound}. The rows are independent, since
$c_1\lambda_1+c_2\lambda_2=0$ applied to $v_1$ gives
$c_1\|v_1\|_{H^1}^2=0$ because $v_1(-1)=0$, and applying the relation to the
constant function $1$ then gives $c_2=0$. The first row of
$\mathbf J(\theta)$ still vanishes identically, so the first row and column of
the kernel vanish and
\begin{equation*}
 \ker\mathbf K_{(0)}^\infty=\linspan\{\mathbf e_1\}\neq\{0\},
 \qquad
 \rank\mathbf K_{(0)}^\infty=1<2=\dim\linspan\{\lambda_1,\lambda_2\}.
\end{equation*}
This is a weak residual row together with a boundary row on a common trial
space, so the weakened form of~\Cref{cor:weak_boundary_joint_positivity} fails
as well.

The one-line reason: the proof of~\Cref{thm:weak_constraint_positivity}
concludes from $F_1\equiv0$ on $\supp\mu_h$ that $F_1\equiv0$ on
$\mathbb R^{d_0+1}$ by the identity theorem; a $C^\infty$ map can be flat, so
that step, and only that step, is what the weakening removes. The hypothesis
as written excludes the example: $\sigma$ vanishes on $|z|\geq1$ while
$\sigma(0)=e^{-1}\neq0$, so it is not real analytic, and since
$\delta_0\in H^1(-1,1)^*$, real analyticity of
$\boldsymbol\xi\mapsto\varphi_{\boldsymbol\xi}$ into $\mathcal X$ would force
$b\mapsto\sigma(b)$ to be real analytic. The degeneracy is a different
mechanism from~\Cref{ex:relu3_fourth_order_obstruction}: there the activation
violates the regularity clause and each feature is locally a nonzero
polynomial that a high-order derivative row annihilates, an order mismatch,
whereas here regularity is amply satisfied, the row is a bounded first-order
weak row, and the features are locally \emph{zero} on the region the row
observes, a support mismatch. The example does not break the existence half of
the theorem: clause $\mathsf W_{\mathrm{reg}}$ still holds, so the first
conclusion of~\Cref{prop:functional_ntk_rank} still yields existence of
$\mathbf K_{(0)}^\infty$ and convergence of $\mathbf K_{(0)}$, here trivially.
Nor does it show that analyticity is the only possible hypothesis: density
over all $\boldsymbol\xi$ does hold, and what fails is that
$\linspan\{\varphi_{\boldsymbol\xi}:\boldsymbol\xi\in\supp\mu_h\}$ lies in the
proper closed subspace $\{f\in\mathcal X:f|_{[1/2,1]}=0\}$, so requiring
density of the features indexed by $\operatorname{int}(\supp\mu_h)$ alone would
restore the conclusion. What is shown is that the analyticity clause cannot
simply be downgraded to $C^\infty$ while the density clause is quantified over
all of $\mathbb R^{d_0+1}$, and that the obstruction is a mismatch between
$\supp\mu_h$ and the region a row observes rather than a defect of compactly
supported activations as such.
\end{example}

\begin{example}[Empty-interior hidden support breaks the exact positivity criterion]
\label{ex:empty_interior_support}
Delete the clause $\operatorname{int}(\supp\mu_h)\neq\varnothing$ from
$\mathsf N_{\mathrm{pd}}$ and $\mathsf W_{\mathrm{pd}}$
in~\Cref{assump:pde_classes} and retain every other hypothesis. The following
system then satisfies all remaining clauses and has linearly independent rows,
yet its limiting NTK is singular.

Let $d_0=d_2=1$, $\Omega=(-\pi,\pi)$, $D=\overline\Omega=[-\pi,\pi]$,
$\sigma=\sin$, and $\gamma>0$. Use the network
of~\Cref{def:ntk_parameterization},
\begin{equation*}
 u_\theta(x)=\frac1{\sqrt{d_1}}\sum_{m=1}^{d_1}a_m\sin(w_mx+\gamma b_m),
\end{equation*}
with the three unit-scaled evaluation rows $\nu_i=\delta_{x_i}$ at
$(x_1,x_2,x_3)=(-\pi,0,\pi)$ and $s_i=1$, so that $P_i(\theta)=u_\theta(x_i)$:
two Dirichlet boundary rows and one interior measurement. Initialize the
neurons i.i.d.\ with independent coordinates
\begin{equation*}
 w_m\sim\Unif\{-1,1\},\qquad
 b_m\sim\Unif[-1,1],\qquad
 a_m\sim\Unif\{-1,1\},
\end{equation*}
so that $\supp\mu_h=\{-1,1\}\times[-1,1]$ is compact with empty interior in
$\mathbb R^2$; for $d_0=1$ this is the unit-norm hidden-weight sphere with a
free bias. Every other hypothesis holds. The law $\mu$ is compactly supported
and width independent as in~\Cref{assump:main}; $\sin$ is real analytic,
nonpolynomial, and smooth; $\gamma>0$; $D$ is compact and the rows are finite
signed regular Borel measures, giving $\mathsf N_{\mathrm{reg}}$. Reading the
same rows as $\lambda_i(f)=f(x_i)$ on $\mathcal X=C(D)$ gives
$\mathsf W_{\mathrm{reg}}$, since $C(D)$ embeds continuously in
$L^1_{\mathrm{loc}}$, point evaluations are bounded on it, and
$\boldsymbol\xi\mapsto\sin(w\,\cdot+\gamma b)$ is $C^2$ (indeed analytic)
into $C(D)$; such rows are
covered by~\Cref{thm:weak_constraint_positivity} on taking
$\mathcal V_i=\mathbb R$, $v_i=s_i=1$, and $a_i(f,t)=t\lambda_i(f)$, exactly as
in the proof of~\Cref{cor:weak_boundary_joint_positivity}. The remaining clauses of
$\mathsf W_{\mathrm{pd}}$ hold as well:
$\boldsymbol\xi\mapsto\varphi_{\boldsymbol\xi,1}=\sin(w\,\cdot+\gamma b)$ is
real analytic into $C(D)$ because $D$ is compact, and the ridge span is dense
by~\citet{Leshno1993}. Finally $\mathbb E[a_m\mid w_m,b_m]=0$, so the
output-weight centering used for GD also holds.

Take $\mathbf c=(1,2,1)^\top$. The associated single-neuron feature
combination is
\begin{equation*}
 H(w,b)=\sum_{i=1}^3c_i\sin(wx_i+\gamma b)
 =\sin(\gamma b-\pi w)+2\sin(\gamma b)+\sin(\gamma b+\pi w)
 =4\sin(\gamma b)\cos^2\!\left(\tfrac{\pi w}{2}\right),
\end{equation*}
by the sum-to-product identity followed by $1+\cos t=2\cos^2(t/2)$. The squared
factor is the whole point: $\cos^2(\pi w/2)$ has a \emph{double} zero at every
odd integer, so the per-neuron map $h_{\mathbf c}(w,b,a)=aH(w,b)$, which
satisfies $\sum_ic_iP_i(\theta)=d_1^{-1/2}\sum_mh_{\mathbf c}(\theta_m)$, has
gradient
\begin{equation*}
 \nabla h_{\mathbf c}
 =\left(
 -4\pi a\sin(\gamma b)\sin\!\left(\tfrac{\pi w}{2}\right)
 \cos\!\left(\tfrac{\pi w}{2}\right),\;
 4a\gamma\cos(\gamma b)\cos^2\!\left(\tfrac{\pi w}{2}\right),\;
 4\sin(\gamma b)\cos^2\!\left(\tfrac{\pi w}{2}\right)
 \right),
\end{equation*}
in the coordinates $(w,b,a)$. Every component retains a factor
$\cos(\pi w/2)$, so $\nabla h_{\mathbf c}$ vanishes identically on
$\supp\mu$. Hence $\mathbf J(\theta_{(0)})^\top\mathbf c=\mathbf0$ almost
surely at every width, so $\mathbf K_{(0)}\mathbf c=\mathbf0$ and
$\mathbf K_{(0)}^\infty\mathbf c=\mathbf0$. Explicitly, the limit exists by
$\mathsf N_{\mathrm{reg}}$ and~\Cref{thm:linearNTK} and equals the per-neuron
gradient expectation; writing $\mathbf f_1=(-1,1,-1)^\top$ and
$\mathbf f_2=(1,0,-1)^\top$, both signs $w=\pm1$ give
$(\sin(wx_i+\gamma b))_i=\sin(\gamma b)\mathbf f_1$,
$(\partial_w\sin(wx_i+\gamma b))_i=\pi\cos(\gamma b)\mathbf f_2$, and
$(\partial_b\sin(wx_i+\gamma b))_i=\gamma\cos(\gamma b)\mathbf f_1$, so that,
with $a^2\equiv1$ and $b\sim\Unif[-1,1]$,
\begin{equation*}
 \mathbf K_{(0)}^\infty
 =\underbrace{\left(\mathbb E[\sin^2(\gamma b)]
 +\gamma^2\mathbb E[\cos^2(\gamma b)]\right)}_{=:A}
 \mathbf f_1\mathbf f_1^\top
 +\underbrace{\pi^2\mathbb E[\cos^2(\gamma b)]}_{=:B}\,
 \mathbf f_2\mathbf f_2^\top.
\end{equation*}
Here $A\geq\min(1,\gamma^2)>0$ because $\sin^2+\cos^2=1$, and $B>0$ because
$\cos(\gamma b)$ vanishes at only finitely many $b\in[-1,1]$. Since
$\mathbf f_1$ and $\mathbf f_2$ are independent,
\begin{equation*}
 \rank\mathbf K_{(0)}^\infty=2,
 \qquad
 \ker\mathbf K_{(0)}^\infty=\linspan\{(1,2,1)^\top\},
\end{equation*}
the null direction being the one orthogonal to both $\mathbf f_1$ and
$\mathbf f_2$. A referee may check this in one line from the entries
$[\mathbf K_{(0)}^\infty]_{11}=[\mathbf K_{(0)}^\infty]_{33}=A+B$,
$[\mathbf K_{(0)}^\infty]_{12}=[\mathbf K_{(0)}^\infty]_{23}=-A$,
$[\mathbf K_{(0)}^\infty]_{13}=A-B$, and
$[\mathbf K_{(0)}^\infty]_{22}=A$: the determinant is zero while the leading
$2\times2$ minor equals $AB>0$.

Distinct Dirac masses are linearly independent, so
$\mathcal N_{\mathrm{nl}}=\{0\}$ and
$\dim\linspan\{\bar{\boldsymbol\nu}_1,\bar{\boldsymbol\nu}_2,
\bar{\boldsymbol\nu}_3\}=3$. The weakened
\Cref{thm:nonlocal_constraint_positivity,prop:nonlocal_ntk_rank} would
therefore assert $\ker\mathbf K_{(0)}^\infty=\{0\}$ and
$\rank\mathbf K_{(0)}^\infty=3$; both fail, and with them the implication
from independence of the measure tuples to positive definiteness. Reading the
same three rows in $C(D)^*$ breaks the weakened
\Cref{thm:weak_constraint_positivity,prop:functional_ntk_rank} identically, and
reading them as order-zero pointwise rows at distinct sites breaks the weakened
\Cref{prop:limiting_ntk_rank}, whose local coefficient families
$\mathbf b_i(x_i)=(1)$ are independent. The finite-dimensional witnesses fail
with them: the moment matrix
$[\mathbf M_2^{\mathrm{nl}}]_{i,(1,\alpha)}=x_i^\alpha$
of~\cref{eqn:nonlocal_moment_matrix} is a Vandermonde matrix at the nodes
$-\pi,0,\pi$, with determinant $(0+\pi)(\pi+\pi)(\pi-0)=2\pi^3\neq0$,
so~\Cref{cor:monomial_moment_rank,prop:analytic_functional_rank} would certify
positive definiteness from a rank-$3$ test.

The deleted clause is exactly what licenses the analytic-continuation step. The
proofs
of~\Cref{thm:weak_constraint_positivity,thm:nonlocal_constraint_positivity}
pass from $\mathbf c\in\ker\mathbf K_{\mathrm{out}}^\infty$ to vanishing of the
analytic function $G_{\mathbf c,1}$ on $\supp\mu_h$ --- here
$G_{\mathbf c,1}=H$ --- invoke the identity theorem because that support
contains a nonempty open set, and only then apply ridge density as
in~\cref{eqn:nonlocal_all_ridges_annihilated}. In this example $H$ does vanish
on $\supp\mu_h$, because $\supp\mu_h$ lies inside the zero set
$\{w\in2\mathbb Z+1\}\cup\{\gamma b\in\pi\mathbb Z\}$, and that zero set has
empty interior; there is therefore no open set to continue from, and the
density step is never reached. No repair is available within the argument: it
is the conclusion, not only the proof, that fails. The failure is also neither
an artifact of the row count nor of a knife-edge law. Appending further rows of
any class keeps $(\mathbf c,\mathbf0)$ in the kernel, because
$\nabla h_{\mathbf c}$ still vanishes on all of $\supp\mu$, and \emph{every}
law carried by $\{|w|=1\}$, with any bias marginal, produces the same null
vector.

Four caveats. First, only the ``pd'' conclusions break: the deleted clause
appears nowhere in $\mathsf N_{\mathrm{reg}}$ or $\mathsf W_{\mathrm{reg}}$, so
existence of $\mathbf K_{(0)}^\infty$ and the convergence
$\mathbf K_{(0)}\to\mathbf K_{(0)}^\infty$ still hold here, and we used them.
Second, the example shows that the clause cannot be \emph{deleted}; it does not
show that the clause is the weakest available hypothesis. A nontrivial
real-analytic function has a Lebesgue-null zero set, so the strictly weaker
requirement $|\supp\mu_h|>0$ already suffices for the same proof. The exact
condition under which the conclusion holds is that $\mu_h$ charge the
complement of the zero set of every feature combination attached to a vector
outside the relation space, but that condition is a restatement of the
conclusion rather than a checkable design criterion, which is why a geometric
sufficient condition is assumed instead. Third, the statement concerns the
kernel at initialization, which is what these theorems characterize and what
the spectral premise of~\Cref{thm:linearPDEsconvergence} requires; the hidden
weights leave $\{-1,1\}$ under GD, and we make no claim about
$\mathbf K_{(k)}$ for $k\geq1$. Fourth, only the positive-dimensional support
uses the periodicity of $\sin$. For an arbitrary real-analytic nonpolynomial
$\sigma$, any $\gamma>0$, and any $d_0$, take $\mu_h=\delta_{\boldsymbol\xi_0}$
and $N\geq d_0+3$ Dirac rows at distinct sites: the limiting NTK is then a sum
of the $d_\theta=d_0+2$ rank-one Gram terms generated by the single parameter
value $\boldsymbol\xi_0$, so $\rank\mathbf K_{(0)}^\infty\leq d_0+2<N$ while
the rows remain independent. Empty interior, not the choice of activation, is
what the construction exploits.
\end{example}

The last example is of a different kind. It concerns~\Cref{assump:main} rather
than a ``pd'' clause, and it breaks existence of the limiting kernel rather
than its richness.

\begin{example}[A heavy-tailed initialization law destroys the deterministic limit]
\label{ex:heavy_tail_no_deterministic_limit}
The initialization clause of~\Cref{assump:main} asks three things of the
per-neuron law $\mu$: that it be i.i.d.\ across neurons, width independent,
and compactly supported. Delete only the third, so that $\mu$ is an arbitrary
width-independent law on $\mathbb R^{d_\theta}$, and retain every clause
of~\Cref{assump:pde_classes}. The existence half
of~\Cref{prop:limiting_ntk_rank}, equivalently statement one
of~\Cref{thm:linearNTK}, then fails already for a single value row.

Take $d_0=d_2=1$, $\gamma=1$, $\Omega=(0,1)$, and the one Dirichlet row
of~\Cref{def:linear_pde} with $\mathbf x_1=1$, $I_{11}=\{0\}$, coefficient
$a_{11\alpha}\equiv1$ at $\alpha=0$, and $s_1=1$, so that $N=1$,
$r_{\max}=0$ and $P_1(\theta)=u_\theta(1)$. Let $\sigma(z)=\exp(z^4)$ and
\begin{equation*}
 \theta_{m,(0)}=(w_m,b_m,a_m)
 \stackrel{\text{i.i.d.}}{\sim}\mathcal N(0,\mathbf I_3),
 \qquad
 t_m=w_m\mathbf x_1+\gamma b_m=w_m+b_m\sim\mathcal N(0,2).
\end{equation*}
Every hypothesis other than compact support holds: $\sigma$ is real analytic
on $\mathbb R$, hence lies in $C_{\mathrm{loc}}^{r_{\max}+1,1}(\mathbb R)$, and
is nonpolynomial; $\gamma>0$;
$\operatorname{int}(\supp\mu_h)=\mathbb R^2\neq\varnothing$; and $\mu$ is
i.i.d.\ across neurons and width independent. Thus $\mathsf P_{\mathrm{reg}}$
and $\mathsf P_{\mathrm{pd}}$ both hold, as does the structural centering
requirement $\mathbb E[a_m\mid\boldsymbol\xi_m]=0$ imposed for gradient
descent, since $a_m$ has mean zero and is independent of $\boldsymbol\xi_m$.
Only $\supp\mu=\mathbb R^3$ is noncompact.

By~\cref{eqn:classical_one_neuron_contribution} the per-neuron map
of~\cref{eqn:predictionmap} is $h_1(\boldsymbol\xi,a)=a\,\sigma(w+\gamma b)$,
so the output-weight entry of the single-neuron Jacobian $\boldsymbol\Psi$
defined in~\Cref{assump:general} is $\partial_ah_1=\sigma(t)$. Since $N=1$,
discarding the nonnegative contributions of the two hidden coordinates gives
\begin{equation*}
 [\mathbf K_{(0)}]_{11}
 =\frac1{d_1}\sum_{m=1}^{d_1}
 \bigl\|\boldsymbol\Psi(\theta_{m,(0)})\bigr\|_F^2
 \;\geq\;\frac1{d_1}\sum_{m=1}^{d_1}X_m,
 \qquad X_m=\exp\!\left(2t_m^4\right).
\end{equation*}
These i.i.d.\ nonnegative summands are not integrable: since $t^2\leq t^4$ for
$|t|\geq1$,
\begin{equation*}
 \mathbb E X_1
 =\frac1{\sqrt{4\pi}}\int_{\mathbb R}
   \exp\!\left(2t^4-\tfrac{t^2}4\right)\,dt
 \geq\frac1{\sqrt{4\pi}}\int_{|t|\geq1}
   \exp\!\left(\tfrac74t^4\right)\,dt
 =\infty,
\end{equation*}
the last integrand being at least $e^{7/4}$ on an unbounded set. Truncation
gives the one-line failure. For each $K\in\mathbb N$ the variables
$\min\{X_m,K\}$ are i.i.d.\ and bounded, so the strong law yields
$\liminf_{d_1\to\infty}[\mathbf K_{(0)}]_{11}\geq\mathbb E\min\{X_1,K\}$
almost surely for every $K$ simultaneously, while
$\mathbb E\min\{X_1,K\}\uparrow\mathbb E X_1=\infty$ by monotone convergence.
Hence $[\mathbf K_{(0)}]_{11}\to\infty$ almost surely along one coupled
i.i.d.\ sequence, and the same truncation with the weak law gives
$\Pr\{[\mathbf K_{(0)}]_{11}\leq M\}\to0$ for every $M$ under separate
resampling at each width, so both convergence modes asserted in statement one
fail. Moreover $\mathbb E[\mathbf K_{(0)}]_{11}=\infty$ at every finite width,
so no deterministic $\mathbf K_{(0)}^\infty$ exists and the Gram
representation in~\Cref{thm:linearNTK} is undefined.

Compact support is used at exactly one point in this chain. Statement one is
proved by~\Cref{thm:generallinearNTK} under the row bounds
of~\cref{eqn:abstract_one_neuron_bounds}, equivalently $(\mathsf G_R)$
in~\cref{eqn:gd_row_condition}, and for the pointwise class those bounds are
not assumed but derived: the proof of~\Cref{prop:linearNTK} extracts them from
compactness of $\supp\mu+\overline{\mathbb B}_R(0)$. Here
$\sup_{z\in\supp\mu}|h_1(z)|=\infty$, so the constant $A_{0,1}$ does not exist
and the derivation has no substitute, the conclusion itself being false. Note
that the divergence comes from the output-weight coordinate alone, which
involves no derivative of $\sigma$ and no differential order, so it is not an
artifact of activation smoothness or of high-order rows.

This sharpens~\Cref{ex:gaussian_moment_obstruction}, which records the
infinite population moment for the same activation: the point here is that the
empirical kernel itself diverges, so the conclusion fails and not merely its
proof, and that every remaining clause has been checked. Three limits should
be stated plainly. First, this is an existence failure and says nothing about
positive definiteness; the rank criteria of~\Cref{prop:limiting_ntk_rank} are
never reached because there is no limiting matrix to test. For the same reason
the identity $\mathbb E_{z\sim\mu}h_1(z)=0$ delivered by conditional centering
is vacuous here, $h_1$ not being $\mu$-integrable; this is a symptom of the
same nonintegrability rather than a second failed hypothesis. Second, the
example shows that the compact-support clause cannot be deleted, not that
compactness is necessary:
$\mathbb E_{z\sim\mu}\|\boldsymbol\Psi(z)\|_F^2<\infty$ would already suffice
for the deterministic limit, which is the relaxation
that~\Cref{ex:example1d} exploits. Compact support is retained because it
additionally supplies the essential-supremum constant $R_K$ for the
finite-width Chernoff bound and the uniform gradient-Lipschitz constant $L_*$
for the local stability bound, neither of which follows from moments alone.
Third, the obstruction is a joint incompatibility between the tail of $\mu$ and
the growth of $\sigma$, not a defect of Gaussian initialization: a Gaussian
$\mu$ with $\tanh$, or with any activation whose relevant derivatives grow
polynomially, leaves the per-neuron terms integrable and the limit intact.
\end{example}

Taken together, these six constructions show that no clause
of~\Cref{assump:main,assump:pde_classes} named above can simply be deleted, or
downgraded from analytic to smooth, while the remaining clauses are left as
written. They do not show that any clause is necessary in the logical sense.
The clauses are sufficient conditions chosen to be checkable at design time,
and in several cases a strictly weaker condition would carry the same proof:
density of $\{\varphi_{\boldsymbol\xi,q}:\boldsymbol\xi\in\supp\mu_h\}$ in
place of analyticity plus global density, $|\supp\mu_h|>0$ in place of
nonempty interior, and a finite second moment of $\boldsymbol\Psi$ in place of
compact support. Nor do the examples say anything quantitative: a feature
family that satisfies every clause but is poorly conditioned still yields a
positive definite kernel, and bounding its smallest eigenvalue
requires~\Cref{prop:quantitative_dntk_bound} rather than the exact nullspace
identities tested here.

\subsection{Taylor-coefficient conditions for linear independence}

The parameter-derivative functions depend on the PDE, initialization,
nonlinearity, architecture, and collocation data, so the linear-independence
condition is problem specific outside the hypotheses of the preceding rank
theorems. For bounded-order pointwise differential constraints, analytic
activations make them analytic in the neuron parameters.
\Cref{prop:analytic_functional_rank} states the exact condition
$\rank\mathbf C_K(\boldsymbol\xi_\star)=N$ for some finite $K$; the
calculation below evaluates the corresponding Taylor coefficients directly.
For weak and nonlocal rows,~\Cref{cor:monomial_moment_rank} gives the matrices
$\mathbf M_K^{\mathrm{weak}}$ and $\mathbf M_K^{\mathrm{nl}}$
in~\cref{eqn:weak_monomial_matrix,eqn:nonlocal_moment_matrix}.

\begin{example}[Positive definiteness for a Poisson PINN]
\label{ex:example1d}
Consider the following \(\gamma=0\) specialization
of~\Cref{def:ntk_parameterization}:
\[
u(x,\theta)=\frac{1}{\sqrt{d_1}}
\sum_{m=1}^{d_1}a_m\tanh(w_mx).
\]
Assume the marginal initialization law of
\(w_m\) has compact support containing an open neighborhood of zero (for
example, a compactly supported density that is positive near zero). Also take
the per-neuron parameter vectors to be i.i.d.\ and the output weights to
have finite second moment. These conditions ensure that the full
 limiting NTK exists; compact initialization as in~\Cref{assump:main} is
 sufficient. Use the PINN loss
in~\Cref{def:pinn_loss}
with inputs \(x=1,2,3,4,5\) for the one-dimensional Poisson problem
\begin{align*}
-\frac{d^2u}{dx^2}&=f(x),\quad x\in\Omega=(1,5),\\
u(1)&=0,\quad u(5)=0.
\end{align*}
In this bias-free model the scalar inputs are all multiples of one another,
so the distinct-direction condition of bias-free positivity results, which
requires that no two inputs be parallel, fails. Nevertheless,
\(\mathbf K_{(0)}^\infty\) is positive definite because an analytic
power-series coefficient calculation proves that the parameter-derivative
functions in the output-weight contribution are linearly independent. This is
the example-specific positive-definiteness certificate described in the pointwise
procedure of~\Cref{sec:pointwise_positivity_certificate}; see also
\hyperref[proof:example1d]{the detailed derivation}.
\end{example}

\begin{example}[Singular NTK with distinct constraints and zero nullspace projection]
\label{ex:compatible_singular_ntk}
Let \(\Omega=(-1,1)\) and use the network
in~\Cref{def:ntk_parameterization} with one-dimensional output and analytic
activation \(\sigma(z)=z+z^2\):
\begin{equation*}
u_\theta(x)=\frac{1}{\sqrt{d_1}}
\sum_{m=1}^{d_1}a_m\sigma(w_mx).
\end{equation*}
Because $\gamma=0$, the formal hidden-bias parameters $b_m$ do not affect
the predictions; their Jacobian columns are zero, so we omit them below.
Initialize \(a_m\) and \(w_m\) independently from the uniform distribution
\(\Unif[-1,1]\). This deliberately polynomial activation
violates the activation clause used by the differential-NTK positivity result.
More decisively, its limiting NTK is singular, so it admits no positive-definiteness
certificate and does not satisfy the corresponding premise of the convergence
theorem. Place one PDE collocation
point at \(x_D=0\) and the two Dirichlet points at \(x_-=-1\) and \(x_+=1\).
Every network output has the form \(A_\theta x+B_\theta x^2\), and therefore
\begin{equation}
u_\theta''(0)=u_\theta(-1)+u_\theta(1).
\label{eqn:compatible_feature_identity}
\end{equation}
Differentiate the identity in~\cref{eqn:compatible_feature_identity} with
respect to every trainable parameter and define
\[
\boldsymbol\Phi_D:=\nabla_\theta u_\theta''(0),\qquad
\boldsymbol\Phi_-:=\nabla_\theta u_\theta(-1),\qquad
\boldsymbol\Phi_+:=\nabla_\theta u_\theta(1).
\]
These gradients obey the relation
\[
\boldsymbol\Phi_D=\boldsymbol\Phi_-+\boldsymbol\Phi_+.
\]
Thus the NTK is singular despite the distinct physical points and the linear
independence of the two boundary parameter-derivative functions.
The data can obey exactly the same relation in the following well-posed
problem:
\[
u''(x)=2,\qquad u(-1)=u(1)=1.
\]
The unique solution is \(u^*(x)=x^2\), which the network represents
for every even \(d_1\), and its unweighted target vector \((2,1,1)^\top\) is
orthogonal to the NTK null vector \((1,-1,-1)^\top\). Consequently, the
fixed-kernel dynamics attain zero limiting loss despite singularity. The
sample-group normalization in~\Cref{def:pde} rescales both prediction
components and data and preserves rank and orthogonality to the
nullspace. This construction uses \(\gamma=0\) and a polynomial activation
to restrict the model and tangent spaces, so it lies
outside~\Cref{prop:limiting_ntk_rank} and the positive-definiteness-certificate premise
of~\Cref{thm:linearPDEsconvergence}.
See~\cref{sec:compatible_singular_ntk_calculation} for the full calculation.
\end{example}

The example shows that kernel rank and target compatibility are distinct:
positive definiteness permits every sampled target, whereas a singular system
can fit targets satisfying its nullspace relations.

\subsection{Constructive positive-definiteness certificates}
\label{sec:constructive_positivity_certificates}

We give finite positive-definiteness certificates and constructive search procedures
for turning the preceding nullspace theorems into checkable tests, using the
terminology of~\Cref{def:positive_definiteness_certificate}. In every procedure,
residual, initial, and boundary
rows are assembled in the order used by the loss.  The target values do not
enter an NTK rank calculation.  Positive loss scales preserve exact rank, but
they must be retained when conditioning or a quantitative eigenvalue bound is
reported.

\subsubsection{Strong-form pointwise constraints}
\label{sec:pointwise_positivity_certificate}

For each distinct physical site $\mathbf x\in\mathcal S$, form the finite
coefficient matrix
\begin{equation}
 \mathbf A_{\mathbf x}
 :=\left[s_i a_{iq\alpha}(\mathbf x)\right]_
 {\substack{i\in I_{\mathbf x}\\(q,\alpha)\in\mathcal C_{\mathbf x}}}.
 \label{eqn:pointwise_certificate_matrix}
\end{equation}
Its \(i\)-th row is \(\mathbf b_i(\mathbf x)^\top\). Hence, under the
hypotheses of~\Cref{prop:limiting_ntk_rank}, full row rank is exactly local
linear independence.
\paragraph{Certificate procedure.}
\begin{enumerate}[leftmargin=*,nosep]
\item Split every vector equation into scalar loss rows, and combine repeated
      occurrences of the same derivative $\partial^\alpha u_q$ at the same
      site.
\item At each site, assemble~\cref{eqn:pointwise_certificate_matrix} jointly
      over all PDE, initial, and boundary rows located there.
\item Exhibit a nonzero
      $|I_{\mathbf x}|\times |I_{\mathbf x}|$ minor of
      $\mathbf A_{\mathbf x}$ for every $\mathbf x\in\mathcal S$.
\end{enumerate}
These minors form a pointwise positive-definiteness certificate: they prove
$\rank\mathbf A_{\mathbf x}=|I_{\mathbf x}|$ and therefore
$\mathbf K_{(0)}^\infty\succ0$. Conversely, an exact nonzero vector in
$\ker\mathbf A_{\mathbf x}^{\top}$ is a structural NTK null vector.  Checking
rows separately is valid at distinct sites, but not for multiple rows at one
site.

When the sitewise hypotheses of~\Cref{prop:limiting_ntk_rank} are unavailable,
the analytic alternative in~\Cref{prop:analytic_functional_rank} applies
provided that the output weights are trainable, every output-weight feature
$F_{iq}$ is real analytic, the hidden law has compact support with nonempty
interior, and the full limiting NTK exists. Under these hypotheses, increase
$K$, assemble $\mathbf C_K(\boldsymbol\xi_\star)$, and exhibit one nonzero
$N\times N$ minor. Equivalently, one may exhibit a finite parameter set
$\Xi$ for which $\mathbf E_\Xi$ has row rank $N$. Failure at one Taylor order
or for one finite set $\Xi$ is inconclusive.

For the five-row Poisson PINN in~\Cref{ex:example1d}, the biased sitewise
theorem is unavailable because $\gamma=0$. The full-dimensional analytic
alternative is not invoked because this example is formulated only in the
active hidden parameter $w$ and assumes interior support in that reduced
parameter space. Its separate direct proof uses the coefficients of
$w,w^3,w^5,w^7,w^9$ in the two boundary features and three residual features.
The resulting $5\times5$ Taylor matrix, displayed in~%
\hyperref[proof:example1d]{the detailed Poisson proof}, has determinant
$6976569344/135\neq0$. This reduced-parameter Taylor argument is an exact
positive-definiteness certificate for that strong-form Poisson example.

\begin{example}[Exact pointwise coefficient certificate]
\label{ex:pointwise_coefficient_certificate}
Assume the network and initialization hypotheses of~%
\Cref{prop:limiting_ntk_rank}; for example, use a biased tanh network with
$\gamma=1$, i.i.d.\ compactly supported per-neuron initialization whose hidden
marginal support has nonempty interior, and bounded conditionally centered
output weights.
For this scalar network on $(0,1)$, place two marked pointwise rows at $x=0$:
\begin{equation*}
 \ell_1(f)=f(0)+f'(0),\qquad
 \ell_2(f)=f(0)-f'(0).
\end{equation*}
The second row can, for example, be read as a Robin boundary observation; the
first may be a co-located residual observation.  With unit row scales and
columns ordered as $(f(0),f'(0))$,
\begin{equation*}
 \mathbf A_0=
 \begin{pmatrix}1&1\\1&-1\end{pmatrix},
 \qquad \det\mathbf A_0=-2\neq0.
\end{equation*}
Thus the two co-located rows are jointly independent and
\Cref{prop:limiting_ntk_rank} proves that their limiting NTK is positive
definite.  Replacing the second row by a copy of the first makes the same
matrix rank one and produces the exact null vector $(1,-1)^\top$.
\end{example}

\subsubsection{Mixed weak residual and boundary constraints}
\label{sec:weak_positivity_certificate}

Concatenate the loss-scaled functionals as
\begin{equation*}
 \Lambda_r=
 \begin{cases}
  \lambda_r^\Omega,&r\in[N_\Omega],\\
  \beta_{r-N_\Omega}^\partial,&r=N_\Omega+1,\ldots,N,
 \end{cases}
 \qquad N=N_\Omega+N_\partial .
\end{equation*}
All rows must be bounded on one common trial space $\mathcal X$.  Under the
analytic hypotheses of~\Cref{cor:monomial_moment_rank}, construct
\begin{equation}
 \mathbf M_K^{\mathrm{weak,mix}}
 :=\left[\Lambda_r(\mathbf e_qp_\alpha)\right]_
 {\substack{r\in[N]\\q\in[d_2],\ |\alpha|\leq K}}
 =
 \begin{bmatrix}
  \left[s_i a_i(\mathbf e_qp_\alpha,v_i)\right]_{i,(q,\alpha)}\\[0.3em]
  \left[\beta_j^\partial(\mathbf e_qp_\alpha)\right]_{j,(q,\alpha)}
 \end{bmatrix}.
 \label{eqn:weak_mixed_monomial_certificate}
\end{equation}
\paragraph{Certificate procedure.}
\begin{enumerate}[leftmargin=*,nosep]
\item Put every residual and boundary row on one common trial space and prove
      that each row is bounded there.
\item For $K=0,1,\ldots$, assemble~%
      \cref{eqn:weak_mixed_monomial_certificate} in exact arithmetic and
      search its $N\times N$ minors.
\item Stop with a rigorous positive-definiteness certificate as soon as one
      minor is proved nonzero.  If the loss is a fixed quadrature rule, use
      the finite jet-atom coefficient matrix described below instead.
\end{enumerate}
By~\cref{eqn:monomial_rank_equivalence}, the nonzero minor is a positive-definiteness
certificate and is eventually equivalent to positive definiteness of the
mixed limiting NTK.
For a tested
boundary row
$\beta_j^\partial(f)=t_j\langle\mathcal Bf,\psi_j\rangle_{\partial\Omega}$,
the lower block is evaluated by applying that expression to
$\mathbf e_qp_\alpha$.  For a strong-form pointwise boundary row
$\beta_j^\partial(f)=t_j\sum_{q,\eta}b_{jq\eta}
\partial^\eta f_q(\mathbf x_j)$, written with $\eta$ for the boundary
multi-index because $\alpha$ indexes the monomial here, it is
\begin{equation*}
 \beta_j^\partial(\mathbf e_qp_\alpha)
 =t_j\sum_{\eta\in J_{jq}}b_{jq\eta}
       \partial^\eta p_\alpha(\mathbf x_j).
\end{equation*}

For a fixed quadrature loss, a smaller exact positive-definiteness certificate is often
available.
Collect all distinct jet atoms from both residual and boundary rows, write
\begin{equation*}
 \mathcal Q_a(f)=\partial^{\alpha_a}f_{q_a}(\mathbf x_a),
 \qquad \Lambda_r=\sum_{a=1}^A B_{ra}\mathcal Q_a,
\end{equation*}
and prove that the retained $\mathcal Q_a$ are independent in
$\mathcal X^*$.  Then~\cref{eqn:quadrature_coefficient_rank} gives the exact
finite test $\rank\mathbf B=N$.  In this finite representation, rank
deficiency is also conclusive.  Distinct triples
$(q_a,\mathbf x_a,\alpha_a)$ alone do not prove independence on an arbitrary
trial space; if the atoms have known relations, first express them in a basis
of their span and rank the resulting compressed coefficient matrix.

\begin{example}[Exact mixed weak residual--boundary certificate]
\label{ex:weak_mixed_monomial_certificate}
Let $\mathcal X=H^1(-1,1)$, take
$v(x)=1-x^2\in H_0^1(-1,1)$, and define one weak residual and two soft
Dirichlet rows by
\begin{equation*}
 \lambda^\Omega(f)=\int_{-1}^1 f'(x)v'(x)\,dx,
 \qquad \beta_-^\partial(f)=f(-1),
 \qquad \beta_+^\partial(f)=f(1).
\end{equation*}
For $p_0=1$, $p_1=x$, and $p_2=x^2$, the degree-two matrix in
\cref{eqn:weak_mixed_monomial_certificate} is
\begin{equation*}
 \mathbf M_2^{\mathrm{weak,mix}}
 =\begin{pmatrix}
  0&0&-8/3\\
  1&-1&1\\
  1&1&1
 \end{pmatrix},
 \qquad
 \det\mathbf M_2^{\mathrm{weak,mix}}=-\frac{16}{3}\neq0.
\end{equation*}
Endpoint evaluation is bounded on $H^1(-1,1)$, and the weak row is bounded by
Cauchy--Schwarz. Take the biased tanh network and compact initialization
specified in~\Cref{cor:continuum_weak_interface_poisson}. The explicit
$H^1$-valued analyticity and polynomial-density argument in the proof of that
corollary, together with the boundedness just checked, verifies the feature
hypotheses used here. The exact minor above and
\Cref{cor:monomial_moment_rank} therefore provide a positive-definiteness certificate
for the joint three-row limiting NTK without appealing to the
positive-definiteness conclusion of that corollary.
\end{example}

\subsubsection{Mixed nonlocal residual and boundary constraints}
\label{sec:nonlocal_positivity_certificate}

Absorb the positive row scales into one measure tuple per row:
\begin{equation*}
 \boldsymbol\mu_r=
 \begin{cases}
  s_r(\nu_{r1},\ldots,\nu_{rd_2}),&r\in[N_\Omega],\\
  t_j(\omega_{j1},\ldots,\omega_{jd_2}),
     &r=N_\Omega+j.
 \end{cases}
\end{equation*}
On compact $D$, form the mixed moment matrix
\begin{equation}
 \mathbf M_K^{\mathrm{nl,mix}}
 :=\left[\int_Dp_\alpha(\mathbf y)\,d\mu_{rq}(\mathbf y)\right]_
 {\substack{r\in[N]\\q\in[d_2],\ |\alpha|\leq K}}.
 \label{eqn:nonlocal_mixed_moment_certificate}
\end{equation}
\paragraph{Certificate procedure.}
\begin{enumerate}[leftmargin=*,nosep]
\item Encode every residual and boundary row as one loss-scaled measure tuple,
      combining repeated atoms before forming a matrix.
\item For finitely supported measures, rank the exact Dirac coefficient
      matrix below.  Otherwise, for $K=0,1,\ldots$, assemble~%
      \cref{eqn:nonlocal_mixed_moment_certificate} and search for a provably
      nonzero $N\times N$ minor.
\item A nonzero minor supplies a positive-definiteness certificate. An exact linear
      dependence of the complete measure tuples certifies singularity; a
      failed finite moment search is only inconclusive.
\end{enumerate}
By~\Cref{cor:monomial_moment_rank,cor:nonlocal_boundary_joint_positivity}, a
nonzero moment minor is an exact positive-definiteness certificate, and
one exists at some finite $K$ exactly when the combined residual and boundary
measure tuples are independent.

For finitely supported measures, take the union of all residual and boundary
nodes, combine repeated atoms, and write
$\mu_{rq}=\sum_{a=1}^Ac_{rqa}\delta_{\mathbf y_a}$.  The exact test then
reduces to
\begin{equation*}
 \rank\left[c_{rqa}\right]_{r,(q,a)}=N,
\end{equation*}
as in~\cref{eqn:nonlocal_dirac_coefficient_matrix}.  This includes a
Dirichlet row as a Dirac mass.  Neumann and Robin derivative observations are
not measures on $C(D)$ and must instead be placed in the weak-functional
procedure on a space where they are bounded.  If residual measures are
absolutely continuous and boundary measures are singular with respect to a
common reference measure, the final clause of
\Cref{cor:nonlocal_boundary_joint_positivity} permits separate checks;
otherwise the combined matrix is necessary.

\begin{example}[Exact mixed nonlocal residual--boundary certificate]
\label{ex:nonlocal_mixed_moment_certificate}
Let $D=[-1,1]$, $d_2=1$, and use the nonlocal residual and Dirichlet rows
\begin{equation*}
 \lambda^\Omega(f)=f(0)-\frac12\int_{-1}^1f(y)\,dy,
 \qquad \beta_-^\partial(f)=f(-1),
 \qquad \beta_+^\partial(f)=f(1).
\end{equation*}
Their scaled measures, with unit scales, are
$\delta_0-\tfrac12\mathrm{Leb}|_{[-1,1]}$, $\delta_{-1}$, and $\delta_1$.
Against $p_0=1$, $p_1=y$, and $p_2=y^2$,
\begin{equation*}
 \mathbf M_2^{\mathrm{nl,mix}}
 =\begin{pmatrix}
  0&0&-1/3\\
  1&-1&1\\
  1&1&1
 \end{pmatrix},
 \qquad
 \det\mathbf M_2^{\mathrm{nl,mix}}=-\frac{2}{3}\neq0.
\end{equation*}
Thus the three measure rows are jointly independent. Under the activation
and initialization hypotheses of~\Cref{thm:nonlocal_constraint_positivity},
the displayed nonzero minor is a positive-definiteness certificate for the joint
limiting NTK.
\end{example}

\subsubsection{Rigorous certificates and numerical diagnostics}
\label{sec:certificate_numerical_caveats}

A symbolic nonzero minor is an exact positive-definiteness certificate. A determinant enclosure
computed with outward-rounded interval arithmetic is also rigorous when the
enclosure excludes zero.  In contrast, floating-point rank, a sampled
smallest singular value, or a Monte Carlo Gram eigenvalue is only a numerical
diagnostic.  Raw monomials can be severely ill-conditioned; replacing them by
an exactly invertible polynomial basis, such as scaled orthogonal polynomials,
preserves rank.

For a diagnostic of the complete parameter-gradient family, draw independent
$z^{(1)},\ldots,z^{(M)}\sim\mu$ and form
\begin{equation*}
 \widehat{\mathbf F}_M\in\mathbb R^{N\times Md_\theta},
 \qquad
 [\widehat{\mathbf F}_M]_{i,(t,p)}=\partial_{z_p}h_i(z^{(t)}).
\end{equation*}
Then the normalized empirical Gram has smallest eigenvalue
$\sigma_{\min}(\widehat{\mathbf F}_M)^2/M$.  Report the sampling law, sample
count, row scales, arithmetic precision, and rank tolerance.  For a fixed
target, also report its projection onto the estimated nullspace.

Failure to find a full-rank truncated moment or Taylor matrix does not prove
singularity, because no universal truncation order is available.  Exact rank
deficiency of a complete finite coefficient representation, such as
$\mathbf A_{\mathbf x}$, $\mathbf B$, or the Dirac coefficient matrix, does
prove singularity. None of these qualitative positive-definiteness certificates by itself
lower-bounds $\lambda_{\min}(\mathbf K_{(0)}^\infty)$; the positive-measure
and feature-variation data in~\Cref{prop:quantitative_dntk_bound} are needed
for that purpose.  Finally, a quadrature rank certificate concerns the stored
discrete loss, and a separate consistency argument is required to connect it
to the continuum weak or nonlocal problem.

%% file: vectorvalued.tex
\section{Vector-valued pointwise constraints}
\label{sec:vector_appendix}

\begin{example}[Coupled linear advection--diffusion system]
\label{ex:vectorvalued}
Let \(\mathbf u=(u_1,u_2)^\top\) on \([0,1]^2\) satisfy
\begin{align*}
 \partial_t u_1+v_{\text{adv}}\partial_x u_1&=f_1, &
 \partial_t u_2-\nu_{\text{diff}}\partial_{xx}u_2-c_{\text{coup}}u_1&=f_2,
\end{align*}
where $v_{\text{adv}},c_{\text{coup}}\in\mathbb R$ and
$\nu_{\text{diff}}>0$.
Impose also initial conditions \(u_1(x,0),u_2(x,0)\) and noncorner side conditions
\(u_1(0,t),u_2(1,t)\), and sample each of the six scalar row families (the two residual rows, the two initial rows, and the two side rows) at pairwise distinct points; a point repeated within one family repeats a row, and the resulting parallel pair gives an exact null vector of \(\mathbf K_{(0)}\) at every width and of \(\mathbf K_{(0)}^\infty\) (\Cref{ex:duplicate_active_subspace}). In coordinates
\((u_1,\partial_x u_1,\partial_t u_1,\partial_{xx}u_2,\partial_t u_2)\), the
two interior rows at the same site are
\((0,v_{\text{adv}},1,0,0)\) and
\((-c_{\text{coup}},0,0,-\nu_{\text{diff}},1)\), and they are independent for
every admissible coefficient triple. At a shared physical location, the initial
constraints are the standard basis vectors in \((u_1,u_2)\), and each side
constraint is nonzero. \Cref{prop:limiting_ntk_rank} asks for independence
of the joint family at each site, not of the interior, initial, and side
families separately. At a site carrying interior and data rows, the rows present match distinct members of the four-row family below, and each $\mathbf b_i(\mathbf x)$ is $s_i>0$ times its member restricted to $\mathcal C_{\mathbf x}$, which drops only coordinates on which every row present vanishes; neither operation affects linear independence, so it suffices to check that family. Its common
coordinates are
\((u_1,\partial_xu_1,\partial_tu_1,u_2,\partial_{xx}u_2,\partial_tu_2)\) and
the rows \((0,v_{\text{adv}},1,0,0,0)\),
\((-c_{\text{coup}},0,0,0,-\nu_{\text{diff}},1)\), \((1,0,0,0,0,0)\), and
\((0,0,0,1,0,0)\) are independent: on the columns $(\partial_tu_1,\partial_tu_2,u_1,u_2)$ they form the unit upper triangular matrix $\begin{pmatrix}1&0&0&0\\0&1&-c_{\text{coup}}&0\\0&0&1&0\\0&0&0&1\end{pmatrix}$, so for every admissible coefficient triple the corresponding $4\times4$ minor of $\mathbf A_{\mathbf x}$, whose rows carry the loss scales, equals the product of the four positive scales $s_i$ and is the nonzero minor required by step~3 of the certificate procedure; the
word ``noncorner'' excludes exactly the corner sites at which an initial row
and a side row of the same component would coincide and be dependent. Thus
the local coefficient vectors are independent at
every site. Under~\Cref{assump:main} and the pointwise clauses of
\Cref{assump:pde_classes}, the rank result
in~\Cref{prop:limiting_ntk_rank} gives \(\mathbf K_{(0)}^\infty\succ0\). The
sitewise procedure in~\Cref{sec:pointwise_positivity_certificate} is therefore
an explicit positive-definiteness certificate for the sampled rows of this coupled pointwise PDE system. It neither requires nor establishes well-posedness of the continuum problem, which as stated is not well posed: \(u_2\) carries a side condition only at \(x=1\), so the problem does not determine \(u_2\) uniquely, and \(u_1(0,t)\) is an inflow condition only when \(v_{\text{adv}}>0\).
More generally, the same criterion applies to vector-valued pointwise
systems. Their limiting NTKs are positive definite exactly when the local
coefficient vectors $\mathbf b_i(\mathbf x)$ are linearly independent at every
sampled physical location; at a location carrying a single row this means only
that its coefficient vector is nonzero.
\end{example}

%% file: pd.tex
\section{Proofs and auxiliary results for NTK nullspaces and positive definiteness}
\label{sec:pd_proof}

\begin{proof}[Proof of~\Cref{lem:positive_definiteness}]
\phantomsection
\label{proof:positive_definiteness}
For every \(\mathbf c\in\mathbb R^N\), the Gram representation gives
\begin{equation*}
 \mathbf c^\top\mathbf K\mathbf c
 =\int\!\left(\sum_{i=1}^Nc_i\Phi_i(z)\right)^2d\mu(z)
 =\left\|\sum_{i=1}^Nc_i\Phi_i\right\|_{L^2(\mu)}^2.
\end{equation*}
Thus \(\mathbf K\succ0\) exactly when the \(\Phi_i\) are independent in
\(L^2(\mu)\). To obtain the support formulation, let
\(g_{\mathbf c}=\sum_i c_i\Phi_i\). If the continuous function
\(g_{\mathbf c}\) is nonzero at a point of \(\supp\mu\), it remains nonzero
on a neighborhood. By definition of the support, this open set has
positive \(\mu\)-measure. Hence \(g_{\mathbf c}=0\) almost surely if and only if it
vanishes on \(\supp\mu\).
\end{proof}

\begin{lemma}[Linear independence across all parameter contributions]
\label{lem:direct_sum_feature_criterion}
Let
\(\mathbf K=\sum_{\ell=1}^{L}\mathbf K^{(\ell)}\), where every
\(\mathbf K^{(\ell)}\) is a positive-semidefinite Gram matrix with
parameter-derivative functions \(\Phi_i^{(\ell)}\), indexed by the constraints, in a
Hilbert space \(\mathcal H_\ell\). Then
\begin{equation}
 \ker(\mathbf K)
 =\bigcap_{\ell=1}^{L}\ker(\mathbf K^{(\ell)})
 =\left\{\mathbf c:
   \sum_{i=1}^{N}c_i\Phi_i^{(\ell)}=0
   \text{ for every }\ell\right\}.
 \label{eqn:direct_sum_kernel_intersection}
\end{equation}
Consequently, \(\mathbf K\) is positive definite if and only if the
combined feature tuples
\(\left(\Phi_i^{(1)},\ldots,\Phi_i^{(L)}\right)\) are linearly independent
in \(\bigoplus_{\ell=1}^{L}\mathcal H_\ell\), equivalently if and only if
the intersection in~\cref{eqn:direct_sum_kernel_intersection} is trivial;
requiring one positive-semidefinite contribution \(\mathbf K^{(\ell)}\) to
be positive definite is sufficient but not necessary.
\end{lemma}
\begin{proof}
For every \(\mathbf c\in\mathbb R^N\),
\begin{equation*}
 \mathbf c^\top\mathbf K\mathbf c
 =\sum_{\ell=1}^{L}
   \left\|\sum_{i=1}^{N}c_i\Phi_i^{(\ell)}\right\|_{\mathcal H_\ell}^{2}.
\end{equation*}
All summands are nonnegative, so the sum is zero exactly when every summand
is zero, that is, exactly when \(\sum_{i=1}^{N}c_i\Phi_i^{(\ell)}=0\) for
every \(\ell\). For a positive-semidefinite matrix \(\mathbf M\),
\(\mathbf c^\top\mathbf M\mathbf c=\|\mathbf M^{1/2}\mathbf c\|_2^2\) vanishes
if and only if \(\mathbf M\mathbf c=\mathbf 0\); applied to \(\mathbf K\) and
to each \(\mathbf K^{(\ell)}\), whose quadratic form is the \(\ell\)th
summand, this gives both equalities
in~\cref{eqn:direct_sum_kernel_intersection}. Since \(\mathbf K\) is the Gram
matrix of the tuples \((\Phi_i^{(1)},\ldots,\Phi_i^{(L)})\) in
\(\bigoplus_{\ell=1}^{L}\mathcal H_\ell\) with the inner product
\(\sum_{\ell}\langle\cdot,\cdot\rangle_{\mathcal H_\ell}\),
\(\mathbf K\succ0\) exactly when these tuples are linearly independent,
equivalently when the intersection is trivial, and
\(\mathbf K\succeq\mathbf K^{(\ell)}\) shows that one positive-definite
contribution suffices. It is not necessary: for \(N=L=2\),
\(\mathcal H_1=\mathcal H_2=\mathbb R\), \(\Phi_1^{(1)}=\Phi_2^{(2)}=1\), and
\(\Phi_2^{(1)}=\Phi_1^{(2)}=0\), both \(\mathbf K^{(1)}=\operatorname{diag}(1,0)\)
and \(\mathbf K^{(2)}=\operatorname{diag}(0,1)\) are singular, while
\(\mathbf K=\mathbf I_2\).
\end{proof}

\begin{proof}[Proof of~\Cref{prop:quantitative_dntk_bound}]
\phantomsection
\label{proof:quantitative_dntk_bound}
Fix \(\|\mathbf c\|_2=1\). Recall that
\begin{gather*}
 \mathbf K_{\mathrm{out}}^\infty
 =\mathbb E_{\boldsymbol\xi\sim\mu_h}
 [\mathbf F(\boldsymbol\xi)\mathbf F(\boldsymbol\xi)^\top],\\
 \|\mathbf p+\mathbf p'\|_2^2\geq\tfrac12\|\mathbf p\|_2^2-\|\mathbf p'\|_2^2
 \quad\text{for all }\mathbf p,\mathbf p'\in\mathbb R^{d_2}.
\end{gather*}
Because the sets \(E_\ell\) are disjoint, these facts give
\begin{align*}
 \mathbf c^\top\mathbf K_{\mathrm{out}}^\infty\mathbf c
 &\geq\sum_{\ell=1}^{L}\int_{E_\ell}
   \|\mathbf F(\boldsymbol\xi)^\top\mathbf c\|_2^2\,d\mu_h(\boldsymbol\xi)\nonumber\\
 &\geq\frac12\sum_{\ell=1}^{L}\mu_h(E_\ell)
   \|\mathbf F(\boldsymbol\xi_\ell)^\top\mathbf c\|_2^2
   -\sum_{\ell=1}^{L}\int_{E_\ell}
   \|(\mathbf F(\boldsymbol\xi)-\mathbf F(\boldsymbol\xi_\ell))^\top\mathbf c\|_2^2
   \,d\mu_h(\boldsymbol\xi)\nonumber\\
 &\geq\frac{\beta}{2}\|\mathbf V\mathbf c\|_2^2
   -\varepsilon^2\sum_{\ell=1}^{L}\mu_h(E_\ell)
 \geq\frac{\beta c_{\mathrm{sep}}^2}{2}-\varepsilon^2.
\end{align*}
Taking the infimum over unit \(\mathbf c\), and using that the limiting NTK
contributions from the remaining trainable parameter groups are positive
semidefinite, proves~%
\cref{eqn:quantitative_dntk_bound}. The final claim follows by substitution.
\end{proof}

The next proposition restates the \(\mathsf P_{\mathrm{pd}}\) part
of~\Cref{prop:limiting_ntk_rank} (nullspace, rank, and positive-definiteness
criterion) together with the output-block
statement~\cref{eqn:exact_output_dntk_nullspace}, which the main text states
only in words, and its proof proves both; existence and convergence under
\(\mathsf P_{\mathrm{reg}}\) come from~\Cref{prop:linearNTK,thm:generallinearNTK},
as recorded in the proof of~\Cref{prop:limiting_ntk_rank} that follows.

\begin{proposition}[Exact limiting-NTK nullspace for a network with hidden biases]
\label{prop:exact_dntk_nullspace}
Use the network and pointwise constraints in~%
\Cref{def:ntk_parameterization,def:scalar_collocation_constraints}.
Assume~\Cref{assump:main} and the pointwise clauses of
\Cref{assump:pde_classes}, but do not assume that the evaluated operator
coefficient vectors at each location are linearly independent. Let
$\mathcal S=\{\mathbf x_i:i\in[N]\}$. For every \(\mathbf x\in\mathcal S\), let
\(I_{\mathbf x}=\{i:\mathbf x_i=\mathbf x\}\), and define the coordinate
index set
\begin{equation*}
 \mathcal C_{\mathbf x}
 =\left\{(q,\alpha):q\in[d_2],
   \alpha\in\bigcup_{i\in I_{\mathbf x}}I_{iq}\right\},
\end{equation*}
assigning zero to entries absent from an individual constraint on
\(\mathcal C_{\mathbf x}\). Define
\[
\mathbf b_i(\mathbf x)
=\bigl(s_i a_{iq\alpha}(\mathbf x)\bigr)_
{(q,\alpha)\in\mathcal C_{\mathbf x}},
\qquad
\mathcal N_{\mathbf x}
=\left\{\mathbf c_{\mathbf x}:
\sum_{i\in I_{\mathbf x}}c_i\mathbf b_i(\mathbf x)=\mathbf0\right\}.
\]
The partition
$[N]=\bigsqcup_{\mathbf x\in\mathcal S}I_{\mathbf x}$ canonically identifies
$\mathbb R^N$ with
$\bigoplus_{\mathbf x\in\mathcal S}\mathbb R^{I_{\mathbf x}}$.
Then the output-weight Gram block is well defined and
\begin{equation}
 \ker(\mathbf K_{\mathrm{out}}^\infty)
 =\bigoplus_{\mathbf x\in\mathcal S}\mathcal N_{\mathbf x},
 \qquad
 \rank(\mathbf K_{\mathrm{out}}^\infty)
 =\sum_{\mathbf x\in\mathcal S}
 \dim\linspan\{\mathbf b_i(\mathbf x):i\in I_{\mathbf x}\}.
 \label{eqn:exact_output_dntk_nullspace}
\end{equation}
The full limiting NTK \(\mathbf K_{(0)}^\infty\), formed from all trainable
parameters, exists and has exactly the same nullspace and rank:
\begin{equation}
 \ker(\mathbf K_{(0)}^\infty)
 =\ker(\mathbf K_{\mathrm{out}}^\infty),
 \qquad
 \rank(\mathbf K_{(0)}^\infty)
 =\sum_{\mathbf x\in\mathcal S}
 \dim\linspan\{\mathbf b_i(\mathbf x):i\in I_{\mathbf x}\}.
 \label{eqn:exact_full_dntk_nullspace}
\end{equation}
Thus, under these hypotheses, linear
independence of the evaluated operator coefficient vectors at every location
is necessary and sufficient for both Gram matrices to be positive definite.
\end{proposition}
\begin{proof}
For scalar constraint row $i$, output component $q$, and one hidden affine
parameter $\boldsymbol\xi=(\mathbf w,b)\sim\mu_h$, the parameter-derivative
function associated with the output-weight coordinate $a_q$ is the entry
$F_{iq}$ of the output feature matrix $\mathbf F=\boldsymbol\Phi_{\mathrm{out}}$:
\begin{equation*}
F_{iq}(\boldsymbol\xi)
=[\mathbf f_i(\boldsymbol\xi)]_q
=s_i(\mathcal A_{iq}\sigma_{\mathbf w,b})(\mathbf x_i),
\qquad
\sigma_{\mathbf w,b}(\mathbf x)=\sigma(\mathbf w^\top\mathbf x+\gamma b).
\end{equation*}
The $d_2$ output-weight families occupy disjoint coordinate blocks. Their
contribution to the limiting NTK is therefore
\begin{equation*}
[\mathbf K_{\mathrm{out}}^\infty]_{ij}
=\sum_{q=1}^{d_2}
\langle F_{iq},F_{jq}\rangle_{L^2(\mu_h)},
\end{equation*}
and, for every $\mathbf c\in\mathbb R^N$,
\begin{equation}
\mathbf c^\top\mathbf K_{\mathrm{out}}^\infty\mathbf c
=\sum_{q=1}^{d_2}
\left\|\sum_{i=1}^N c_iF_{iq}\right\|_{L^2(\mu_h)}^2.
\label{eqn:dntk_output_quadratic}
\end{equation}

The compact hidden support makes these parameter-derivative functions square-integrable,
so the display also proves existence of this block.
Suppose the quadratic form in~%
\cref{eqn:dntk_output_quadratic} is zero. For every output coordinate
$q$, define the real-analytic function
\begin{equation*}
G_{\mathbf c,q}(\mathbf w,b)=\sum_{i=1}^N c_i s_i
(\mathcal A_{iq}\sigma_{\mathbf w,b})(\mathbf x_i).
\end{equation*}
It vanishes $\mu_h$-almost surely. Continuity
extends this equality to $\supp(\mu_h)$, which contains a nonempty open
subset. Hence the
identity theorem for real-analytic functions
gives $G_{\mathbf c,q}\equiv0$ on $\mathbb R^{d_0+1}$. Because $\gamma>0$, the change of
variable $v=\gamma b$ ranges over all $v\in\mathbb R$.

Group the scalar constraints by their distinct physical sites. At a fixed
$\mathbf x$, the corresponding part of $G_{\mathbf c,q}$ is the following finite
sum of activation derivatives multiplied by monomials in the hidden weights:
\begin{equation*}
G_{\mathbf c,q,\mathbf x}(\mathbf w,v)
=\sum_{i:\,\mathbf x_i=\mathbf x}c_i s_i
\sum_{\alpha\in I_{iq}}a_{iq\alpha}(\mathbf x)
\sigma^{(|\alpha|)}(\mathbf w^\top\mathbf x+v)\mathbf w^\alpha.
\end{equation*}
After combining like terms, every formally nonzero $G_{\mathbf c,q,\mathbf x}$ is in
the standard form of~\citet{Jin2026}. Omitting trivial groups,
the independence lemma in~\Cref{lem:jinwu_sigma_independence} applies across
the remaining distinct sites. Since
\(\sum_{\mathbf x\in\mathcal S}G_{\mathbf c,q,\mathbf x}(\mathbf w,v)
=G_{\mathbf c,q}(\mathbf w,v/\gamma)=0\) for all
\((\mathbf w,v)\in\mathbb R^{d_0+1}\), and the omitted groups are formally
zero, the remaining groups satisfy a linear relation with all coefficients
equal to one; the lemma excludes such a relation unless no group remains.
Thus every grouped polynomial is formally zero, which means that, for every
$q$, $\alpha$, and $\mathbf x\in\mathcal S$,
\begin{equation*}
\sum_{i:\,\mathbf x_i=\mathbf x}c_i s_i
a_{iq\alpha}(\mathbf x)=0.
\end{equation*}
Equivalently, \(\mathbf c_{\mathbf x}\in\mathcal N_{\mathbf x}\) for every
\(\mathbf x\in\mathcal S\).
Conversely, by~\cref{eqn:dntk_output_quadratic} and
\(\mathbf K_{\mathrm{out}}^\infty\succeq0\),
\(\mathbf c\in\ker\mathbf K_{\mathrm{out}}^\infty\) as soon as every
\(G_{\mathbf c,q}\) vanishes, and, through the grouped expansion
\(G_{\mathbf c,q}(\mathbf w,b)
=\sum_{\mathbf x\in\mathcal S}G_{\mathbf c,q,\mathbf x}(\mathbf w,\gamma b)\),
these local coefficient cancellations make every \(G_{\mathbf c,q}\) vanish
identically.
The equivalence proves the first nullspace equality
in~\cref{eqn:exact_output_dntk_nullspace}. Rank--nullity supplies its rank
expression.

\Cref{prop:linearNTK,thm:generallinearNTK} give existence of the full limiting
NTK.
It remains to check that the other parameter coordinates neither remove nor
create dependencies caused by local operator coefficient relations. If
\(\mathbf c\in\bigoplus_{\mathbf x\in\mathcal S}\mathcal N_{\mathbf x}\), then for every
vector-valued function \(\mathbf u\) sufficiently differentiable through the
largest differential order among the constraints, and hence for every
network realization,
\begin{equation}
 \sum_{i=1}^{N}c_iP_i(\theta)
 =\sum_{\mathbf x\in\mathcal S}
 \sum_{(q,\alpha)\in\mathcal C_{\mathbf x}}
 \left(\sum_{i\in I_{\mathbf x}}c_i s_i
 a_{iq\alpha}(\mathbf x)\right)
 \partial^\alpha u_q(\mathbf x;\theta)=0.
 \label{eqn:local_operator_relation_predictions}
\end{equation}
Differentiating gives \(\mathbf J(\theta)^\top\mathbf c=0\) at every finite
width and every parameter value. Thus such a \(\mathbf c\) also belongs to
the kernel of the full limiting NTK whenever it exists.
Conversely, that matrix is the output-weight contribution plus
positive-semidefinite terms from the remaining trainable coordinates. Hence
its kernel is contained in
\(\ker(\mathbf K_{\mathrm{out}}^\infty)\). This argument establishes~%
\cref{eqn:exact_full_dntk_nullspace}. The final assertion follows because a
finite family is linearly independent exactly when its relation space is zero.
\end{proof}

\begin{proof}[Proof of~\Cref{prop:limiting_ntk_rank}]
\phantomsection
\label{proof:limiting_ntk_rank}
\Cref{prop:linearNTK,thm:generallinearNTK} give existence and empirical
convergence.
Under clause $\mathsf P_{\mathrm{pd}}$, apply
\Cref{prop:exact_dntk_nullspace}. Its local relation spaces and span dimensions
give~\cref{eqn:exact_dntk_rank}; the positive-definiteness criterion follows
by rank--nullity.
\end{proof}

\begin{proof}[Proof of~\Cref{thm:weak_constraint_positivity}]
\phantomsection
\label{proof:weak_constraint_positivity}
The complete one-neuron map
\[
\mathcal G(\boldsymbol\xi,\mathbf a)
=\sum_{q=1}^{d_2}a_q\varphi_{\boldsymbol\xi,q}
\]
is real analytic into \(\mathcal X\), being a finite sum of the coordinates
\(a_q\) times the analytic maps
\(\boldsymbol\xi\mapsto\varphi_{\boldsymbol\xi,q}\) of clause
\(\mathsf W_{\mathrm{pd}}\); for this step the \(C^2\) regularity of clause
\(\mathsf W_{\mathrm{reg}}\) already suffices. Compact initialization and
\Cref{lem:bounded_functional_row_regularity} therefore verify the condition
in~\Cref{assump:general}; \Cref{thm:generallinearNTK} proves existence of
\(\mathbf K_{(0)}^\infty\) and the stated empirical convergence.

Recall the relation map
\(\mathcal R_\lambda\mathbf c
=\sum_{i=1}^N c_i\lambda_i\in\mathcal X^*\) of the main text.
For the derivative with respect to output coordinate \(q\) of one neuron's
output weight, define
\begin{equation*}
 F_{iq}(\boldsymbol\xi)
 =\lambda_i(\varphi_{\boldsymbol\xi,q}).
\end{equation*}
Linearity of \(\lambda_i\) and the network parameterization give
\begin{equation*}
 \frac{\partial P_i(\theta)}
 {\partial (\mathbf a_m)_q}
 =\frac{1}{\sqrt{d_1}}F_{iq}(\boldsymbol\xi_m).
\end{equation*}
Consequently, the limiting output-weight contribution satisfies
\begin{align}
 [\mathbf K_{\mathrm{out}}^\infty]_{ij}
 &=\sum_{q=1}^{d_2}\int
 F_{iq}(\boldsymbol\xi)F_{jq}(\boldsymbol\xi)
 \,d\mu_h(\boldsymbol\xi),
 \label{eqn:weak_output_kernel}\\
 \mathbf c^\top\mathbf K_{\mathrm{out}}^\infty\mathbf c
 &=\sum_{q=1}^{d_2}\int
 \left|(\mathcal R_\lambda\mathbf c)
 (\varphi_{\boldsymbol\xi,q})\right|^2
 \,d\mu_h(\boldsymbol\xi).
 \label{eqn:weak_output_quadratic}
\end{align}
In particular, this matrix is positive semidefinite without any
independence assumption.

Suppose first that
\(\mathbf c\in\ker\mathbf K_{\mathrm{out}}^\infty\). Every nonnegative term in~%
\cref{eqn:weak_output_quadratic} is then zero, so, for every \(q\),
\begin{equation}
 g_{\mathbf c,q}(\boldsymbol\xi)
 =(\mathcal R_\lambda\mathbf c)(\varphi_{\boldsymbol\xi,q})=0
 \qquad\mu_h\text{-almost surely}.
 \label{eqn:weak_feature_relation_ae}
\end{equation}
The map into \(\mathcal X\) is real analytic by assumption, and composition
with the bounded functional \(\mathcal R_\lambda\mathbf c\) shows that
\(g_{\mathbf c,q}\) is a real-analytic scalar function. In particular it
is continuous. If it were nonzero at a point of
\(\supp(\mu_h)\), it would remain nonzero on a
neighborhood having positive \(\mu_h\)-measure, contradicting~%
\cref{eqn:weak_feature_relation_ae}. It therefore vanishes on the support,
and hence, since \(\operatorname{int}(\supp\mu_h)\neq\varnothing\) by clause
\(\mathsf W_{\mathrm{pd}}\), on a nonempty open set. The identity theorem for
real-analytic
functions gives
\begin{equation*}
 (\mathcal R_\lambda\mathbf c)(\varphi_{\boldsymbol\xi,q})=0
 \qquad
 \text{for every }\boldsymbol\xi\in\mathbb R^{d_0+1}
 \text{ and }q\in[d_2].
\end{equation*}
Thus \(\mathcal R_\lambda\mathbf c\) vanishes on the linear span of all coordinate
ridge features. This set is dense in \(\mathcal X\), so continuity of
\(\mathcal R_\lambda\mathbf c\) implies \(\mathcal R_\lambda\mathbf c=0\) on
\(\mathcal X\). This approximation argument proves
\(\ker\mathbf K_{\mathrm{out}}^\infty\subseteq\mathcal N_{\mathrm{weak}}\).
The reverse inclusion follows immediately from~%
\cref{eqn:weak_output_quadratic}; hence
\begin{equation}
 \ker\mathbf K_{\mathrm{out}}^\infty=\mathcal N_{\mathrm{weak}}.
 \label{eqn:weak_output_kernel_nullspace}
\end{equation}

It remains to compare the output-weight block with the total NTK.
The full limiting matrix equals this contribution plus positive-semidefinite
terms from the other coordinates.
Therefore
\(\ker\mathbf K_{(0)}^\infty\subseteq
\ker\mathbf K_{\mathrm{out}}^\infty\).
Conversely, if \(\mathbf c\in\mathcal N_{\mathrm{weak}}\), then at every finite
width and every parameter value,
\begin{equation*}
 \sum_{i=1}^N c_iP_i(\theta)
 =(\mathcal R_\lambda\mathbf c)(\mathbf u(\cdot;\theta))=0.
\end{equation*}
Differentiating this relation gives
\(\mathbf J(\theta)^\top\mathbf c=0\), so \(\mathbf c\) lies in the empirical
full NTK's kernel and hence in the kernel of its population limit.
Together with~\cref{eqn:weak_output_kernel_nullspace}, this argument proves both
nullspace identities in~\cref{eqn:weak_exact_ntk_nullspace}. Rank--nullity
then supplies
\begin{equation*}
 \rank\mathbf K_{(0)}^\infty
 =N-\dim\mathcal N_{\mathrm{weak}}
 =\dim\linspan\{\lambda_1,\ldots,\lambda_N\}.
\end{equation*}

Finally, in the coercive special case, a relation among the weak
functionals implies, with
\(v_{\mathbf c}=\sum_i c_i s_i v_i\),
\begin{equation*}
 a(u,v_{\mathbf c})=0\qquad\text{for every }u\in\mathcal X.
\end{equation*}
Taking \(u=v_{\mathbf c}\) and using coercivity yields
\(c_a\|v_{\mathbf c}\|_{\mathcal X}^2\leq0\), so
\(v_{\mathbf c}=0\). Linear independence of the \(v_i\), together with
\(s_i>0\), gives \(\mathbf c=0\). Thus
\(\mathcal N_{\mathrm{weak}}=\{0\}\), which proves the final assertion.
\end{proof}

\begin{proof}[Proof of~\Cref{thm:nonlocal_constraint_positivity}]
\phantomsection
\label{proof:nonlocal_constraint_positivity}
Because analyticity implies \(\sigma\in C_{\mathrm{loc}}^2\),
\Cref{prop:finite_measure_row_regularity} verifies the condition in
\Cref{assump:general}; \Cref{thm:generallinearNTK} proves existence of
\(\mathbf K_{(0)}^\infty\) and the stated empirical convergence.

For \(\boldsymbol\xi=(\mathbf w,b)\), use the scalar ridge feature
\(\sigma_{\mathbf w,b}(\mathbf y)
=\sigma(\mathbf w^\top\mathbf y+\gamma b)\)
of~\Cref{prop:quantitative_dntk_bound}, so that
\(\varphi_{\boldsymbol\xi,q}=\mathbf e_q\sigma_{\mathbf w,b}\)
in~\Cref{def:ntk_parameterization}, and define the output-weight features
\begin{equation*}
 F_{iq}(\boldsymbol\xi)
 =s_i\int_D\sigma_{\mathbf w,b}(\mathbf y)
 \,d\nu_{iq}(\mathbf y).
\end{equation*}
 Because \(D\) is compact and each measure has finite total variation, the
 integrals are finite. We next justify that
 \(\boldsymbol\xi\mapsto F_{iq}(\boldsymbol\xi)\) is real analytic. Fix
 \(\boldsymbol\xi_0=(\mathbf w_0,b_0)\). The affine arguments
 \(\mathbf w_0^\top\mathbf y+\gamma b_0\), \(\mathbf y\in D\), lie in a compact
 interval \(I\). A real-analytic function
 on \(\mathbb R\) has compatible local holomorphic extensions, and compactness
 gives a holomorphic extension \(\widetilde\sigma\) to an open set
 \(U\subset\mathbb C\) containing \(I\). There is then a neighborhood
 \(V\subset\mathbb C^{d_0+1}\) of \(\boldsymbol\xi_0\) such that every affine
 argument for \(\boldsymbol\xi\in V\) and \(\mathbf y\in D\) lies in \(U\).
 The resulting integrand is holomorphic in \(\boldsymbol\xi\) and uniformly
 bounded on compact subsets of \(V\). Dominated integration against the
 finite total-variation measure (alternatively, Fubini's theorem with Morera's
 theorem in each complex variable) preserves
 holomorphy near \(\boldsymbol\xi_0\). Restriction to real arguments proves
 the asserted analyticity. Because
 \(\boldsymbol\xi_0\) was arbitrary, the conclusion holds on all of
 \(\mathbb R^{d_0+1}\).

As in~\cref{eqn:weak_output_kernel}, differentiation with respect to the
output weights gives
\begin{align}
 [\mathbf K_{\mathrm{out}}^\infty]_{ij}
 &=\sum_{q=1}^{d_2}\int
 F_{iq}(\boldsymbol\xi)F_{jq}(\boldsymbol\xi)
 \,d\mu_h(\boldsymbol\xi),\nonumber\\
 \mathbf c^\top\mathbf K_{\mathrm{out}}^\infty\mathbf c
 &=\sum_{q=1}^{d_2}\int
 \left|\int_D\sigma_{\mathbf w,b}(\mathbf y)
 \,d\zeta_{\mathbf c,q}(\mathbf y)\right|^2
 \,d\mu_h(\boldsymbol\xi),
 \label{eqn:nonlocal_output_quadratic}
\end{align}
where
\begin{equation}
 \zeta_{\mathbf c,q}=\sum_{i=1}^N c_i s_i\nu_{iq}.
 \label{eqn:combined_nonlocal_measure}
\end{equation}
Thus this Gram matrix is always positive semidefinite.

Suppose \(\mathbf c\in\ker\mathbf K_{\mathrm{out}}^\infty\). For each \(q\),
define the real-analytic function
\begin{equation*}
 G_{\mathbf c,q}(\boldsymbol\xi)
 =\int_D\sigma_{\mathbf w,b}(\mathbf y)
 \,d\zeta_{\mathbf c,q}(\mathbf y).
\end{equation*}
It vanishes \(\mu_h\)-almost surely by~%
\cref{eqn:nonlocal_output_quadratic}. Continuity extends this equality to
\(\supp(\mu_h)\), and the support contains a nonempty open
set. The real-analytic identity theorem therefore gives
\begin{equation}
 \int_D\sigma(\mathbf w^\top\mathbf y+\gamma b)
 \,d\zeta_{\mathbf c,q}(\mathbf y)=0
 \qquad\text{for every }(\mathbf w,b)\in\mathbb R^{d_0+1}.
 \label{eqn:nonlocal_all_ridges_annihilated}
\end{equation}
Because \(\gamma>0\), the shifts \(\gamma b\) range over all real numbers.
The uniform universal-approximation theorem for a nonpolynomial activation
with a bias~\citep{Leshno1993} says that the linear span of the functions in~%
\cref{eqn:nonlocal_all_ridges_annihilated} is dense in \(C(D)\). Integration
against the finite signed measure \(\zeta_{\mathbf c,q}\) defines a bounded
functional on that space. It therefore vanishes on all of \(C(D)\), and
uniqueness in the Riesz--Markov representation theorem yields
\(\zeta_{\mathbf c,q}=0\). The conclusion holds for every \(q\), proving
\(\mathbf c\in\mathcal N_{\mathrm{nl}}\). Conversely, if
\(\mathbf c\in\mathcal N_{\mathrm{nl}}\), every combined measure in~%
\cref{eqn:combined_nonlocal_measure} is zero, and the quadratic form in~%
\cref{eqn:nonlocal_output_quadratic} is zero. Hence
\begin{equation}
 \ker\mathbf K_{\mathrm{out}}^\infty=\mathcal N_{\mathrm{nl}}.
 \label{eqn:nonlocal_output_kernel_nullspace}
\end{equation}

The full limiting NTK is a sum of positive-semidefinite parameter-coordinate
contributions, so its kernel is contained in the output-weight kernel. In
the reverse direction, \(\mathbf c\in\mathcal N_{\mathrm{nl}}\) implies, at
every finite width and every parameter value \(\theta\),
\begin{equation*}
 \sum_{i=1}^N c_iP_i(\theta)
 =\sum_{q=1}^{d_2}\int_D u_q(\mathbf y;\theta)
 \,d\zeta_{\mathbf c,q}(\mathbf y)=0.
\end{equation*}
Differentiating gives \(\mathbf J(\theta)^\top\mathbf c=0\). Thus
\(\mathcal N_{\mathrm{nl}}\subseteq\ker\mathbf K_{(0)}^\infty\), and~%
\cref{eqn:nonlocal_output_kernel_nullspace} proves the two nullspace
identities in~\cref{eqn:nonlocal_exact_ntk_nullspace}. Rank--nullity yields
the displayed rank formula, and the positive-definiteness criterion follows.
\end{proof}

\subsection{Compatible training error for a singular NTK}
\label{sec:compatible_singular_ntk_calculation}
\begin{proof}[Calculation for \Cref{ex:compatible_singular_ntk}]
Before the normalization assigned to each loss group in~%
\Cref{def:pde}, set
$\widetilde P_D(\theta)=u_\theta''(0)$,
$\widetilde P_-(\theta)=u_\theta(-1)$, and
$\widetilde P_+(\theta)=u_\theta(1)$. For a scalar-output neuron with weights
$(a_m,w_m)$, define its width-rescaled parameter-gradient
blocks by
$\boldsymbol\phi_\bullet(a_m,w_m)
:=\sqrt{d_1}\,\nabla_{(a_m,w_m)}\widetilde P_\bullet(\theta)$.
The hidden-bias derivative is zero because $\gamma=0$, so we omit that
coordinate below. Writing a generic block as $(a,w)$ gives
\begin{align*}
\boldsymbol\phi_D(a,w)&=(2w^2,4aw),\\
\boldsymbol\phi_-(a,w)&=(-w+w^2,a(-1+2w)),\\
\boldsymbol\phi_+(a,w)&=(w+w^2,a(1+2w)).
\end{align*}
Hence $\boldsymbol\phi_D=\boldsymbol\phi_-+\boldsymbol\phi_+$ pointwise for
every $(a,w)$, separately for the
output-weight and hidden-weight coordinates. This identity is therefore a linear
dependence among the full parameter-gradient vectors used by the NTK, rather
than a dependence confined to the contribution from one parameter group.

For independent $a,w\sim\Unif[-1,1]$, direct integration gives, in
the order $(D,-,+)$, the unweighted population Gram matrix
\begin{equation*}
\widetilde{\mathbf K}_{(0)}^\infty=\frac{1}{45}
\begin{pmatrix}
116&58&58\\
58&59&-1\\
58&-1&59
\end{pmatrix}.
\end{equation*}
Its eigenvalues are $0$, $4/3$, and $58/15$, and its nullspace is spanned by
$\mathbf n=(1,-1,-1)^\top$. In particular, the two boundary parameter-gradient
vectors are linearly independent, and the displayed identity is the only
dependence among the three parameter-gradient vectors.
The target $\widetilde{\mathbf v}_T=(2,1,1)^\top$ satisfies
$\mathbf n^\top\widetilde{\mathbf v}_T=0$. Every network prediction obeys the
same identity, so every initial error is also orthogonal to $\mathbf n$ and
belongs to $\Range(\widetilde{\mathbf K}_{(0)}^\infty)$; the loss dynamics with
the kernel held fixed at $\widetilde{\mathbf K}_{(0)}^\infty$ therefore contract to zero for
a suitable step size.

The continuum compatibility is not merely a relation among the three target
entries. The Dirichlet problem has the unique solution $u^*(x)=x^2$. For every
even width, the network represents it exactly by taking half of the hidden weights
equal to $1$, half equal to $-1$, and all output weights equal to
$d_1^{-1/2}$, because
\begin{equation*}
\frac12\left(\sigma(x)+\sigma(-x)\right)=x^2.
\end{equation*}
Finally, the residual group has $N_r=1$ row and the boundary group has two
rows, so the normalization in~%
\Cref{def:pde} multiplies the last two prediction components and targets
by $1/\sqrt2$. Define the rescaled features by
$\boldsymbol\phi_D^{\mathrm{norm}}:=\boldsymbol\phi_D$ and
$\boldsymbol\phi_\pm^{\mathrm{norm}}:=\boldsymbol\phi_\pm/\sqrt2$, so
$\boldsymbol\phi_D^{\mathrm{norm}}
=\sqrt2(\boldsymbol\phi_-^{\mathrm{norm}}
+\boldsymbol\phi_+^{\mathrm{norm}})$. The null
vector becomes $(1,-\sqrt2,-\sqrt2)^\top$, while both the normalized target
and every normalized prediction remain orthogonal to it.
\end{proof}

\begin{lemma}[Target projection for an architecture linear in its output weights]
\label{lem:output_linear_target_projection}
For the network in~\Cref{def:ntk_parameterization} and linear prediction
components under the applicable ``reg'' clause
of~\Cref{assump:pde_classes}, which makes the Jacobian and the tangent Gram
matrix defined at every parameter value, let \(\mathbf J_{\mathrm{out}}(\theta)\) be the columns of the full
Jacobian associated with the output weights, and set
$\mathbf a_{\mathrm{stack}}=(\mathbf a_1^\top,\ldots,
\mathbf a_{d_1}^\top)^\top$. At every finite width and
parameter value,
\begin{equation*}
 \mathbf v_P(\theta)
 =\mathbf J_{\mathrm{out}}(\theta)\mathbf a_{\mathrm{stack}},
 \qquad
 \mathbf v_P(\theta)\in\Range(\mathbf K(\theta)).
\end{equation*}
Consequently, let \(\Pi_{\ker\mathbf K(\theta)}\) denote the orthogonal
projector onto \(\ker\mathbf K(\theta)\); then
\begin{equation}
 \Pi_{\ker\mathbf K(\theta)}
   \left(\mathbf v_T-\mathbf v_P(\theta)\right)
 =\Pi_{\ker\mathbf K(\theta)}\mathbf v_T.
 \label{eqn:target_residual_projection_identity}
\end{equation}
For a limiting-kernel nullspace generated by the local operator coefficient
relations in~%
\Cref{prop:exact_dntk_nullspace}, the same conclusion holds because~%
\cref{eqn:local_operator_relation_predictions} annihilates every network prediction.
Without either structural guarantee, the generally valid formula projects the
initial training error rather than the target alone.
\end{lemma}
\begin{proof}
Linearity of every prediction component
in the network output makes the prediction vector homogeneous and linear in
$\mathbf a_{\mathrm{stack}}$; its coefficient matrix is
exactly \(\mathbf J_{\mathrm{out}}(\theta)\). Moreover,
\(\Range(\mathbf J(\theta))=
\Range(\mathbf J(\theta)\mathbf J(\theta)^\top)=
\Range(\mathbf K(\theta))\). Hence the prediction has zero
projection onto \(\ker\mathbf K(\theta)\), which proves~%
\cref{eqn:target_residual_projection_identity}. The claim for a limiting
nullspace forced by local operator relations follows directly from~%
\cref{eqn:local_operator_relation_predictions}.
\end{proof}

\subsection{Poisson proof by linear independence of parameter-derivative functions}

\begin{proof}
    \phantomsection
    \label{proof:example1d}
    The stated marginal-law hypothesis covers the Gram contribution studied
    below. To establish the full limiting NTK using
    all trainable parameters, use the example's further hypotheses that the
    per-neuron parameter vectors are i.i.d.\ across neurons and that the output
    weights have finite second moment;~\Cref{assump:main} is one sufficient
    choice. These hypotheses
    make every individual Gram contribution integrable, so the strong-law
    argument in the \hyperref[proof:ntk_deterministic]{deterministic-NTK proof}
    gives existence.
    We study the parameter-derivative functions associated with $a_m$, whose
    linear independence follows from the invertibility
    of the coefficient matrix constructed from their Taylor coefficients.
    Because nonzero row scaling preserves this property, divide by the
    positive loss-row scales and multiply the three Poisson-residual features
    by $-1$. Denote $w=w_m$. The resulting derivatives fall into two groups.
    The boundary pair is \(g_1(w)=\tanh(w)\) and \(g_2(w)=\tanh(5w)\) on the
    support of the hidden-weight law. For the interior constraints, define
    \(\chi_x(w)=w^2\tanh^{(2)}(xw)\), where \(\tanh^{(2)}\) is the second
    derivative of the hyperbolic tangent. The three collocation sites then
    give \(g_3=\chi_2\), \(g_4=\chi_3\), and \(g_5=\chi_4\) on the support of
    the hidden-weight law.
    
    All five parameter-derivative functions are analytic in \(w\).
  Because, by the assumption in~\Cref{ex:example1d}, the support of the
  hidden-weight law contains an open neighborhood of $w=0$, if a linear
  combination \(\sum_{i=1}^5c_ig_i\) vanishes on the support, then every
  coefficient of its power series at \(w=0\) vanishes. The two boundary-feature series are
  \begin{align*}
    \tanh(w) &= w - \frac{w^3}{3} + \frac{2w^5}{15} - \frac{17w^7}{315} + \frac{62w^9}{2835} + O(w^{11}), \\
    \tanh(5w) &= 5w - \frac{125w^3}{3} + \frac{6250w^5}{15} - \frac{1328125w^7}{315} + \frac{121093750w^9}{2835} + O(w^{11}).
  \end{align*}
  The corresponding interior expansions are
  \begin{align*}
    \chi_2(w) &= -4w^3 + \frac{64w^5}{3} - \frac{1088w^7}{15} + \frac{63488w^9}{315} + O(w^{11}), \\
    \chi_3(w) &= -6w^3 + 72w^5 - \frac{2754w^7}{5} + \frac{120528w^9}{35} + O(w^{11}), \\
    \chi_4(w) &= -8w^3 + \frac{512w^5}{3} - \frac{34816w^7}{15} + \frac{8126464w^9}{315} + O(w^{11}).
  \end{align*}
  Setting to zero the coefficients of \(w,w^3,w^5,w^7,w^9\) in
  \(\sum_{i=1}^5c_ig_i\) (every even coefficient vanishes because each
  \(g_i\) is odd) produces a five-equation system. The
  coefficients of $w$ and $w^3$ give
  \begin{align*}
      c_1 + 5c_2 &= 0, \\
      -\frac{c_1}{3} - \frac{125c_2}{3} - 4c_3 - 6c_4 - 8c_5 &= 0.
  \end{align*}
  The remaining powers yield
  \begin{align*}
      \frac{2c_1}{15} + \frac{6250c_2}{15} + \frac{64c_3}{3} + 72c_4 + \frac{512c_5}{3} &= 0, \\
      -\frac{17c_1}{315} - \frac{1328125c_2}{315} - \frac{1088c_3}{15} - \frac{2754c_4}{5} - \frac{34816c_5}{15} &= 0, \\
      \frac{62c_1}{2835} + \frac{121093750c_2}{2835} + \frac{63488c_3}{315} + \frac{120528c_4}{35} + \frac{8126464c_5}{315} &= 0.
  \end{align*}
  The coefficient matrix of the combined system has determinant
  $6976569344/135\neq0$. Therefore it has the unique solution
  $\mathbf{c}=\mathbf0$, and the five
  parameter-derivative functions are linearly independent as functions of $w$.
  Hence the population output-weight contribution is nonsingular. Under the
  finite-moment condition above, the full limiting NTK
  exists and is positive definite because it is the sum of this block and
  positive-semidefinite terms from the other parameters.
\end{proof}

%% file: optimization_aux.tex
\section{Auxiliary optimization results}

\subsection{Initialization error}

\begin{lemma}[Initial error bound]
\label{lem:initial_residual_bound}
Under~\Cref{assump:main}, suppose $(\mathsf G_R)$ holds and
\(\mathbb E[\mathbf a\mid\boldsymbol\xi]=\mathbf0\), set
\(\mathbf h(z)=(h_i(z))_{i=1}^N\),
\(H_0^2=\mathbb E_{z\sim\mu}\|\mathbf h(z)\|_2^2\leq A_{0,*}^2\), and
\(\mathbf r_{(0)}=\mathbf v_T-\mathbf v_P(\theta_{(0)})\). Every initialization
realization whose per-neuron parameter vectors lie in the initialization
support (and hence almost every realization) satisfies
\begin{equation}
\|\mathbf r_{(0)}\|_2
\leq \|\mathbf v_T\|_2+\sqrt{d_1}\,A_{0,*}.
\label{eqn:deterministic_r0_bound}
\end{equation}
Moreover, for every \(0<\varepsilon<1\),
\begin{equation*}
\Pr\!\left\{
\|\mathbf r_{(0)}\|_2
\leq \|\mathbf v_T\|_2
+H_0+A_{0,*}\sqrt{2\log\frac1\varepsilon}
\right\}\geq1-\varepsilon.
\end{equation*}
\end{lemma}

\Cref{lem:initial_residual_bound} is proved at the start
of~\Cref{sec:convergenceproof}. By~\cref{eqn:deterministic_r0_bound}, every
realized \(\mathbf r_{(0)}\) is finite under bounded initialization,
but the direct deterministic estimate can grow as \(\sqrt{d_1}\). Centering
the output weights gives the width-independent high-probability control used
in~\Cref{thm:linearPDEsconvergence}. For a fixed number of loss groups, suppose
each row is divided by the square root of its group's sample count. Under
uniform bounds on the unnormalized per-neuron prediction terms and target
values, \(A_{0,*}\) and \(\|\mathbf v_T\|_2\) remain \(O(1)\), uniformly in the
group sample counts, as the
discretization is refined.

\subsection{Structural dependencies and compatible singular kernels}
\label{sec:compatible_singular_kernels}

\begin{definition}[Structural spaces and compatibility]
\label{def:structural_spaces_compatibility}
For the pointwise, weak, and finite-measure classes, let
$\mathcal X_{\mathrm{row}}$ be, respectively,
$C^{r_{\max}}(\mathbb R^{d_0};\mathbb R^{d_2})$, the stated trial space
$\mathcal X$, or $C(D;\mathbb R^{d_2})$, and let
$\lambda_i:\mathcal X_{\mathrm{row}}\to\mathbb R$ denote the $i$th
loss-scaled linear row, so that $P_i(\theta)=\lambda_i(\mathbf u(\cdot;\theta))$
(for weak and finite-measure rows this is $\lambda_i$
of~\Cref{def:linear_pde}, that is, $s_i\ell_i$ in the notation
of~\Cref{thm:weak_constraint_positivity,thm:nonlocal_constraint_positivity}).
Define
\begin{align*}
 \mathcal O&:\mathcal X_{\mathrm{row}}\to\mathbb R^N,
 & [\mathcal O f]_i&=\lambda_i(f),\\
 \mathcal N_{\mathrm{str}}
 &=\left\{\mathbf c\in\mathbb R^N:
 \sum_{i=1}^Nc_i\lambda_i=0\text{ on }\mathcal X_{\mathrm{row}}\right\},
 & \mathcal U_{\mathrm{str}}&=\mathcal N_{\mathrm{str}}^\perp.
\end{align*}
A target is \emph{compatible} if
$\mathbf v_T\in\mathcal U_{\mathrm{str}}$, equivalently if it is orthogonal to
every exact row relation. A \emph{structural-nullspace certificate} proves
$\ker\mathbf K_{(0)}^\infty=\mathcal N_{\mathrm{str}}$. For a nonzero
prescribed active subspace $\mathcal U$ with orthonormal basis matrix
$\mathbf U$, a \emph{compressed positive-definiteness certificate} verifies
\mbox{$\mathbf U^\top\mathbf K_{(0)}^\infty\mathbf U\succ0$}.
\end{definition}

\begin{lemma}[Compatible targets and exact compression]
\label{lem:structural_compatibility_compression}
For the structural spaces in~\Cref{def:structural_spaces_compatibility},
under the applicable ``reg'' clause of~\Cref{assump:pde_classes}, so that
$\mathbf u(\cdot;\theta)\in\mathcal X_{\mathrm{row}}$ for every $\theta$,
$\Range(\mathcal O)=\mathcal U_{\mathrm{str}}$. In particular, if one
admissible field $f_\star\in\mathcal X_{\mathrm{row}}$ generates all targets,
$\mathbf v_T=\mathcal O f_\star$, then the target is compatible. Moreover,
every network prediction lies in $\mathcal U_{\mathrm{str}}$ and
\begin{equation*}
 \mathcal N_{\mathrm{str}}\subseteq\ker\mathbf K(\theta)
 \qquad\text{for every }\theta.
\end{equation*}
If a structural-nullspace certificate holds,
$\mathcal U_{\mathrm{str}}\neq\{\mathbf0\}$, and $\mathbf U$ is an orthonormal
basis matrix for $\mathcal U_{\mathrm{str}}$, then
$\mathbf U^\top\mathbf K_{(0)}^\infty\mathbf U\succ0$, with smallest
eigenvalue equal to the smallest positive eigenvalue of
$\mathbf K_{(0)}^\infty$. Whether or not a structural-nullspace certificate
holds, for every target, every $\theta$, and every orthonormal basis matrix
$\mathbf U$ of $\mathcal U_{\mathrm{str}}$, the loss decomposes exactly as
\begin{equation}
 \mathcal L(\theta)
 =\|\mathbf U^\top(\mathbf v_T-\mathbf v_P(\theta))\|_2^2
  +\|\Pi_{\mathcal N_{\mathrm{str}}}\mathbf v_T\|_2^2,
 \qquad
 \mathbf K_{\mathcal U}(\theta)
 =\mathbf U^\top\mathbf K(\theta)\mathbf U.
 \label{eqn:structural_loss_compression}
\end{equation}
The second term is an irreducible constant and vanishes exactly for compatible
targets. Hence, for a compatible target, the full residual belongs to
$\mathcal U_{\mathrm{str}}$ and the original loss is exactly the compressed
loss; for an incompatible target, gradient descent still optimizes the same
active loss but cannot remove the nullspace term.
If $\mathcal U_{\mathrm{str}}=\{\mathbf0\}$, the loss decomposition still
holds, but there is no positive active eigenvalue and no nontrivial compressed
GD problem.
\end{lemma}
\begin{proof}
The definition of $\mathcal N_{\mathrm{str}}$ gives
$\mathcal N_{\mathrm{str}}=\Range(\mathcal O)^\perp$; the range is a subspace
of finite-dimensional $\mathbb R^N$ and hence is closed, proving the first
identity. Taking $f=f_\star$ proves compatibility of a jointly generated
target, while taking $f=\mathbf u(\cdot;\theta)$ proves that every prediction
is active. Differentiating
$\mathbf c^\top\mathbf v_P(\theta)=0$ gives
$\mathbf J(\theta)^\top\mathbf c=0$, and hence the empirical-kernel
inclusion. If the population kernel has kernel
$\mathcal N_{\mathrm{str}}$, its quadratic form is strictly positive on the
orthogonal complement, which proves compressed positive definiteness. The
spectral assertion follows by restricting the eigendecomposition to the
positive eigenspace. Since every prediction is active, orthogonal decomposition
of the residual gives~\cref{eqn:structural_loss_compression}; its nullspace
component is $\Pi_{\mathcal N_{\mathrm{str}}}\mathbf v_T$. Finally, the
compressed Jacobian is $\mathbf U^\top\mathbf J(\theta)$, giving the final
kernel identity.
\end{proof}

The nonsingular result in~\Cref{thm:linearPDEsconvergence} yields contraction
for arbitrary data. For a singular design,
\Cref{lem:structural_compatibility_compression} gives
$\mathcal N_{\mathrm{str}}\subseteq\ker\mathbf K_{(0)}^\infty$ under the
applicable ``reg'' clause of~\Cref{assump:pde_classes}, and under the
corresponding ``pd'' clause
\Cref{prop:limiting_ntk_rank,prop:functional_ntk_rank,prop:nonlocal_ntk_rank}
identify $\ker\mathbf K_{(0)}^\infty$ with $\mathcal N_{\mathrm{str}}$, which
is a structural-nullspace certificate. Under a ``reg'' clause alone the
inclusion can be strict (\Cref{ex:weak_analyticity_necessity}), and the
compressed kernel on $\mathcal U_{\mathrm{str}}$ is then singular.
Target compatibility is a separate data condition, not a consequence of
kernel singularity; by~\Cref{lem:structural_compatibility_compression}, it is
automatic when one admissible exact field generates all rows of the target and
otherwise must be checked as
$\mathbf v_T\perp\mathcal N_{\mathrm{str}}$. The same lemma proves that the
loss and NTK then compress exactly to
$\mathcal U_{\mathrm{str}}=\mathcal N_{\mathrm{str}}^\perp$ and, when this
space is nontrivial and a structural-nullspace certificate holds, that the
population compressed kernel is positive definite. Under the corresponding compressed regularity, width, and step-size
conditions, \Cref{cor:active_subspace_convergence} therefore gives the same
geometric loss decay even though the full kernel is singular. In the rank-zero
edge case, a compatible target has identically zero residual, so there is no
optimization problem.

\begin{corollary}[Convergence when all errors lie in a fixed subspace]
\label{cor:active_subspace_convergence}
Fix \(0<\delta<1\). Let \(\mathcal U\subseteq\mathbb R^N\) be a fixed
\(r\)-dimensional subspace, where $1\leq r\leq N$, and let
\(\mathbf U\in\mathbb R^{N\times r}\) have orthonormal columns that span
\(\mathcal U\). Suppose
\begin{equation}
\mathbf v_T-\mathbf v_P(\theta)\in\mathcal U
\qquad\text{for every }\theta\text{ in an open neighborhood of }
\mathcal T_R(\theta_{(0)}).
\label{eqn:fixed_compatible_subspace}
\end{equation}
Set \(\mathbf v_{T,\mathcal U}=\mathbf U^\top\mathbf v_T\), assume
\Cref{assump:main} and
\(\mathbb E[\mathbf a\mid\boldsymbol\xi]=\mathbf0\), and require
$(\mathsf G_R)$ for the compressed prediction
\(\mathbf U^\top\mathbf v_P\). Form its per-neuron regularity constants and
the corresponding initial-residual bound \(\overline r_{\delta,\mathcal U}\),
with target \(\mathbf v_{T,\mathcal U}\), as in~\Cref{thm:linearPDEsconvergence}.
Also define
\(\mathbf K_{\mathcal U}^\infty=\mathbf U^\top\mathbf K_{(0)}^\infty\mathbf U\)
and require a compressed positive-definiteness certificate as in
\Cref{def:structural_spaces_compatibility}, namely
\(\lambda_{\mathcal U}=\lambda_{\min}(\mathbf K_{\mathcal U}^\infty)>0\).
In the width and step-size conditions of~\Cref{thm:linearPDEsconvergence},
replace
\(\lambda_0,\Lambda,R_K,A_{0,*},H_0,B_*,L_*\) by their compressed
counterparts, \(\overline r_\delta\) by
\(\overline r_{\delta,\mathcal U}\), and \(N\) by \(r\) in the Chernoff
logarithm. Assume that the resulting inequalities hold. Then, with probability at least
\(1-\delta\), every iterate lies in
\(\mathcal T_R(\theta_{(0)})\) and, for every \(k\in\mathbb N_0\),
\begin{equation*}
\mathcal L(\theta_{(k)})
\leq(1-\eta\lambda_{\mathcal U})^k\mathcal L(\theta_{(0)}).
\end{equation*}
The full \(N\times N\) kernel may be singular. In particular, let
$\mathcal N=\mathcal N_{\mathrm{str}}$ and assume the applicable ``pd''
clause of~\Cref{assump:pde_classes}, so that
$\ker\mathbf K_{(0)}^\infty=\mathcal N$, a structural-nullspace certificate:
the relation set in each kernel identity
of~\Cref{prop:limiting_ntk_rank,prop:functional_ntk_rank,prop:nonlocal_ntk_rank}
lies in $\mathcal N$, and
$\mathcal N\subseteq\ker\mathbf K_{(0)}^\infty$
by~\Cref{lem:structural_compatibility_compression}. Choose
$\mathcal U=\mathcal N^\perp$. If $\dim\mathcal U\geq1$, then
\Cref{lem:structural_compatibility_compression} verifies
\cref{eqn:fixed_compatible_subspace} and the compressed certificate for every
compatible target; it also shows that targets generated jointly by one
admissible exact field are compatible automatically. If
$\mathcal U=\{\mathbf0\}$, every compatible residual is identically zero, so
there is no nontrivial optimization statement. If instead $\dim\mathcal U\geq1$
and $\mathbf v_T$ is incompatible, then $\Pi_{\mathcal U}\mathbf v_T$ is
compatible and has the same $\mathbf v_{T,\mathcal U}$, hence the same
$\overline r_{\delta,\mathcal U}$; by~\cref{eqn:structural_loss_compression}
the two losses differ by the constant $\|\Pi_{\mathcal N}\mathbf v_T\|_2^2$,
so they have the same gradient and generate the same GD path. The displayed
bound for $\Pi_{\mathcal U}\mathbf v_T$ therefore gives, on the same event and
for every $k\in\mathbb N_0$,
$\mathcal L(\theta_{(k)})-\|\Pi_{\mathcal N}\mathbf v_T\|_2^2
\leq(1-\eta\lambda_{\mathcal U})^k
\bigl(\mathcal L(\theta_{(0)})-\|\Pi_{\mathcal N}\mathbf v_T\|_2^2\bigr)$.
Requiring subspace membership only at initialization, without such a global
relation, is not sufficient for this corollary.
\end{corollary}
\begin{proof}
Under~\cref{eqn:fixed_compatible_subspace},
\(\|\mathbf v_T-\mathbf v_P(\theta)\|_2^2
=\|\mathbf U^\top(\mathbf v_T-\mathbf v_P(\theta))\|_2^2\) throughout the parameter
neighborhood. Therefore the original objective and gradient are exactly those
of the compressed prediction map, whose NTK is
\(\mathbf U^\top\mathbf K(\theta)\mathbf U\). Starting from the same initialization, the
two update paths consequently coincide up to
the first possible exit from the neighborhood. Apply~%
\Cref{thm:linearPDEsconvergence} to this compressed system; its
localization induction rules out leaving that set.
\end{proof}

\subsection{Fixed-kernel limiting loss}

Consider the fixed-kernel residual iteration
$\mathbf r^{\mathrm{lin}}_{(k+1)}=(\mathbf I_N-2\eta\mathbf K_{(0)})
\mathbf r^{\mathrm{lin}}_{(k)}$ with $\mathbf r^{\mathrm{lin}}_{(0)}=\mathbf r_{(0)}$.
If $\mathbf K_{(0)}\neq\mathbf0$ and
$0<\eta<1/\lambda_{\max}(\mathbf K_{(0)})$, its limiting loss is
\begin{equation*}
\mathcal L_\infty^{\mathrm{lin}}
=\left\|\Pi_{\ker(\mathbf K_{(0)})}
\mathbf r_{(0)}\right\|_2^2.
\end{equation*}
When $\mathbf K_{(0)}=\mathbf0$, this expression remains valid because the
update leaves every residual unchanged.
For any architecture that is linear in its output
weights,~\Cref{lem:output_linear_target_projection} shows that
\(\mathbf v_P(\theta_{(0)})\in\Range(\mathbf K_{(0)})\), so
this nullspace projection equals that of \(\mathbf v_T\). Otherwise,
substitute the initial error into the displayed formula.

\subsection{Necessary conditions: a loss floor and a width lower bound}
\label{sec:necessary_conditions}

The guarantees above are sufficient conditions. Two elementary necessary
conditions delimit what cannot be repaired: part~(i) by any width, step size,
or optimizer, and part~(ii) by any step size or optimizer whose iterates stay
in a fixed bounded parameter set, such as $\mathcal T_R(\theta_{(0)})$, at a
width below $N/d_\theta$. Both apply verbatim to vector-valued systems.

\begin{proposition}[Two obstructions to geometric decay]
\label{prop:necessary_conditions}
Let the rows be as in~\Cref{def:linear_pde} and satisfy the applicable
``reg'' clause of~\Cref{assump:pde_classes}, so that
$\mathbf u(\cdot;\theta)\in\mathcal X_{\mathrm{row}}$ for every $\theta$, and
let $\mathcal N_{\mathrm{str}}$ be the structural nullspace
of~\Cref{def:structural_spaces_compatibility}.
\begin{enumerate}[label=(\roman*),nosep,leftmargin=*]
\item \emph{Structural loss floor.} For every $\theta$,
\begin{equation*}
 \mathcal L(\theta)\geq\|\Pi_{\mathcal N_{\mathrm{str}}}\mathbf v_T\|_2^2 .
\end{equation*}
Hence, for an incompatible target, no parameter sequence, whether produced by
gradient descent at any width and step size or by any other method, drives
the loss below this positive constant.
\item \emph{Counting bound.} Let $\mathcal W\subset\mathbb R^{d_1d_\theta}$ be
bounded and suppose $\theta\mapsto\mathbf v_P(\theta)$ is Lipschitz on
$\overline{\mathcal W}$; under $(\mathsf G_R)$ this holds, with constant
$B_*$, for the closed radius-$R$ tube $\mathcal W=\mathcal T_R(\theta_{(0)})$
whenever $\theta_{(0)}\in(\supp\mu)^{d_1}$, which holds almost surely
under~\Cref{assump:main}. If $N>d_1d_\theta$, then
$\mathbf v_P(\overline{\mathcal W})$ is a compact Lebesgue-null subset of
$\mathbb R^N$, so for Lebesgue-almost every target $\mathbf v_T\in\mathbb R^N$,
\begin{equation*}
 \inf_{\theta\in\mathcal W}\mathcal L(\theta)>0 .
\end{equation*}
In particular, for every fixed initialization
$\theta_{(0)}\in(\supp\mu)^{d_1}$ and Lebesgue-almost every target, no
trajectory that remains in $\mathcal T_R(\theta_{(0)})$ satisfies
$\mathcal L(\theta_{(k)})\to0$. When $d_1<N/d_\theta$, the conclusion
of~\Cref{thm:linearPDEsconvergence} fails for every target and every
initialization, because $\rank\mathbf K(\theta)=\rank\mathbf J(\theta)
\leq d_1d_\theta<N$ for every $\theta$ rules out
$\mathbf K_{(k)}\succeq\frac{\lambda_0}{2}\mathbf I_N$. Its localization and
decay parts fail as well: Fubini's theorem applied to the $\sigma$-compact set
$\{(\theta_{(0)},\mathbf v_T):\theta_{(0)}\in(\supp\mu)^{d_1},\
\mathbf v_T\in\mathbf v_P(\mathcal T_R(\theta_{(0)}))\}$ shows that, for almost
every target, localization in $\mathcal T_R(\theta_{(0)})$ together with
$\mathcal L(\theta_{(k)})\to0$ has probability zero over the initialization.
\end{enumerate}
\end{proposition}
\begin{proof}
Part (i) is the decomposition~\cref{eqn:structural_loss_compression}, whose
second term does not depend on $\theta$. For part (ii), a map that is
Lipschitz on a subset of $\mathbb R^{d_1d_\theta}$ does not increase Hausdorff
dimension, so $\mathbf v_P(\overline{\mathcal W})$ has Hausdorff dimension at
most $d_1d_\theta<N$ and therefore Lebesgue measure zero in $\mathbb R^N$; the
image is compact because $\overline{\mathcal W}$ is compact and the map is
continuous. For $\mathbf v_T$ outside this null set,
$\mathbf v_T\neq\mathbf v_P(\theta)$ for every
$\theta\in\overline{\mathcal W}$, so the continuous function
$\theta\mapsto\mathcal L(\theta)$ attains a positive minimum on
$\overline{\mathcal W}$, which bounds the infimum over $\mathcal W$ from
below. A trajectory confined to $\mathcal W$ with
$\mathcal L(\theta_{(k)})\to0$ would contradict this bound. When
$\theta_{(0)}\in(\supp\mu)^{d_1}$, $(\mathsf G_R)$ bounds the gradient of
every $h_i$ by $B_i$ on each ball $\overline{\mathbb B}_R(\theta_{m,(0)})$ of
the tube, so the mean-value inequality on the convex set
$\mathcal T_R(\theta_{(0)})$ and the Cauchy--Schwarz inequality over the
neurons make $\mathbf v_P$ Lipschitz there with constant $B_*$. The Fubini
step uses that the set in question is the continuous image of the closed set
$\{(\theta_{(0)},\theta):\theta_{(0)}\in(\supp\mu)^{d_1},\
\theta\in\mathcal T_R(\theta_{(0)})\}$, that each $\theta_{(0)}$-section of it
is Lebesgue-null by the Lipschitz bound above, and that $(\supp\mu)^{d_1}$ has
full probability under~\Cref{assump:main}.
\end{proof}

The counting bound is linear in $N$, whereas the sufficient width in
\cref{eqn:linear_pde_convergence_conditions} carries the factors
$\lambda_0^{-3}$, $L_*^2$, $\Lambda$, and $\overline r_\delta^2$; the gap
between the two is open. Already the matrix-Chernoff term
of~\cref{eqn:linear_pde_convergence_conditions} implies
$d_1\geq(N/d_\theta)\log(2N/\delta)/c_{\mathrm{Ch}}$, because
$N\lambda_0\leq\operatorname{tr}\mathbf K_{(0)}^\infty
=\mathbb E\|\boldsymbol\Psi\|_F^2\leq d_\theta R_K$; the counting bound is
therefore a statement about every method rather than a sharpening of the
theorem. Vector-valued outputs enter through
$d_\theta=d_0+d_2+1$ on one side and through the number $N$ of scalar rows on
the other, and at the level of the limiting kernel they add no source of
singularity beyond an exact row relation: for pointwise
rows, \Cref{prop:limiting_ntk_rank} shows that under clause
$\mathsf P_{\mathrm{pd}}$ the limiting kernel of a coupled system is singular
only through a co-located dependency, which part~(i) then turns into a
positive loss floor for every incompatible target.

\begin{example}[A coupled system with a co-located dependency]
\label{ex:vector_dependency_floor}
Let $d_2=2$ and prescribe at one point $\mathbf x_0$ the two rows
$u_1+u_2=y_1$ and $2u_1+2u_2=y_2$ with loss scales $s_1,s_2>0$, so that
$P_1(\theta)=s_1(u_1+u_2)(\mathbf x_0;\theta)$ and
$P_2(\theta)=2s_2(u_1+u_2)(\mathbf x_0;\theta)$. The local coefficient
vectors $s_1(1,1)$ and $2s_2(1,1)$ are dependent, so
$\rank\mathbf K_{(0)}^\infty\leq1$, with equality under clause
$\mathsf P_{\mathrm{pd}}$ by~\Cref{prop:limiting_ntk_rank}, and the
sitewise certificate of~\Cref{sec:pointwise_positivity_certificate} fails
before any training. Indeed $2s_2P_1(\theta)-s_1P_2(\theta)=0$ for every
$\theta$, so $\mathbf c=(2s_2,-s_1)^\top\in\mathcal N_{\mathrm{str}}$, and
with $\mathbf v_T=(s_1y_1,s_2y_2)^\top$ part~(i) of
\Cref{prop:necessary_conditions} gives, for every width and every $\theta$,
\begin{equation*}
 \mathcal L(\theta)\geq
 \frac{(\mathbf c^\top\mathbf v_T)^2}{\|\mathbf c\|_2^2}
 =\frac{s_1^2s_2^2(2y_1-y_2)^2}{4s_2^2+s_1^2},
\end{equation*}
which is positive whenever $y_2\neq2y_1$. For the compatible data $y_2=2y_1$
the floor vanishes, the active subspace is one-dimensional, and, under clause
$\mathsf P_{\mathrm{pd}}$, conditional centering
$\mathbb E[\mathbf a\mid\boldsymbol\xi]=\mathbf0$, and the compressed width
and step-size conditions, \Cref{cor:active_subspace_convergence} applies to
the compressed single-row problem.
\end{example}

%% file: convergence.tex
\section{Proof of gradient descent convergence}
\label{sec:convergenceproof}
This section first outlines~\Cref{thm:linearPDEsconvergence} and then
establishes its supporting lemmas.
\subsection{Proof skeleton}
We begin with an abstract result under the row-regularity condition
in~\Cref{assump:general}.
\Cref{prop:linearNTK} verifies its hypotheses for finite-order strong-form
differential constraints; the theorem also covers weak and nonlocal constraints
that directly satisfy the stated regularity conditions. We first prove the
initial-residual bound stated in~\Cref{lem:initial_residual_bound}.
\begin{proof}[Proof of~\Cref{lem:initial_residual_bound}]
Let
\(\mathbf h_m=(h_1(\theta_{m,(0)}),\ldots,
h_N(\theta_{m,(0)}))^\top\). Then
\(\mathbf v_P(\theta_{(0)})=d_1^{-1/2}\sum_m\mathbf h_m\) and
\(\|\mathbf h_m\|_2\leq A_{0,*}\), which proves~%
\cref{eqn:deterministic_r0_bound} by the triangle inequality. The
conditional-centering condition gives \(\mathbb E\mathbf h_m=0\); with
independence of the per-neuron parameter vectors, this yields
\begin{equation*}
 \mathbb E\|\mathbf v_P(\theta_{(0)})\|_2^2
 =\mathbb E\|\mathbf h_1\|_2^2=H_0^2\leq A_{0,*}^2,
 \qquad
 \mathbb E\|\mathbf v_P(\theta_{(0)})\|_2\leq H_0.
\end{equation*}
If \(A_{0,*}=0\), the prediction is zero almost surely and the claim is immediate.
Suppose \(A_{0,*}>0\). Changing one per-neuron parameter vector changes
\(\|d_1^{-1/2}\sum_m\mathbf h_m\|_2\) by at most
\(2A_{0,*}/\sqrt{d_1}\). McDiarmid's inequality therefore gives
\begin{equation*}
 \Pr\!\left\{
 \|\mathbf v_P(\theta_{(0)})\|_2
 >H_0+t\right\}
 \leq\exp\!\left(-\frac{t^2}{2A_{0,*}^2}\right).
\end{equation*}
Set \(t=A_{0,*}\sqrt{2\log(1/\varepsilon)}\) and use
\(\|\mathbf r_{(0)}\|_2\leq\|\mathbf v_T\|_2+
\|\mathbf v_P(\theta_{(0)})\|_2\).
\end{proof}
\begin{remark}
For a fixed number of loss groups, divide each row by the square root of its
group's sample count. If the unnormalized per-neuron contributions and target
values are uniformly bounded, \(A_{0,*}\) and \(\|\mathbf v_T\|_2\) are
\(O(1)\), uniformly in the group sample counts. Hence the initial-residual estimate in~%
\Cref{lem:initial_residual_bound} gives width- and sample-count-independent
high-probability control of \(\|\mathbf r_{(0)}\|_2\). Without per-constraint
normalization,
the corresponding aggregate quantities can grow with \(N\).
\end{remark}
\begin{theorem}[Convergence of gradient descent under abstract row regularity]
\label{thm:generalconvergence}
Suppose the condition in~\Cref{assump:general} holds and use gradient descent
as in~\Cref{def:gradient_descent}.
  Require a positive-definiteness certificate as in
  \Cref{def:positive_definiteness_certificate}, and let
  $0<\lambda_0\leq\lambda_{\min}(\mathbf K_{(0)}^{\infty})$ be a certified
  lower bound. Set
  $\mathbf r_{(0)}=\mathbf v_T-\mathbf v_P(\theta_{(0)})$. Define the squared
  operator-norm envelope for one neuron's initialization Jacobian contribution by
  \[
   R_K:=\esssup_{z\sim\mu}\|\boldsymbol\Psi(z)\|_2^2
   \leq B_*^2.
  \]
  Also set
  \begin{equation*}
    c_{\mathrm{Ch}}
    :=\frac14+\frac34\log\frac34>0
    \qquad
    (c_{\mathrm{Ch}}^{-1}\approx29.207).
  \end{equation*}
  Let $\Lambda\geq\lambda_0$ be a deterministic upper bound on
  $\lambda_{\max}(\mathbf K(\theta))$ throughout the per-neuron parameter
  neighborhood in~%
  \Cref{assump:general}, and let $0<\delta<1$. If
\begin{equation*}
  d_1\geq \frac{R_K}{c_{\mathrm{Ch}}\lambda_0}
  \log\frac{N}{\delta},
\end{equation*}
then the matrix-Chernoff initialization event
$\{\lambda_{\min}(\mathbf K_{(0)})\geq3\lambda_0/4\}$
of~\Cref{prop:positive_definitenessfinitewidth} has probability at least $1-\delta$.
On this event, suppose additionally that the realized initialization satisfies
\(\theta_{(0)}\in(\supp\mu)^{d_1}\), which holds almost surely
under~\Cref{assump:main}, and
\begin{equation*}
d_1\geq\max\!\left\{
  (320+128\sqrt6)\frac{L_*^2\Lambda\|\mathbf r_{(0)}\|_2^2}
    {\lambda_0^3},
  \frac{16B_*^2\|\mathbf r_{(0)}\|_2^2}{R^2\lambda_0^2}
\right\}.
\end{equation*}
If $0<\eta\Lambda\leq1/2$, every iterate remains in the fixed parameter set
\begin{equation*}
 \mathcal T_R(\theta_{(0)})
 :=\left\{\theta:\max_{m\in[d_1]}
 \|\theta_m-\theta_{m,(0)}\|_2\leq R\right\},
\end{equation*}
and, for every $k\in\mathbb N_0$,
\[
 \mathbf K_{(k)}\succeq\frac{\lambda_0}{2}\mathbf I_N,
 \qquad
 \mathcal L(\theta_{(k)})
 \leq(1-\eta\lambda_0)^k\mathcal L(\theta_{(0)}).
\]
One may take $\Lambda=B_*^2$
by~\Cref{prop:positive_definiteness_via_parameters}; since
$\lambda_0\leq\lambda_{\min}(\mathbf K_{(0)}^\infty)\leq R_K\leq B_*^2$, this
choice also gives
$\lambda_0\leq\Lambda$, so the contraction factor is positive. The
Jacobian-perturbation width condition also implies the Taylor-remainder
condition
\begin{equation*}
 d_1\geq(3+2\sqrt2)\frac{L_*^2\|\mathbf r_{(0)}\|_2^2}
 {\lambda_0^2}.
\end{equation*}
Indeed, $\Lambda\geq\lambda_0$ and
$320+128\sqrt6>3+2\sqrt2$. For a fixed collection of PINN loss groups, divide each
row by the square root of its group's sample count. Then $B_*$ and $L_*$ are
independent of the number of points when the corresponding unnormalized
constraint-wise bounds are uniform.
\end{theorem}
\begin{proof}[Proof of~\Cref{thm:linearPDEsconvergence}]
Apply~\Cref{lem:initial_residual_bound} with
\(\varepsilon=\delta/2\). On the resulting event,
\(\|\mathbf r_{(0)}\|_2\leq\overline r_\delta\). Invoke~%
\Cref{thm:generalconvergence} with its initialization failure probability set
to \(\delta/2\), \(\Lambda=B_*^2\), the squared operator-norm envelope
\(R_K\) of the single-neuron Jacobian, and the deterministic envelope \(\overline r_\delta\) in every
condition that depends on the initial training error. The union bound gives
joint confidence at least \(1-\delta\). The localization width condition
ensures, through the induction in the proof skeleton
of~\Cref{thm:generalconvergence} below, that every per-neuron displacement is
at most \(R\), which closes that induction. Consequently, every parameter block remains in the fixed
neighborhood on which the constraint-wise constants are valid.
\end{proof}
\begin{proof}[Proof skeleton of~\Cref{thm:generalconvergence}]
    \phantomsection
    \label{proof:skeleton}
Set \(c_\eta=1-\eta\lambda_0\). We prove jointly by strong induction that
\begin{equation}
 \theta_{(j)}\in\mathcal T_R(\theta_{(0)})
 \quad\text{and}\quad
 \mathcal L(\theta_{(j)})\leq c_\eta^j\mathcal L(\theta_{(0)})
 \qquad(0\leq j\leq n).
 \label{eqn:inductionhypo}
\end{equation}
The assertion is immediate at \(n=0\).

Suppose it holds through \(n=k\). The per-neuron part of~%
\Cref{lem:stronginduction1}, using only the Jacobians at
\(\theta_{(0)},\ldots,\theta_{(k)}\), which satisfy the displacement bound,
gives
\begin{equation*}
 \max_m\|\theta_{m,(k+1)}-\theta_{m,(0)}\|_2
 \leq\frac{4B_*\|\mathbf r_{(0)}\|_2}{\sqrt{d_1}\lambda_0}
 \leq R.
\end{equation*}
Thus \(\theta_{(k+1)}\in\mathcal T_R(\theta_{(0)})\). Convexity places the
full update segment there as well. The local gradient-Lipschitz bound is
therefore available throughout this set.

The global-displacement part of~\Cref{lem:stronginduction1}, the estimate for the
Jacobian perturbation in~%
\Cref{prop:NTKposdef_eachiter}, and the initialization event imply
\[
 \frac{\lambda_0}{2}\mathbf I_N
 \preceq\mathbf K_{(k)}
 \preceq\Lambda\mathbf I_N.
\]
Using~\Cref{prop:NTKtaylor} for the residual $\mathbf r_{(k)}$
from~\Cref{def:pinn_loss} gives the exact error recursion
\begin{equation*}
 \mathbf r_{(k+1)}
 =(\mathbf I_N-2\eta\mathbf K_{(k)})\mathbf r_{(k)}-\mathbf R_{(k)}.
\end{equation*}
\Cref{lem:spectralonestep} now yields
\[
 \|\mathbf r_{(k+1)}\|_2
 \leq\sqrt{c_\eta}\,\|\mathbf r_{(k)}\|_2.
\]
Squaring and applying the induction hypothesis proves the loss assertion at
\(k+1\), and the per-neuron estimate proves the displacement assertion.
These two conclusions close the argument.
\end{proof}

\subsection{Taylor decomposition of consecutive predictions}
    The following proposition uses the fundamental theorem of calculus along
    the update segment to decompose the prediction increment into the Gram
    term and a gradient-integral remainder. The same blockwise Lipschitz
    constant therefore controls kernel stability and this nonlinearity.
\begin{proposition}[Prediction difference between consecutive iterates]
\label{prop:NTKtaylor}
    Under the condition in~\Cref{assump:general} and the gradient descent algorithm setup in~%
    \Cref{def:gradient_descent}, suppose the segment
    \([\theta_{(n)},\theta_{(n+1)}]\) lies in a convex set on which
\(\theta\mapsto\mathbf v_P(\theta)\) is continuously differentiable; apart from
this and the squared loss, the proof uses neither the
representation~\cref{eqn:predictionmap} nor the
bounds~\cref{eqn:abstract_one_neuron_bounds}. Under~\Cref{assump:general} this
holds when \(\theta_{(0)}\in(\supp\mu)^{d_1}\) and the segment lies in the
prescribed per-neuron
    parameter neighborhood.
    Then the difference in prediction between consecutive iterations admits
    the decomposition
\begin{align*}
    \mathbf v_P(\theta_{(n+1)})-\mathbf v_P(\theta_{(n)})
    &=2\eta\mathbf K_{(n)}\mathbf r_{(n)}+\mathbf R_{(n)},
\end{align*}
where
\(\Delta\theta_{(n)}:=\theta_{(n+1)}-\theta_{(n)}
=-\eta\nabla\mathcal L(\theta_{(n)})\) and
\begin{equation}
 \mathbf R_{(n)}
 =\int_0^1
 \left[\mathbf J(\theta_{(n)}+s\Delta\theta_{(n)})-\mathbf J_{(n)}\right]
 \Delta\theta_{(n)}\,ds.
 \label{eqn:gradient_integral_remainder}
\end{equation}
\end{proposition}
\begin{proof}
The fundamental theorem of calculus on the update segment gives
\begin{align*}
 \mathbf v_P(\theta_{(n+1)})-\mathbf v_P(\theta_{(n)})
 &=\int_0^1\mathbf J(\theta_{(n)}+s\Delta\theta_{(n)})
   \Delta\theta_{(n)}\,ds\\
 &=\mathbf J_{(n)}\Delta\theta_{(n)}+\mathbf R_{(n)}.
\end{align*}
By the chain rule,
\[
 \nabla\mathcal L(\theta_{(n)})
 =2\mathbf J_{(n)}^\top
   \left(\mathbf v_P(\theta_{(n)})-\mathbf v_T\right).
\]
Consequently,
\[
 \mathbf J_{(n)}\Delta\theta_{(n)}
 =2\eta\mathbf J_{(n)}\mathbf J_{(n)}^\top
   \left(\mathbf v_T-\mathbf v_P(\theta_{(n)})\right),
\]
which is the stated decomposition.
\end{proof}
\subsection{Positive definiteness along gradient descent} \label{sec:losscontrol4}
We preserve the spectral gap by controlling the
$N\times(d_1d_\theta)$ Jacobian directly,
rather than its Gram matrix. This comparison avoids one factor
\(B_*^2/\lambda_0\) in the overparameterization condition.
\begin{proposition}[Lower bound on the NTK eigenvalue along training] \label{prop:NTKposdef_eachiter}
    Under the condition in~\Cref{assump:general} and gradient descent in~%
    \Cref{def:gradient_descent}, suppose
    \(\theta_{(0)},\ldots,\theta_{(k)}\) lie in the prescribed per-neuron
    parameter neighborhood and the fixed design has a positive-definiteness certificate
    as in~\Cref{def:positive_definiteness_certificate}, with
    \(0<\lambda_0\leq\lambda_{\min}(\mathbf K_{(0)}^\infty)\) a certified
    lower bound. Let \(\Lambda\geq\sup_{\theta\in\mathcal T_R(\theta_{(0)})}
    \lambda_{\max}(\mathbf K(\theta))\) be a deterministic upper bound. Fix
    \(0<\delta<1\) and set
    \(c_{\mathrm{Ch}}=\frac14+\frac34\log\frac34\). If the induction hypothesis in~%
    \cref{eqn:inductionhypo} holds through \(k\), \(0<\eta\Lambda\leq1/2\), and
    \begin{equation*}
    d_1\geq\max\!\left\{
      \frac{R_K}{c_{\mathrm{Ch}}\lambda_0}\log\frac{N}{\delta},
      (320+128\sqrt6)
      \frac{L_*^2\Lambda\|\mathbf r_{(0)}\|_2^2}{\lambda_0^3}
    \right\},
    \end{equation*}
    then
    \(\lambda_{\min}(\mathbf K_{(k)})\geq\lambda_0/2\) on the
    initialization event \(\{\lambda_{\min}(\mathbf K_{(0)})\geq3\lambda_0/4\}\)
of~\Cref{prop:positive_definitenessfinitewidth} intersected with the event
\(\theta_{(0)}\in(\supp\mu)^{d_1}\), which holds almost surely
under~\Cref{assump:main}; by the first width term
and~\Cref{prop:positive_definitenessfinitewidth}, this intersection has
probability at least \(1-\delta\).
\end{proposition}
\begin{proof}[Proof of~\Cref{prop:NTKposdef_eachiter}]
On the initialization event,
\[
 \sigma_{\min}(\mathbf J_{(0)})
 =\sqrt{\lambda_{\min}(\mathbf K_{(0)})}
 \geq\frac{\sqrt3}{2}\sqrt{\lambda_0}.
\]
The uniform Lipschitz bound for each neuron's gradient and the block structure of
the Jacobian columns associated with different neurons give, as
in~\cref{eqn:direct_jacobian_perturbation},
\begin{equation*}
 \|\mathbf J_{(k)}-\mathbf J_{(0)}\|_2
 \leq\frac{L_*}{\sqrt{d_1}}
       \|\theta_{(k)}-\theta_{(0)}\|_2.
\end{equation*}
Indeed, the squared operator norm of a horizontal concatenation is at most the
sum of the squared norms of its neuronal components.

For $k\geq1$, sum the global one-step estimate
of~\Cref{lem:parameterdifferencestep6} over \(j=0,\ldots,k-1\): each
\(\theta_{(j)}\in\mathcal T_R(\theta_{(0)})\), so
\(\lambda_{\max}(\mathbf K_{(j)})\leq\Lambda\), and the induction hypothesis
gives \(\|\mathbf r_{(j)}\|_2\leq(1-\eta\lambda_0)^{j/2}\|\mathbf r_{(0)}\|_2\).
On the initialization event,
\(\Lambda\geq\lambda_{\max}(\mathbf K_{(0)})\geq3\lambda_0/4\), so
\(0<\eta\lambda_0\leq\frac43\eta\Lambda\leq\frac23\), and the geometric-series
argument in the proof of~\Cref{lem:stronginduction1} gives (the case $k=0$ is
trivial),
\begin{equation*}
 \|\theta_{(k)}-\theta_{(0)}\|_2
 \leq\frac{4\sqrt\Lambda}{\lambda_0}\|\mathbf r_{(0)}\|_2.
\end{equation*}
Because
\[
 320+128\sqrt6
 =\frac{64}{(\sqrt3-\sqrt2)^2},
\]
the width condition implies
\[
 \|\mathbf J_{(k)}-\mathbf J_{(0)}\|_2
 \leq\frac{\sqrt3-\sqrt2}{2}\sqrt{\lambda_0}.
\]
Singular-value perturbation therefore yields
\[
 \sigma_{\min}(\mathbf J_{(k)})
 \geq\frac{\sqrt2}{2}\sqrt{\lambda_0}.
\]
Squaring proves
\(\lambda_{\min}(\mathbf K_{(k)})\geq\lambda_0/2\).
\end{proof}

%% file: convergence2.tex
\subsection{Consequences of strong induction and a small learning step}
This subsection derives spectral controls used in~\Cref{thm:generalconvergence}:
a trajectory-length bound, a width-sensitive Taylor remainder, and a joint
one-step contraction estimate. Throughout, the prescribed per-neuron parameter
neighborhood is \(\mathcal T_R(\theta_{(0)})\) of~\cref{eqn:abstract_product_tube},
and, as stated after~\cref{eqn:abstract_one_neuron_bounds}, we work on the
probability-one event \(\theta_{(0)}\in(\supp\mu)^{d_1}\), on which the bounds
in~\cref{eqn:abstract_one_neuron_bounds} apply to every neuron of every
\(\theta\in\mathcal T_R(\theta_{(0)})\).
\subsubsection{Preparations for the analysis}
\begin{lemma}[One-step parameter motion] \label{lem:parameterdifferencestep6}
Under the row-regularity condition in~\Cref{assump:general} and gradient descent in~%
\Cref{def:gradient_descent}, let $n\geq1$ be an integer. If
$\theta_{(n-1)}$ lies in the prescribed per-neuron parameter neighborhood and
$\lambda_{\max}(\mathbf K_{(n-1)})\leq\Lambda$, then
\begin{equation*}
 \|\theta_{(n)}-\theta_{(n-1)}\|_2^2
 \leq4\eta^2\Lambda\|\mathbf r_{(n-1)}\|_2^2.
\end{equation*}
Moreover, each unit satisfies
\begin{equation}
 \|\theta_{m,(n)}-\theta_{m,(n-1)}\|_2
 \leq\frac{2\eta B_*}{\sqrt{d_1}}
       \|\mathbf r_{(n-1)}\|_2.
 \label{eqn:one_step_block_motion}
\end{equation}
\end{lemma}
\begin{proof}
The gradient update and $\mathbf K_{(n-1)}=\mathbf J_{(n-1)}
\mathbf J_{(n-1)}^\top$ give the exact identity
\begin{align*}
\|\theta_{(n)}-\theta_{(n-1)}\|_2^2
&=4\eta^2\|\mathbf J_{(n-1)}^\top\mathbf r_{(n-1)}\|_2^2\\
&=4\eta^2\mathbf r_{(n-1)}^\top\mathbf K_{(n-1)}
\mathbf r_{(n-1)},
\end{align*}
and the spectral bound proves the global estimate.
For the per-neuron claim, the block Jacobian and corresponding change are
\begin{align*}
 \mathbf J_{(n-1),m}
 &=d_1^{-1/2}\boldsymbol\Psi(\theta_{m,(n-1)}),\\
 \theta_{m,(n)}-\theta_{m,(n-1)}
 &=2\eta\mathbf J_{(n-1),m}^\top\mathbf r_{(n-1)}.
\end{align*}
Because \(\theta_{(n-1)}\in\mathcal T_R(\theta_{(0)})\), the row bounds
\(\|\nabla h_i\|_2\leq B_i\) in~\cref{eqn:abstract_one_neuron_bounds} apply at
\(\theta_{m,(n-1)}\); summing them over \(i\in[N]\) and
using~\cref{eqn:abstract_aggregate_constants} together with
\(\|\cdot\|_2\leq\|\cdot\|_F\) gives
\(\|\boldsymbol\Psi(\theta_{m,(n-1)})\|_2\leq B_*\), which proves~%
\cref{eqn:one_step_block_motion}.
\end{proof}
\subsubsection{First consequence: parameter displacement from initialization}
The one-step estimate in~\Cref{lem:parameterdifferencestep6} and the strong-induction
bound in~\cref{eqn:inductionhypo} control the parameter displacement from
initialization. The proof of~\Cref{prop:NTKposdef_eachiter} uses the resulting
inequality.
\begin{lemma}[Parameter displacement from initialization] \label{lem:stronginduction1}
    Under the conditions in~\Cref{thm:generalconvergence} and gradient descent
    in~\Cref{def:gradient_descent}, suppose
    \(\theta_{(0)},\ldots,\theta_{(k)}\) lie in the prescribed per-neuron
    parameter neighborhood. If the
    induction hypothesis~\cref{eqn:inductionhypo} holds through \(k\),
    then, with \(x_\eta=\eta\lambda_0\),
    \begin{align*}
      \|\theta_{(k+1)}-\theta_{(0)}\|_2
      &\leq
      \frac{2(1+\sqrt{1-x_\eta})\sqrt\Lambda}{\lambda_0}
      \|\mathbf r_{(0)}\|_2
      \leq\frac{4\sqrt\Lambda}{\lambda_0}\|\mathbf r_{(0)}\|_2,
      \\
      \max_m\|\theta_{m,(k+1)}-\theta_{m,(0)}\|_2
      &\leq
      \frac{2(1+\sqrt{1-x_\eta})B_*}{\sqrt{d_1}\lambda_0}
      \|\mathbf r_{(0)}\|_2
      \leq\frac{4B_*}{\sqrt{d_1}\lambda_0}\|\mathbf r_{(0)}\|_2.
   \end{align*}
\end{lemma}
\begin{proof}
By~\cref{eqn:inductionhypo},
\(\|\mathbf r_{(n)}\|_2\leq(1-x_\eta)^{n/2}\|\mathbf r_{(0)}\|_2\).
Summing the global one-step estimate in~%
\Cref{lem:parameterdifferencestep6} gives
\[
 \|\theta_{(k+1)}-\theta_{(0)}\|_2
 \leq
 2\eta\sqrt\Lambda\|\mathbf r_{(0)}\|_2
 \sum_{n=0}^{k}(1-x_\eta)^{n/2}.
\]
Applying the same argument to~\cref{eqn:one_step_block_motion}, followed by a
maximum over \(m\), produces the analogous expression with
\(\sqrt\Lambda\) replaced by \(B_*/\sqrt{d_1}\). Finally,
\[
 \frac{\eta}{1-\sqrt{1-x_\eta}}
 =\frac{1+\sqrt{1-x_\eta}}{\lambda_0}
 \leq\frac2{\lambda_0},
\]
which proves both claims. All one-step estimates use only the iterates through
\(k\), each of which satisfies the per-neuron displacement bound.
\end{proof}

\subsubsection{Second consequence: control of the gradient-integral remainder}
\label{sec:third_term_control}
The gradient-Lipschitz bounds in~\cref{eqn:abstract_one_neuron_bounds} control
the Taylor remainder
without introducing a separate Hessian constant.
\begin{lemma}[Taylor remainder bound] \label{lem:residualtermstep}
    Under the conditions in~\Cref{thm:generalconvergence} and the update
    in~\Cref{def:gradient_descent}, suppose the update segment
    \([\theta_{(k)},\theta_{(k+1)}]\) lies in the prescribed per-neuron
    parameter neighborhood. Then
    \begin{equation*}
        \left\| \mathbf{R}_{(k)} \right\|_2
        \leq \frac{2L_*}{\sqrt{d_1}}\,\eta^2\Lambda
        \left\|\mathbf r_{(k)}\right\|_2^2.
    \end{equation*}
\end{lemma}
\begin{proof}
For any two points in this convex set, the decomposition
of the Jacobian by neurons and the gradient-Lipschitz bounds
in~\cref{eqn:abstract_one_neuron_bounds} give
\begin{equation}
 \|\mathbf J(\theta')-\mathbf J(\theta)\|_2
 \leq\frac{L_*}{\sqrt{d_1}}\|\theta'-\theta\|_2.
 \label{eqn:jacobian_lipschitz_global}
\end{equation}
Indeed, after squaring, the norm of the horizontal concatenation is at most the
sum of the squared operator norms of its blocks.
Using~\cref{eqn:gradient_integral_remainder} and~%
\cref{eqn:jacobian_lipschitz_global},
set $\Delta\theta_{(k)}=\theta_{(k+1)}-\theta_{(k)}$ and obtain
\[
 \|\mathbf R_{(k)}\|_2
 \leq\int_0^1\frac{sL_*}{\sqrt{d_1}}
       \|\Delta\theta_{(k)}\|_2^2\,ds
 =\frac{L_*}{2\sqrt{d_1}}\|\Delta\theta_{(k)}\|_2^2.
\]
Finally, the one-step estimate in~\Cref{lem:parameterdifferencestep6} gives
\[
 \|\Delta\theta_{(k)}\|_2^2
 \leq4\eta^2\Lambda\|\mathbf r_{(k)}\|_2^2,
\]
which proves the claim.
\end{proof}
\subsubsection{Third consequence: one-step contraction of the residual}
\begin{lemma}[Spectral one-step descent]\label{lem:spectralonestep}
Suppose the induction hypothesis~\cref{eqn:inductionhypo} holds through
iteration $k$, the update segment $[\theta_{(k)},\theta_{(k+1)}]$ lies in
$\mathcal T_R(\theta_{(0)})$,
$\lambda_{\min}(\mathbf K_{(k)})\geq\lambda_0/2$, and
$\lambda_{\max}(\mathbf K_{(k)})\leq\Lambda$. Under the width and step-size
conditions of~\Cref{thm:generalconvergence},
\begin{equation*}
 \|\mathbf r_{(k+1)}\|_2
 \leq\sqrt{1-\eta\lambda_0}\,\|\mathbf r_{(k)}\|_2.
\end{equation*}
\end{lemma}
\begin{proof}
Set \(x_\eta=\eta\lambda_0\) and \(y_\eta=\eta\Lambda\). Because
\[
 \frac{\lambda_0}{2}\mathbf I_N
 \preceq\mathbf K_{(k)}
 \preceq\Lambda\mathbf I_N,
 \qquad y_\eta\leq\frac12,
\]
all eigenvalues of \(\mathbf I_N-2\eta\mathbf K_{(k)}\) are nonnegative and
\begin{equation}
 \|\mathbf I_N-2\eta\mathbf K_{(k)}\|_2\leq1-x_\eta.
 \label{eqn:linearized_error_contraction}
\end{equation}
The width condition controlling the Jacobian perturbation in~%
\Cref{thm:generalconvergence} implies
\[
 d_1\geq(3+2\sqrt2)
 \frac{L_*^2\|\mathbf r_{(0)}\|_2^2}{\lambda_0^2}.
\]
If \(\mathbf r_{(k)}=0\), the update and remainder both vanish, so the result is
immediate. Otherwise, because
\(\sqrt{3+2\sqrt2}=1+\sqrt2\), the induction hypothesis and the Taylor-remainder
bound in~\Cref{lem:residualtermstep} imply
\begin{align}
 \frac{\|\mathbf R_{(k)}\|_2}{\|\mathbf r_{(k)}\|_2}
 &\leq\frac{2L_*\|\mathbf r_{(0)}\|_2}{\sqrt{d_1}}\eta^2\Lambda\nonumber\\
 &\leq2(\sqrt2-1)x_\eta y_\eta
 \leq(\sqrt2-1)x_\eta.
 \label{eqn:relative_remainder_bound}
\end{align}
\Cref{prop:NTKtaylor} with \(n=k\), together with
\(\mathbf r(\theta)=\mathbf v_T-\mathbf v_P(\theta)\), gives
\[
 \mathbf r_{(k+1)}
 =(\mathbf I_N-2\eta\mathbf K_{(k)})\mathbf r_{(k)}-\mathbf R_{(k)}.
\]
For \(0<x_\eta\leq1/2\),
\[
 \sqrt{1-x_\eta}-(1-x_\eta)
 =x_\eta\frac{\sqrt{1-x_\eta}}{1+\sqrt{1-x_\eta}}
 \geq(\sqrt2-1)x_\eta.
\]
Combining this inequality with~%
\cref{eqn:linearized_error_contraction,eqn:relative_remainder_bound} proves the
claim.
\end{proof}

%% file: tanh_compatible_experiment.tex
\section{A singular NTK with a compatible target for \texorpdfstring{$\tanh$}{tanh}}
\label{sec:tanh_compatible_shared_point}

This example shows that gradient flow with a singular kernel held fixed at initialization can still drive the loss to zero when the
target satisfies the same linear relation as the predictions. It uses neither
a polynomial activation nor inconsistent data. On the interval $[0,2]$,
consider the fully trainable network
\begin{equation}
  u_\theta(x)
  =c+\frac{1}{\sqrt{d_1}}\sum_{m=1}^{d_1} a_m\tanh(w_mx),
  \qquad \theta=(c,(a_m,w_m)_{m=1}^{d_1})\in\mathbb R^{2d_1+1}.
  \label{eqn:tanh-compatible-network}
\end{equation}
We pair this architecture with the following second-order initial-value problem (IVP)
for an ordinary differential equation (ODE):
\begin{equation}
  \mathcal P u:=u''+u'+u=f,
  \qquad u(0)=0,
  \qquad u'(0)=1,
  \label{eqn:tanh-compatible-ivp}
\end{equation}
where
\begin{equation}
  u^*(x)=\tanh x,
  \qquad
  f(x)=\tanh x+\sech^2x
       -2\tanh x\sech^2x.
  \label{eqn:tanh-compatible-manufactured}
\end{equation}
Here $\sech(x)=1/\cosh(x)$. Continuity of the ODE coefficients
in~\eqref{eqn:tanh-compatible-ivp} makes the IVP uniquely solvable. Direct
substitution of the formula in~\eqref{eqn:tanh-compatible-manufactured}
verifies $u^*$.
Thus all three loss targets correspond to this exact function. Moreover,
$u^*$ belongs to the network class at every width: take $c=0$, $w_m=1$, and
$a_m=d_1^{-1/2}$ for every neuron $m$.

For one ODE residual point $x_f$, define the example prediction and target
vectors by
\begin{equation}
  \mathbf v_P^{\text{ex}}(\theta;x_f)
  =\begin{pmatrix}
    (\mathcal P u_\theta)(x_f)\\
    u_\theta(0)\\
    u'_\theta(0)
  \end{pmatrix},
  \qquad
  \mathbf v_T^{\text{ex}}(x_f)
  =\begin{pmatrix}f(x_f)\\0\\1\end{pmatrix}.
  \label{eqn:tanh-compatible-prediction-target}
\end{equation}
The entries in~\eqref{eqn:tanh-compatible-prediction-target} correspond, in
order, to the differential equation, value, and derivative constraints.
Following the target-minus-prediction convention of the main text, set
\begin{equation}
  \mathbf r^{\text{ex}}(\theta;x_f)
  =\mathbf v_T^{\text{ex}}(x_f)-\mathbf v_P^{\text{ex}}(\theta;x_f),
  \qquad
  \mathcal L_{\text{ex}}(\theta;x_f)
  =\|\mathbf r^{\text{ex}}(\theta;x_f)\|_2^2.
  \label{eqn:tanh-compatible-loss}
\end{equation}
Write the three components of the residual in~\eqref{eqn:tanh-compatible-loss}
as $r_{\mathcal P}^{\text{ex}}$, $r_{\text{val}}^{\text{ex}}$, and
$r_{\text{der}}^{\text{ex}}$, in the displayed order.
Define the three feature gradients with respect to all trainable parameters:
\begin{align}
  \boldsymbol\Phi_{\mathcal P,x_f}(\theta)
  &:=\nabla_\theta(\mathcal P u_\theta)(x_f),
  &\boldsymbol\Phi_{\text{val}}(\theta)&:=\nabla_\theta u_\theta(0),
  &\boldsymbol\Phi_{\text{der}}(\theta)&:=\nabla_\theta u'_\theta(0),
  \notag\\
  \mathbf J^{\text{ex}}(\theta;x_f)
  &:=\begin{pmatrix}
    \boldsymbol\Phi_{\mathcal P,x_f}(\theta)^\top\\
    \boldsymbol\Phi_{\text{val}}(\theta)^\top\\
    \boldsymbol\Phi_{\text{der}}(\theta)^\top
  \end{pmatrix},
  &\mathbf K^{\text{ex}}(\theta;x_f)
  &:=\mathbf J^{\text{ex}}(\theta;x_f)
       \mathbf J^{\text{ex}}(\theta;x_f)^\top.
  \label{eqn:tanh-compatible-jacobian-kernel}
\end{align}
These vectors lie in
$\mathbb R^{2d_1+1}$. Stacking them as
in~\eqref{eqn:tanh-compatible-jacobian-kernel} gives a prediction Jacobian in
$\mathbb R^{3\times(2d_1+1)}$ and a tangent Gram matrix in
$\mathbb R^{3\times3}$.
Oddness of $\tanh$ yields
\begin{equation}
  u_\theta''(0)=0
  \quad\text{and}\quad
  \nabla_\theta u_\theta''(0)=\mathbf0
  \qquad\text{for every $\theta$ and every width $d_1$}.
  \label{eqn:tanh-compatible-zero-second-derivative}
\end{equation}
The identities in~\eqref{eqn:tanh-compatible-zero-second-derivative} imply that,
when we evaluate the ODE residual, value, and derivative constraints at $x_f=0$,
\begin{equation}
  \boldsymbol\Phi_{\mathcal P,0}(\theta)
  =\boldsymbol\Phi_{\text{val}}(\theta)+\boldsymbol\Phi_{\text{der}}(\theta),
  \qquad
  \mathbf n:=\frac{1}{\sqrt3}(1,-1,-1)^\top
  \in\ker \mathbf K^{\text{ex}}(\theta;0),
  \label{eqn:tanh-compatible-feature-relation}
\end{equation}
because
$\mathbf J^{\text{ex}}(\theta;0)^\top\mathbf n=\mathbf0$. This feature relation
in~\eqref{eqn:tanh-compatible-feature-relation} follows
from the absence of hidden biases and the odd activation. The three distinct
constraints evaluate the value, the first derivative, and the differential
operator. Meanwhile, the target vector is
$\mathbf v_T^{\text{ex}}(0)
=(f(0),u^*(0),u^{*\prime}(0))^\top=(1,0,1)^\top$, so its projection onto
$\mathbf n$ is zero. Throughout this section, \emph{compatible} means exactly this orthogonality: the target obeys the relation $\mathbf n^\top\mathbf v_P^{\text{ex}}(\theta;0)=0$ that every prediction obeys. This is not the structural compatibility of~\Cref{def:structural_spaces_compatibility}. Applied to $1$, $x$, and $x^2$, the three rows take the values $(1,1,0)^\top$, $(1,0,1)^\top$, and $(2,0,0)^\top$, which are linearly independent. Hence $\mathcal N_{\mathrm{str}}=\{\mathbf0\}$, every target is structurally compatible, and the null vector $\mathbf n\in\ker\mathbf K^{\text{ex}}(\theta;0)$ comes from the architecture, not from a row relation. \Cref{lem:structural_compatibility_compression} makes structural compatibility automatic for a target generated by one exact field, but this orthogonality is not automatic. If $v\in C^2$ generates $\mathbf v=((\mathcal Pv)(0),v(0),v'(0))^\top$, then $\mathbf n^\top\mathbf v=v''(0)/\sqrt3$, which vanishes here because $u^{*\prime\prime}(0)=\tanh''(0)=0$. A field with $v''(0)\neq0$, such as $v(x)=x+x^2/2$, which also satisfies $v(0)=0$ and $v'(0)=1$, gives $\|\mathbf v-\mathbf v_P^{\text{ex}}(\theta;0)\|_2^2\geq v''(0)^2/3$ at every $\theta$. In fact, the three residual components obey
$r_{\mathcal P}^{\text{ex}}(\theta;0)
 =r_{\text{val}}^{\text{ex}}(\theta;0)+r_{\text{der}}^{\text{ex}}(\theta;0)$ throughout
nonlinear training.

The population kernel for $x_f=0$ is particularly transparent. Draw
$a_m,w_m\stackrel{\text{i.i.d.}}{\sim}\Unif[-1,1]$ and set $c=0$. Let
$\mathbf K_{(0)}^{\text{ex},\infty}(x_f)$ denote the deterministic
infinite-width limit of $\mathbf K^{\text{ex}}(\theta_{(0)};x_f)$ under this
initialization.
At $x=0$, the gradient of the value constraint is nonzero only in the $c$
coordinate, whereas the gradient of the derivative constraint has per-neuron
coordinates $(w_m,a_m)/\sqrt{d_1}$. Hence
\begin{equation}
  \mathbf K_{(0)}^{\text{ex},\infty}(0)
  =\begin{pmatrix}
      5/3&1&2/3\\
      1&1&0\\
      2/3&0&2/3
    \end{pmatrix},
  \qquad
  \spec\!\left(\mathbf K_{(0)}^{\text{ex},\infty}(0)\right)
  =\left\{0,\frac{5-\sqrt7}{3},\frac{5+\sqrt7}{3}\right\}.
  \label{eqn:tanh-compatible-shared-kernel}
\end{equation}
The calculation in~\eqref{eqn:tanh-compatible-shared-kernel} shows that this
$3\times3$ kernel has rank two. Its quadratic form is strictly positive on
$(\linspan\{\mathbf n\})^\perp$, which contains every compatible error vector.

\paragraph{Moving only the ODE residual point restores positive definiteness}
Keep the value and derivative observations at zero and move the ODE residual
point to $x_f=s>0$. It suffices to inspect the gradients with respect to the
output bias and output weights. The value observation and equation row both
add $1$ to the former. Within the latter gradient, the derivative constraint
contributes $w$, whereas the ODE residual constraint contributes
\begin{equation}
  g_s(w)
  =w^2\tanh''(sw)+w\tanh'(sw)+\tanh(sw).
  \label{eqn:tanh-compatible-output-feature}
\end{equation}
The output feature in~\eqref{eqn:tanh-compatible-output-feature} has the
following expansion at the origin:
\begin{equation}
  g_s(w)=(1+s)w-\left(2s+s^2+\frac{s^3}{3}\right)w^3+O(w^5).
  \label{eqn:tanh-compatible-output-feature-series}
\end{equation}
For $s>0$, $g_s$ is therefore not proportional to $w$. Suppose a linear
combination of the three full parameter-gradient vectors with coefficients
$(\chi_{\mathcal P},\chi_{\text{val}},\chi_{\text{der}})$ vanishes. The
output-bias coordinate gives $\chi_{\mathcal P}+\chi_{\text{val}}=0$, whereas
the output-weight coordinates give
$\chi_{\mathcal P}g_s(w)+\chi_{\text{der}}w=0$ in $L^2(\Unif[-1,1])$. The
expansion in~\eqref{eqn:tanh-compatible-output-feature-series} forces
$\chi_{\mathcal P}=\chi_{\text{der}}=0$ and then
$\chi_{\text{val}}=0$. The Gram-matrix contribution from the global output bias
and the per-neuron output weights is therefore positive definite, and so is the
full limiting NTK\@. The same argument gives an empirical kernel of rank three,
$\mathbf K^{\text{ex}}(\theta_{(0)};s)$, almost surely for $d_1\geq2$:
the exceptional samples form the zero set of a nonzero analytic
determinant.

For a concrete population calculation, take $s=1$. The two nonconstant
entries needed below are
\begin{equation}
  k_{\mathcal P\mathcal P}(s)
  :=1+\int_0^1\!\left(g_s(w)^2+\frac13g_s'(w)^2\right)\,dw,
  \qquad
  k_{\mathcal P,\text{der}}(s)
  :=\int_0^1\!wg_s(w)\,dw+\frac13g_s(1).
  \label{eqn:tanh-compatible-shifted-integrals}
\end{equation}
The integrals in~\eqref{eqn:tanh-compatible-shifted-integrals} use a prime on
$g_s$ to denote differentiation with respect to $w$. Deterministic fixed-node
Gauss--Legendre quadrature then gives
\begin{equation}
  \begin{aligned}
    \mathbf K_{(0)}^{\text{ex},\infty}(1)
    &=\begin{pmatrix}
      k_{\mathcal P\mathcal P}(1)&1&k_{\mathcal P,\text{der}}(1)\\
      1&1&0\\
      k_{\mathcal P,\text{der}}(1)&0&2/3
    \end{pmatrix}
    \approx
    \begin{pmatrix}
      1.68126802&1&0.48341254\\
      1&1&0\\
      0.48341254&0&2/3
    \end{pmatrix}.
  \end{aligned}
  \label{eqn:tanh-compatible-shifted-kernel}
\end{equation}
The matrix in~\eqref{eqn:tanh-compatible-shifted-kernel} has eigenvalues
\begin{equation*}
  \spec\!\left(\mathbf K_{(0)}^{\text{ex},\infty}(1)\right)
  \approx\{0.11916210,0.74496231,2.48381029\}.
\end{equation*}
These decimals do not establish positive definiteness;
the preceding linear-independence argument does so independently.

For a chosen $x_f$, abbreviate
$\mathbf r^{\text{ex}}_{(0)}(x_f)
:=\mathbf r^{\text{ex}}(\theta_{(0)};x_f)$ and
$\mathbf K^{\text{ex}}_{(0)}(x_f)
:=\mathbf K^{\text{ex}}(\theta_{(0)};x_f)$. Under gradient flow with this
initial kernel held fixed,
\begin{equation}
  \mathcal L_{\text{ex}}^{\text{fixed}}(t;x_f)
  =\mathbf r^{\text{ex}}_{(0)}(x_f)^\top
  \exp\!\left(-4\mathbf K^{\text{ex}}_{(0)}(x_f)t\right)
  \mathbf r^{\text{ex}}_{(0)}(x_f).
  \label{eqn:tanh-compatible-fixed-kernel-flow}
\end{equation}
For $x_f=0$ and $d_1\geq1$, the bias-coordinate value row and the nonzero
derivative row give rank two with probability one. Therefore,
on this event, $\ker\mathbf K^{\text{ex}}_{(0)}(0)=\linspan\{\mathbf n\}$. The
exact residual relation makes the initial residual orthogonal to this
nullspace. The deterministic linear-independence argument above is, in the
terminology of~\Cref{def:positive_definiteness_certificate}, a
positive-definiteness certificate for the population kernel
$\mathbf K_{(0)}^{\text{ex},\infty}(1)$. It is not the analytic alternative
of~\Cref{sec:pointwise_positivity_certificate}: that output-weight criterion is
stated for the network of~\Cref{def:ntk_parameterization}, which has no global
output bias, and it fails here because the value row's per-neuron features
vanish identically, so positivity rests on the global output bias. For
$x_f=1$ and $d_1\geq2$, the preceding analytic-determinant argument shows that
$\mathbf K^{\text{ex}}_{(0)}(1)\succ0$ almost surely. Thus the
fixed-kernel formula in~\eqref{eqn:tanh-compatible-fixed-kernel-flow} implies loss convergence
to zero with probability one in both cases when $d_1\geq2$.
For the two population matrices above, positive definiteness alone does not order
their rates. The smallest nonzero eigenvalue for $x_f=0$ is $0.785$ to three
decimals, whereas the smallest eigenvalue for $x_f=1$ is $0.119$. With the
normalized step rule $\eta=0.2/\lambda_{\max}$, the two limiting matrices
multiply the squared residual-mode contribution in the slowest non-null spectral
direction by approximately $0.769$ and $0.962$ per step, respectively. For
$x_f=1$, this certificate gives a positive-definite kernel, so the loss with
the kernel held fixed at initialization converges to zero for every initial
error. At $x_f=0$, the
same conclusion holds for the manufactured target above: $u^{*\prime\prime}(0)=0$
makes it orthogonal to $\mathbf n$, so the residual remains in
$\mathbf n^\perp$.

\paragraph{Scope}
All convergence statements in this section concern dynamics with the kernel held fixed, either at its initial value $\mathbf K^{\text{ex}}_{(0)}(x_f)$ or at its infinite-width limit $\mathbf K^{\text{ex},\infty}_{(0)}(x_f)$. Apart from the kernel certificates above, the example does not verify the hypotheses of \Cref{thm:linearPDEsconvergence} (for $x_f=1$) or \Cref{cor:active_subspace_convergence} (for $x_f=0$), including their width and step-size conditions. It therefore gives no finite-width guarantee for gradient descent on the nonlinear loss $\mathcal L_{\text{ex}}$. The exact singularity for $x_f=0$ uses the hidden-bias-free form
in~\eqref{eqn:tanh-compatible-network}; a generic trainable hidden bias removes the
identity $u_\theta''(0)=0$. This relation depends jointly on that architecture,
the operator $u''+u'+u$, and evaluation of all three
constraints at zero. Repeated physical locations alone need not cause
singularity. The global output bias contributes
a fixed rank-one term to the kernel; equivalently, one may distribute it over
i.i.d.\ per-neuron parameter vectors as
$c=d_1^{-1/2}\sum_{m=1}^{d_1}c_m$ if all per-neuron
parameter vectors must be identically distributed. Positive diagonal rescaling
of the three loss components preserves the number of nonzero
eigenvalues. It also leaves
unchanged whether the weighted target obeys the corresponding weighted prediction
relation, although it can change the nonzero eigenvalues
and hence the convergence rates.

%% file: nonlinear_depth.tex
\section{Nonlinear constraints and depth: obstructions to the present proof
architecture}
\label{sec:nonlinear_and_depth}

Every guarantee above concerns prediction components that are linear in the
network output, on the two-layer network of~\Cref{def:ntk_parameterization}.
This section records what happens when either restriction is dropped. The
nonlinear part is a negative result and is proved here: with one genuinely
nonlinear polynomial row, under conditional centering and second-moment isotropy of the output weights (conditions~\textup{(\(\mathsf C\))} and~\textup{(\(\mathsf C_{\mathrm{iso}}\))} below) and the output-block positive-definiteness certificate for the base observations (\Cref{lem:base_family_certificates}), the tangent Gram matrix at initialization still has a
limit, but that limit is a random matrix; a larger width exponent removes the
randomness only by sending the initial predictions to zero, after which the
rescaled kernel is that of the rows linearized at zero. The depth part isolates one unconditional lemma, one
conditional substitution proposition, and two conjectures, stated so that the
hypotheses they would need are explicit. No numbered result of the main text, and no proof of one, depends on this
section; the Depth and Nonlinear constraints paragraphs of the concluding
section summarize the results and conjectures stated here.

\paragraph{Relation to prior work.}
That the nonlinear regime is qualitatively different is not a new observation.
\citet{Zhou2024} ask when the PINN tangent kernel converges at all for
possibly nonlinear differential operators and identify homogeneity-dependent
width-scaling thresholds; \citet{Bonfanti2024} study the difficulties specific
to the nonlinear regime and already find that there the tangent kernel is a
random matrix at initialization and is not constant during training. We claim no
priority for the qualitative statement that, under the parameterization
of~\Cref{def:ntk_parameterization}, a genuinely nonlinear row can make the
infinite-width limit of the tangent Gram matrix at initialization random.
What~\Cref{sec:composite_rows,sec:nonlinear_nondeterminism,sec:nonlinear_scaling} add is a form
sharp enough to be used as a no-go: an exact finite-width identity that
computes each kernel entry in full, so that no cancellation between
contributions remains to be ruled out; an explicit \(L^2(\mu)\) formula for the
limiting variance of the entry pairing a product row with an affine row
(\Cref{cor:product_row_nondeterminism}), which turns its nondegeneracy into a
checkable condition rather than an assumption; an affine-or-random
dichotomy that needs no activation-specific or initialization-specific
hypothesis beyond conditional centering, second-moment isotropy of the
output weights, and this paper's output-block positive-definiteness certificate (\Cref{def:positive_definiteness_certificate}) applied to the base observations (\Cref{lem:base_family_certificates}); and a
trichotomy in the width exponent showing that, for each monomial block, the
limiting object does not depend on the exponent at all.

\subsection{Composite rows and an exact tangent Gram identity}
\label{sec:composite_rows}

\begin{definition}[Width exponent]
\label{def:exponent_normalization}
For \(\varrho>0\), consider the one-parameter family
\begin{equation}
 \mathbf u(\mathbf x;\theta)
 =\frac1{d_1^{\varrho}}\sum_{m=1}^{d_1}
  \sigma(\mathbf w_m^\top\mathbf x+\gamma b_m)\mathbf a_m,
 \label{eqn:exponent_network}
\end{equation}
with \(\theta_m\), \(d_\theta\), and the trained parameters exactly as
in~\Cref{def:ntk_parameterization}. The choice \(\varrho=1/2\)
is~\cref{eqn:network_main} and is the only member used elsewhere in this paper.
For a nonlinear constraint of degree \(p\geq2\) we single out the
\emph{degree-matched} exponent \(\varrho_p=(2p-1)/(2p)\), for which the overall normalization \(d_1^{-2p\varrho_p}=d_1^{-(2p-1)}\) of a degree-\((p,p)\) kernel block supplies exactly one factor \(d_1^{-1}\) for each of the \(2p-1\) neuron sums that the block formally contains: the \(2p-2\) prediction factors and the sum over neurons in the tangent Gram entry.
\end{definition}

\begin{definition}[Composite rows over a base family of linear observations]
\label{def:composite_nonlinear_rows}
Fix \(n\in\mathbb N\) and a family \(\mathcal L=[n]\) (an index set, unrelated to the
loss \(\mathcal L(\theta)\) of~\Cref{def:pinn_loss}) of scalar linear
observations of the network, each satisfying the row-regularity condition
of~\Cref{assump:general} at \(\varrho=1/2\), with per-neuron maps
\(h_\ell(\boldsymbol\xi,\mathbf a)=\mathbf a^\top\mathbf f_\ell(\boldsymbol\xi)\)
as in~\cref{eqn:predictionmap} and constants \(A_{0,\ell},B_\ell,L_\ell<\infty\)
as in~\cref{eqn:abstract_one_neuron_bounds}. Under~\cref{eqn:exponent_network},
the value of observation \(\ell\) is
\begin{equation}
 Q_\ell(\theta)=\frac1{d_1^{\varrho}}\sum_{m=1}^{d_1}h_\ell(\theta_m),
 \qquad
 \mathbf Q(\theta)=(Q_1(\theta),\ldots,Q_n(\theta))^\top .
 \label{eqn:base_family_representation}
\end{equation}
A \emph{composite row} is \(P_i(\theta)=\pi_i(\mathbf Q(\theta))\) for a fixed
\(\pi_i\in C^1(\mathbb R^n)\); the loss scale \(s_i\) of~\Cref{def:linear_pde}
is absorbed into \(\pi_i\), so \(P_i\) is the full loss-scaled row and
\(\mathbf K(\theta)\) of~\cref{eqn:loss_kernel_main} is the full tangent Gram
matrix. The row is \emph{affine} when \(\pi_i\) is affine and \emph{genuinely
nonlinear} otherwise, and it is a \emph{degree-\(p\) polynomial row} when
\(\pi_i\) is a polynomial of degree \(p\). For a multiset
\(J=(j_1,\ldots,j_q)\) with entries in \(\mathcal L\) we write
\(\mathcal M_J(\theta)=\prod_{s=1}^qQ_{j_s}(\theta)\) for the associated
monomial and call \(q=|J|\) its degree; every polynomial row is a finite linear
combination of monomials, and by bilinearity every kernel entry is the
corresponding combination of the \emph{monomial blocks}
\(K^{J,J'}=\sum_{m}\langle\nabla_{\theta_m}\mathcal M_J,
\nabla_{\theta_m}\mathcal M_{J'}\rangle\).
\end{definition}

An affine \(\pi_i\) recovers a
row of~\Cref{def:linear_pde} whenever the base family spans the corresponding
observation, and a constant term in \(\pi_i\) changes neither the Jacobian nor
the tangent Gram matrix. The viscous Burgers equation \(\partial_tu+\partial_x(u^2/2)=\nu\partial_{xx}u\), a scalar conservation law with quadratic flux,
gives the degree-two row
\(\pi_i(\mathbf q)=s_i(q_1+q_2q_3-\nu q_4)\), where \(Q_1,\ldots,Q_4\) are the
unscaled observations \(\partial_tu\), \(u\), \(\partial_xu\),
\(\partial_{xx}u\) at the collocation point \(\mathbf x_i\); a \(p\)-th-power
boundary row is \(\pi_i(\mathbf q)=s_iq_\ell^{\,p}\).

Throughout this section we use two conditions on the initialization
of~\Cref{assump:main}.
\begin{enumerate}[leftmargin=*,nosep]
\item[\textup{(\(\mathsf C\))}]
\emph{Conditional output-weight centering:}
\(\mathbb E[\mathbf a\mid\boldsymbol\xi]=\mathbf 0\) for \(\mu\)-almost every
neuron parameter, which implies \(\mathbb E_{z\sim\mu}h_\ell(z)=0\) for every
\(\ell\in\mathcal L\). This is the condition already imposed
after~\Cref{assump:general} and used by~\Cref{lem:initial_residual_bound}
and~\Cref{thm:linearPDEsconvergence}.
\item[\textup{(\(\mathsf C_{\mathrm{iso}}\))}]
In addition, \(\mathbb E[\mathbf a\mathbf a^\top\mid\boldsymbol\xi]
=c_{\mathbf a}\mathbf I_{d_2}\) \(\mu_h\)-almost surely for some
\(c_{\mathbf a}>0\). Every initialization whose output-weight coordinates are
i.i.d., symmetric, nondegenerate, and drawn independently of
\(\boldsymbol\xi\) satisfies it. This is an addition
to~\Cref{assump:main}, which imposes no second-moment structure, but it is not
a restrictive one: the \(\Unif[-1,1]\) output weights used throughout the
experiments of~\Cref{sec:empirical_validation} satisfy it with
\(c_{\mathbf a}=1/3\), as does the Rademacher law or any other centered, bounded, nondegenerate coordinate
law. It fails when the output-weight coordinates are correlated given \(\boldsymbol\xi\), when their conditional variance \(\mathbb E[a_q^2\mid\boldsymbol\xi]\) varies with \(\boldsymbol\xi\) (for instance when \(\mathbf a\) is drawn with a \(\boldsymbol\xi\)-dependent scale), or when they have
unequal variances given \(\boldsymbol\xi\).
\end{enumerate}

Write \(\boldsymbol\Psi_{\mathcal L}(z)\in\mathbb R^{n\times d_\theta}\) for the
base-family analogue of \(\boldsymbol\Psi\), namely
the matrix whose \(\ell\)-th row is \(\nabla h_\ell(z)^\top\),
and set
\begin{equation}
 \begin{aligned}
 \widehat{\mathbf K}_{\mathcal L}(\theta)
 &=\frac1{d_1}\sum_{m=1}^{d_1}
   \boldsymbol\Psi_{\mathcal L}(\theta_m)
   \boldsymbol\Psi_{\mathcal L}(\theta_m)^\top,\\
 [\mathbf K_{\mathcal L}^\infty]_{\ell\ell'}
 &=\mathbb E_{z\sim\mu}\langle\nabla h_\ell(z),\nabla h_{\ell'}(z)\rangle,
 \qquad
 [\boldsymbol\Sigma_{\mathcal L}]_{\ell\ell'}
 =\langle h_\ell,h_{\ell'}\rangle_{L^2(\mu)} .
 \end{aligned}
 \label{eqn:base_family_matrices}
\end{equation}
Here \(\mathbf K_{\mathcal L}^\infty\) is the limiting NTK supplied by
statement one of~\Cref{thm:linearNTK} for the auxiliary \emph{linear} design
\(\mathcal L\), and under~\textup{(\(\mathsf C\))} the matrix
\(\boldsymbol\Sigma_{\mathcal L}\) is the infinite-width prediction covariance
of that design, that is, the covariance of the Gaussian limit in item~2
of~\Cref{lem:base_family_limits}. It is the analogue, for the law~\(\mu\), of
the Gaussian-process covariance that~\citet{Lee2019} compute for Gaussian
weights, evaluated on the base observations; it is not given by their
recursion, which treats the first-layer pre-activations as Gaussian, and the
two need not coincide under the compactly supported laws
of~\Cref{assump:main}.

\begin{lemma}[Exact composite tangent Gram identity]
\label{lem:composite_kernel_identity}
Let the rows be composite as in~\Cref{def:composite_nonlinear_rows} under the
exponent \(\varrho\). Then for every \(\theta\) whose neuron blocks all lie in
\(\supp\mu+\overline{\mathbb B}_R(\mathbf0)\),
where~\cref{eqn:abstract_one_neuron_bounds} provides \(\nabla h_\ell\) (almost
surely, every \(\theta\in\mathcal T_R(\theta_{(0)})\) qualifies), every
\(m\in[d_1]\), and every
\(i,j\in[N]\),
\begin{equation}
 \nabla_{\theta_m}P_i(\theta)
 =\frac1{d_1^{\varrho}}
  \boldsymbol\Psi_{\mathcal L}(\theta_m)^\top\nabla\pi_i(\mathbf Q(\theta)),
 \qquad
 [\mathbf K(\theta)]_{ij}
 =d_1^{1-2\varrho}\,
  \nabla\pi_i(\mathbf Q(\theta))^\top
  \widehat{\mathbf K}_{\mathcal L}(\theta)\,
  \nabla\pi_j(\mathbf Q(\theta)).
 \label{eqn:composite_row_exact_kernel}
\end{equation}
In particular, for monomials \(J=(j_1,\ldots,j_q)\) and
\(J'=(l_1,\ldots,l_{q'})\) of degrees \(q\) and \(q'\),
\begin{equation}
 K^{J,J'}(\theta)
 =d_1^{1-2\varrho}
  \sum_{r=1}^{q}\sum_{t=1}^{q'}
  \Bigl(\prod_{s\neq r}Q_{j_s}(\theta)\Bigr)
  \Bigl(\prod_{s\neq t}Q_{l_s}(\theta)\Bigr)
  \bigl[\widehat{\mathbf K}_{\mathcal L}(\theta)\bigr]_{j_rl_t}.
 \label{eqn:polynomial_kernel_identity}
\end{equation}
Both identities hold at every width and every such parameter value; no expectation,
limit, or approximation is involved.
\end{lemma}
\begin{proof}
Only the \(m\)-th summand of~\cref{eqn:base_family_representation} depends on
\(\theta_m\), so \(\nabla_{\theta_m}Q_\ell(\theta)
=d_1^{-\varrho}\nabla h_\ell(\theta_m)\), and the chain rule gives the first
display. Since the neuron blocks exhaust every trainable parameter and
\(\mathbf K=\mathbf J\mathbf J^\top\) by~\cref{eqn:loss_kernel_main},
\begin{equation*}
 [\mathbf K(\theta)]_{ij}
 =\sum_{m=1}^{d_1}\langle\nabla_{\theta_m}P_i,\nabla_{\theta_m}P_j\rangle
 =d_1^{-2\varrho}\nabla\pi_i(\mathbf Q)^\top
  \Bigl(\sum_{m=1}^{d_1}\boldsymbol\Psi_{\mathcal L}(\theta_m)
   \boldsymbol\Psi_{\mathcal L}(\theta_m)^\top\Bigr)\nabla\pi_j(\mathbf Q),
\end{equation*}
which is the second. For the monomial form, apply the second identity with \(\pi_i\) and \(\pi_j\) replaced by the monomial maps \(\pi_J(\mathbf z)=\prod_{s=1}^{q}z_{j_s}\) and \(\pi_{J'}(\mathbf z)=\prod_{s=1}^{q'}z_{l_s}\) on \(\mathbb R^n\). Then \(\mathcal M_J=\pi_J\circ\mathbf Q\) and \(\mathcal M_{J'}=\pi_{J'}\circ\mathbf Q\), and
\(\nabla\pi_J(\mathbf Q(\theta))
=\sum_{r=1}^q\bigl(\prod_{s\neq r}Q_{j_s}(\theta)\bigr)\mathbf e_{j_r}\), with the same formula for \(\pi_{J'}\). Here \(\mathbf e_\ell\) is the \(\ell\)-th standard basis vector of \(\mathbb R^n\), not of \(\mathbb R^{d_2}\).
\end{proof}

The identity already displays the obstruction. The empirical average
\(\widehat{\mathbf K}_{\mathcal L}\) is the object the linear theory controls:
at initialization it is a mean of i.i.d.\ bounded matrices. The factors
\(\nabla\pi_i(\mathbf Q(\theta))\) are not averages. They are fixed functions of
the base observations \(\mathbf Q(\theta)\), which every neuron shares, and which
at initialization are sums of i.i.d.\ terms normalized by \(d_1^{\varrho}\).
At \(\varrho=1/2\) and under~\textup{(\(\mathsf C\))} such a sum is governed by
the central limit theorem, not by the strong law (item~2
of~\Cref{lem:base_family_limits}); under~\textup{(\(\mathsf C\))} with
\(\varrho>1/2\) it tends to zero (\Cref{rem:normalized_kernel_linearization}),
and without~\textup{(\(\mathsf C\))} its leading term is the strong-law mean
\(d_1^{1-\varrho}m_\ell\) of item~3.

\begin{lemma}[Limits of the two empirical objects]
\label{lem:base_family_limits}
Assume~\Cref{assump:main} and let the base family satisfy
\Cref{assump:general}. Put
\(\widetilde Q_\ell(\theta)=d_1^{-1/2}\sum_mh_\ell(\theta_m)\), so that
\(\mathbf Q(\theta)=d_1^{1/2-\varrho}\widetilde{\mathbf Q}(\theta)\). At
initialization, as \(d_1\to\infty\):
\begin{enumerate}[leftmargin=*,nosep]
\item \(\widehat{\mathbf K}_{\mathcal L}(\theta_{(0)})
 \to\mathbf K_{\mathcal L}^\infty\) in probability, and almost surely under the
 nested-width convention of~\Cref{thm:linearNTK};
\item under~\textup{(\(\mathsf C\))},
 \(\widetilde{\mathbf Q}(\theta_{(0)})\Rightarrow
 \mathbf Z\sim\mathcal N_n(\mathbf0,\boldsymbol\Sigma_{\mathcal L})\), jointly
 with the convergence in item~\textup{1};
\item without~\textup{(\(\mathsf C\))}, with
 \(m_\ell=\mathbb E_{z\sim\mu}h_\ell(z)\), one has
 \(d_1^{\varrho-1}Q_\ell(\theta_{(0)})\to m_\ell\), almost surely along
 nested widths as in~\Cref{thm:linearNTK} and in probability under fresh
 resampling.
\end{enumerate}
Moreover \(|m_\ell|\leq A_{0,\ell}\),
\(|[\boldsymbol\Sigma_{\mathcal L}]_{\ell\ell'}|\leq A_{0,\ell}A_{0,\ell'}\)
and \(|[\mathbf K_{\mathcal L}^\infty]_{\ell\ell'}|\leq B_\ell B_{\ell'}\).
\end{lemma}
\begin{proof}
By~\Cref{assump:main} the vectors \(\theta_{m,(0)}\) are i.i.d.\ and lie in
\(\supp\mu\) almost surely. The matrices
\(\boldsymbol\Psi_{\mathcal L}(\theta_{m,(0)})
\boldsymbol\Psi_{\mathcal L}(\theta_{m,(0)})^\top\) are i.i.d.\ and bounded
entrywise by \(B_\ell B_{\ell'}\) through Cauchy--Schwarz
and~\cref{eqn:abstract_one_neuron_bounds}, so item~1 is statement one
of~\Cref{thm:linearNTK} applied to the design \(\mathcal L\), with the same
convention governing the mode of convergence. For item~2 the random vectors
\((h_\ell(\theta_{m,(0)}))_{\ell}\) are i.i.d., bounded by \((A_{0,\ell})_\ell\),
hence square integrable, and centered by~\textup{(\(\mathsf C\))}; the
multivariate Lindeberg--L\'evy theorem gives the stated Gaussian limit, which is
degenerate exactly when \(\boldsymbol\Sigma_{\mathcal L}\) is singular. Joint
convergence follows from Slutsky's theorem, which requires no independence
between the two components, and indeed they are strongly dependent, being built
from the same neurons; it applies because the second limit is
\emph{deterministic}. Item~3 is the strong law for the same bounded i.i.d.\@
summands along nested widths, and the weak law otherwise. The three bounds are immediate
from~\cref{eqn:abstract_one_neuron_bounds}.
\end{proof}

\subsection{Nondeterminism of the infinite-width kernel}
\label{sec:nonlinear_nondeterminism}

Throughout this subsection \(\varrho=1/2\), the standard parameterization
of~\Cref{def:ntk_parameterization}.

\begin{proposition}[Random limit of the tangent Gram matrix for composite rows]
\label{prop:nonlinear_ntk_nondeterminism}
Assume~\Cref{assump:main}, let the base family satisfy~\Cref{assump:general}
and~\textup{(\(\mathsf C\))}, let \(\varrho=1/2\), and let \(P_1,\ldots,P_N\) be
composite rows. Then, as \(d_1\to\infty\):
\begin{enumerate}[leftmargin=*,nosep]
\item jointly in all entries,
\begin{equation}
 \mathbf K_{(0)}\Longrightarrow
 \mathbf K^\infty_{\boldsymbol\pi}(\mathbf Z)
 :=\mathbf D\boldsymbol\pi(\mathbf Z)\,\mathbf K_{\mathcal L}^\infty\,
   \mathbf D\boldsymbol\pi(\mathbf Z)^\top,
 \qquad
 \mathbf Z\sim\mathcal N_n(\mathbf0,\boldsymbol\Sigma_{\mathcal L}),
 \label{eqn:nonlinear_kernel_weak_limit}
\end{equation}
where the rows of \(\mathbf D\boldsymbol\pi(\mathbf q)\in\mathbb R^{N\times n}\)
are \(\nabla\pi_i(\mathbf q)^\top\);
\item if \(\pi_i\) and \(\pi_j\) are affine with
 \(\nabla\pi_i\equiv\boldsymbol\beta_i\) and
 \(\nabla\pi_j\equiv\boldsymbol\beta_j\), the corresponding limit is the
 deterministic constant
 \(\boldsymbol\beta_i^\top\mathbf K_{\mathcal L}^\infty\boldsymbol\beta_j\),
 which is the entry \([\mathbf K_{(0)}^\infty]_{ij}\) supplied
 by~\Cref{thm:linearNTK} for those rows;
\item if \([\mathbf K^\infty_{\boldsymbol\pi}(\mathbf Z)]_{ij}\) is not almost
 surely constant, then \([\mathbf K_{(0)}]_{ij}\) converges in probability to no
 constant whatsoever, so at \(\varrho=1/2\) no deterministic infinite-width kernel exists for that
 entry, either almost surely along nested widths or in probability;
\item if every \(\pi_i\) is a polynomial, then
\begin{equation}
 \mathbb E\bigl[[\mathbf K_{(0)}]_{ij}\bigr]
 \to\mathbb E\bigl[[\mathbf K^\infty_{\boldsymbol\pi}(\mathbf Z)]_{ij}\bigr],
 \qquad
 \Var\bigl([\mathbf K_{(0)}]_{ij}\bigr)
 \to\Var\bigl([\mathbf K^\infty_{\boldsymbol\pi}(\mathbf Z)]_{ij}\bigr).
 \label{eqn:nonlinear_kernel_variance_limit}
\end{equation}
Hence whenever the right-hand variance is positive the entrywise standard
deviation of \(\mathbf K_{(0)}\) is \(\Theta(1)\) in \(d_1\), in contrast with
the \(O(d_1^{-1/2})\) standard deviation of the affine case.
\end{enumerate}
\end{proposition}
\begin{proof}
\emph{Item 1.} At \(\varrho=1/2\) we have
\(\mathbf Q=\widetilde{\mathbf Q}\) and the prefactor
in~\cref{eqn:composite_row_exact_kernel} equals one. By items~1 and~2
of~\Cref{lem:base_family_limits},
\(\bigl(\mathbf Q(\theta_{(0)}),\widehat{\mathbf K}_{\mathcal L}(\theta_{(0)})
\bigr)\Rightarrow(\mathbf Z,\mathbf K_{\mathcal L}^\infty)\) in
\(\mathbb R^n\times\mathbb R^{n\times n}\). The map
\((\mathbf q,\mathbf G)\mapsto\mathbf D\boldsymbol\pi(\mathbf q)\mathbf G
\mathbf D\boldsymbol\pi(\mathbf q)^\top\) is continuous because each \(\pi_i\)
is \(C^1\); the continuous mapping theorem
and~\cref{eqn:composite_row_exact_kernel} give the claim.

\emph{Item 2} is immediate, \(\nabla\pi_i\) and \(\nabla\pi_j\) being constant, together
with \(\mathbb E_{z\sim\mu}\langle\nabla(\boldsymbol\beta_i^\top\mathbf h)(z),
\nabla(\boldsymbol\beta_j^\top\mathbf h)(z)\rangle
=\boldsymbol\beta_i^\top\mathbf K_{\mathcal L}^\infty\boldsymbol\beta_j\), where \(\mathbf h=(h_\ell)_{\ell\in\mathcal L}\).

\emph{Item 3.} Convergence in probability to \(c\) implies convergence in
distribution to \(\delta_c\); weak limits are unique, so
\([\mathbf K^\infty_{\boldsymbol\pi}(\mathbf Z)]_{ij}=c\) almost surely,
contrary to hypothesis.

\emph{Item 4.} Fix \(r\geq1\). Rosenthal's inequality for independent
mean-zero summands together with \(|h_\ell|\leq A_{0,\ell}\) on \(\supp\mu\)
gives a constant \(C_r\) with
\begin{equation*}
 \mathbb E|Q_\ell(\theta_{(0)})|^{2r}
 \leq C_r\bigl[(\mathbb Eh_\ell^2)^r+d_1^{1-r}\mathbb E|h_\ell|^{2r}\bigr]
 \leq2C_rA_{0,\ell}^{2r},
\end{equation*}
uniformly in \(d_1\); since \(\|\mathbf Q\|_2^{2r}\leq n^{r}\sum_\ell
|Q_\ell|^{2r}\) and \(r\geq1\) is arbitrary, Lyapunov's inequality shows that
every moment of \(\|\mathbf Q(\theta_{(0)})\|_2\) is bounded
uniformly in \(d_1\). Also
\(\|\widehat{\mathbf K}_{\mathcal L}\|_2\leq
\operatorname{tr}\widehat{\mathbf K}_{\mathcal L}\leq\sum_\ell B_\ell^2\)
almost surely, the first inequality because
\(\widehat{\mathbf K}_{\mathcal L}\succeq\mathbf0\). If \(\pi_i\) has degree
\(p_i\) then \(\|\nabla\pi_i(\mathbf q)\|_2\leq c_i(1+\|\mathbf q\|_2)^{p_i-1}\),
so by~\cref{eqn:composite_row_exact_kernel}
\(\sup_{d_1}\mathbb E|[\mathbf K_{(0)}]_{ij}|^4<\infty\). A family bounded in
\(L^4\) is uniformly integrable, and so is the family of its squares; weak
convergence then transfers the first two moments. In the affine case
\([\mathbf K_{(0)}]_{ij}\) is an average of \(d_1\) i.i.d.\ bounded terms, whence
variance \(O(d_1^{-1})\).
\end{proof}

The nondegeneracy hypothesis in item~3 is not an extra assumption in disguise:
under~\textup{(\(\mathsf C_{\mathrm{iso}}\))} and this paper's output-block certificate applied to the base family (\Cref{lem:base_family_certificates}), it holds for the diagonal entry of every genuinely nonlinear
polynomial row.

Centering cannot be dropped from item~1. Without~\textup{(\(\mathsf C\))},
item~3 of~\Cref{lem:base_family_limits} at \(\varrho=1/2\) gives
\(d_1^{-1/2}Q_\ell(\theta_{(0)})\to m_\ell\) in probability. Take
\(\pi_i(\mathbf q)=q_aq_b\) with \(a\neq b\), \((m_a,m_b)\neq(0,0)\) and
\(\mathbf K_{\mathcal L}^\infty\succ0\). Then~\cref{eqn:composite_row_exact_kernel}
gives \([\mathbf K_{(0)}]_{ii}\geq\lambda_{\min}(\widehat{\mathbf K}_{\mathcal
L}(\theta_{(0)}))\,(Q_a^2+Q_b^2)\), which tends to \(\infty\) in probability,
so \(\mathbf K_{(0)}\) has no weak limit; \Cref{prop:uncentered_branch} gives
the exact order \(d_1^{(q+q')/2-1}\) of each monomial block.

\begin{lemma}[The linear certificate controls both base matrices]
\label{lem:base_family_certificates}
Assume~\textup{(\(\mathsf C_{\mathrm{iso}}\))} and let
\(\mathbf K_{\mathcal L,\mathrm{out}}^\infty\) denote the output-weight block of
\(\mathbf K_{\mathcal L}^\infty\), that is
\([\mathbf K_{\mathcal L,\mathrm{out}}^\infty]_{\ell\ell'}
=\sum_{q=1}^{d_2}\langle[\mathbf f_\ell]_q,[\mathbf f_{\ell'}]_q
\rangle_{L^2(\mu_h)}\). Then
\begin{equation}
 \boldsymbol\Sigma_{\mathcal L}
 =c_{\mathbf a}\mathbf K_{\mathcal L,\mathrm{out}}^\infty,
 \qquad
 \mathbf K_{\mathcal L}^\infty\succeq
 \mathbf K_{\mathcal L,\mathrm{out}}^\infty .
 \label{eqn:base_family_certificate_identity}
\end{equation}
Consequently \(\mathbf K_{\mathcal L,\mathrm{out}}^\infty\succ0\) implies both
\(\boldsymbol\Sigma_{\mathcal L}\succ0\) and
\(\mathbf K_{\mathcal L}^\infty\succ0\). That hypothesis is the output-block
positive-definiteness certificate of
\Cref{def:positive_definiteness_certificate} applied to the \emph{base family},
not to the composite design; when the base observations are pointwise
differential rows and the clauses \(\mathsf P_{\mathrm{reg}}\),
\(\mathsf P_{\mathrm{pd}}\) of~\Cref{assump:pde_classes} hold,
\Cref{prop:limiting_ntk_rank} certifies it exactly when the evaluated
coefficient vectors of the base observations are linearly independent at every
sampled location, which holds for the base family
\(\{\partial_tu,u,\partial_xu,\partial_{xx}u\}\) of the conservation-law row.
\end{lemma}
\begin{proof}
Because \(\partial h_\ell/\partial a_q=[\mathbf f_\ell(\boldsymbol\xi)]_q\), the
\(d_2\) output-weight coordinates contribute exactly
\(\mathbf K_{\mathcal L,\mathrm{out}}^\infty\)
to~\cref{eqn:base_family_matrices}, that contribution depending on \(z\) only
through \(\boldsymbol\xi\), while the remaining \(d_0+1\) coordinates contribute
a positive-semidefinite matrix; this is the second relation. For the first,
\begin{equation*}
 [\boldsymbol\Sigma_{\mathcal L}]_{\ell\ell'}
 =\mathbb E\bigl[\mathbf f_\ell(\boldsymbol\xi)^\top\mathbf a\mathbf a^\top
   \mathbf f_{\ell'}(\boldsymbol\xi)\bigr]
 =\mathbb E_{\mu_h}\bigl[\mathbf f_\ell^\top
   \mathbb E[\mathbf a\mathbf a^\top\mid\boldsymbol\xi]\mathbf f_{\ell'}\bigr]
 =c_{\mathbf a}\mathbb E_{\mu_h}\bigl[\mathbf f_\ell^\top
   \mathbf f_{\ell'}\bigr].\qedhere
\end{equation*}
\end{proof}

\begin{corollary}[Dichotomy: affine rows concentrate, nonlinear rows do not]
\label{cor:nonlinear_diagonal_dichotomy}
Assume the hypotheses of~\Cref{prop:nonlinear_ntk_nondeterminism} with every
\(\pi_i\) a polynomial, and assume \(\boldsymbol\Sigma_{\mathcal L}\succ0\) and
\(\mathbf K_{\mathcal L}^\infty\succ0\); by
\Cref{lem:base_family_certificates}, under~\textup{(\(\mathsf C_{\mathrm{iso}}\))}, the single condition
\(\mathbf K_{\mathcal L,\mathrm{out}}^\infty\succ0\) suffices. Then for every
\(i\in[N]\),
\begin{equation*}
 [\mathbf K_{(0)}]_{ii}\ \text{has a deterministic weak limit}
 \iff\pi_i\ \text{is affine}.
\end{equation*}
In particular, if row \(i\) is genuinely nonlinear then \(\mathbf K_{(0)}\)
converges in distribution to a genuinely random matrix,
\(\Var([\mathbf K_{(0)}]_{ii})\) converges to a strictly positive limit, and
there is no population kernel to which a matrix Chernoff argument such
as~\Cref{prop:positive_definitenessfinitewidth} could transfer.
\end{corollary}
\begin{proof}
If \(\pi_i\) is affine the limit is deterministic by item~2
of~\Cref{prop:nonlinear_ntk_nondeterminism}. Conversely, write
\(\mathbf A=(\mathbf K_{\mathcal L}^\infty)^{1/2}\), which is invertible.
By~\cref{eqn:nonlinear_kernel_weak_limit},
\([\mathbf K_{(0)}]_{ii}\Rightarrow\|\mathbf A\nabla\pi_i(\mathbf Z)\|_2^2\).
Suppose this limit equals a constant \(c^2\) almost surely. Since
\(\boldsymbol\Sigma_{\mathcal L}\succ0\), the law of \(\mathbf Z\) has a
positive density on \(\mathbb R^n\), and
\(\mathbf z\mapsto\|\mathbf A\nabla\pi_i(\mathbf z)\|_2^2\) is continuous, so
the open set on which it differs from \(c^2\) would have positive probability
unless it is empty; hence \(\|\mathbf A\nabla\pi_i(\mathbf z)\|_2^2=c^2\) for
\emph{every} \(\mathbf z\in\mathbb R^n\). Let \(p_1,\ldots,p_n\) be the
components of the polynomial map \(\mathbf A\nabla\pi_i\), so
\(\sum_kp_k^2\equiv c^2\), and let \(d=\max_k\deg p_k\). If \(d\geq1\), write
\(\hat p_k\) for the degree-\(d\) homogeneous part of \(p_k\); the degree-\(2d\)
homogeneous part of \(\sum_kp_k^2\) is \(\sum_k\hat p_k^2\), which must vanish
identically because \(\sum_kp_k^2\) is constant and \(2d>0\). A finite sum of
squares of real polynomials vanishes identically only if every summand does, so
\(\hat p_k\equiv0\) for all \(k\), contradicting \(d\geq1\). Hence
\(\mathbf A\nabla\pi_i\) is constant and, \(\mathbf A\) being invertible,
\(\pi_i\) is affine. The final assertion combines this with item~4
of~\Cref{prop:nonlinear_ntk_nondeterminism}.
\end{proof}

\begin{corollary}[Product row against a linear row]
\label{cor:product_row_nondeterminism}
Assume the hypotheses of~\Cref{prop:nonlinear_ntk_nondeterminism}. Let
\(\pi_i(\mathbf q)=\boldsymbol\beta_i^\top\mathbf q+q_aq_b\) for some
\(a,b\in\mathcal L\) and \(\boldsymbol\beta_i\in\mathbb R^n\), and let
\(\pi_j\) be affine with \(\nabla\pi_j\equiv\boldsymbol\beta_j\). Put
\(\boldsymbol\kappa_j=\mathbf K_{\mathcal L}^\infty\boldsymbol\beta_j\). Then
\begin{equation}
 [\mathbf K_{(0)}]_{ij}\Longrightarrow
 \mathcal N\bigl(\boldsymbol\beta_i^\top\boldsymbol\kappa_j,\ \varsigma_{ij}^2\bigr),
 \qquad
 \varsigma_{ij}^2=\bigl\|[\boldsymbol\kappa_j]_ah_b
  +[\boldsymbol\kappa_j]_bh_a\bigr\|_{L^2(\mu)}^2 .
 \label{eqn:product_row_limit_variance}
\end{equation}
The limit is nondegenerate exactly when
\([\boldsymbol\kappa_j]_ah_b+[\boldsymbol\kappa_j]_bh_a\neq0\) in
\(L^2(\mu)\); the linear part \(\boldsymbol\beta_i\), which carries the
\(\partial_tu\) and \(-\nu\partial_{xx}u\) terms of a conservation-law residual,
shifts the mean and cannot remove the fluctuation. If \(a\neq b\) and
\(h_a,h_b\) are linearly independent in \(L^2(\mu)\), then \(\varsigma_{ij}^2>0\)
whenever \(([\boldsymbol\kappa_j]_a,[\boldsymbol\kappa_j]_b)\neq(0,0)\).
\end{corollary}
\begin{proof}
Here \(\nabla\pi_i(\mathbf z)=\boldsymbol\beta_i+z_b\mathbf e_a+z_a\mathbf e_b\),
so~\cref{eqn:nonlinear_kernel_weak_limit} gives
\([\mathbf K_{(0)}]_{ij}\Rightarrow\boldsymbol\beta_i^\top\boldsymbol\kappa_j
+[\boldsymbol\kappa_j]_aZ_b+[\boldsymbol\kappa_j]_bZ_a\), an affine function of
a Gaussian vector and hence Gaussian. With
\(\mathbf v=[\boldsymbol\kappa_j]_b\mathbf e_a+[\boldsymbol\kappa_j]_a
\mathbf e_b\), the variance is
\(\mathbf v^\top\boldsymbol\Sigma_{\mathcal L}\mathbf v
=\mathbb E_{z\sim\mu}([\boldsymbol\kappa_j]_ah_b(z)
+[\boldsymbol\kappa_j]_bh_a(z))^2\)
by~\cref{eqn:base_family_matrices}.
\end{proof}

\begin{example}[Explicit nondegeneracy for a convective row]
\label{ex:convective_row_nondegeneracy}
Take \(d_2=1\), assume clause \(\mathsf P_{\mathrm{pd}}\)
of~\Cref{assump:pde_classes}, so that \(\sigma\) is real analytic and
nonpolynomial, \(\gamma>0\), and \(\operatorname{int}(\supp\mu_h)
\neq\varnothing\), and assume~\textup{(\(\mathsf C_{\mathrm{iso}}\))}. Let
\(\ell_0,\ell_1\in\mathcal L\) index the value and the \(\partial_{x_1}\)
observation at one interior collocation point \(\mathbf x_i\), so that the
convective term of a quadratic-flux conservation law is
\(Q_{\ell_0}Q_{\ell_1}\), the product row of~\Cref{cor:product_row_nondeterminism}
with indices \(\ell_0,\ell_1\). Then \(h_{\ell_0}\) and \(h_{\ell_1}\) are
linearly independent in \(L^2(\mu)\). Indeed, write a neuron as
\(z=(\boldsymbol\xi,a)\) with \(\boldsymbol\xi=(\mathbf w,b)\) and scalar output
weight \(a\). A vanishing combination \(c_0h_{\ell_0}+c_1h_{\ell_1}=0\) reads
\(\mathbb E[a^2G(\boldsymbol\xi)^2]=0\) with
\(G(\mathbf w,b)=\widetilde G(\mathbf w,\gamma b)\), where
\(\widetilde G(\mathbf w,v)=c_0\sigma(\mathbf w^\top\mathbf x_i+v)
+c_1w_1\sigma'(\mathbf w^\top\mathbf x_i+v)\);
by~\textup{(\(\mathsf C_{\mathrm{iso}}\))} this forces \(G=0\)
\(\mu_h\)-almost surely, hence by continuity on \(\supp\mu_h\), which contains a
nonempty open set, and the identity theorem for real-analytic functions gives
\(G\equiv0\) on \(\mathbb R^{d_0+1}\). Since \(\gamma>0\), the map
\(b\mapsto\gamma b\) is onto \(\mathbb R\), so \(\widetilde G\equiv0\) on
\(\mathbb R^{d_0+1}\). But \(\widetilde G\) is a function of the
form~\cref{eqn:jinwu_specialized_polynomial} with a single physical site and
index set contained in \(\{\mathbf0,\mathbf e_1\}\); for a single site the
conclusion of~\Cref{lem:jinwu_sigma_independence} is that the associated
function is not identically zero, which forbids \(\widetilde G\equiv0\) unless
\(c_0=c_1=0\). This settles the \(L^2\) half of the nondegeneracy condition
in~\Cref{cor:product_row_nondeterminism}; the condition on
\(\boldsymbol\kappa_j\) depends on the particular affine row \(j\) and is not
verified here, which is why the statement that holds without that condition is the diagonal one
in~\Cref{cor:nonlinear_diagonal_dichotomy}.
\end{example}

\begin{remark}[Why the linear proof architecture is unavailable]
\label{rem:nonlinear_mechanism}
When every row is affine, with \(\nabla\pi_i\equiv\boldsymbol\beta_i\), \(\nabla_{\theta_m}P_i=d_1^{-1/2}\boldsymbol\Psi_{\mathcal L}(\theta_m)^\top\boldsymbol\beta_i\)
depends on neuron \(m\) alone, the per-neuron Gram blocks are i.i.d.,
\(\mathbf K_{(0)}\) is an empirical mean, and both the law of large numbers in
\Cref{thm:linearNTK} and the matrix Chernoff bound
of~\Cref{prop:positive_definitenessfinitewidth} apply verbatim. For a composite
row, \cref{eqn:composite_row_exact_kernel} multiplies the same block by the
global factor \(\nabla\pi_i(\mathbf Q(\theta))\), which all \(d_1\) neurons
share. The blocks remain exchangeable but are no longer independent, and the
shared factor is precisely the quantity governed by the central limit theorem:
it fluctuates at scale \(\Theta(1)\) and does not self-average at any width.
Conditionally on \(\mathbf Q(\theta_{(0)})\) that factor is frozen and
\(\widehat{\mathbf K}_{\mathcal L}(\theta_{(0)})\) is the only remaining source
of randomness; the conditioning also destroys the independence of its summands,
so a quenched analysis would have to re-establish its concentration rather than
inherit it. Such an analysis is not excluded here, and we do not carry one out.
\end{remark}

\begin{remark}[No deterministic spectral floor in the limit]
\label{rem:nonlinear_no_spectral_floor}
The eigenvalue floor that drives~\Cref{thm:linearPDEsconvergence} is lost as
well, not only the population kernel. Suppose row \(i\) is quadratic with
\(\nabla\pi_i(\mathbf z)=\boldsymbol\beta_i+2\mathbf C_i\mathbf z\) and
\(\boldsymbol\beta_i\in\Range(\mathbf C_i)\), and let
\(\boldsymbol\Sigma_{\mathcal L}\succ0\). The hypothesis on
\(\boldsymbol\beta_i\) holds for a pure product row
\(\pi_i(\mathbf q)=q_aq_b\); it fails for a residual row whose linear part lies
outside \(\Range(\mathbf C_i)\), such as
\(\pi_i(\mathbf q)=s_i(q_1+q_2q_3-\nu q_4)\), for which this remark says
nothing. Under it, \(\{\mathbf z:\nabla\pi_i(\mathbf z)=\mathbf0\}\) is a
nonempty affine set, so for every \(\varepsilon>0\) the open set
\(\{\mathbf z:\|\nabla\pi_i(\mathbf z)\|_2<\varepsilon\}\) is nonempty and is
charged by \(\mathbf Z\). Since \(\lambda_{\min}\) is continuous, the continuous
mapping theorem and~\cref{eqn:nonlinear_kernel_weak_limit} give
\(\lambda_{\min}(\mathbf K_{(0)})\Rightarrow
\lambda_{\min}(\mathbf K^\infty_{\boldsymbol\pi}(\mathbf Z))\), and the latter
is at most \(\|\mathbf A\nabla\pi_i(\mathbf Z)\|_2^2\) with
\(\mathbf A=(\mathbf K_{\mathcal L}^\infty)^{1/2}\), the smallest eigenvalue of
a symmetric matrix being at most any diagonal entry. Hence \(0\) lies in the
support of the limiting law of \(\lambda_{\min}\), and the portmanteau theorem
applied to an open set gives, \emph{for each} \(\varepsilon>0\), a constant
\(p_\varepsilon>0\) with
\begin{equation*}
 \liminf_{d_1\to\infty}
 \Pr\{\lambda_{\min}(\mathbf K_{(0)})<\varepsilon\}\geq p_\varepsilon .
\end{equation*}
Here one may take \(p_\varepsilon=\Pr\{\lambda_{\min}(\mathbf K^\infty_{\boldsymbol\pi}(\mathbf Z))<\varepsilon\}\), which tends to \(\Pr\{\lambda_{\min}(\mathbf K^\infty_{\boldsymbol\pi}(\mathbf Z))=0\}\) as \(\varepsilon\downarrow0\); that limit is zero for some designs and equals one for others, for instance whenever \(N>n\) or two rows have everywhere parallel gradients, in which case \(\lambda_{\min}(\mathbf K_{(0)})=0\) at every width; what
matters is that it does not decrease with \(d_1\), so no deterministic
\(\lambda_0>0\) can play the role it plays in~\Cref{thm:linearPDEsconvergence}.
A \(p\)-th-power boundary row \(\pi_i(\mathbf q)=s_iq_\ell^{\,p}\), \(p\geq2\), is
the extreme case: \([\mathbf K_{(0)}]_{ii}\Rightarrow
s_i^2p^2Z_\ell^{2p-2}[\mathbf K_{\mathcal L}^\infty]_{\ell\ell}\), whose law
accumulates at \(0\) whenever
\([\boldsymbol\Sigma_{\mathcal L}]_{\ell\ell}>0\).
\end{remark}

\subsection{The width exponent and the polynomial case}
\label{sec:nonlinear_scaling}

We now return to general \(\varrho\) and ask whether a different width exponent
restores a deterministic limit. It does so only trivially: for \(\varrho>1/2\) every block tends to \(0\), and for a monomial block of total degree at least three \(0\) is the only deterministic limit that any exponent \(\varrho\geq1/2\) produces (\Cref{thm:polynomial_scaling_trichotomy,cor:no_deterministic_polynomial_ntk}). The
reason is structural: by~\cref{eqn:polynomial_kernel_identity} the exponent
multiplies a single deterministic prefactor and leaves the limiting object
untouched. What a larger exponent does change is the relative size of blocks
of different degrees; \Cref{rem:normalized_kernel_linearization} records the
consequence for rows with a nonzero linear part.

\begin{theorem}[Trichotomy in the width exponent]
\label{thm:polynomial_scaling_trichotomy}
Assume~\Cref{assump:main}, let the base family satisfy~\Cref{assump:general}
and~\textup{(\(\mathsf C\))}, and fix monomials \(J,J'\) of degrees
\(q,q'\geq1\). Define the homogeneous polynomial of degree \(q+q'-2\)
\begin{equation}
 \Phi_{J,J'}(\mathbf z)
 =\sum_{r=1}^{q}\sum_{t=1}^{q'}
  \Bigl(\prod_{s\neq r}z_{j_s}\Bigr)\Bigl(\prod_{s\neq t}z_{l_s}\Bigr)
  [\mathbf K_{\mathcal L}^\infty]_{j_rl_t}.
 \label{eqn:limit_polynomial}
\end{equation}
Then, with \(\mathbf Z\sim\mathcal N_n(\mathbf0,
\boldsymbol\Sigma_{\mathcal L})\),
\begin{equation}
 d_1^{-(q+q')(1/2-\varrho)}K^{J,J'}(\theta_{(0)})
 \Longrightarrow\Phi_{J,J'}(\mathbf Z),
 \qquad
 Q_\ell(\theta_{(0)})=d_1^{1/2-\varrho}\widetilde Q_\ell(\theta_{(0)}),
 \label{eqn:trichotomy_scaling}
\end{equation}
and \(\Phi_{J,J'}\) does not depend on \(\varrho\). Consequently:
\begin{enumerate}[leftmargin=*,nosep]
\item if \(\varrho>1/2\), then \(Q_\ell(\theta_{(0)})\to0\) and
 \(K^{J,J'}(\theta_{(0)})\to0\) in probability, the latter at the order
 \(d_1^{(q+q')(1/2-\varrho)}\), which is exact whenever \(\Phi_{J,J'}\) does
 not vanish identically on \(\Range(\boldsymbol\Sigma_{\mathcal L})\);
\item if \(\varrho=1/2\), then \(Q_\ell(\theta_{(0)})\Rightarrow Z_\ell\) and
 \(K^{J,J'}(\theta_{(0)})\Rightarrow\Phi_{J,J'}(\mathbf Z)\), both of order
 one, and the kernel limit is deterministic if and only if \(\Phi_{J,J'}\) is
 constant on \(\Range(\boldsymbol\Sigma_{\mathcal L})\). For \(q=q'=1\) both
 products in~\cref{eqn:limit_polynomial} are empty, so
 \(\Phi_{J,J'}\equiv[\mathbf K_{\mathcal L}^\infty]_{j_1l_1}\) and the limit is
 always deterministic; for \(q+q'\geq3\) a constant \(\Phi_{J,J'}\) must vanish
 on \(\Range(\boldsymbol\Sigma_{\mathcal L})\), so the limit is random unless
 it is the constant \(0\);
\item if \(\varrho<1/2\), then \(|Q_\ell(\theta_{(0)})|\to\infty\) in
 probability whenever \([\boldsymbol\Sigma_{\mathcal L}]_{\ell\ell}>0\), and
 \(|K^{J,J'}(\theta_{(0)})|\to\infty\) in probability unless \(\Phi_{J,J'}\)
 vanishes identically on \(\Range(\boldsymbol\Sigma_{\mathcal L})\).
\end{enumerate}
\end{theorem}
\begin{proof}
Substituting \(Q_\ell=d_1^{1/2-\varrho}\widetilde Q_\ell\)
in~\cref{eqn:polynomial_kernel_identity} produces \((q-1)+(q'-1)\) factors
\(d_1^{1/2-\varrho}\), so the total prefactor is
\(d_1^{1-2\varrho+(q+q'-2)(1/2-\varrho)}=d_1^{(q+q')(1/2-\varrho)}\), and the
remaining expression is
\(\sum_{r,t}(\prod_{s\neq r}\widetilde Q_{j_s})
(\prod_{s\neq t}\widetilde Q_{l_s})[\widehat{\mathbf K}_{\mathcal L}]_{j_rl_t}\).
By~\Cref{lem:base_family_limits} and the continuous mapping theorem applied to
the polynomial map \((\mathbf z,\mathbf G)\mapsto\sum_{r,t}(\prod_{s\neq
r}z_{j_s})(\prod_{s\neq t}z_{l_s})G_{j_rl_t}\), this
converges weakly to \(\Phi_{J,J'}(\mathbf Z)\),
giving~\cref{eqn:trichotomy_scaling}. Item~1 follows by Slutsky's theorem.
For item~3, write \(Y_n=d_1^{-(q+q')(1/2-\varrho)}K^{J,J'}(\theta_{(0)})\),
so that \(Y_n\Rightarrow\Phi_{J,J'}(\mathbf Z)\). A polynomial that does not
vanish identically on \(V=\Range(\boldsymbol\Sigma_{\mathcal L})\) has a
Lebesgue-null zero set in \(V\), and the law of \(\mathbf Z\) has a density
on \(V\), so \(\Phi_{J,J'}(\mathbf Z)\neq0\) almost surely. Fix \(T\) and
\(\varepsilon>0\) and set \(\varepsilon_n=Td_1^{-(q+q')(1/2-\varrho)}\to0\).
Once \(\varepsilon_n<\varepsilon\), the portmanteau theorem for the open set
\(\{|\cdot|>\varepsilon\}\) gives
\(\liminf_n\Pr\{|K^{J,J'}(\theta_{(0)})|>T\}
=\liminf_n\Pr\{|Y_n|>\varepsilon_n\}
\geq\Pr\{|\Phi_{J,J'}(\mathbf Z)|>\varepsilon\}\),
and letting \(\varepsilon\downarrow0\) shows that
\(|K^{J,J'}(\theta_{(0)})|\to\infty\) in probability. The divergence of
\(Q_\ell\) is the same argument with \(Z_\ell\) in place of
\(\Phi_{J,J'}(\mathbf Z)\).

For item~2 the prefactor is \(1\). Write
\(V=\Range(\boldsymbol\Sigma_{\mathcal L})\), so the law of \(\mathbf Z\) is
supported on \(V\) and equivalent to Lebesgue measure there. If
\(\Phi_{J,J'}\equiv c\) on \(V\) then \(\Phi_{J,J'}(\mathbf Z)=c\) almost
surely; conversely, if \(\Phi_{J,J'}(\mathbf Z)=c\) almost surely then the
continuous function \(\Phi_{J,J'}-c\) vanishes Lebesgue-almost everywhere on
\(V\) and hence identically on \(V\). When \(q+q'\geq3\), \(\Phi_{J,J'}\) is
homogeneous of degree \(D=q+q'-2\geq1\) and \(V\) is a linear subspace, so
\(t\mathbf z\in V\) and \(c=\Phi_{J,J'}(t\mathbf z)=t^Dc\) for all
\(\mathbf z\in V\) and \(t\in\mathbb R\); taking \(t=0\) forces \(c=0\). When
\(q=q'=1\) the polynomial is the stated constant.
\end{proof}

\begin{corollary}[No exponent yields a nonzero deterministic polynomial kernel]
\label{cor:no_deterministic_polynomial_ntk}
Under the hypotheses of~\Cref{thm:polynomial_scaling_trichotomy}, let
\(q+q'\geq3\) and let \(\varrho\geq1/2\) be fixed. If
\(K^{J,J'}(\theta_{(0)})\) converges in probability to a deterministic limit,
that limit is \(0\). At the degree-matched exponent \(\varrho_p=(2p-1)/(2p)\)
with \(q=q'=p\geq2\) one has, at every width,
\begin{equation}
 K^{J,J'}(\theta_{(0)})=d_1^{-(p-1)}\,\Phi^{(d_1)}_{J,J'},
 \qquad
 Q_\ell(\theta_{(0)})=d_1^{-(p-1)/(2p)}\widetilde Q_\ell(\theta_{(0)}),
 \label{eqn:degree_matched_rates}
\end{equation}
where \(\Phi^{(d_1)}_{J,J'}\Rightarrow\Phi_{J,J'}(\mathbf Z)\); so both the
kernel block and the prediction converge to zero, the displayed orders being
attained whenever \(\Phi_{J,J'}\) does not vanish identically on
\(\Range(\boldsymbol\Sigma_{\mathcal L})\) and
\([\boldsymbol\Sigma_{\mathcal L}]_{\ell\ell}>0\). The remaining exponents,
\(\varrho<1/2\), already make the prediction \(Q_\ell\) itself diverge whenever \([\boldsymbol\Sigma_{\mathcal L}]_{\ell\ell}>0\) (item~3 of~\Cref{thm:polynomial_scaling_trichotomy}).
\end{corollary}
\begin{proof}
For \(\varrho>1/2\) the prefactor in~\cref{eqn:trichotomy_scaling} tends to
\(0\) while the residual factor is bounded in probability, so the limit is
\(0\). For \(\varrho=1/2\), convergence in probability to \(c\) implies
convergence in distribution to \(\delta_c\), so
\(\Phi_{J,J'}(\mathbf Z)=c\) almost surely, and the dichotomy established in
item~2 of~\Cref{thm:polynomial_scaling_trichotomy} forces \(c=0\). The
displayed rates are \(2p(1/2-\varrho_p)=-(p-1)\) and
\(1/2-\varrho_p=-(p-1)/(2p)\).
\end{proof}

\begin{remark}[The rescaled kernel for \(\varrho>1/2\)]
\label{rem:normalized_kernel_linearization}
For \(\varrho>1/2\) every block tends to zero, the affine ones included,
because~\cref{eqn:composite_row_exact_kernel} carries the factor
\(d_1^{1-2\varrho}\); a step size scaled up by \(d_1^{2\varrho-1}\), as in the
learning-rate scaling of~\citet{Jiang2026}, sees the rescaled matrix
\(d_1^{2\varrho-1}\mathbf K(\theta_{(0)})\) instead. Under the hypotheses
of~\Cref{thm:polynomial_scaling_trichotomy},
\(\mathbf Q(\theta_{(0)})=d_1^{1/2-\varrho}\widetilde{\mathbf Q}(\theta_{(0)})
\to\mathbf0\) in probability, because \(\widetilde{\mathbf Q}(\theta_{(0)})\)
converges in distribution, and
\(\widehat{\mathbf K}_{\mathcal L}(\theta_{(0)})\to\mathbf K_{\mathcal L}^\infty\)
in probability by~\Cref{lem:base_family_limits}. Since every
\(\nabla\pi_i\) is continuous, \cref{eqn:composite_row_exact_kernel} and the
continuous mapping theorem give
\begin{equation*}
 d_1^{2\varrho-1}[\mathbf K(\theta_{(0)})]_{ij}
 \to\nabla\pi_i(\mathbf0)^\top\mathbf K_{\mathcal L}^\infty\nabla\pi_j(\mathbf0)
 \qquad\text{in probability}.
\end{equation*}
This limit is deterministic, but it is the Gram matrix of the rows linearized
at the zero prediction, \(\mathbf Q\mapsto\nabla\pi_i(\mathbf0)^\top\mathbf Q\):
every monomial block of total degree at least three drops out, in agreement
with~\Cref{cor:no_deterministic_polynomial_ntk}, and a row without a linear
part, such as the product \(Q_1Q_2\), contributes a zero row and column. A
larger exponent therefore restores determinism at initialization only by
removing the nonlinearity from the kernel there. Along training, fitting
nonzero targets moves the predictions away from zero, the factors
\(\nabla\pi_i(\mathbf Q(\theta_{(k)}))\) move with them, and nothing in this
section controls the kernel there. This is the regime
of~\citet{Sirignano2026}, whose limit dynamics start from zero output and are
driven by a deterministic kernel that depends on the evolving solution.
\end{remark}

\begin{proposition}[The uncentered branch]
\label{prop:uncentered_branch}
Drop~\textup{(\(\mathsf C\))} and let
\(m_\ell=\mathbb E_{z\sim\mu}h_\ell(z)\). Then, almost surely along nested
widths as in~\Cref{thm:linearNTK} and in probability under fresh resampling,
\begin{equation}
 d_1^{\,1-(q+q')(1-\varrho)}K^{J,J'}(\theta_{(0)})
 \longrightarrow
 \sum_{r=1}^{q}\sum_{t=1}^{q'}
 \Bigl(\prod_{s\neq r}m_{j_s}\Bigr)\Bigl(\prod_{s\neq t}m_{l_s}\Bigr)
 [\mathbf K_{\mathcal L}^\infty]_{j_rl_t},
 \qquad
 d_1^{\varrho-1}Q_\ell(\theta_{(0)})\to m_\ell .
 \label{eqn:uncentered_limit}
\end{equation}
At \(\varrho=\varrho_p\) and \(q=q'=p\) the kernel prefactor is exactly \(1\)
and the limit is deterministic and generically nonzero, while for a block of
total degree \(q+q'<2p\) the prefactor is \(d_1^{(q+q')/(2p)-1}\to0\). This
branch is nevertheless inadmissible on two counts: the prediction grows as
\(d_1^{1-\varrho_p}=d_1^{1/(2p)}\); and forcing an order-one prediction by
and forcing an order-one prediction by taking \(\varrho=1\) makes every block, the affine ones included, of order \(d_1^{-1}\). That uniform factor is not the obstruction by itself: a step size scaled up by \(d_1\), as in~\Cref{rem:normalized_kernel_linearization}, removes it, and by~\cref{eqn:uncentered_limit} at \(\varrho=1\) every rescaled block \(d_1K^{J,J'}(\theta_{(0)})\) then has a deterministic limit, which is generically nonzero for every \((q,q')\). Under that step size, however, each neuron's parameters move by an amount of order \(\eta\) per step instead of \(O(\eta d_1^{-1/2})\), so the per-neuron localization behind~\Cref{thm:linearPDEsconvergence} is unavailable.
\(d_1^{-1}\). It also contradicts the centering hypothesis used
by~\Cref{lem:initial_residual_bound} and~\Cref{thm:linearPDEsconvergence}.
\end{proposition}
\begin{proof}
Write \(Q_\ell=d_1^{1-\varrho}\bar Q_\ell\) with
\(\bar Q_\ell=d_1^{-1}\sum_mh_\ell(\theta_m)\to m_\ell\), almost surely along
nested widths and in probability under fresh resampling, by item~3
of~\Cref{lem:base_family_limits}. Substituting
in~\cref{eqn:polynomial_kernel_identity} gives the prefactor
\(d_1^{1-2\varrho+(q+q'-2)(1-\varrho)}=d_1^{(q+q')(1-\varrho)-1}\), and the
stated limit follows in the same two modes from the continuous mapping
theorem, since \(\widehat{\mathbf K}_{\mathcal L}\to
\mathbf K_{\mathcal L}^\infty\) in the same mode by item~1
of~\Cref{lem:base_family_limits}. At \(\varrho=\varrho_p\) the
exponent is \((q+q')/(2p)-1\), which is \(0\) for \(q=q'=p\).
\end{proof}

\begin{remark}[What degree matching does and does not achieve]
\label{rem:degree_matching}
\Cref{prop:uncentered_branch} isolates the one regime in which the
degree-matched exponent behaves as intended. Without the \(p\)-th-power
boundary transform, a residual--boundary block of a degree-\(p\) constraint has
total degree \(q+q'=p+1\), hence order \(d_1^{(1-p)/(2p)}\to0\) against an
order-one interior block; raising the boundary row to the \(p\)-th power
restores \(q+q'=2p\) for every block between top-degree monomials and equalizes their orders; a block involving a lower-degree monomial has \(q+q'<2p\) and, by~\Cref{prop:uncentered_branch}, tends to \(0\). That accounting
is correct \emph{only} without centering. Under~\textup{(\(\mathsf C\))} the
ordering reverses, because the block order
\(d_1^{(q+q')(1/2-\varrho)}\) \emph{decreases} in total degree when
\(\varrho>1/2\); the transform still equalizes the orders, but at the common
value \(d_1^{-(p-1)}\), that is, it equalizes them at zero. At \(\varrho=1/2\)
with centering there is no hierarchy at all, since the prefactor is \(1\) for
every \((q,q')\), so the transform is not needed for its stated purpose; what
fails there is determinism, not order. Finally, if one allows a
width-dependent mean \(m_\ell\asymp d_1^{-\delta}\) with
\(\delta\in[0,1/2]\), outside~\Cref{assump:main}, then an order-one prediction
forces \(\varrho=1-\delta\) and the block order collapses to
\(d_1^{2\delta-1}\) independently of \(q\) and \(q'\); a nonvanishing kernel
then forces \(\delta=1/2\) and hence \(\varrho=1/2\). This is not exact
centering: the mean enters at the scale of the fluctuations, so, whenever
\(c_\ell=\lim_{d_1\to\infty}d_1^{1/2}m_\ell\) exists and a triangular-array
central limit theorem applies, \(Q_\ell(\theta_{(0)})\Rightarrow
c_\ell+Z_\ell\), a shifted version of the Gaussian-fluctuation regime
of~\Cref{sec:nonlinear_nondeterminism}, which treats only \(c_\ell=0\). The
obstruction is centering together with degree, not the exponent. We record this
last observation as a scaling heuristic only; it is used nowhere else.
\end{remark}

\begin{remark}[Reachability of the uncentered branch]
\label{rem:uncentered_reachability}
\Cref{prop:uncentered_branch} requires \(m_\ell=\mathbb E_{z\sim\mu}h_\ell(z)
\neq0\). In general
\(m_\ell=\mathbb E\bigl[\mathbb E[\mathbf a\mid\boldsymbol\xi]^\top
\mathbf f_\ell(\boldsymbol\xi)\bigr]\), and
\(m_\ell=\mathbb E[\mathbf a]^\top\mathbb E[\mathbf f_\ell]\) when the output
weight is drawn independently of \(\boldsymbol\xi\). Giving the output weight
a nonzero mean is therefore not sufficient. If the activation is odd and the
hidden law \(\mu_h\) is symmetric about the origin, then \(\mathbf f_\ell\) is
odd in \(\boldsymbol\xi\), so \(\mathbb E[\mathbf f_\ell]=\mathbf0\) and
\(m_\ell=0\) for every output-weight law drawn independently of
\(\boldsymbol\xi\); the prediction is then pure fluctuation and, at the
degree-matched exponent, decays as \(d_1^{1/2-\varrho_p}=d_1^{-(p-1)/(2p)}\),
the rate in~\cref{eqn:degree_matched_rates}, rather than growing. With output
weights drawn independently of \(\boldsymbol\xi\), reaching the branch
therefore requires an asymmetric hidden law or a nonodd activation;
\Cref{assump:main} does not impose that independence, and a conditional mean
\(\mathbb E[\mathbf a\mid\boldsymbol\xi]\) that is not even in
\(\boldsymbol\xi\) can reach the branch as well.
\end{remark}

\begin{remark}[Numerical confirmation]
\label{rem:nonlinear_numerics}
The predictions
of~\Cref{cor:nonlinear_diagonal_dichotomy,cor:no_deterministic_polynomial_ntk,%
prop:uncentered_branch,rem:uncentered_reachability} were checked directly for
\(\sigma=\tanh\), \(\gamma=1\), \(d_0=d_2=1\), collocation points
\(x_1=0.3\) and \(x_2=-0.7\), per-neuron parameters
\(\Unif[-1,1]^3\) in the centered runs, widths
\(d_1\in\{64,256,1{,}024,4{,}096,16{,}384\}\), and \(100\)
independent initializations per width. The base observations are the values
\(Q_1=u(x_1)\) and \(Q_2=u(x_2)\) with loss scale \(1\); the linear row is
\(Q_1\), the product row is \(Q_1Q_2\), and each Gram entry below is the
diagonal entry of the tangent Gram matrix at initialization for that row.
Over that \(256\)-fold range of widths, \(d_1\operatorname{Var}\) of the
linear-row entry stays between \(4.55\times10^{-2}\) and
\(6.26\times10^{-2}\), while the variance of the product-row entry stays
between \(1.83\times10^{-2}\) and \(3.95\times10^{-2}\), with no trend
comparable to the \(256\)-fold decay of a concentrating quantity; at
\(d_1=16{,}384\) the product-row variance is about \(5.0\times10^{3}\) times
the variance of the linear-row entry, consistent with the dichotomy
of~\Cref{cor:nonlinear_diagonal_dichotomy}, under which this ratio diverges as
\(d_1\to\infty\). At \(p=2\) and \(\varrho_2=3/4\) with centered output
weights, the seed mean of \(d_1\) times the degree-\((2,2)\) block
\(K^{J,J}\), \(J=(1,2)\), stays between \(0.124\) and \(0.153\). Since
\cref{eqn:degree_matched_rates} is exact at every width, this quantity is the
seed mean of \(\Phi^{(d_1)}_{J,J}\), which on the same initializations equals
the \(\varrho=1/2\) product-row entry; the factor \(d_1^{-1}\) is therefore
not measured independently, and what the run checks is that
\(\Phi^{(d_1)}_{J,J}\) stays of order one. The uncentered runs draw the output
weights from \(\Unif[0,2]\) (mean \(1\)). With the hidden bias law also
shifted to \(\Unif[0,2]\), which makes the hidden law asymmetric
as~\Cref{rem:uncentered_reachability} requires for \(m_\ell\neq0\), the seed
mean of \(|u(x_1;\theta_{(0)})|\) at \(\varrho_2=3/4\) divided by
\(d_1^{1/4}=d_1^{1-\varrho_2}\) stays between \(0.651\) and \(0.658\),
consistent with the growth rate of~\Cref{prop:uncentered_branch}; with the symmetric hidden bias law
\(\Unif[-1,1]\) it instead decays, from \(1.7\times10^{-1}\) at \(d_1=64\) to
\(4.7\times10^{-2}\) at \(d_1=16{,}384\), as that remark predicts. The run is
the nonlinear kernel-variance experiment of the accompanying code, under its
tracked configuration.
\end{remark}

\begin{remark}[Scope of the negative results]
\label{rem:nonlinear_scope}
\Cref{prop:nonlinear_ntk_nondeterminism,cor:nonlinear_diagonal_dichotomy,%
thm:polynomial_scaling_trichotomy,cor:no_deterministic_polynomial_ntk} concern
the tangent Gram matrix at initialization. They say nothing about
\(\mathbf K(\theta_{(k)})\) for \(k\geq1\), where \(\mathbf Q(\theta)\) has
drifted and is no longer a sum of i.i.d.\ terms, and they do not assert that
gradient descent fails on nonlinear PINNs; in particular they do not contradict
the global convergence analysis for nonlinear PDEs of~\citet{Sirignano2026},
which works at an exponent \(\varrho>1/2\) with zero initial output, where
by their own limit theorem, not by~\Cref{rem:normalized_kernel_linearization} (whose hypotheses their full-support initialization and population residual do not meet), the rescaled kernel at
initialization is deterministic and along training depends on the evolving
solution, nor the stability-based error analyses
of~\citet{Mishra2023,DeRyck2024}, which do not route through a tangent
kernel. What they rule out is the specific proof architecture used
here, which follows the two-layer analysis of~\citet{Du2019} and the
limiting-kernel picture of~\citet{Jacot2018,Lee2019} as specialized to PINNs
by~\citet{Wang2022}: a deterministic population kernel, a width threshold
obtained by transferring a population eigenvalue through a matrix concentration
inequality, and a spectral floor propagated along the trajectory. At
\(\varrho=1/2\), under a single genuinely nonlinear polynomial row and the hypotheses of~\Cref{cor:nonlinear_diagonal_dichotomy}, none of the
three exists in the required form; at \(\varrho>1/2\) the first exists only
for the rows linearized at zero. What remains open, and is not excluded by anything above, is
whether, for a design with no row to which~\Cref{rem:nonlinear_no_spectral_floor} applies (that remark puts \(0\) in the support of the limiting law of \(\lambda_{\min}\) as soon as one row is a pure product or a \(p\)-th power), a random but uniformly positive definite limiting kernel would still
yield convergence, and whether layerwise or coordinatewise scalings outside the
one-parameter family of~\Cref{def:exponent_normalization} behave differently.
\end{remark}

\subsection{Toward depth: the assumptions, the implication, and what remains
conjectural}
\label{sec:depth_extension}

Every result above is proved for two-layer networks: for the width-exponent family of~\Cref{def:exponent_normalization} where $\varrho$ appears, and otherwise for the parameterization
of~\Cref{def:ntk_parameterization}. This subsection isolates, for a reader who
wants to know precisely where the depth-$H$ difficulty lies, one fact that is
proved unconditionally at every depth, one regularity condition on the full
network Jacobian that would suffice, the substitution proposition that this
condition implies, and the two statements that remain conjectural. We claim no
depth-$H$ theorem: apart from~\Cref{lem:layerwise_decomposition}, every
displayed hypothesis below is \emph{assumed}, not derived.

\subsubsection{The depth-\texorpdfstring{$H$}{H} network}

\begin{definition}[Depth-$H$ network]
\label{def:depth_h_network}
Fix a depth $H\geq1$ and hidden widths $d^{[1]},\ldots,d^{[H]}$, and set
$d^{[0]}=d_0$ and $d^{[H+1]}=d_2$, the input and output dimensions
of~\Cref{def:ntk_parameterization}; the bracketed superscript is a layer
index, so no $d^{[l]}$ with $l\in[H]$ is to be read as one of the two-layer
symbols $d_0,d_1,d_2$, and $d^{[1]}=d_1$ when $H=1$. For $l\in[H]$ let
$\mathbf W^{[l]}\in\mathbb R^{d^{[l]}\times d^{[l-1]}}$ and
$\mathbf b^{[l]}\in\mathbb R^{d^{[l]}}$, and let
$\mathbf W^{[H+1]}\in\mathbb R^{d^{[H+1]}\times d^{[H]}}$. With the activation
$\sigma$ and the fixed bias scale $\gamma\geq0$
of~\Cref{def:ntk_parameterization}, define
\begin{equation}
\begin{aligned}
 \mathbf t^{[1]}(\mathbf x)
 &=\mathbf W^{[1]}\mathbf x+\gamma\mathbf b^{[1]},\\
 \mathbf t^{[l]}(\mathbf x)
 &=\frac1{\sqrt{d^{[l-1]}}}\mathbf W^{[l]}\mathbf y^{[l-1]}(\mathbf x)
   +\gamma\mathbf b^{[l]},
 \qquad l=2,\ldots,H,\\
 \mathbf y^{[l]}(\mathbf x)
 &=\sigma\bigl(\mathbf t^{[l]}(\mathbf x)\bigr),
 \qquad l\in[H],\\
 \mathbf u(\mathbf x;\theta)
 &=\frac1{\sqrt{d^{[H]}}}\mathbf W^{[H+1]}\mathbf y^{[H]}(\mathbf x),
\end{aligned}
\label{eqn:depth_network}
\end{equation}
with $\sigma$ applied entrywise,
$\theta^{[l]}=(\mathbf W^{[l]},\mathbf b^{[l]})$ for $l\in[H]$,
$\theta^{[H+1]}=\mathbf W^{[H+1]}$, and
$\theta=(\theta^{[1]},\ldots,\theta^{[H+1]})\in\mathbb R^{D_\theta}$ with
$D_\theta=\sum_{l\in[H]}d^{[l]}(d^{[l-1]}+1)+d^{[H+1]}d^{[H]}$. Every listed
parameter is trained. The rows of~\Cref{def:linear_pde}, the loss, the
residual, and the definitions
$[\mathbf J(\theta)]_{i,:}=\nabla_\theta P_i(\theta)^\top$ and
$\mathbf K(\theta)=\mathbf J(\theta)\mathbf J(\theta)^\top$
of~\Cref{def:pinn_loss} apply to~\cref{eqn:depth_network} unchanged, with
$\mathbf J(\theta)\in\mathbb R^{N\times D_\theta}$ in place of the two-layer shape
$\mathbb R^{N\times d_1d_\theta}$ recorded there, which is the only clause of
that definition that refers to the architecture. Write
$d_{\min}=\min_{l\in[H]}d^{[l]}$, let $\mathbf J^{[l]}(\theta)$ denote the
column block of $\mathbf J(\theta)$ indexed by the coordinates of
$\theta^{[l]}$, and set
\begin{equation}
 \mathbf K^{[l]}(\theta)
 =\mathbf J^{[l]}(\theta)\mathbf J^{[l]}(\theta)^\top
 \in\mathbb R^{N\times N},
 \qquad l\in[H+1].
 \label{eqn:layer_jacobian_blocks}
\end{equation}
At initialization the coordinates of $\theta_{(0)}$ are drawn independently
across layers, those within layer $l$ being i.i.d.\ from a width-independent,
compactly supported scalar law $\mu^{[l]}$.
\end{definition}

At $H=1$, \cref{eqn:depth_network} reads
$\mathbf u(\mathbf x;\theta)
=d_1^{-1/2}\mathbf W^{[2]}\sigma(\mathbf W^{[1]}\mathbf x
+\gamma\mathbf b^{[1]})$, which is~\cref{eqn:network_main} written in matrix
form: the $m$th row of $(\mathbf W^{[1]},\mathbf b^{[1]})$ is
$\boldsymbol\xi_m=(\mathbf w_m,b_m)$, the $m$th column of $\mathbf W^{[2]}$ is
$\mathbf a_m$, and the first layer carries no width factor, exactly as
in~\Cref{def:ntk_parameterization}. The initialization clause is the special
case of~\Cref{assump:main} in which
$\mu=(\mu^{[1]})^{\otimes(d_0+1)}\otimes(\mu^{[2]})^{\otimes d_2}$, a law on
$\mathbb R^{d_\theta}$ with independent coordinates, and
$\int z\,d\mu^{[H+1]}(z)=0$ is the depth analogue of the conditional
output-weight centering $\mathbb E[\mathbf a\mid\boldsymbol\xi]=\mathbf0$,
which the two-layer analysis imposes only for gradient descent. The
partition~\cref{eqn:layer_jacobian_blocks} at $H=1$ is the hidden/output
partition behind~\cref{eqn:block_pd_certificate_bound}.

Two structural features of~\cref{eqn:depth_network} are visible immediately.
First, unit $m$ of layer $l\geq2$ carries the parameter block
$\bigl((\mathbf W^{[l]})_{m,:},b^{[l]}_m\bigr)\in\mathbb R^{d^{[l-1]}+1}$,
whose dimension \emph{grows} with the width of the layer below, whereas at
$H=1$ every unit carries the fixed number $d_\theta=d_0+d_2+1$ of coordinates.
The per-unit map $h_i$ of~\cref{eqn:predictionmap} therefore has no depth-$H$
domain: for $H\geq2$ the row-regularity condition of~\Cref{assump:general}
cannot even be typed, let alone verified. Second, for $H\geq2$ the prediction
$P_i(\theta)$ is not a sum of independent per-unit terms, because the hidden
layers couple units; the strong law that produces $\mathbf K_{(0)}^\infty$
in~\Cref{thm:linearNTK} and the matrix-Chernoff step
of~\Cref{prop:positive_definitenessfinitewidth} both lose their hypothesis.

What replaces~\cref{eqn:predictionmap} is a bilinear contraction of two
empirical Gram matrices attached to a single layer: a forward matrix built
from $\mathbf y^{[l-1]}$ and a backward matrix built from
$\partial\mathbf u/\partial\mathbf t^{[l]}$. For a scalar output
($d^{[H+1]}=1$) and value rows $P_i(\theta)=s_iu(\mathbf x_i;\theta)$ only ($r_{\max}=0$; a general order-zero row $P_i(\theta)=s_ia_{i1\alpha}(\mathbf x_i)u(\mathbf x_i;\theta)$ with $\alpha=0$ is covered by replacing $s_i$ with $s_ia_{i1\alpha}(\mathbf x_i)$ in $\mathbf S$), a direct computation
from~\cref{eqn:depth_network} gives, for $l\in[H]$, the Schur-product form
\begin{equation}
 \mathbf K^{[l]}
 =\mathbf S\Bigl[
   \bigl(\widehat{\mathbf G}^{[l-1]}
     +\gamma^2\mathbf 1\mathbf 1^\top\bigr)
   \odot\widehat{\mathbf B}^{[l]}\Bigr]\mathbf S,
 \qquad
 \begin{aligned}
 \widehat{\mathbf G}^{[l-1]}_{ij}
 &=\tfrac1{d^{[l-1]}}\bigl\langle
   \mathbf y^{[l-1]}(\mathbf x_i),\mathbf y^{[l-1]}(\mathbf x_j)
   \bigr\rangle,\\
 \widehat{\mathbf B}^{[l]}_{ij}
 &=\Bigl\langle
   \tfrac{\partial u(\mathbf x_i)}{\partial\mathbf t^{[l]}},
   \tfrac{\partial u(\mathbf x_j)}{\partial\mathbf t^{[l]}}
   \Bigr\rangle,
 \end{aligned}
 \label{eqn:depth_schur_product}
\end{equation}
where $\odot$ is the entrywise product, $\mathbf S=\operatorname{diag}(s_i)$,
$\mathbf 1\in\mathbb R^N$ is the all-ones vector, and
$\widehat{\mathbf G}^{[0]}_{ij}=\langle\mathbf x_i,\mathbf x_j\rangle$ without
the width factor; the output block, which carries no bias, is
$\mathbf K^{[H+1]}=\mathbf S\widehat{\mathbf G}^{[H]}\mathbf S$. This is the
structure analyzed by~\citet{Du2019b}, up to their bias-free scaling, in which
$\gamma=0$ and the term $\gamma^2\mathbf 1\mathbf 1^\top$ is absent, their
uniform hidden width, and their placement of the normalizing factors inside
the hidden layers rather than on the output map. For $r_{\max}\geq1$ both
factors must be enlarged to the jet of order $r_{\max}$: the forward factor
records $\partial^\beta_{\mathbf x}\mathbf y^{[l-1]}(\mathbf x_i)$ and the
backward factor the corresponding sensitivities, for $|\beta|\leq r_{\max}$,
and the Schur product is replaced by a contraction of the two jet blocks
against the row coefficients $a_{iq\alpha}(\mathbf x_i)$. This jet enlargement
is the PINN-specific part of the depth problem; it is absent
from~\citet{Du2019b,AllenZhu2019,Zou2020}, all of which treat an ordinary
supervised loss at differential order zero.

The residue of independence is thinner than this presentation may suggest, and
it is worth stating exactly, because it decides which layer a concentration
argument may even be about. Let $\mathcal F_{<l}$ be the $\sigma$-algebra
generated by $\theta^{[1]},\ldots,\theta^{[l-1]}$. Conditionally on
$\mathcal F_{<l}$, the summands of $\widehat{\mathbf G}^{[l]}$ indexed by the
units $p\in[d^{[l]}]$ of layer $l$ are i.i.d., for every $l$ and for the jet
enlargement as well, since $\partial^\beta_{\mathbf x}\mathbf y^{[l]}_p$
depends only on the parameters
$\bigl((\mathbf W^{[l]})_{p,:},b^{[l]}_p\bigr)$ of unit $p$ and on the
$\mathcal F_{<l}$-measurable jets of $\mathbf y^{[l-1]}$ at the sampled sites.
For $l<H$ and $\sigma$ not affine we know of no conditioning that does the
same for $\widehat{\mathbf B}^{[l]}$. By~\cref{eqn:depth_network},
$\partial u/\partial t^{[l]}_p
=\sigma'(t^{[l]}_p)\,\partial u/\partial y^{[l]}_p$, and
$\partial u/\partial y^{[l]}_p$ is computed through $\sigma'(\mathbf t^{[l+1]})$,
where $\mathbf t^{[l+1]}
=(d^{[l]})^{-1/2}\mathbf W^{[l+1]}\mathbf y^{[l]}+\gamma\mathbf b^{[l+1]}$
depends on \emph{every} unit of layer $l$; conditioning on the layers above
therefore does not in general decouple the summands over $p$. (For affine
$\sigma$ the coupling disappears, since $\partial u/\partial\mathbf t^{[l+1]}$
is then measurable with respect to $\theta^{[l+2]},\ldots,\theta^{[H+1]}$.) At
$l=H$ the chain collapses:
$\partial u_q/\partial t^{[H]}_p
=(d^{[H]})^{-1/2}\mathbf W^{[H+1]}_{q,p}\sigma'(t^{[H]}_p)$ depends only on the
single enlarged unit block
$\bigl((\mathbf W^{[H]})_{p,:},b^{[H]}_p,\mathbf W^{[H+1]}_{:,p}\bigr)$ and on
$\mathcal F_{<H}$, and so do its spatial derivatives of order at most
$r_{\max}$; in the setting of~\cref{eqn:depth_schur_product} the factor
$(d^{[H]})^{-1/2}$ makes $\widehat{\mathbf B}^{[H]}$ an average over those
blocks. Consequently exactly two blocks are, conditionally on
$\mathcal F_{<H}$, averages over the units of layer $H$ of i.i.d.\ terms:
$\mathbf K^{[H+1]}$, which is $\mathbf S\widehat{\mathbf G}^{[H]}\mathbf S$ in
that setting, and $\mathbf K^{[H]}$, whose forward factor
($\widehat{\mathbf G}^{[H-1]}+\gamma^2\mathbf 1\mathbf 1^\top$ in that setting)
is $\mathcal F_{<H}$-measurable. At $H=1$ the field $\mathcal F_{<1}$ is
trivial, the network is~\cref{eqn:network_main}, and these two blocks are the
hidden and output blocks of the two-layer analysis, to which the strong law
of~\Cref{thm:linearNTK} and the matrix-Chernoff argument
of~\Cref{prop:positive_definitenessfinitewidth} apply. For $H\geq2$, even for
those two the conditional law is not the unconditional one and the
conditioning field is itself random, so neither tool applies to
$\mathbf K^{[l]}$ at any $l$; what the structure supplies is the only visible route to
clause~(iv) of~\Cref{def:depth_row_regularity} below, and the reason to look
for the certified layer among $l_\star\in\{H,H+1\}$. \Citet{Du2019b} work with
the block of the last hidden layer, a choice they present as a simplification
costing a minor degradation of the rate.

\subsubsection{What is proved at every depth: the layerwise decomposition}

\begin{lemma}[Layerwise decomposition of the tangent Gram matrix]
\label{lem:layerwise_decomposition}
Let $\mathbf J(\theta)\in\mathbb R^{N\times D_\theta}$ be any prediction Jacobian and
partition its columns into blocks
$\mathbf J^{[1]},\ldots,\mathbf J^{[H+1]}$, for instance
the layer blocks of~\Cref{def:depth_h_network}, and set $\mathbf K^{[l]}=\mathbf J^{[l]}\mathbf J^{[l]\top}$ as in~\cref{eqn:layer_jacobian_blocks}. Then, for every $\theta$,
\begin{equation}
 \mathbf K(\theta)=\sum_{l=1}^{H+1}\mathbf K^{[l]}(\theta),
 \qquad
 \mathbf K^{[l]}(\theta)\succeq0,
 \label{eqn:layerwise_sum}
\end{equation}
and consequently
\begin{equation}
 \ker\mathbf K(\theta)=\bigcap_{l=1}^{H+1}\ker\mathbf K^{[l]}(\theta),
 \qquad
 \lambda_{\min}(\mathbf K(\theta))
 \geq\sum_{l=1}^{H+1}\lambda_{\min}(\mathbf K^{[l]}(\theta))
 \geq\max_{l\in[H+1]}\lambda_{\min}(\mathbf K^{[l]}(\theta)).
 \label{eqn:layerwise_kernel_and_gap}
\end{equation}
In particular $\mathbf K(\theta)\succ0$ as soon as
$\mathbf K^{[l_\star]}(\theta)\succ0$ for a single $l_\star$: certifying one
layer certifies the whole kernel. The same statements hold for the limiting
objects whenever the blockwise limits exist.
\end{lemma}

\begin{proof}
Splitting the product $\mathbf J\mathbf J^\top$ along the column partition
gives~\cref{eqn:layerwise_sum}, and each
$\mathbf K^{[l]}=\mathbf J^{[l]}\mathbf J^{[l]\top}$ is a Gram matrix, hence
positive semidefinite. For $\mathbf c\in\mathbb R^N$,
\[
 \mathbf c^\top\mathbf K(\theta)\mathbf c
 =\sum_{l=1}^{H+1}\|\mathbf J^{[l]}(\theta)^\top\mathbf c\|_2^2 ,
\]
a sum of nonnegative terms, so it vanishes if and only if every term vanishes;
and for a positive semidefinite $\mathbf M$ one has $\mathbf M\mathbf c=0$ if
and only if
$\mathbf c^\top\mathbf M\mathbf c=\|\mathbf M^{1/2}\mathbf c\|_2^2=0$, which
turns that equivalence into the kernel identity
in~\cref{eqn:layerwise_kernel_and_gap}. The eigenvalue bound is
superadditivity of $\lambda_{\min}$ over a sum of symmetric matrices, a case
of Weyl's inequality; the second inequality holds because every omitted
summand is nonnegative, each $\mathbf K^{[l]}$ being positive semidefinite. For the limiting objects, suppose each $\mathbf K^{[l]}$ converges, almost surely or in probability, to $\mathbf K^{[l],\infty}$. Then $\mathbf K$ converges in the same sense to $\sum_{l=1}^{H+1}\mathbf K^{[l],\infty}$, each $\mathbf K^{[l],\infty}$ is positive semidefinite because the positive semidefinite cone is closed, and the two arguments above used only $\mathbf c^\top\mathbf K\mathbf c=\sum_l\mathbf c^\top\mathbf K^{[l]}\mathbf c$ with positive semidefinite summands, so they apply to the limits unchanged.
\end{proof}

The lemma is unconditional: no assumption on $\sigma$, on the initialization,
on the rows, or on the widths enters it, and it holds at every iterate, not
only at initialization. At $H=1$ its sum form is
exactly~\cref{eqn:full_output_hidden_decomposition}, and the eigenvalue
inequality~\cref{eqn:block_pd_certificate_bound} is its consequence,
so~\Cref{def:positive_definiteness_certificate} extends verbatim: a depth-$H$
positive-definiteness certificate is a proof that
$\mathbf K^{[l_\star],\infty}\succ0$ for one layer $l_\star$, which certifies the full limiting kernel whenever the remaining blockwise limits exist; the positivity of no other layer need
be examined. This is also what licenses~\citet{Du2019b} to work with the block
of the last hidden layer alone. Two cautions. The bound transfers positivity
but not conditioning: the upper bound obeys only
$\lambda_{\max}(\mathbf K)\leq\sum_l\lambda_{\max}(\mathbf K^{[l]})$, so the
envelope $\Lambda$ of~\Cref{thm:generalconvergence} remains a statement about
\emph{all} layers. And the one-layer criterion is sufficient but not
necessary: $\mathbf K$ can be positive definite with every block singular,
because $\bigcap_l\ker\mathbf K^{[l]}$ can be $\{\mathbf 0\}$ although no single $\ker\mathbf K^{[l]}$ is.

\subsubsection{The assumptions}

The next definition is the heart of the matter. Because~\Cref{assump:general}
cannot be typed at depth, its role must be taken by a condition on the
\emph{full network Jacobian} over a neighborhood of initialization. Each clause
is named and matched to the two-layer object it generalizes. Nothing in it is
constructed: the event $\mathcal E_{(0)}$, the constants, the limiting matrix
and the threshold are all posited.

\begin{definition}[Depth-$H$ row regularity, condition $(\mathsf G^H_\tau)$]
\label{def:depth_row_regularity}
Consider~\Cref{def:depth_h_network} with a fixed finite family of loss-scaled
rows, and fix $\tau>0$ and $0<\delta<1$. Define the trust region as the
closed Euclidean ball
\begin{equation}
 \mathcal T^H_\tau(\theta_{(0)})
 =\bigl\{\theta:\ \|\theta-\theta_{(0)}\|_2\leq\tau\bigr\},
 \label{eqn:depth_trust_region}
\end{equation}
whose radius does not grow with the widths. A region that let every layer
block move by $\tau\sqrt{d_{\min}}$ in Frobenius norm would be too large: a
single unit could then absorb a whole layer's displacement budget, and already
at $H=1$ clause~(ii) below would force $L_1$ to grow like $\sqrt{d_1}$
whenever $\sigma''\not\equiv0$, and clause~(i) would force $\Lambda_1$ to grow
like $d_1$ for rows of order $r_{\max}\geq1$. We say that the rows satisfy
$(\mathsf G^H_\tau)$ if there are an initialization event $\mathcal E_{(0)}$
with $\Pr(\mathcal E_{(0)})\geq1-\delta/2$, constants
$\Lambda_H\geq\lambda_0>0$, $L_H<\infty$ and $\overline r^H_\delta<\infty$
independent of $d^{[1]},\ldots,d^{[H]}$, a layer index $l_\star\in[H+1]$, a
deterministic matrix $\mathbf K^{[l_\star],\infty}$, posited to be the limit in probability of $\mathbf K^{[l_\star]}_{(0)}$ as all hidden widths tend to infinity, and a width threshold
$\mathcal N_{\mathrm{init}}
=\mathcal N_{\mathrm{init}}(\lambda_0,N,H,r_{\max},\delta)$, which may also depend on the fixed rows, $\sigma$, $\gamma$ and the laws $\mu^{[l]}$ but not on the widths, such that
clauses (i), (ii), and (iv) hold on $\mathcal E_{(0)}$, clause (iii) is a
deterministic statement about the posited limit, and clause (v) is a
separate event whose intersection with $\mathcal E_{(0)}$ has probability at
least $1-\delta$.
\begin{enumerate}[label=\textup{(\roman*)},leftmargin=*,nosep]
\item \textup{(Uniform Jacobian envelope.)}
 $\displaystyle\sup_{\theta\in\mathcal T^H_\tau(\theta_{(0)})}
 \|\mathbf J(\theta)\|_2^2\leq\Lambda_H$; in particular
 $\lambda_{\max}(\mathbf K(\theta))\leq\Lambda_H$ throughout the trust region.
 This generalizes the aggregate gradient bound $B_*^2$
 of~\cref{eqn:abstract_aggregate_constants}, equivalently the choice
 $\Lambda=B_*^2$ of~\Cref{thm:generalconvergence}, which at $H=1$ follows from
 the per-unit bounds $B_i$ by blockwise concatenation \emph{on the per-unit
 tube}.
\item \textup{(Jacobian Lipschitz modulus with an explicit width factor.)}
 For all $\theta,\theta'\in\mathcal T^H_\tau(\theta_{(0)})$,
 \begin{equation}
  \|\mathbf J(\theta)-\mathbf J(\theta')\|_2
  \leq\frac{L_H}{\sqrt{d_{\min}}}\|\theta-\theta'\|_2 .
  \label{eqn:depth_jacobian_lipschitz}
 \end{equation}
 This generalizes~\cref{eqn:jacobian_lipschitz_global}, whose modulus
 $L_*d_1^{-1/2}$ is \emph{derived} from the per-unit constants $L_i$ by
 blockwise separability of the Jacobian columns. At depth the columns
 belonging to different units are not separable in this way: the $H=1$ grouping $\theta_m=(\boldsymbol\xi_m,\mathbf a_m)$ has no depth analogue, since for $H\geq2$ the columns of a unit of layer $l$ depend on parameters of other units, through $\mathbf y^{[l-1]}$ when $l\geq2$ and through $\partial\mathbf u/\partial\mathbf t^{[l]}$, which involves every unit of layer $l$ and the layers above, when $l<H$; so the factor
 $d_{\min}^{-1/2}$ is part of the hypothesis.
\item \textup{(Certified limiting gap at one layer.)}
 $\lambda_{\min}(\mathbf K^{[l_\star],\infty})\geq\lambda_0>0$. This is the
 depth analogue of~\Cref{def:positive_definiteness_certificate} together with
 the certified bound $0<\lambda_0\leq\lambda_{\min}(\mathbf K_{(0)}^\infty)$
 of~\Cref{thm:linearPDEsconvergence}; by~\Cref{lem:layerwise_decomposition} it
 certifies the full limiting kernel provided the remaining blockwise limits
 exist, only $\mathbf K^{[l_\star],\infty}$ being posited here. The clause
 \emph{posits} the limiting matrix: at depth it is not produced by a strong
 law over units, as it is in~\Cref{thm:linearNTK}.
 \Cref{prop:depth_h_substitution} does not consume this clause; it only ties
 $\lambda_0$ to a limiting object.
\item \textup{(Initialization concentration at the certified layer.)} If
 $d_{\min}\geq\mathcal N_{\mathrm{init}}$ then, on $\mathcal E_{(0)}$,
 \begin{equation}
  \lambda_{\min}\bigl(\mathbf K^{[l_\star]}_{(0)}\bigr)
  \geq\frac{3\lambda_0}{4}.
  \label{eqn:depth_init_concentration}
 \end{equation}
 The norm bound
 $\bigl\|\mathbf K^{[l_\star]}_{(0)}-\mathbf K^{[l_\star],\infty}\bigr\|_2
 \leq\lambda_0/4$ is one sufficient condition, and is the form that a
 layerwise propagation argument of the kind used by~\citet{Du2019b} would
 deliver; we state the eigenvalue bound because that is what the proof
 consumes, and because the two-layer tool being replaced,
 \Cref{prop:positive_definitenessfinitewidth}, produces
 the full-kernel counterpart $\lambda_{\min}(\mathbf K_{(0)})\geq3\lambda_0/4$ of~\cref{eqn:depth_init_concentration} directly by a matrix Chernoff bound on
 the independent positive-semidefinite per-unit matrices $\mathbf G_m$ and
 never passes through the norm bound. At depth that independence fails, as the
 discussion after~\cref{eqn:depth_schur_product} records, so the required
 input is a layerwise propagation of the initialization error, conditional at
 each layer, rather than one concentration inequality; and by the same
 discussion only $l_\star\in\{H,H+1\}$ presents a conditionally independent
 average to propagate into.
\item \textup{(Initial residual envelope.)}
 $\Pr\{\|\mathbf r_{(0)}\|_2\leq\overline r^H_\delta\}\geq1-\delta/2$, with
 $\overline r^H_\delta$ independent of the widths. This replaces
 \Cref{lem:initial_residual_bound}, whose McDiarmid step uses the decomposition
 of $\mathbf v_P(\theta_{(0)})$ into $d_1$ independent blocks, each changing
 the norm by at most $2A_{0,*}/\sqrt{d_1}$. At depth such a decomposition
 holds only conditionally on the layers below the output, and for rows of
 order $r_{\max}\geq1$ it has no width-independent bounded-difference
 constant. For centered output weights the clause follows from clause~(i)
 instead (\Cref{sec:depth_conjectural}).
\end{enumerate}
\end{definition}

Clauses (i), (ii) and (v) are statements about the whole network on a
high-probability event; clauses (iii) and (iv) are statements about one layer.
This is the sharpest way to put the difference from the two-layer case:
\Cref{assump:general} is verified by inspecting \emph{one neuron} on the fixed
compact tube $\supp\mu+\overline{\mathbb B}_R(\mathbf0)$, and its aggregate
constants $A_{0,*},B_*,L_*$ are width-independent \emph{by compactness},
whereas $\mathcal T^H_\tau(\theta_{(0)})$, though of fixed radius, lies in a
space whose dimension grows with the widths, and for $H\geq2$ its units are
coupled, so no compactness argument is available and width-independence of
$\Lambda_H$ and $L_H$ must be \emph{proved} on $\mathcal E_{(0)}$, for instance
from operator-norm control of the initialization matrices,
$\|\mathbf W^{[l]}_{(0)}\|_2=O(\sqrt{d^{[l]}}+\sqrt{d^{[l-1]}})$ for centered
$\mu^{[l]}$, as in~\citet{Du2019b}.

At $H=1$ the two conditions are comparable. The per-unit tube
$\mathcal T_R(\theta_{(0)})$ is a product of radius-$R$ per-unit balls, so the
Euclidean ball $\mathcal T^1_R(\theta_{(0)})$ lies inside it. Hence, on the
event $\theta_{(0)}\in(\supp\mu)^{d_1}$, $(\mathsf G_R)$ supplies clauses~(i)
and~(ii) on $\mathcal T^1_R(\theta_{(0)})$ with $\Lambda_1=B_*^2$
(\Cref{prop:positive_definiteness_via_parameters}) and $L_1=L_*$
(\cref{eqn:jacobian_lipschitz_global}). With, in addition,
$\overline r^1_\delta=\overline r_\delta$, $l_\star$ the output block,
$\mathbf K^{[2],\infty}=\mathbf K_{\mathrm{out}}^\infty$,
$\lambda_0\leq\lambda_{\min}(\mathbf K_{\mathrm{out}}^\infty)$ (a block bound,
which~\Cref{thm:linearPDEsconvergence} admits but which may be strictly
smaller than the full-kernel gap $\lambda_{\min}(\mathbf K_{(0)}^\infty)$ used
in~\Cref{cor:classwise_gd_convergence}), and $\mathcal N_{\mathrm{init}}$ the
matrix-Chernoff threshold $R_Kc_{\mathrm{Ch}}^{-1}\lambda_0^{-1}\log(2N/\delta)$
of~\Cref{thm:linearPDEsconvergence}, the constants, the first two width
conditions and the arithmetic of the two-layer proof are reproduced
termwise.

\subsubsection{The implication}

\begin{proposition}[Depth-$H$ substitution into the two-layer argument]
\label{prop:depth_h_substitution}
Consider~\Cref{def:depth_h_network} and gradient descent as
in~\Cref{def:gradient_descent}, and suppose the rows satisfy
$(\mathsf G^H_\tau)$ of~\Cref{def:depth_row_regularity}, with constants
$\Lambda_H\geq\lambda_0>0$, $L_H$, $\overline r^H_\delta$, layer $l_\star$ and
threshold $\mathcal N_{\mathrm{init}}$, and that
$\theta\mapsto\mathbf v_P(\theta)$ is continuously differentiable on a convex
neighborhood of $\mathcal T^H_\tau(\theta_{(0)})$ (for rows of type (i) this
follows from clause $\mathsf P_{\mathrm{reg}}$ of~\Cref{assump:pde_classes}).
If
\begin{equation}
 d_{\min}\geq\max\left\{
 \mathcal N_{\mathrm{init}},\
 (320+128\sqrt6)
   \frac{L_H^2\Lambda_H(\overline r^H_\delta)^2}{\lambda_0^3}
 \right\},
 \quad
 4\sqrt{\Lambda_H}\,\overline r^H_\delta\leq\tau\lambda_0,
 \quad 0<\eta\Lambda_H\leq\frac12,
 \label{eqn:depth_width_conditions}
\end{equation}
then, with probability at least $1-\delta$, simultaneously for every
$k\in\mathbb N_0$,
\begin{equation}
 \theta_{(k)}\in\mathcal T^H_\tau(\theta_{(0)}),
 \qquad
 \mathbf K_{(k)}\succeq\frac{\lambda_0}{2}\mathbf I_N,
 \qquad
 \mathcal L(\theta_{(k)})
 \leq(1-\eta\lambda_0)^k\mathcal L(\theta_{(0)}).
 \label{eqn:depth_conclusion}
\end{equation}
\end{proposition}

We stress what this proposition is and is not. It is a substitution lemma: its
hypotheses are precisely the quantities that the two-layer proof consumes, and
its content is that the two-layer bookkeeping survives the substitution. It is
not a depth-$H$ convergence theorem, because every clause
of~\Cref{def:depth_row_regularity} is assumed. In particular nothing here
claims that $\Lambda_H$, $L_H$, $\mathcal N_{\mathrm{init}}$ or
$\overline r^H_\delta$ are finite, width-independent, or of any particular size
in $H$ and $r_{\max}$; see~\Cref{conj:depth_jacobian_constants} for the first
three and~\Cref{sec:depth_conjectural} for the last.

Nor does it contain the two-layer theorem at $H=1$. With the choices above,
$\tau=R$ and $d_{\min}=d_1$, the first two width conditions
of~\cref{eqn:depth_width_conditions} are those
of~\cref{eqn:linear_pde_convergence_conditions}, and the last two assertions
of~\cref{eqn:depth_conclusion} are the last two
of~\cref{eqn:linear_convergence_main}. The localization differs:
\Cref{thm:linearPDEsconvergence} needs only a width condition to localize every
unit, $\max_{m\in[d_1]}\|\theta_{m,(k)}-\theta_{m,(0)}\|_2\leq R$, which is
bought from the per-neuron one-step bound~\cref{eqn:one_step_block_motion},
whereas the proposition needs the radius condition
$4B_*\overline r_\delta\leq R\lambda_0$ and keeps the iterates in a Euclidean
ball of radius $R$. It also certifies a single block. The
proposition reproduces the two-layer bookkeeping; it does not subsume the
two-layer theorem.

\begin{proof}[Proof sketch]
Intersecting the two events in~\Cref{def:depth_row_regularity}(iv) and (v)
leaves probability at least $1-\delta$; work on that intersection and set
$c_\eta=1-\eta\lambda_0$ and $x_\eta=\eta\lambda_0$. Exactly as
in~\cref{eqn:inductionhypo}, we prove by strong induction that
$\theta_{(j)}\in\mathcal T^H_\tau(\theta_{(0)})$ and
$\mathcal L(\theta_{(j)})\leq c_\eta^j\mathcal L(\theta_{(0)})$ for
$0\leq j\leq n$; the case $n=0$ is immediate. Each step below names the
two-layer ingredient and the object that replaces it.

\emph{\textup{(S1)} Taylor decomposition.} \Cref{prop:NTKtaylor} is
architecture-free: the fundamental theorem of calculus along the update segment
and the chain rule give
$\mathbf r_{(k+1)}=(\mathbf I_N-2\eta\mathbf K_{(k)})\mathbf r_{(k)}
-\mathbf R_{(k)}$ with the gradient-integral
remainder~\cref{eqn:gradient_integral_remainder}. Its only structural input is
that the segment $[\theta_{(k)},\theta_{(k+1)}]$ lie in a convex set on which
$\theta\mapsto\mathbf v_P(\theta)$ is continuously differentiable. This is
the differentiability hypothesis of the proposition; for rows of type (i) it
holds at every depth under clause $\mathsf P_{\mathrm{reg}}$
of~\Cref{assump:pde_classes}, since finitely many compositions of
$C^{r_{\max}+1,1}_{\mathrm{loc}}$ maps are again $C^{r_{\max}+1,1}$ on compact
sets, so $\mathbf J$ exists and is continuous; clause (ii) then makes it
Lipschitz on the trust region. The per-unit product
tube~\cref{eqn:abstract_product_tube} is replaced
by~\cref{eqn:depth_trust_region}, a Euclidean ball and hence convex.
Nothing else changes.

\emph{\textup{(S2)} Parameter displacement.} The global one-step estimate
of~\Cref{lem:parameterdifferencestep6} uses only
$\|\Delta\theta_{(n)}\|_2=2\eta\|\mathbf J_{(n)}^\top\mathbf r_{(n)}\|_2
\leq2\eta\sqrt{\Lambda_H}\|\mathbf r_{(n)}\|_2$, that is, clause (i) alone;
summing the geometric series as in~\Cref{lem:stronginduction1} gives
$\|\theta_{(k+1)}-\theta_{(0)}\|_2
\leq4\sqrt{\Lambda_H}\|\mathbf r_{(0)}\|_2/\lambda_0$. The \emph{per-neuron}
one-step estimate~\cref{eqn:one_step_block_motion}, which uses
$\|\boldsymbol\Psi(\theta_{m,(n)})\|_2\leq B_*$ and has no depth analogue, is
not needed: by clause (v) and the radius condition
in~\cref{eqn:depth_width_conditions},
$4\sqrt{\Lambda_H}\|\mathbf r_{(0)}\|_2/\lambda_0\leq\tau$, so
$\theta_{(k+1)}\in\mathcal T^H_\tau(\theta_{(0)})$, and convexity places the
whole update segment there. This is the step at which the shape of the
localization changes: a per-unit displacement bound is traded for the
width-independent Euclidean bound
$4\sqrt{\Lambda_H}\,\overline r^H_\delta/\lambda_0$ on the whole parameter
vector, and the per-unit conclusion is lost.

\emph{\textup{(S3)} Remainder bound.} \Cref{lem:residualtermstep} is reproved
verbatim with~\cref{eqn:depth_jacobian_lipschitz} in place
of~\cref{eqn:jacobian_lipschitz_global}:
\[
 \|\mathbf R_{(k)}\|_2
 \leq\frac{L_H}{2\sqrt{d_{\min}}}\|\Delta\theta_{(k)}\|_2^2
 \leq\frac{2L_H}{\sqrt{d_{\min}}}\eta^2\Lambda_H\|\mathbf r_{(k)}\|_2^2 .
\]
This is the substitution that is not a free lunch: at $H=1$ the factor
$d_1^{-1/2}$ is derived from blockwise separability, whereas here it is
clause (ii).

\emph{\textup{(S4)} Spectral lower bound along training.} Here
\Cref{lem:layerwise_decomposition} earns its place. Write
$\sigma_{\min}(\mathbf A)$ for the $N$th largest singular value of
$\mathbf A\in\mathbb R^{N\times D'}$, so that
$\lambda_{\min}(\mathbf A\mathbf A^\top)=\sigma_{\min}(\mathbf A)^2$; clause
(iv) forces $D'\geq N$ for the block $l_\star$, since otherwise
$\sigma_{\min}(\mathbf J^{[l_\star]})=0$. By clause (iv),
$\sigma_{\min}(\mathbf J^{[l_\star]}_{(0)})
=\lambda_{\min}(\mathbf K^{[l_\star]}_{(0)})^{1/2}
\geq\tfrac{\sqrt3}{2}\sqrt{\lambda_0}$. Since
$\mathbf J^{[l_\star]}(\theta)-\mathbf J^{[l_\star]}(\theta')$ is a column
block of $\mathbf J(\theta)-\mathbf J(\theta')$, and since for any matrix
$\mathbf M$ with column blocks $\mathbf M^{[l]}$ and any
$\mathbf c\in\mathbb R^N$ one has
$\|\mathbf M^{[l]\top}\mathbf c\|_2\leq\|\mathbf M^\top\mathbf c\|_2$, its
spectral norm is no larger, so~\cref{eqn:depth_jacobian_lipschitz} and (S2)
give
\[
 \bigl\|\mathbf J^{[l_\star]}_{(k)}-\mathbf J^{[l_\star]}_{(0)}\bigr\|_2
 \leq\frac{L_H}{\sqrt{d_{\min}}}\cdot
 \frac{4\sqrt{\Lambda_H}}{\lambda_0}\|\mathbf r_{(0)}\|_2
 \leq\frac{\sqrt3-\sqrt2}{2}\sqrt{\lambda_0},
\]
the last inequality being the second width condition
in~\cref{eqn:depth_width_conditions} together with clause (v), because
$64/(\sqrt3-\sqrt2)^2=320+128\sqrt6$. Singular-value perturbation yields
$\sigma_{\min}(\mathbf J^{[l_\star]}_{(k)})
\geq\tfrac{\sqrt2}{2}\sqrt{\lambda_0}$, hence
$\lambda_{\min}(\mathbf K^{[l_\star]}_{(k)})\geq\lambda_0/2$, and
\cref{eqn:layerwise_kernel_and_gap} promotes this to
$\lambda_{\min}(\mathbf K_{(k)})\geq\lambda_0/2$. The upper bound
$\lambda_{\max}(\mathbf K_{(k)})\leq\Lambda_H$ comes from clause (i), a
statement about all layers; the layerwise lemma does not supply it. This step
is the exact analogue of~\Cref{prop:NTKposdef_eachiter}, with a one-layer
certificate in place of the full-kernel one.

\emph{\textup{(S5)} Spectral one-step descent.} With
$\tfrac{\lambda_0}{2}\mathbf I_N\preceq\mathbf K_{(k)}
\preceq\Lambda_H\mathbf I_N$ and $\eta\Lambda_H\leq\tfrac12$, every eigenvalue
of $\mathbf I_N-2\eta\mathbf K_{(k)}$ is nonnegative and its norm is at most
$1-x_\eta$. For the remainder, the induction hypothesis gives
$\|\mathbf r_{(k)}\|_2\leq c_\eta^{k/2}\|\mathbf r_{(0)}\|_2
\leq\|\mathbf r_{(0)}\|_2\leq\overline r^H_\delta$, so (S3) together with
$\eta\Lambda_H\leq\tfrac12$ gives
\[
 \|\mathbf R_{(k)}\|_2
 \leq\frac{\eta L_H}{\sqrt{d_{\min}}}\|\mathbf r_{(k)}\|_2^2
 \leq\frac{\eta L_H\overline r^H_\delta}{\sqrt{d_{\min}}}
     \|\mathbf r_{(k)}\|_2 .
\]
Because $320+128\sqrt6>3+2\sqrt2$ and $\Lambda_H\geq\lambda_0$, the second
width condition implies
$d_{\min}\geq(3+2\sqrt2)L_H^2(\overline r^H_\delta)^2/\lambda_0^2$, and
$\sqrt{3+2\sqrt2}=1+\sqrt2=(\sqrt2-1)^{-1}$, so
$L_H\overline r^H_\delta/\sqrt{d_{\min}}\leq(\sqrt2-1)\lambda_0$ and therefore
$\|\mathbf R_{(k)}\|_2\leq(\sqrt2-1)x_\eta\|\mathbf r_{(k)}\|_2$. The
concluding inequality in the proof of~\Cref{lem:spectralonestep}, whose only
inputs are the recursion of (S1), the spectral bracket, this remainder bound and $x_\eta\leq\eta\Lambda_H\leq\tfrac12$ (by $\Lambda_H\geq\lambda_0$), then gives
$\|\mathbf r_{(k+1)}\|_2\leq\sqrt{c_\eta}\,\|\mathbf r_{(k)}\|_2$.

\emph{\textup{(S6)} Closing the induction.} Squaring (S5) proves the loss
assertion at $k+1$ and (S2) proves the localization assertion, which closes the
strong induction and gives~\cref{eqn:depth_conclusion}. Every arithmetic
constant is the two-layer one, with $(B_*^2,L_*,d_1,\overline r_\delta)$
replaced throughout by $(\Lambda_H,L_H,d_{\min},\overline r^H_\delta)$, the
matrix-Chernoff threshold by $\mathcal N_{\mathrm{init}}$, and the per-unit
localization condition by the radius condition.
\end{proof}

\subsubsection{What remains conjectural}
\label{sec:depth_conjectural}

Whether $(\mathsf G^H_\tau)$ \emph{holds} for PINN rows at depth is open. Two
ingredients are missing: a quantitative bound on the constants of clauses (i),
(ii) and (iv), and a positivity criterion for clause (iii). We state each as a
conjecture and then record what is and is not known about it. Clause (v) is
not a third missing ingredient once the output weights are centered,
$\int z\,d\mu^{[H+1]}(z)=0$, as both conjectures assume. The rows are linear
and the output layer carries no bias, so
$\mathbf v_P(\theta_{(0)})=\mathbf J^{[H+1]}(\theta_{(0)})\theta^{[H+1]}_{(0)}$,
the first factor being a function of the first $H$ layers alone. Clause (i) at
$\theta_{(0)}$ places $\mathcal E_{(0)}$ inside the event
$\{\|\mathbf J^{[H+1]}(\theta_{(0)})\|_F^2\leq N\Lambda_H\}$, which is
determined by those layers, and with $m_2=\int z^2\,d\mu^{[H+1]}(z)$ a
conditional Chebyshev bound gives
$\Pr\bigl(\mathcal E_{(0)}\cap\{\|\mathbf r_{(0)}\|_2\leq\|\mathbf v_T\|_2
+(2m_2N\Lambda_H/\delta)^{1/2}\}\bigr)\geq1-\delta$. That width-independent
envelope is the only use the proof of~\Cref{prop:depth_h_substitution} makes of
clause (v).

\begin{conjecture}[Depth and differential-order dependence of the constants]
\label{conj:depth_jacobian_constants}
Consider~\Cref{def:depth_h_network} with the pointwise rows
of~\Cref{def:linear_pde} of largest order $r_{\max}$, under clause
$\mathsf P_{\mathrm{reg}}$ of~\Cref{assump:pde_classes} and with compactly
supported $\mu^{[1]},\ldots,\mu^{[H+1]}$, with $\int z\,d\mu^{[l]}(z)=0$ for
$l=2,\ldots,H+1$. For every $\tau>0$ and
$0<\delta<1$ there are an event $\mathcal E_{(0)}$ with
$\Pr(\mathcal E_{(0)})\geq1-\delta/2$ and a constant $C_0<\infty$ depending
only on $\sigma$, $\gamma$, $r_{\max}$, $\tau$, a common bound
$M\geq\sup\{|z|:z\in\supp\mu^{[l]},\ l\in[H+1]\}$,
$\overline\Omega$ and the row data, but not on $H$ or on the widths, such that
clauses (i) and (ii) of~\Cref{def:depth_row_regularity} hold, whenever
$d_{\min}\geq C_0\log(2(H+1)/\delta)$, on
$\mathcal E_{(0)}$ with
\begin{equation}
 \max\bigl\{\Lambda_H,\;L_H^2\bigr\}\leq C_0^{(1+r_{\max})H},
 \label{eqn:depth_constant_rate}
\end{equation}
and, given a $\lambda_0$ as in clause (iii), the threshold
$\mathcal N_{\mathrm{init}}$ of clause (iv) obeys the same
exponential rate in $(1+r_{\max})H$, times a factor polynomial in $N$,
$\lambda_0^{-1}$ and $\log(1/\delta)$.
\end{conjecture}

\begin{remark}[Why the differential order enters the exponent]
\label{rem:depth_faa_di_bruno}
The differential order is what
separates~\Cref{conj:depth_jacobian_constants} from the corresponding
statement at order zero. A strong-form row has Jacobian row
$s_i\sum_{q=1}^{d_2}\sum_{\alpha\in I_{iq}}a_{iq\alpha}(\mathbf x_i)\,
\nabla_\theta\partial^\alpha_{\mathbf x}u_q(\mathbf x_i;\theta)^\top$, a fixed
linear combination, with coefficients from the row data, of terms with
$|\alpha|\leq r_{\max}$; it therefore suffices to consider one term
$\nabla_\theta\partial^\alpha_{\mathbf x}u_q(\mathbf x_i;\theta)$, and
in~\cref{eqn:depth_network} the spatial derivative obeys
\begin{equation}
 \partial_{x_j}\mathbf y^{[l]}
 =\operatorname{diag}\bigl(\sigma'(\mathbf t^{[l]})\bigr)
 \frac1{\sqrt{d^{[l-1]}}}\mathbf W^{[l]}\,\partial_{x_j}\mathbf y^{[l-1]},
 \qquad l=2,\ldots,H,
 \label{eqn:depth_x_derivative_chain}
\end{equation}
with $\partial_{x_j}\mathbf y^{[1]}
=\operatorname{diag}(\sigma'(\mathbf t^{[1]}))\mathbf W^{[1]}\mathbf e_j$. Each
of the $|\alpha|$ spatial derivatives therefore inserts one further scaled
weight factor into every layer of the chain, and Fa\`a di Bruno expands
$\partial^\alpha$ into a sum over the ways of distributing those derivatives
among the $H$ compositions, with $\sigma^{(p)}$, $p\leq r_{\max}$, at each
node; the parameter gradient then appends the usual backward chain. Bounding
the result term by term multiplies $H$ layer factors, each of which now carries
up to $r_{\max}$ weight insertions, which is the route to the rate
in~\cref{eqn:depth_constant_rate}. Three things are unproved. The bound itself:
a layerwise induction of the kind used by~\citet{Du2019b} at order zero is the
natural route, but the jet enlargement described
below~\cref{eqn:depth_schur_product} has no counterpart there.
Width-independence: unlike the two-layer case it cannot be read off from
compactness of a fixed tube, because the ball~\cref{eqn:depth_trust_region}
lies in a space whose dimension grows with the widths and, for $H\geq2$, its
units are coupled. And sharpness: we do not know that
an exponential rate in $(1+r_{\max})H$ is necessary, nor even that an
exponential rate in $H$ is necessary at $r_{\max}=0$ for laws normalized to
the activation, as the factor $c_\sigma=(\mathbb E_{z\sim N(0,1)}\sigma(z)^2)^{-1}$
of~\citet{Du2019b} normalizes them; without such a normalization,
which~\Cref{def:depth_h_network} does not impose, the envelope can grow
exponentially in $H$ already at $\theta_{(0)}$. On that last point the
comparison with~\citet{Du2019b} must be drawn carefully. Their $2^{O(H)}$ arises in both phases of the fully connected analysis. At
initialization it is the constant in the propagation of the initialization
error through the layers, for the Gram matrix
$\varepsilon_H\lesssim2^{O(H)}\varepsilon_1$, and this is the analogue of the
threshold $\mathcal N_{\mathrm{init}}$ in clause (iv). During training it is
the layerwise amplification of the weight displacement into the change of the
layer outputs and of the Gram matrix, and of the weight norms into the size of
the gradient. This produces the $n^4\lambda_{\min}^{-4}$ term of their width
and the factor $2^{-O(H)}$ in their step size, and it is the analogue of the
modulus $L_H$ and the envelope $\Lambda_H$, the quantities
that~\cref{eqn:depth_x_derivative_chain} governs. Because these are upper
bounds, they are not evidence that any of the three must grow exponentially;
Du et al.\ add that fully connected networks need not depend exponentially on
$H$, and the factor is replaced by $\mathrm{poly}(H)$ under residual
scaling. The qualitative
smoothness requirement is, by contrast, unchanged with depth: finitely many
compositions of $C^{r_{\max}+1,1}_{\mathrm{loc}}$ maps are again
$C^{r_{\max}+1,1}$ on compact sets, so clause $\mathsf P_{\mathrm{reg}}$
of~\Cref{assump:pde_classes} remains the right regularity to impose. What
degrades with depth is quantitative, not qualitative.
\end{remark}

\begin{conjecture}[Positivity criterion at depth]
\label{conj:depth_positivity_criterion}
Consider~\Cref{def:depth_h_network} with the pointwise rows
of~\Cref{def:linear_pde}, under the depth-$H$ analogue of clause
$\mathsf P_{\mathrm{pd}}$ of~\Cref{assump:pde_classes}: $\sigma$ real analytic
and nonpolynomial, $\gamma>0$, $\int z\,d\mu^{[l]}(z)=0$ for $l=2,\ldots,H$,
and hidden-parameter laws $\mu^{[l]}$ whose
supports have nonempty interior. For every $H\geq1$ there is a layer
$l_\star\in\{H,H+1\}$ for which $\mathbf K^{[l_\star]}_{(0)}$ converges in
probability, as $d_{\min}\to\infty$ uniformly over the width profile
$(d^{[1]},\ldots,d^{[H]})$, to a deterministic matrix
$\mathbf K^{[l_\star],\infty}$, and
\begin{equation}
 \ker\mathbf K^{[l_\star],\infty}
 =\Bigl\{\mathbf c\in\mathbb R^N:
 \sum_{i\in I_{\mathbf x}}c_i\mathbf b_i(\mathbf x)=\mathbf0
 \ \text{ at every sampled site }\mathbf x\Bigr\},
 \label{eqn:depth_positivity_kernel}
\end{equation}
the sitewise coefficient nullspace of~\cref{eqn:exact_dntk_rank}. Consequently
clause (iii) of~\Cref{def:depth_row_regularity} holds with some $\lambda_0>0$
exactly when every local family
$\{\mathbf b_i(\mathbf x):i\in I_{\mathbf x}\}$ is linearly independent, at
every depth.
\end{conjecture}

\begin{remark}[Why the input-separation criterion is the wrong one]
\label{rem:depth_positivity_criterion}
\Citet{Du2019b} certify positivity of their limiting Gram matrix for value
rows, for an analytic nonpolynomial activation, under the hypothesis that no
two training inputs are parallel. That is the wrong criterion for PINN rows.
\Cref{prop:limiting_ntk_rank} makes positivity a \emph{sitewise coefficient
rank} condition, as in~\cref{eqn:exact_dntk_rank}: site geometry is neither
necessary nor sufficient for degeneracy. It is not necessary: a row with
$\mathbf b_i(\mathbf x_i)=\mathbf0$ has $P_i(\theta)=0$ for every $\theta$,
hence gives $\mathbf K(\theta)$ a zero row, however well separated the sites
are (\Cref{sec:prior_work_comparison}). It is not sufficient, and two worked
examples place the obstruction in the coefficients instead:
\Cref{ex:shared_point_rank_obstruction} exhibits two rows at one shared site
that are singular for one choice of operator coefficients and positive definite
for another, and~\Cref{ex:vector_coupling_rank_obstruction} exhibits three
coupled vector rows at a single site with
$\nabla_\theta P_3=\nabla_\theta P_1+\nabla_\theta P_2$. The second example
transfers to depth unconditionally, and it is worth isolating what it proves:
an exact linear dependence among the loss-scaled rows \emph{as functionals of}
$\mathbf u$ is an identity in $\theta$, so the associated null direction lies
in $\ker\mathbf K(\theta)$ at every width, every depth, every architecture and
every iterate, and no input-separation hypothesis can repair it.
\Cref{conj:depth_positivity_criterion} asserts that the positive direction
transfers as well. Its restriction of $l_\star$ to $\{H,H+1\}$ is not a
convenience: by the discussion after~\cref{eqn:depth_schur_product} these are
the only blocks that are, conditionally on the layers below, averages over the
units of one layer of i.i.d.\ terms, hence the only blocks for which the conditional-independence route to the finite-width concentration of clause~(iv) is available. The restriction concerns that concentration, not the existence of a limit: for value rows and Gaussian initialization, the layerwise induction by which~\citet{Jacot2018} identify the deep NTK yields the sequential infinite-width limit of every block, but it supplies no finite-width rate. Of the two, $l_\star=H+1$ is the more
convenient, for the reason the two-layer certificates use
$\mathbf K^\infty_{\mathrm{out}}$: the output block's Jacobian involves only
the order-$r_{\max}$ jet of the layer-$H$ features, so the criterion is again a
statement about the span of a feature family evaluated against the rows. The
gap is precise. The two-layer proof separates features by analyticity in the
parameters of a \emph{single} neuron together with
$\operatorname{int}(\supp\mu_h)\neq\varnothing$; at depth the layer-$H$ feature
is a composition whose law is determined by the layers below, and that
propagated law is controlled in the depth literature only through moments.
For Gaussian initialization, positivity at pairwise-distinct sites is
nevertheless known at every depth: \citet{Carvalho2025} prove the
value-observation NTK strictly positive definite at distinct inputs with bias,
and at nonzero, pairwise nonproportional inputs without it, for continuous,
almost everywhere differentiable, nonpolynomial activations;
\citet[Theorem~2]{Jin2026} prove the infinite-width covariance and DNTK
positive definite for $\gamma>0$, smooth nonpolynomial or RePU $\sigma$, and
one linear differential row per site whose coefficients do not all vanish
there; and~\citet{Du2019b} certify their last-hidden-layer block for analytic
nonpolynomial activations under non-parallel inputs in the bias-free scaling.
None of these treats the compactly supported laws
of~\Cref{def:depth_h_network} or co-located and coupled rows, and we therefore
regard~\Cref{conj:depth_positivity_criterion} as open for $H\geq2$, even at
$r_{\max}=0$; at $H=1$ it holds with $l_\star=2$,
by~\cref{eqn:exact_output_dntk_nullspace}.
\end{remark}

\begin{remark}[What the literature establishes at depth]
\label{rem:depth_literature}
For an ordinary supervised loss at differential order zero an analogue of the depth-$H$
program has been carried out, but only under Gaussian initialization, in bias-free parameterizations and under input-separation hypotheses; for~\Cref{def:depth_h_network} its positivity step remains open even at $r_{\max}=0$, as~\Cref{rem:depth_positivity_criterion} records. \Citet{Du2019b} prove global linear convergence of
gradient descent for deep fully connected networks, ResNets and convolutional
ResNets on the squared loss over value observations, in a bias-free
parameterization with a uniform hidden width $m$, standard Gaussian
initialization, normalized inputs $\|\mathbf x_i\|_2=1$, targets $|y_i|=O(1)$,
and an activation that is Lipschitz with a Lipschitz derivative, their guiding
example being softplus and their positivity proposition additionally requiring
analyticity and nonpolynomiality. They use exactly the layerwise decomposition
of~\Cref{lem:layerwise_decomposition}, work with the block of the last hidden
layer, localize by the averaged Frobenius norm
$m^{-1/2}\|\mathbf W^{[l]}_{(k)}-\mathbf W^{[l]}_{(0)}\|_F$ rather than by a
per-neuron tube, and require for the fully connected architecture a width
$m=\Omega\bigl(2^{O(H)}\max\{n^4\lambda_{\min}^{-4},\,n\delta^{-1},\,
n^2\lambda_{\min}^{-2}\log(Hn/\delta)\}\bigr)$ and a step size
$\eta=O\bigl(\lambda_{\min}2^{-O(H)}n^{-2}\bigr)$, where, in their notation, $n$ is the sample size (our $N$) and
$\lambda_{\min}$ is the least eigenvalue of their limiting Gram matrix of the
weights of the last hidden layer, the analogue of $\mathbf K^{[H],\infty}$ and
not of the output block $\mathbf K^{[H+1],\infty}$; they attribute the exponential factor to the instability of the fully
connected architecture, which in their analysis amplifies perturbations
through the layers both at initialization and during training, and reduce it
to $\mathrm{poly}(H)$ by the residual scaling.
\Citet{AllenZhu2019} and~\citet{Zou2020} prove convergence for deep ReLU
networks under separation conditions on the training inputs, each training a proper subset of the layers (the output layer is held fixed) from a He-type Gaussian initialization, with width and iteration count polynomial in the depth already for fully connected networks. For ReLU at that initialization the fully connected architecture therefore does not by itself force the $2^{O(H)}$ of~\citet{Du2019b}, as their own discussion anticipates; the two settings are not directly comparable (ReLU and input separation versus a smooth analytic activation and $\lambda_{\min}$), and since ReLU violates clause $\mathsf P_{\mathrm{reg}}$ these rates do not decide the sharpness question of~\Cref{rem:depth_faa_di_bruno}. Both results are again for an
ordinary empirical loss. \Citet{Jacot2018} and~\citet{Lee2019} identify and
freeze the deep NTK in the infinite-width limit, once more for value
observations. None of these results covers a row of positive differential
order, and none supplies the jet structure described
below~\cref{eqn:depth_schur_product}. On the PINN side, \citet{Wang2022}
write the finite-width joint residual--boundary NTK for an arbitrary
architecture but prove its infinite-width limit and its constancy along
gradient flow only for a one-hidden-layer network on a one-dimensional
Poisson problem (the constancy only on a finite horizon $[0,T]$ and under assumed width-uniform bounds on the parameters and on the time-integrated residuals along the flow), and note that multiple hidden layers are not covered by
their theory. At depth, \citet[Theorem~2]{Jin2026} prove, for standard-Gaussian parameters, a positive bias scale and sequentially infinite widths, that the limiting output covariance and DNTK are positive definite at pairwise-distinct sites carrying one nonzero linear differential row each. This is the Gaussian analogue of clause~(iii) of~\Cref{def:depth_row_regularity} at the output layer, in their parameterization with an output bias; it does not reach the compactly supported laws or the repeated sites of~\Cref{conj:depth_positivity_criterion}, and it is not a training statement. The depth-$H$ training statement we are aware of assumes
what~\Cref{def:depth_row_regularity} makes explicit rather than proving it:
\citet{Gan2025} prove uniform convergence in
probability of the operator-transformed kernel for a deep bias-free network
with a single fixed linear differential operator and state a training theorem
that assumes limiting-Gram positivity and uniform gradient continuity. The
kernel-convergence question for general, possibly nonlinear operators is
studied by~\citet{Zhou2024}, but for two-layer pointwise PINNs under a width
scaling $d_1^{-\varrho}$, so that result is not a depth-$H$ statement either. We are
aware of no finite-width discrete-gradient-descent guarantee for depth-$H$
PINNs with joint residual and boundary rows of arbitrary fixed order; that is
what proofs
of~\Cref{conj:depth_jacobian_constants,conj:depth_positivity_criterion},
together with~\Cref{prop:depth_h_substitution}, would deliver for centered
output weights, clause (v) then following from clause (i)
(\Cref{sec:depth_conjectural}).
\end{remark}

%% file: theorem_audit_results.tex
\section{Experimental protocols and finite-width diagnostics}
\label{sec:completed_empirical_appendix}

This appendix documents the experiments summarized
in~\cref{sec:empirical_validation}. The study comprises 770 PDE training
runs, 1,080 compatible-target two-row control runs, and 7,700 initialization
measurements from 100 independent seed banks per PDE, with widths paired by
nested prefixes; two small audits, a three-row joint-nullspace control and a
per-iterate monotonicity check, complete the design. The tracked aggregates
store the displayed summaries with content hashes, whereas per-run records of
the registered campaign are not retained (\Cref{sec:completed_reproduction});
we retain narrow or slowly converging cases.
The campaign snapshot stores configuration, source, and result hashes.
\Cref{tab:completed_empirical_matrix} summarizes the study components.

\subsection{Matched architecture and experimental matrix}

Every experiment uses
\begin{equation*}
 \mathbf u(\mathbf x;\theta)
 =\frac{1}{\sqrt{d_1}}\sum_{m=1}^{d_1}
 \mathbf a_m\tanh(\mathbf w_m^\top\mathbf x+b_m),
 \qquad
 (w_{m\ell},b_m,a_{mq})\stackrel{\text{i.i.d.}}{\sim}
 \Unif[-1,1],
\end{equation*}
with all hidden weights, hidden biases, and output weights trained and no
global output bias. This setup is exactly~\cref{eqn:network_main} with
$\sigma=\tanh$ and $\gamma=1$.
Constraints use the group normalization in~\Cref{def:pde}, and full-batch GD minimizes
$\|\mathbf v_P(\theta)-\mathbf v_T\|_2^2$ with
$\eta=0.2/\lambda_{\max}(\mathbf K_{(0)})$. Computation uses float64 and
deterministic nested initialization prefixes.

\begin{table}[H]
\centering
\caption{Experimental matrix. Independent row designs use two compatible
targets with a shared initial kernel, so optimization-run counts exceed
distinct-kernel counts.}
\label{tab:completed_empirical_matrix}
\small
\begin{tabular}{@{}lrrp{5.7cm}@{}}
\toprule
Study & Instances & Runs & Grid \\
\midrule
Rank/conditioning controls
 & 600 kernels & 1,080
 & six families, five row designs, widths $\{16,256,4{,}096,16{,}384\}$, five paired seeds \\
PDE training diagnostics
 & 11 reference problems & 770
 & 11 PDEs, widths $\{4,16,64,256,1{,}024,4{,}096,16{,}384\}$, ten seeds, 3,000 GD steps \\
Initialization sweep
 & 7,700 kernels & 7,700
 & 11 PDEs, the same seven widths, 100 independent seed banks \\
Joint-nullspace audit
 & 72 kernels & 36
 & three designs, four perturbation sizes, widths $\{32,128\}$, three seeds, 5,000 GD steps \\
Per-iterate audit
 & 30 kernels & 30
 & one two-row design, widths $\{256,1{,}024\}$, three row mixings, five seeds, 3,000 GD steps \\
\bottomrule
\end{tabular}
\end{table}

\subsection{PDE benchmarks and manufactured data}
\label{sec:completed_pde_specifications}

All spatial domains are $[-1,1]$ or $[-1,1]^2$, and time-dependent problems
use $(t,x)\in[0,1]\times[-1,1]$. We obtain the forcing terms and all boundary
or initial data by substituting the manufactured solutions below into the
listed operators. The same discrete functional defines each target and the
training loss; this choice is especially relevant for the weak, nonlocal,
and fractional rows. We fix the case-specific collocation locations
independently of initialization and list them below.
The first seven cases
use automatic differentiation to evaluate the displayed pointwise derivatives
at the listed sites; they introduce no PDE-operator quadrature.
\Cref{tab:completed_pde_benchmarks} gives the full suite.

\begin{table}[t]
\centering
\caption{Eleven PDE benchmarks in the 770-run training study. Here $N$ counts
scalar constraints after splitting vector equations into rows, and $d_2$ is
the network output dimension. Applying the displayed value or derivative
operator to each manufactured solution yields the corresponding data.}
\label{tab:completed_pde_benchmarks}
\small
\setlength{\tabcolsep}{3pt}
\begin{tabular}{@{}>{\raggedright\arraybackslash}p{2.45cm}rrp{8.15cm}@{}}
\toprule
Problem & $N$ & $d_2$ & PDE and sampled data \\
\midrule
Poisson 1D & 7 & 1
 & $-u_{xx}=f$; five interior and two endpoint Dirichlet rows. \\
Variable elliptic 2D & 13 & 1
 & $-\nabla\!\cdot(a\nabla u)+0.35u=f$ with
 $a(x,y)=2+0.3x+0.2y$; five interior rows and eight Robin rows
 $\partial_nu+\beta u=g$, where $\beta=0.75+0.08x-0.05y$. \\
Heat 1D & 17 & 1
 & $u_t-0.35u_{xx}=f$; six interior, five initial, and six endpoint
 Dirichlet rows. \\
Transport 1D & 15 & 1
 & $u_t+c(t,x)u_x+0.25u=f$ with
 $c(t,x)=0.65+0.12x+0.08t$; six interior, five initial, and four
 inflow rows at $x=-1$. \\
Wave 1D & 22 & 1
 & $u_{tt}-0.8^2u_{xx}=f$; six interior rows, five co-located
 initial displacement/velocity pairs, and six endpoint Dirichlet rows. \\
Biharmonic 2D & 21 & 1
 & $\Delta^2u=f$; five interior rows and eight co-located clamped pairs
 $u=g_0$, $\partial_nu=g_1$. \\
Stokes 2D & 32 & 3
 & $-0.2\Delta\mathbf v+\nabla p=\mathbf f$ and
 $\nabla\!\cdot\mathbf v=0$; five shared interior locations, one pressure
 anchor, and both velocity components at eight boundary locations. \\
Weak Poisson 1D & 6 & 1
 & Four rows $\int_{-1}^{1}u'v_n'\,dx$ with
 $v_n'=\sqrt{(2n+1)/2}\,P_n$, $n=1,\ldots,4$, plus two endpoint
 Dirichlet rows; 16-node Gauss--Legendre quadrature. \\
Nonlocal diffusion 1D & 5 & 1
 & Three rows at $x_i\in\{-0.5,0,0.5\}$ of
 $\int_{-1}^{1}\kappa(x_i,y)\{u(x_i)-u(y)\}\,dy=f(x_i)$, where
 $\kappa(x,y)=\exp\!\left[-\tfrac12\left(\frac{x-y}{0.5}\right)^2\right]$, plus two endpoint
 Dirichlet rows; 18-node Gauss--Legendre quadrature. \\
Integro-differential 1D & 6 & 1
 & Five rows of $u'(x)+(0.4+0.1x)u(x)
 -0.55\int_{-1}^{1}\exp\!\left[-\tfrac12\left(\frac{x-y}{0.42}\right)^2\right]u(y)\,dy=f(x)$ and one
 Dirichlet row at $x=-1$; 18-node Gauss--Legendre quadrature. \\
Caputo diffusion 1D & 17 & 1
 & ${}^{C}D_t^{0.6}u-0.3u_{xx}=f$; six fractional residual, five
 initial, and six endpoint Dirichlet rows; 16-node transformed quadrature
 for the Caputo derivative. \\
\bottomrule
\end{tabular}
\end{table}

\begin{table}[H]
\centering
\caption{Finite-horizon GD coverage by observation class. ``Loss decreased''
means only $\mathcal L_{3{,}000}<\mathcal L_0$; it neither asserts monotonicity
at unsaved iterates nor verifies the theorem's sufficient width and step-size
conditions. Each problem contributes seven widths and ten initialization
seeds. Weak and integral/nonlocal entries refer to the stored quadrature
functionals.}
\label{tab:completed_operator_class_gd}
\small
\setlength{\tabcolsep}{4pt}
\begin{tabular}{@{}>{\raggedright\arraybackslash}p{2.8cm}
                    >{\raggedright\arraybackslash}p{6.2cm}rr@{}}
\toprule
Observation class & Included problems and rows & Runs & Loss decreased \\
\midrule
Strong-form local
 & Seven scalar or vector strong-form systems: point evaluations and
 derivatives through order four, including the biharmonic and Stokes systems
 & 490 & 490/490 \\
Weak variational
 & Weak Poisson with four fixed test functions and endpoint traces
 & 70 & 70/70 \\
Integral/nonlocal
 & Nonlocal diffusion, an integro-differential equation, and transformed
 Caputo-time rows
 & 210 & 210/210 \\
\midrule
All classes & Eleven fixed finite constraint systems & 770 & 770/770 \\
\bottomrule
\end{tabular}
\end{table}

The strong-form row samples derivative orders only through four; it is not an
experiment over every finite order. The theorem instead applies to each
chosen fixed finite order when the activation has the corresponding
smoothness, with constants that may depend on that order and the design.
Likewise, the last two rows provide optimization evidence for their exact
discrete prediction maps, not a continuum-consistency result.

The grouping in~\cref{tab:completed_operator_class_gd} counts observation
classes: seven strong-form, one weak, and three integral or nonlocal problems.
It is not the certificate-route split (seven designs certified by the
pointwise procedure, three by the weak jet-atom procedure, and one by the
nonlocal signed-measure procedure;
\cref{sec:constructive_positivity_certificates}). The integro-differential and Caputo designs
are observed as nonlocal but certified through their stored quadrature
functionals, which accounts for the difference between the two splits. The
stored route for each design is recorded in its reference manifest in the
accompanying code.

The scalar manufactured solutions, in the order
of~\cref{tab:completed_pde_benchmarks}, are
\begin{align*}
u^*_{\text{P}}(x)
 &=\sin(1.3x+0.2)+0.17\cos(2.1x-0.1),\\
u^*_{\text{VE}}(x,y)
 &=\sin(0.8x+0.3)\cos(1.1y-0.2)+0.13\sin(xy+0.4),\\
u^*_{\text{H}}(t,x)
 &=e^{-0.7t}\{\sin(1.2x+0.25)+0.2\cos(0.7x-0.1)\},\\
u^*_{\text{T}}(t,x)
 &=\sin(0.9x-0.8t+0.35)+0.16\cos(1.3x+0.4t).
\end{align*}
The next four cases use
\begin{align*}
u^*_{\text{W}}(t,x)
 &=\cos(1.1t+0.1)\sin(0.9x+0.4)
   +0.15\sin(0.7t+1.3x),\\
u^*_{\text{B}}(x,y)
 &=\sin(0.8x+0.2)\cos(1.05y-0.3)+0.11\cos(0.6xy+0.2),\\
u^*_{\text{WP}}(x)
 &=\sin(1.15x+0.3)+0.19\cos(0.65x-0.2),\\
u^*_{\text{NL}}(x)
 &=\sin(0.95x+0.23)+0.21\cos(1.45x-0.17).
\end{align*}
The final pair is
\begin{align*}
u^*_{\text{ID}}(x)
 &=e^{0.2x}\sin(1.1x+0.4)+0.3\cos(0.6x-0.1),\\
u^*_{\text{C}}(t,x)
 &=(1+t^2)\sin(0.9x+0.2)+0.1e^{-0.3t}\cos(0.7x-0.1).
\end{align*}
For Stokes, $\mathbf u^*=(v_1^*,v_2^*,p^*)$ is
\begin{align*}
v_1^*(x,y)&=1.1\sin(0.8x+0.2)\cos(1.1y-0.3)+0.1x,\\
v_2^*(x,y)&=-0.8\cos(0.8x+0.2)\sin(1.1y-0.3)-0.1y,\\
p^*(x,y)&=0.3\cos(0.7x-0.4y+0.1)+0.08x-0.05y.
\end{align*}
Thus every PDE target is compatible by construction. The fractional claim
concerns the displayed transformed-quadrature prediction map; we infer no
continuum fractional-space positivity conclusion from that experiment.
The stored discrete functionals act on $C^1([-1,1])$ for weak Poisson,
$C([-1,1])$ for nonlocal diffusion, $C^1([-1,1])$ for the
integro-differential problem, and $C^2([0,1]\times[-1,1])$ for Caputo
diffusion.

\paragraph{Collocation points and loss groups}
We scale every named group below by the inverse square root of its number of
scalar rows. Time-dependent problems use coordinate order $(t,x)$; the other
pairs are $(x,y)$. The fixed point sets are:
\begin{itemize}[leftmargin=*,itemsep=2pt,topsep=3pt]
\item \textbf{Poisson:} interior
$\{-0.79,-0.34,0.07,0.43,0.82\}$ and boundary $\{-1,1\}$,
forming groups of sizes $5$ and $2$.
\item \textbf{Variable elliptic:} interior
\[
 \{(-0.71,-0.48),(-0.38,0.26),(0.06,-0.17),(0.41,0.63),(0.77,-0.31)\},
\]
and Robin boundary points
\[
 \{(-1,-0.64),(-1,0.23),(1,-0.19),(1,0.72),
   (-0.57,-1),(0.18,-1),(-0.24,1),(0.69,1)\},
\]
with outward coordinate normals and group sizes $5$ and $8$.
\item \textbf{Heat:} interior points
\[
 \{(0.13,-0.72),(0.24,0.18),(0.39,-0.27),(0.56,0.69),
   (0.73,-0.51),(0.91,0.34)\},
\]
initial locations $x\in\{-0.83,\allowbreak -0.29,\allowbreak 0.14,\allowbreak 0.58,\allowbreak 0.91\}$, and boundary
points
$\{(0.17,-1),\allowbreak(0.48,-1),\allowbreak(0.86,-1),\allowbreak(0.29,1),\allowbreak(0.67,1),\allowbreak(0.94,1)\}$,
forming groups of sizes $6$, $5$, and $6$.
\item \textbf{Transport:} interior points
\[
 \{(0.11,-0.66),(0.22,0.09),(0.37,-0.24),(0.53,0.73),
   (0.76,-0.39),(0.92,0.41)\},
\]
initial locations $x\in\{-0.78,\allowbreak -0.31,\allowbreak 0.12,\allowbreak 0.49,\allowbreak 0.87\}$, and inflow
points $\{(0.16,-1),\allowbreak(0.43,-1),\allowbreak(0.71,-1),\allowbreak(0.93,-1)\}$, forming groups
of sizes $6$, $5$, and $4$.
\item \textbf{Wave:} interior points
\[
 \{(0.12,-0.58),(0.28,0.21),(0.44,-0.33),(0.61,0.74),
   (0.79,-0.47),(0.93,0.38)\},
\]
initial locations $x\in\{-0.82,\allowbreak -0.36,\allowbreak 0.09,\allowbreak 0.53,\allowbreak 0.88\}$, and boundary
points
$\{(0.18,-1),\allowbreak(0.52,-1),\allowbreak(0.87,-1),\allowbreak(0.31,1),\allowbreak(0.68,1),\allowbreak(0.95,1)\}$.
The interior, initial-displacement, initial-velocity, and boundary groups
have sizes $6$, $5$, $5$, and $6$.
\item \textbf{Biharmonic:} interior points
\[
 \{(-0.68,-0.43),(-0.29,0.24),(0.08,-0.16),
   (0.39,0.67),(0.74,-0.28)\},
\]
and boundary points
\[
 \{(-1,-0.61),(-1,0.32),(1,-0.22),(1,0.73),
   (-0.52,-1),(0.19,-1),(-0.27,1),(0.66,1)\}.
\]
The interior, boundary-value, and boundary-normal groups have sizes $5$,
$8$, and $8$; the normal is the outward coordinate normal.
\item \textbf{Stokes:} shared interior locations
\[
 \{(-0.66,-0.41),(-0.31,0.28),(0.07,-0.14),
   (0.42,0.64),(0.76,-0.27)\},
\]
pressure anchor $(0.07,-0.14)$, and velocity boundary locations
\[
 \{(-1,-0.56),(-1,0.37),(1,-0.18),(1,0.69),
   (-0.47,-1),(0.21,-1),(-0.25,1),(0.63,1)\}.
\]
The momentum, continuity, pressure-anchor, and velocity-boundary groups
contain $10$, $5$, $1$, and $16$ scalar rows.
\item \textbf{Weak Poisson:} the four weak rows form one group and use the
standard 16-node Gauss--Legendre nodes on $[-1,1]$; the two endpoint values
form the boundary group. In detail,
\begin{equation}
 v_n(x)=\int_{-1}^{x}\sqrt{\frac{2n+1}{2}}P_n(s)\,ds,
 \qquad
 \lambda_n(u)=\sum_{r=1}^{16}\omega_r u'(y_r)
 \sqrt{\frac{2n+1}{2}}P_n(y_r),
 \quad n=1,\ldots,4.
 \label{eqn:completed_weak_poisson_rows}
\end{equation}
The weak-row formula in~\eqref{eqn:completed_weak_poisson_rows} uses the
Gauss--Legendre nodes and weights $(y_r,\omega_r)$ on $[-1,1]$; in particular,
$v_n(-1)=v_n(1)=0$.
\item \textbf{Nonlocal diffusion:} the anchors $\{-0.5,0,0.5\}$ form the
three-row residual group, with a standard 18-node Gauss--Legendre rule on
$[-1,1]$; the two endpoint values form the boundary group.
\item \textbf{Integro-differential:} residual anchors
$\{-0.76,-0.33,0.08,0.47,0.81\}$ form one five-row group using the same
18-node rule, and $u(-1)$ is a separate one-row boundary group.
\item \textbf{Caputo diffusion:} fractional-residual anchors
\[
 \{(0.16,-0.68),(0.29,0.17),(0.43,-0.26),(0.61,0.72),
   (0.78,-0.44),(0.93,0.36)\},
\]
initial locations $x\in\{-0.81,\allowbreak -0.32,\allowbreak 0.11,\allowbreak 0.55,\allowbreak 0.89\}$, and boundary
points
$\{(0.19,-1),\allowbreak(0.51,-1),\allowbreak(0.84,-1),\allowbreak(0.34,1),\allowbreak(0.69,1),\allowbreak(0.96,1)\}$,
forming groups of sizes $6$, $5$, and $6$.
\end{itemize}

For completeness, if $(z_r,\omega_r)_{r=1}^{16}$ are the Gauss--Legendre
nodes and weights on $[0,1]$, the stored Caputo row at $(t,x)$ uses
\begin{equation}
 {}^C D_t^\alpha u(t,x)
 \approx \frac{t^{1-\alpha}}{(1-\alpha)\Gamma(1-\alpha)}
 \sum_{r=1}^{16}\omega_r
 u_t\!\left(t\left(1-z_r^{1/(1-\alpha)}\right),x\right),
 \qquad \alpha=0.6.
 \label{eqn:completed_caputo_quadrature}
\end{equation}
The change of variables in~\eqref{eqn:completed_caputo_quadrature} cancels the
endpoint singularity before Gaussian quadrature. Together
with~\cref{tab:completed_pde_benchmarks}, these points,
groups, and manufactured solutions specify every constraint used in the
eleven-PDE study.

\subsection{Compatible two-row controls}

The six control constructions use the pairs
in~\cref{tab:completed_control_functionals} before coefficient normalization and
row mixing.
\begin{table}[t]
\centering
\caption{Independent base functionals for the six two-row controls.}
\label{tab:completed_control_functionals}
\small
\begin{tabular}{@{}>{\raggedright\arraybackslash}p{3.0cm}p{9.7cm}@{}}
\toprule
Family & Independent base functionals $(\ell_1,\ell_2)$ \\
\midrule
Pointwise Poisson
 & $(-u''(0.07),u(0.07))$. \\
Vector Stokes
 & $(-0.2\Delta v_1+p_x,-0.2\Delta v_2+p_y)$ at
 $(x,y)=(-0.66,-0.41)$. \\
Weak Poisson
 & The $n=1$ and $n=2$ rows $\int_{-1}^{1}u'v_n'\,dx$ under the
 16-node Gauss--Legendre rule. \\
Nonlocal diffusion
 & The displayed nonlocal residuals at anchors $x=-0.5$ and $x=0.5$
 under the 18-node rule. \\
Weak boundary--residual
 & The Dirichlet value $u(-1)$ and the $n=1$ weak-Poisson row. \\
Nonlocal boundary--residual
 & The Dirichlet value $u(-1)$ and the nonlocal residual at $x=-0.5$. \\
\bottomrule
\end{tabular}
\end{table}
We divide each coefficient vector by its Euclidean norm and apply the
same scaling to its manufactured target. Weak and nonlocal rows are the exact
stored quadrature functionals, so the associated conclusions
concern the optimized discrete loss.

For every control family, let $(\ell_1,\ell_2)$ be the normalized independent
base pair. The row matrix either retains these rows or applies
\begin{equation}
 \mathbf C_\varepsilon
 =\begin{pmatrix}
 1&0\\
 (1+\varepsilon^2)^{-1/2}&
 \varepsilon(1+\varepsilon^2)^{-1/2}
 \end{pmatrix},
 \qquad \varepsilon\in\{1,0.1,0.01,0\}.
 \label{eqn:completed_control_matrix}
\end{equation}
For manufactured data, we apply the same transformed functionals to the
manufactured solution. For the independent baseline and for
$\varepsilon>0$, a second compatible choice adds the antisymmetric
contrast $\tau(-1,1)^\top$, with $\tau=0.5$, to the two raw row targets. We then
apply the common control-group scale $1/\sqrt2$. For the mixed designs,
the perturbation increasingly excites the weak spectral direction as
$\varepsilon$ decreases. At $\varepsilon=0$ the two rows and their
manufactured values are identical. Consequently every reported target obeys
all exact linear relations of its prediction map.

\subsection{Constraint independence and numerical rank}

For $\varepsilon>0$, $\mathbf C_\varepsilon$ is invertible, whereas at
$\varepsilon=0$ it is singular with left-null vector $(1,-1)^\top$; equivalently,
$\mathbf C_0^\top(1,-1)^\top=0$. The pointwise
coefficient vectors, weak functionals, and signed measures of the base pairs are
independent by construction. Thus the rank formulas
in~\Cref{prop:limiting_ntk_rank,prop:functional_ntk_rank,%
prop:nonlocal_ntk_rank} predict nonsingular
two-row Gram matrices for the nonredundant designs and singular ones for the
duplicates.
These controls are deliberately constructed so that their ranks follow
algebraically before any network is initialized. Their purpose is to check
whether the finite-width kernels and optimization trajectories reproduce the
predicted effects of exact and near row dependence, not to show that arbitrary
PDE discretizations are nonredundant.

\Cref{tab:completed_rank_results} reports the comparison.

\begin{table}[H]
\centering
\caption{Structural and observed finite-width Gram status. The output-block
entry reports the single design-side rank certificate; it is not a second
hidden-block test. The 960 qualifying runs use 480 independent kernels and two
compatible targets. The Gram-status and certificate tallies are values of the
registered campaign; the tracked aggregate stores the GD-outcome counts.}
\label{tab:completed_rank_results}
\small
\setlength{\tabcolsep}{3pt}
\begin{tabular}{@{}>{\raggedright\arraybackslash}p{2.1cm}rr
                    >{\raggedright\arraybackslash}p{3.6cm}
                    >{\raggedright\arraybackslash}p{5.0cm}@{}}
\toprule
Rows & Kernels & GD runs & Gram status (pred./obs.) & GD outcome \\
\midrule
Independent & 480 & 960 & nonsingular/\allowbreak nonsingular in 480 of 480
 & Output-block certificate passed in 480 of 480; loss below its initial value
 in 959 of 960 \\
Exact duplicate & 120 & 120 & singular/singular in 120 of 120
 & loss reduced $>100\times$ in 120 of 120 \\
\bottomrule
\end{tabular}
\end{table}

The structural labels in the first half of the status column are algebraic
and do not depend on a numerical tolerance. For the observed label, we
symmetrize the Gram matrix and declare an eigenvalue nonzero when it exceeds
$10^{-12}$ times that matrix's largest eigenvalue; this relative tolerance was
fixed before aggregation and was not selected by family or optimization
outcome. No output-block eigenvalue cutoff is used as a structural test;
instead, the stored exact rank witness certifies that block and hence the full
kernel without a hidden-block PD check. Exact duplication, in contrast, makes the two Jacobian rows
identical and forces singularity independently of floating-point
classification. The agreement in all 600 instances is an implementation
check across six constraint classes; the analytic theorems, rather than the
chosen tolerance, establish the population nullspaces.

\subsection{Positive-definite and compatible singular GD}

All 480 manufactured-data cases whose initial NTK passes the numerical
definiteness test reduce
their training loss by at least a factor of ten within 3,000 steps. Across the
960 runs with either compatible target, 959 finish below their starting loss
and 812 achieve a tenfold reduction; the sole failure is a
contrast run with $d_1=16$. All 720 cases with $d_1\geq256$
improve, and 612 attain $\mathcal L_{3{,}000}/\mathcal L_0\leq0.1$. At
$d_1=16{,}384$, each of the 240 trajectories lowers its training loss. Among them,
200 achieve at least a tenfold reduction, and the median final-to-initial ratio
is $6.89\times10^{-7}$ (the counts 812, 612, and 200 are per-run tallies of
the registered campaign that the tracked aggregate does not store). Contrast cases with severe ill-conditioning can nevertheless
retain most of their starting loss over this finite budget, as indicated by
their slow spectral modes.

For duplicate rows, both predictions and targets lie in
\begin{equation}
 \mathcal U_{\text{dup}}=\{(z,z)^\top:z\in\mathbb R\},
 \qquad
 \mathbf v_T-\mathbf v_P(\theta)\in\mathcal U_{\text{dup}}
 \quad\text{for every }\theta.
 \label{eqn:completed_duplicate_active_subspace}
\end{equation}
The subspace identity in~\eqref{eqn:completed_duplicate_active_subspace} holds
globally. Although the full two-by-two NTK is singular, its compression to
$\mathcal U_{\text{dup}}$ has a positive eigenvalue. This invariant property
is the global invariant-subspace hypothesis
in~\Cref{cor:active_subspace_convergence}, rather than compatibility imposed
only at initialization. All 120 singular runs finish at less than one percent
of their initial loss; 118 reach absolute loss below $10^{-12}$, including all
90 runs with $d_1\geq256$. The remaining two cases with $d_1=16$ still improve
by factors of approximately 694 and 416 (registered-campaign values outside
the tracked aggregate). Under the theorem's other hypotheses, a full
positive-definiteness certificate would remove the target-compatibility
restriction; the duplicate-row results show that such a certificate is
unnecessary when the complete error trajectory lies in an active subspace on which the
compressed positive-definiteness certificate
of~\Cref{def:structural_spaces_compatibility}
holds.

Compatibility is essential to this conclusion. If the two duplicate rows
were assigned the contrast target used for the independent systems, then its
antisymmetric component would be orthogonal to
$\mathcal U_{\mathrm{dup}}$. With the experiment's contrast amplitude
$\tau=0.5$ and common $1/\sqrt{2}$ row scaling, the exact attainable-loss
decomposition gives
\begin{equation*}
 \|\mathbf v_P(\theta)-\mathbf v_T\|_2^2
 =\|\Pi_{\mathcal U_{\mathrm{dup}}}
       (\mathbf v_P(\theta)-\mathbf v_T)\|_2^2+\tau^2
 \geq \tau^2=0.25.
\end{equation*}
No incompatible-target trajectory is included in the reported counts. Thus
the duplicate experiment demonstrates convergence for compatible singular
systems; it does not empirically establish necessity of the compatibility
condition, which the three-row audit below supplies through its
incompatible-target pairs.

In the generated paragraph below, $\epsilon$ is the size of the third-row
perturbation $A+B+\epsilon C$ of that audit and is unrelated to the
row-mixing parameter $\varepsilon$ of the two-row controls. Its rank table
also shows that at $\epsilon=10^{-5}$ the relative tolerance $10^{-12}$
underestimates the rank of the rank-three nonlocal design, which is why the
tolerance discussion above is restricted to the 600 two-row instances.
\input{experiments/theorem_audit/joint_nullspace_controls}

\begin{figure}[t]
  \centering
  \includegraphics[width=\linewidth,clip]{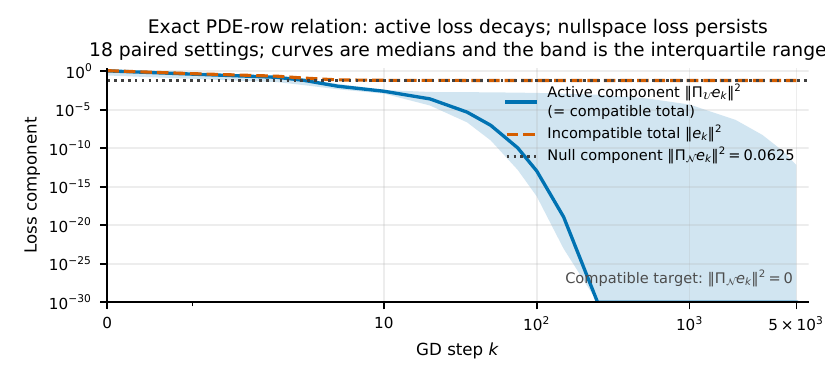}
  \caption{Exact structural nullspace control in three PDE settings. For each
  setting the three loss-scaled prediction rows are $A$, $B$, and $A+B$, so
  every prediction $\mathbf v_P(\theta)\in\mathbb R^3$ satisfies
  $\mathbf z^\top\mathbf v_P(\theta)=0$ for
  $\mathbf z=(1,1,-1)^\top/\sqrt{3}$. Define
  $\mathcal N=\linspan\{\mathbf z\}$,
  $\mathcal U=\mathcal N^\perp$, and the training error
  $\mathbf e_k=\mathbf v_P(\theta_k)-\mathbf v_T$; the plotted loss is
  $\mathcal L_k=\|\mathbf e_k\|_2^2$. A compatible target has
  $\Pi_{\mathcal N}\mathbf e_k=0$. Its paired incompatible target is displaced
  by $0.25\mathbf z$, hence
  $\|\Pi_{\mathcal N}\mathbf e_k\|_2^2=0.25^2=0.0625$ at every step, while
  both targets induce the same parameter updates and therefore share the same
  active loss $\|\Pi_{\mathcal U}\mathbf e_k\|_2^2$. The 18 paired settings
  comprise a co-located vector PDE, a weak-residual--boundary PDE, and a
  nonlocal-residual--boundary PDE, with widths $d_1\in\{32,128\}$ and three
  seeds. Full-batch GD uses
  $\eta=0.2/\lambda_{\max}(\mathbf K_{(0)})$ for $5{,}000$ steps. Curves are
  medians across the 18 settings; the blue band is the interquartile range.
  Compatible nullspace loss is exactly zero and therefore is stated rather than
  drawn on the logarithmic axis. Active losses below $10^{-30}$ are placed at
  the display boundary. At all 19 saved checkpoints, the paired total-loss
  offset agrees with $0.0625$ within $4.44\times10^{-16}$. The weak and
  nonlocal rows are the fixed stored quadrature functionals, so this is a
  discrete-loss diagnostic rather than a continuum-error result.}
  \label{fig:joint_nullspace_loss_decomposition}
\end{figure}

\subsection{Conditioning and the GD rate}

The two singular values of the coefficient transformation
in~\cref{eqn:completed_control_matrix} imply
\begin{equation}
 \lambda_{\min}(\mathbf C_\varepsilon\mathbf C_\varepsilon^\top)
 =1-\frac{1}{\sqrt{1+\varepsilon^2}}
 =\frac{\varepsilon^2}{2}+O(\varepsilon^4).
 \label{eqn:completed_epsilon_squared_conditioning}
\end{equation}

The expansion in~\eqref{eqn:completed_epsilon_squared_conditioning} shows that
reducing $\varepsilon$ preserves rank for every $\varepsilon>0$ and creates an
increasingly weak row direction. Write
$\rho_0:=\lambda_{\min}(\mathbf K_{(0)})/\lambda_{\max}(\mathbf K_{(0)})$ for
the normalized initial minimum eigenvalue and $\kappa_0=1/\rho_0$ for the
initial condition number plotted
in~\cref{fig:theorem_supportive_evidence}(b). In the registered campaign the
median $\rho_0$ decreased as $\varepsilon$ went from $1$ to $0.1$ to $0.01$ in
all 24 family--width settings, and the compatible contrast-target loss ratio
worsened in 23 of 24; the only exception was the nonlocal-diffusion instance at
$d_1=16$. Those 24 settings are recorded in the registered campaign but lie
outside the tracked aggregate, which retains the largest width only.
\Cref{tab:completed_family_conditioning} reports that largest-width control,
where the expected ordering holds in both quantities for all six families. The
independent Poisson baseline without row mixing, likewise recorded in the
original campaign but outside the tracked aggregate, had
$\rho_0=3.621\times10^{-1}$ and
$\mathcal L_{3{,}000}/\mathcal L_0=4.94\times10^{-32}$.

\begin{table}[H]
\centering
\caption{Conditioning and finite-horizon loss at $d_1=16{,}384$. Entries are
five-seed medians after $T=3{,}000$ GD steps, with $\mathcal L_T/\mathcal L_0$
for the compatible contrast target; every entry is taken from the tracked
aggregate that also generates~\cref{fig:theorem_supportive_evidence}(b). Here
$\rho_0$ is the normalized initial minimum eigenvalue defined above. Every row
design is independent and every contrast target is compatible. The weak
boundary--residual entry at $\varepsilon=1$ is exactly zero in float64.}
\label{tab:completed_family_conditioning}
\scriptsize
\setlength{\tabcolsep}{3pt}
\begin{tabular}{@{}l rr rr rr@{}}
\toprule
& \multicolumn{2}{c}{$\varepsilon=1$}
& \multicolumn{2}{c}{$\varepsilon=0.1$}
& \multicolumn{2}{c}{$\varepsilon=0.01$} \\
\cmidrule(lr){2-3}\cmidrule(lr){4-5}\cmidrule(lr){6-7}
Family & $\rho_0$ & $\mathcal L_T/\mathcal L_0$
       & $\rho_0$ & $\mathcal L_T/\mathcal L_0$
       & $\rho_0$ & $\mathcal L_T/\mathcal L_0$ \\
\midrule
Pointwise Poisson
 & $8.778\times10^{-2}$ & $8.03\times10^{-31}$
 & $2.416\times10^{-3}$ & $4.17\times10^{-4}$
 & $2.667\times10^{-5}$ & $1.73\times10^{-1}$ \\
Vector Stokes
 & $2.037\times10^{-1}$ & $5.22\times10^{-32}$
 & $2.665\times10^{-3}$ & $1.10\times10^{-3}$
 & $2.637\times10^{-5}$ & $8.00\times10^{-1}$ \\
Weak Poisson
 & $5.007\times10^{-2}$ & $6.55\times10^{-29}$
 & $5.949\times10^{-4}$ & $5.93\times10^{-2}$
 & $5.956\times10^{-6}$ & $3.41\times10^{-1}$ \\
Nonlocal diffusion
 & $3.871\times10^{-3}$ & $6.36\times10^{-30}$
 & $2.359\times10^{-5}$ & $5.93\times10^{-1}$
 & $1.974\times10^{-7}$ & $9.33\times10^{-1}$ \\
Weak boundary--residual
 & $2.703\times10^{-1}$ & $0$
 & $3.301\times10^{-3}$ & $1.04\times10^{-4}$
 & $3.195\times10^{-5}$ & $3.70\times10^{-1}$ \\
Nonlocal boundary--residual
 & $1.344\times10^{-3}$ & $1.21\times10^{-2}$
 & $1.663\times10^{-5}$ & $3.51\times10^{-1}$
 & $1.687\times10^{-7}$ & $3.71\times10^{-1}$ \\
\bottomrule
\end{tabular}
\end{table}

Two features of~\cref{tab:completed_family_conditioning} qualify how these
numbers may be read. First, between $\varepsilon=0.1$ and $\varepsilon=0.01$
the tabulated medians of $\rho_0$ fall by factors between $90$ and $120$ across
the six families, consistent with the $\varepsilon^2$ scaling
in~\eqref{eqn:completed_epsilon_squared_conditioning}; because these are
medians of per-run ratios rather than a fitted exponent, we report the
comparison as a diagnostic and not as an estimated rate. Second, $\rho_0$ does
not order the finite-horizon outcome across families. At $\varepsilon=0.01$ the
pointwise Poisson, weak boundary--residual, and vector Stokes controls have
median $\kappa_0$ of $3.75\times10^{4}$, $3.13\times10^{4}$, and
$3.79\times10^{4}$ but loss ratios of $1.73\times10^{-1}$,
$3.70\times10^{-1}$, and $8.00\times10^{-1}$; at the ill-conditioned end,
nonlocal boundary--residual at $\kappa_0=5.93\times10^{6}$ finishes at
$3.71\times10^{-1}$ while nonlocal diffusion at $\kappa_0=5.07\times10^{6}$
finishes at $9.33\times10^{-1}$.

Part of the apparent trend in $\kappa_0$ is fixed by the step-size rule rather
than measured. Since $\eta=0.2/\lambda_{\max}(\mathbf K_{(0)})$, every mode of
the frozen residual recursion
$\widetilde{\mathbf r}_{(k+1)}=(\mathbf I-2\eta\mathbf K_{(0)})
\widetilde{\mathbf r}_{(k)}$ contracts by a factor in $[0.6,1-0.4\rho_0]$, so
the frozen loss obeys
$\widetilde{\mathcal L}_{3{,}000}/\mathcal L_0\leq(1-0.4\rho_0)^{6{,}000}$,
which is an arithmetic consequence of the normalization and not an empirical
finding. It is also not a statement about the plotted quantity. The vertical
axis of~\cref{fig:theorem_supportive_evidence}(b) is the observed nonlinear
loss, and three of the eighteen plotted medians---pointwise Poisson, vector
Stokes, and weak Poisson at $\varepsilon=1$---lie many orders of magnitude
above this frozen bound, because those runs reach float64 roundoff long before
step 3,000. Where the bound does hold informatively it does not separate the
families: at $\varepsilon=0.01$ it exceeds $0.92$ for all six, while the
observed medians range from $1.73\times10^{-1}$ to $9.33\times10^{-1}$. We
therefore draw no rate or timescale conclusion from $\kappa_0$. The five
$\varepsilon=1$ medians below the $10^{-16}$ display floor
are drawn at the boundary by the plotting script, so the
small-$\varepsilon$ end of each curve carries the information.

\subsection{Width, NTK stability, and scope}

The PDE training study covers 770 cases outside the two-row controls. Every
run reduces the sampled training loss, and 767 improve it by at least a
factor of ten (registered-campaign tallies; per-run records are not
retained). \Cref{tab:completed_width_results} reports pooled medians that
summarize the transition toward the frozen-kernel regime. For each numerical
population reference, we integrate the output-weight coordinates analytically
and estimate the hidden-parameter expectation with eight independently
scrambled Sobol replicates. Each uses $32{,}768$ fine samples and
its nested first $16{,}384$ samples as a coarse approximation. The base seed is
$1{,}729$, and the chunk size is $2{,}048$. The conservative gap
subtracts $2.365$ times the between-replicate standard error and the
coarse--fine operator-norm difference; its relative numerical tolerance is
$10^{-8}$.

Let $\widehat{\mathbf K}^{\infty}$ denote this numerical population reference,
let $\mathbf K_{(k)}$ be the empirical Gram matrix at GD step $k$, and let
$\mathcal C$ be the set of saved checkpoints. The frozen residual starts from
$\widetilde{\mathbf r}_{(0)}=\mathbf r_{(0)}$ and evolves at every GD step by
$\widetilde{\mathbf r}_{(k+1)}=(\mathbf I-2\eta\mathbf K_{(0)})
\widetilde{\mathbf r}_{(k)}$. With $T=3{,}000$, the three relative diagnostics
in~\cref{tab:completed_width_results} are
\begin{equation*}
 \frac{\|\mathbf K_{(0)}-\widehat{\mathbf K}^{\infty}\|_{\mathrm{op}}}
 {\max\{\|\widehat{\mathbf K}^{\infty}\|_{\mathrm{op}},\mathrm{tiny}_{64}\}},
 \qquad
 \max_{k\in\mathcal C}
 \frac{\|\mathbf K_{(k)}-\mathbf K_{(0)}\|_{\mathrm{op}}}
 {\max\{\|\mathbf K_{(0)}\|_{\mathrm{op}},\mathrm{tiny}_{64}\}},
 \qquad
 \frac{\|\mathbf r_{(T)}-\widetilde{\mathbf r}_{(T)}\|_2}
 {\max\{\|\widetilde{\mathbf r}_{(T)}\|_2,\mathrm{tiny}_{64}\}},
\end{equation*}
respectively, where $\mathrm{tiny}_{64}$ is the smallest positive normal
float64 number. Each ratio is computed per run before taking the pooled median.

\begin{table}[H]
\centering
\caption{Width diagnostics pooled over 110 PDE runs per width, transcribed
from the registered 770-run campaign; the tracked aggregate stores this
transcription with its content hash. All three quantities
are relative; initial-kernel error and maximum drift use operator norm. Frozen
discrepancy compares the nonlinear residual with the trajectory obtained by
holding the empirical Gram matrix at initialization. Gap retained counts runs that preserve at least half
the numerical population-reference gap at every saved checkpoint, not at every
GD iterate.}
\label{tab:completed_width_results}
\small
\begin{tabular}{@{}rrrrr@{}}
\toprule
$d_1$ & \shortstack{Initial kernel\\error}
 & \shortstack{Maximum kernel\\drift}
 & \shortstack{Final frozen\\discrepancy}
 & \shortstack{Gap\\retained} \\
\midrule
16 & $2.23\times10^{-1}$ & $6.11\times10^{-1}$ & $9.08\times10^{-1}$ & 17/110 \\
256 & $5.46\times10^{-2}$ & $4.84\times10^{-2}$ & $2.74\times10^{-1}$ & 110/110 \\
$4{,}096$ & $1.56\times10^{-2}$ & $3.33\times10^{-3}$ & $2.25\times10^{-2}$ & 110/110 \\
$16{,}384$ & $8.09\times10^{-3}$ & $8.62\times10^{-4}$ & $6.91\times10^{-3}$ & 110/110 \\
\bottomrule
\end{tabular}
\end{table}

Across 100 initialization seeds per PDE, log--log slopes of the median
operator error over widths $64$--$16{,}384$ range from $-0.512$ to $-0.469$,
consistent with square-root concentration; these slopes are also
registered-campaign values. These five-scale fits do not
establish a sharp asymptotic exponent. More fundamentally, the
experiment does not certify the deterministic derivative envelopes or the
explicit sufficient-width and step-size inequalities
in~\cref{eqn:linear_pde_convergence_conditions}. The observations cover 3,000
GD
steps with broad diagnostics every 100 steps, so they support finite-horizon
spectral stability and training-loss contraction rather than an all-iteration proof.
Finally, the optimized quantity is sampled constraint loss; we draw no PDE
generalization or continuum-solution conclusion from it.

\Cref{tab:completed_width_results} reports four of the seven trained widths;
$d_1\in\{4,64,1{,}024\}$ were trained and are counted in the 770-run totals but
are not summarized here. At $d_1=4$ the necessary condition
$d_1(d_0+1+d_2)\geq N$ fails for the heat, wave, biharmonic, Stokes, and Caputo
designs. The campaign admits rank-infeasible widths, so those 50 runs are
trained and counted in~\cref{tab:completed_operator_class_gd}; their Jacobians
cannot have full row rank, and we make no positivity or convergence claim for
them. Finally, both runners record, per run, whether the loss is nonincreasing
across saved checkpoints, but the retained 770-run and 1,080-run aggregates
keep no monotonicity count. Consequently we report no checkpoint-monotonicity
frequency for the 770-run or 1,080-run campaigns, and producing one would
require rerunning the registered campaign; the per-iterate audit
of~\Cref{sec:per_iterate_monotonicity} is a separate reduced design recorded
at every step.

\subsection{Per-iterate monotonicity}
\label{sec:per_iterate_monotonicity}

\Cref{thm:linearPDEsconvergence} asserts a decrease at every iterate, whereas
the campaigns above store the loss only at saved checkpoints. This audit closes
that gap on a reduced design for which every step can be recorded.

The network is the two-layer map of~\Cref{def:ntk_parameterization} with
$\sigma=\tanh$, $\gamma=1$, $d_0=d_2=1$, and per-neuron parameters drawn
i.i.d.\ from $\Unif[-1,1]^3$. The two rows are the value row
$\ell_1(u)=u(x_\star)$ and the mixed row
$\ell_2^\varepsilon=(\ell_1+\varepsilon\ell_1')/\sqrt{1+\varepsilon^2}$, where
$\ell_1'(u)=u'(x_\star)$ is the derivative row, at
$x_\star=0.35$, so $\varepsilon$ is the row-mixing parameter used in the
control study. Training is full-batch gradient descent on the unscaled loss
of~\Cref{def:pinn_loss} for $3{,}000$ steps at
$\eta=0.2/\lambda_{\max}(\mathbf K_{(0)})$, with a contrast target that excites
the weak direction. Widths are $d_1\in\{256,1024\}$, row mixings are
$\varepsilon\in\{1,0.1,0.01\}$, and five seeds are run per cell, giving thirty
runs.

An iterate is \emph{scored} while the loss retains double-precision
significance, that is while
$\mathcal L(\theta_{(k)})>10^{-26}\mathcal L(\theta_{(0)})$. The cutoff is
necessary and not conservative: the residual is a difference of two quantities
that agree to machine precision once the loss falls that far, so the recorded
per-step ratios quantize to small integers and carry no information. It is the
same effect that the display floor of~\Cref{fig:theorem_supportive_evidence}
accommodates.

\begin{table}[H]
\centering
\caption{Per-iterate audit. Violations are steps with
$\mathcal L(\theta_{(k+1)})>(1+10^{-12})\mathcal L(\theta_{(k)})$ among scored iterates,
pooled over five seeds. The last two columns are the largest observed per-step
loss ratio and the frozen-kernel prediction
$(1-2\eta\lambda_{\min}(\mathbf K_{(0)}))^2$, reported for the first seed.}
\label{tab:per_iterate_monotonicity}
\small
\begin{tabular}{@{}llrrrr@{}}
\toprule
$d_1$ & $\varepsilon$ & Violations & Scored iterates
 & Worst ratio & Predicted \\
\midrule
$256$   & $1$    & $0$ & $2{,}845$  & --- & --- \\
$256$   & $0.1$  & $0$ & $15{,}000$ & $0.998359$ & $0.998371$ \\
$256$   & $0.01$ & $0$ & $15{,}000$ & $0.999983$ & $0.999983$ \\
$1{,}024$ & $1$    & $0$ & $3{,}015$  & --- & --- \\
$1{,}024$ & $0.1$  & $0$ & $15{,}000$ & $0.998387$ & $0.998389$ \\
$1{,}024$ & $0.01$ & $0$ & $15{,}000$ & $0.999983$ & $0.999983$ \\
\midrule
\multicolumn{2}{@{}l}{Total} & $0$ & $65{,}860$ & & \\
\bottomrule
\end{tabular}
\end{table}

No violation occurs in any scored iterate, and in the four cells that retain
a worst per-step ratio that ratio is below one, so the $10^{-12}$ tolerance
is inactive there. At $\varepsilon=1$ the loss reaches
the precision cutoff before step $3{,}000$, so fewer iterates are scored and no
worst-ratio figure is quoted; the ratios recorded past that point are
cancellation artifacts rather than measurements. At the two smaller mixings the
full horizon is scored and the largest per-step ratio matches the frozen-kernel
factor to a relative discrepancy between $2.8\times10^{-9}$ (at $d_1=1{,}024$,
$\varepsilon=0.01$) and $1.2\times10^{-5}$ (at $d_1=256$, $\varepsilon=0.1$),
which is the sharpest agreement with a predicted rate anywhere in these
experiments.

Two limits. The design is two rows at one collocation point, not one of the
eleven PDE benchmarks, so this audit speaks to the per-iterate claim and not to
coverage. And the agreement with $(1-2\eta\lambda_{\min})^2$ is agreement with
the \emph{frozen} initial kernel, not with the certified constants
of~\cref{eqn:linear_pde_convergence_conditions}; it is consistent with the
kernel drift reported above being small at these widths, and is not a
verification of the theorem's sufficient constants.
The run is the per-iterate monotonicity experiment of the accompanying code,
under its tracked configuration.

\subsection{Reproduction}
\label{sec:completed_reproduction}

The tracked aggregate of the control study and the plotting script in the
accompanying code reproduce \cref{fig:theorem_supportive_evidence} with one
command. The script validates the displayed counts and five-seed medians and
records the hash of the original 1,080-row source table. Summary statistics
cannot reconstruct that source table.

The prepare and submit commands documented with the code rerun the full
campaign through its campaign driver. After such a run, the control
aggregation script selects and aggregates compatible controls, while the
evidence-assessment script produces the operator-class, width,
Jacobian-drift, neuron-motion, frozen-dynamics, and rank-tolerance summaries.
The current repository does not contain the original
per-run 770-PDE, 1,080-control, or 7,700-initialization records; consequently,
recomputing those broad-study summaries requires rerunning the registered
campaign. This limitation does not affect regeneration of the main figure
from its tracked aggregate.

%% file: experiments/theorem_audit/joint_nullspace_controls.tex
\paragraph{Three-row joint-nullspace controls}
We tested rows $A$, $B$, and $A+B+\epsilon C$ for a co-located two-output pointwise design, a weak-quadrature residual jointly with a boundary value, and a finite-measure nonlocal residual jointly with a boundary value.  The audit contains 72 initial kernels (3 cases, 4 values of $\epsilon$, 2 widths, and 3 seeds) in float64.  The parenthesized counts below are agreements with the exact coefficient-design rank $r_\star$; $r_\tau$ is numerical rank at relative tolerance $\tau$, and $\rho_z=\|Kz\|/(\|K\|\|z\|)$.  The displayed rank set records all values observed over the configured widths and seeds.
\begin{center}\scriptsize
\renewcommand{\arraystretch}{1.05}
\setlength{\tabcolsep}{2.8pt}
\begin{tabular}{@{}lcccccc@{}}
\toprule
Design & $\epsilon$ & $r_\star$ & $r_{10^{-6}}$ & $r_{10^{-10}}$ & $r_{10^{-12}}$ & max. $\rho_z$ \\
\midrule
Co-located & $0$ & $2$ & $2\;(6/6)$ & $2\;(6/6)$ & $2\;(6/6)$ & $2.86\times 10^{-16}$ \\
Co-located & $10^{-1}$ & $3$ & $3\;(6/6)$ & $3\;(6/6)$ & $3\;(6/6)$ & $5.06\times 10^{-3}$ \\
Co-located & $10^{-3}$ & $3$ & $2\;(0/6)$ & $3\;(6/6)$ & $3\;(6/6)$ & $3.48\times 10^{-5}$ \\
Co-located & $10^{-5}$ & $3$ & $2\;(0/6)$ & $2\;(0/6)$ & $3\;(6/6)$ & $3.46\times 10^{-7}$ \\
Weak+$\partial$ & $0$ & $2$ & $2\;(6/6)$ & $2\;(6/6)$ & $2\;(6/6)$ & $2.59\times 10^{-16}$ \\
Weak+$\partial$ & $10^{-1}$ & $3$ & $3\;(6/6)$ & $3\;(6/6)$ & $3\;(6/6)$ & $9.45\times 10^{-3}$ \\
Weak+$\partial$ & $10^{-3}$ & $3$ & $2\;(0/6)$ & $3\;(6/6)$ & $3\;(6/6)$ & $8.96\times 10^{-5}$ \\
Weak+$\partial$ & $10^{-5}$ & $3$ & $2\;(0/6)$ & $2\;(0/6)$ & $3\;(6/6)$ & $8.96\times 10^{-7}$ \\
Nonlocal+$\partial$ & $0$ & $2$ & $2\;(6/6)$ & $2\;(6/6)$ & $2\;(6/6)$ & $1.15\times 10^{-16}$ \\
Nonlocal+$\partial$ & $10^{-1}$ & $3$ & $3\;(6/6)$ & $3\;(6/6)$ & $3\;(6/6)$ & $6.46\times 10^{-4}$ \\
Nonlocal+$\partial$ & $10^{-3}$ & $3$ & $2\;(0/6)$ & $3\;(6/6)$ & $3\;(6/6)$ & $6.42\times 10^{-6}$ \\
Nonlocal+$\partial$ & $10^{-5}$ & $3$ & $2\;(0/6)$ & $2\;(0/6)$ & $2\;(0/6)$ & $6.41\times 10^{-8}$ \\
\bottomrule
\end{tabular}
\end{center}
For the exact relation ($\epsilon=0$), we additionally ran 18 paired settings, hence 36 trajectories: manufactured compatible targets and targets displaced only along the known left null direction. The predicted incompatible loss floor was $6.25\times 10^{-2}$. Across all 19 saved steps, the largest error in the predicted paired loss offset was $4.44\times 10^{-16}$; the largest final paired parameter difference was $2.09\times 10^{-14}$.
\begin{center}\small
\setlength{\tabcolsep}{4pt}
\begin{tabular}{@{}lrrrr@{}}
\toprule
Design & pairs & max. $\mathcal L_T^{\mathrm{comp}}$ & max. $\mathcal L_T^{\mathrm{inc}}$ & max. offset error \\
\midrule
Co-located & 6 & $3.16\times 10^{-32}$ & $6.25\times 10^{-2}$ & $1.39\times 10^{-17}$ \\
Weak+$\partial$ & 6 & $5.24\times 10^{-32}$ & $6.25\times 10^{-2}$ & $4.16\times 10^{-17}$ \\
Nonlocal+$\partial$ & 6 & $1.10\times 10^{-10}$ & $6.25\times 10^{-2}$ & $3.47\times 10^{-17}$ \\
\bottomrule
\end{tabular}
\end{center}
These controls diagnose the exact stored finite prediction maps.  The weak and nonlocal rows use fixed 16- and 18-node quadrature, respectively; they do not establish continuum discretization error.  Numerical rank depends on the displayed tolerance, and the GD runs do not certify the theorem's deterministic width or $B_*$ step-size bounds.